\documentclass[letterpaper,12pt,oneside,final]{book}

\makeatletter
\def\bstctlcite{\@ifnextchar[{\@bstctlcite}{\@bstctlcite[@auxout]}}
\def\@bstctlcite[#1]#2{\@bsphack
 \@for\@citeb:=#2\do{%
   \edef\@citeb{\expandafter\@firstofone\@citeb}%
   \if@filesw\immediate\write\csname #1\endcsname{\string\citation{\@citeb}}\fi}%
 \@esphack}
\makeatother

\newcommand\Langue{english}            

\usepackage{ifthen}
\usepackage[utf8]{inputenc}
\usepackage{lmodern}      
\usepackage[T1]{fontenc}  

\ifthenelse{\equal{\Langue}{english}}{
	\usepackage[french,english]{babel}
}{
	\usepackage[english,french]{babel} 
}

\usepackage{graphicx}
\usepackage{epstopdf}  
\graphicspath{{./images/}{./dia/}{./gnuplot/}}
\usepackage{flafter,placeins}
\usepackage[dvipsnames]{xcolor}
\usepackage{amsmath,soulutf8,longtable,colortbl,setspace,xspace,url,pdflscape,cite}
\usepackage{amsfonts}
\usepackage{bm}
\usepackage{csquotes}
\usepackage{enumitem}
\usepackage{tikz}
\usetikzlibrary{positioning,chains}
\usepackage{subcaption}
\usepackage{booktabs}
\usepackage{makecell}

\newcommand{\mat}[1]{\mathbf{#1}}

\newcommand{\ewprod}{\odot}
\newcommand{\reals}{\mathbb{R}}

\newcommand{\x}{\mathbf x}

\newcommand{\y}{\mathbf y}

\DeclareUnicodeCharacter{2212}{-}

\newcommand{\bmx}[0]{\begin{bmatrix}}
\newcommand{\emx}[0]{\end{bmatrix}}

\newcommand{\vect}[1]{\mathbf{#1}}
\newcommand{\vects}[1]{\boldsymbol{#1}}

\newcommand{\TT}[0]{\vects{\theta}}

\newcommand{\todo}[1]{{\Large\textcolor{red}{#1}}}

\usepackage[nolist]{acronym}
\usepackage{fancyhdr}
\fancypagestyle{pagenumber}{\fancyhf{}\fancyhead[R]{\thepage}}

\makeatletter
\let\ps@plain=\ps@pagenumber
\makeatother
\usepackage{hyperref}
\usepackage{caption}  
\makeatletter
\providecommand*{\toclevel@compteur}{0}
\makeatother
\RequirePackage[\Langue]{MemoireThese}
\newcommand\monTitre{Generalizing in the Real World with Representation Learning}
\newcommand\monPrenom{Tegan}
\newcommand\monNom{Maharaj}
\newcommand\monDepartement{génie informatique et génie logiciel}  
\newcommand\maDiscipline{Génie informatique}
\newcommand\monDiplome{D}        
\newcommand\anneeDepot{2022}    
\newcommand\moisDepot{Mai}       

\newcommand\PageGarde{N}         
\newcommand\AnnexesPresentes{N}  
\newcommand\mesMotsClef{Deep Learning, 
Representation Learning,
Generalization,
Regularization,
Neural Networks,
Model Evaluation,
Responsible AI Development,
Specification,
Alignment,
Spatio-temporal data,
Distributional shift,
Epidemiology, 
}

\DeclareMathOperator*{\argmin}{arg\,min}
\newcommand\monJury{\PresidentJury{M}{Desmarais}{Michel}\\
\DirecteurRecherche{M}{Pal}{Christopher}\\
\MembreJury{F}{Agrawal}{Aishwarya}\\
\MembreExterneJury{F}{Mitchell}{Margaret}}

\ifthenelse{\equal{\monDiplome}{M}}{
\newcommand\monSujet{Mémoire de maîtrise}
\newcommand\monDipl{Maîtrise ès sciences appliquées}
}{
\newcommand\monSujet{Thèse de doctorat}
\newcommand\monDipl{Philosophi\ae{} Doctor}
}
\hypersetup{
  pdftitle={\monTitre},
  pdfsubject={\monSujet},
  pdfauthor={\monPrenom{} \monNom},
  pdfkeywords={\mesMotsClef},
  bookmarksnumbered,
  pdfstartview={FitV},
  hidelinks,
  linktoc=all
}
\usepackage{nicefrac}
\usepackage{hyphenat}
\begin{document}
\bstctlcite{IEEEexample:BSTcontrol}

\frontmatter
\ifthenelse{\equal{\PageGarde}{O}}{\addtocounter{page}{1}}{}
\thispagestyle{empty}%
\begin{center}%
\vspace*{\stretch{0.1}}
\textbf{POLYTECHNIQUE MONTRÉAL}\\
affiliée à l'Université de Montréal\\
\vspace*{\stretch{1}}
\textbf{\monTitre}\\
\vspace*{\stretch{1}}
\textbf{\MakeUppercase{\monPrenom~\monNom}}\\
Département de~{\monDepartement}\\
\vspace*{\stretch{1}}
\ifthenelse{\equal{\monDiplome}{M}}{Mémoire présenté}{Thèse présentée} en vue de l'obtention du diplôme de~\emph{\monDipl}\\
\maDiscipline\\
\vskip 0.4in
\moisDepot~\anneeDepot
\end{center}%
\vspace*{\stretch{1}}
\copyright~\monPrenom~\monNom, \anneeDepot.
\newpage\thispagestyle{empty}%
\begin{center}%

\vspace*{\stretch{0.1}}
\textbf{POLYTECHNIQUE MONTRÉAL}\\
affiliée à l'Université de Montréal\\
\vspace*{\stretch{2}}
Ce\ifthenelse{\equal{\monDiplome}{M}}{~mémoire intitulé}{tte thèse intitulée} :\\
\vspace*{\stretch{1}}
\textbf{\monTitre}\\
\vspace*{\stretch{1}}
présenté\ifthenelse{\equal{\monDiplome}{M}}{}{e}
par~\textbf{\mbox{\monPrenom~\MakeUppercase{\monNom}}}\\
en vue de l'obtention du diplôme de~\emph{\mbox{\monDipl}}\\
a été dûment accepté\ifthenelse{\equal{\monDiplome}{M}}{}{e} par le jury d'examen constitué de :\end{center}
\vspace*{\stretch{2}}
\monJury
\pagestyle{pagenumber}%

\newpage

\begin{displayquote}
\strut

\vspace{1in}
{\small ``We want AI agents that can discover like we can, not which contain what we have discovered.''}

\hspace{0.5in}{\footnotesize -- Rich Sutton, \textit{The Bitter Lesson} (2019)}

\vspace{1in}
{\small ``There is something at work in my soul, which I do not understand.''}

\hspace{0.5in}{\footnotesize -- Mary Shelley, \textit{Frankenstein}, (1818)}

\vspace{1in}
{\small ``Curiously enough, the only thing that went through the mind of the bowl of petunias as it fell was Oh no, not again. Many people have speculated that if we knew exactly why the bowl of petunias had thought that we would know a lot more about the nature of the Universe than we do now.''}

\hspace{0.5in}{\footnotesize -- Douglas Adams, \textit{The Hitchhiker's Guide to the Galaxy}, (1979)}
\end{displayquote}

\newpage
\ifthenelse{\equal{\Langue}{english}}{
	\chapter*{DEDICATION}\thispagestyle{headings}
	\addcontentsline{toc}{compteur}{DEDICATION}
}{
	\chapter*{DÉDICACE}\thispagestyle{headings}
	\addcontentsline{toc}{compteur}{DÉDICACE}
}

\begin{flushleft}
  \itshape
  This thesis is dedicated to all those once and future of Mila,\\
  for being part of that emergent wonderland\\
  imperfect utopia of creative and collaborative science.\\
  ~\\
  Especially\\
  to Chris,\\ 
  ~~~
  for taking a chance on me,\\
  and to David, \\~~~
  for believing in me.\\
  ~\\
\end{flushleft}

\newpage

\ifthenelse{\equal{\Langue}{english}}{
	\chapter*{ACKNOWLEDGEMENTS}\thispagestyle{headings}
	\addcontentsline{toc}{compteur}{ACKNOWLEDGEMENTS}
}{
	\chapter*{REMERCIEMENTS}\thispagestyle{headings}
	\addcontentsline{toc}{compteur}{REMERCIEMENTS}
}

I have the great good fortune to write the final drafts of this thesis as an Assistant Professor at the University of Toronto. A huge thank you to all my new colleagues for welcoming me to the Faculty of Information, Schwartz Reisman Institute, and Vector - I'm excited and proud to start the next chapter of my career in such an illustrious and interdisciplinary place. 

It's easy looking back from my dream job to say all the work in this thesis (and much more done during my degree that didn't fit in here) was worth it. But perhaps more surprisingly, it would also have been easy for me to do that at almost any point in the last 6 years. Things were hard sometimes, but I learned so much, and met so many amazing people in this journey - while I doubted myself at times, I don't think I ever doubted that I was doing the best thing I could be. It's the first time I've felt that way for years at a time, and it's a pretty great feeling. 
For this, I am grateful to more people than I can cover here, but I'm going to give it the old college try. 

I am first and foremost grateful to my family for their invaluable love and support - including all my extended family and family by choice. From a young age you've made me feel the world is full of kind, interesting, intelligent, creative people who will be there for each other and for me. I'm lucky to have you all. I'm especially grateful to my parents, Laurie and Ken, for always taking the time to describe, explain, educate, and discuss; to Keir, Aaricia, Amiani, Amy, Kathleen, Carlos, Chuck, Corinne, Jordan, Ben, and Lisa, for making me feel at home wherever I am; and to David, Don, Val, Susan, Clint, and Gretchen for welcoming me into your families. 


I'm grateful to Yoshua Bengio, for setting up a lab truly founded on the best of science - curiosity-driven, egalitarian, and humanitarian. 
I knew I wanted to apply to Mila (then Lisa) after reading Representation Learning [Yoshua Bengio, Aaron Courville, Pascal Vincent], but I might never have done so without the encouragement of several wonderful people - thank you Guillaume Alain, Umut Şimşekli, Orhan Firat, Chris Maddison, and Jamie Kiros for taking the time to talk with an interested Master's student. Thank you especially to Jeff Clune, for making me believe it was possible, and for almost a decade of long-distance friendship. Even after all that encouragement, I might have given up on the whole thing if not for Kyle Gill - Thank you.

I'm so grateful to everyone at Mila, for helping me feel at home in the field of AI and in the wider world. This is a special place. Particular shoutouts to:
 \\  $\cdot$ Chris Beckham for breaking the Mila ice with me;
 \\  $\cdot$ Harm deVries, David Krueger, Asja Fischer, Daniel Jiwoong Im, Faruk Ahmed, Negar Rostamzadeh, Tim Cooijmans, Nicolas Ballas, Samira Shabanian, and Olivier Mastropietro for welcoming me to the Main Lab and for all the good times;
 \\ $\cdot$ Nicolas Ballas for being an incredibly competent and kind postdoc;
 \\ $\cdot$ Chiheb Trabelsi for your generous good spirit and high fives;
 \\ $\cdot$ Olexa Bilaniuk for long conversations about interesting esoterica, and for YEARS of patient, informative, helpful help everyone;
 \\ $\cdot$ Alex Lamb for excellent absurdity, and to Alex, Rosemary Nan Ke, Anirudh Goyal, and the rest of the late night crew for being such good company,
 \\ $\cdot$ Tabasco and Aunt Dai's, for letting us close out your evenings with such exciting discussions as ``is gradient descent like a butterfly'', ``how swamp coolers work and how many we need for the NeurIPS deadline this year'' and ``the list of things Schmidhuber was right about'';
 \\ $\cdot$ to Caglar Gulcehre, Sherjil Ozair, Dmitry Serdyuk, David Warde Farley, Kyle Kaster, Hugo Larochelle, Martin Arjovsky, Marcin Moczulski, Julian Serban, Kyunghyun Cho, and many others for setting such a high standard of in-depth thoughful mailing list replies, knowledge sharing, and open collaboration;
 \\ $\cdot$ to Aaron Courville and Simon Lacoste Julien for years of mentorship, \& of thoughtful and fun neighbourly conversation; Doina Precup for clear and insightful teaching, and Joelle Pineau for leadership and excellent practical advice;
  \\ $\cdot$ Pascal Lamblin, Frederic Bastien, Mathieu Germain, Linda Penethiere, Myriam Cote, Guillaume Alain, and Emelie Brunet for going far above and beyond to Do The Right Thing;
  \\ $\cdot$ the GPU spreadsheet for many lessons in resource allocation and diplomacy, RIP and never come back (thank you Mathieu!);
  \\ $\cdot$ my fellow inaugural lab reps Dzimitry Bahdanau, Bart van Merriënboer, Tristan Sylvain, Vincent Dumoulin, Sina Honari, and David Krueger, and all the lab reps over the years: Thank you for your service, for making it happen and making it fun, for our discussions and collegial disagreement. Mila wouldn't be the same without you.
  \\ $\cdot$ all the developers of Theano, I hope history will know the importance of your contributions to deep learning. I'm proud of my small contribution (thank you Pascal and CCW!) to this cornerstone of our field;
  \\ $\cdot$ all the residents of the Coin Noir, may all your future workspaces spread as a forest of good cheer with many opportunities for tea;
  \\ $\cdot$ to Benjamin Akera, Lena Ezzine, Charles Onu, and Sumana Basu for great music, conversations, and friendship.

To Mati Roy, Philippe Regnier, Stephanie Pataraccia, Nicolas Lacombe, Amy Zhang, Nasim Rahaman, and everyone at the Macroscope, thank you for the adventure in making community, for the good discussions, fun challenges, delicious dinners and questionable cookies; Suwitra Wonwongwaree for sharing your meal, home, and country with us, and for being serious; to Will Whitney for being the best productivity buddy; Carles Gelada and Emmanuel for jam sessions; Abi See for DDR, comedy, and company; Benjamin Akera for instant friendship (just add books!); to Amy Zhang, for being my date, for planning with me, for all the late nights and camping trips, for *cats and all the love; to Amiani Johns, for festivals and music, for dancing, for olive goo, board games, hikes and skis and camps and discussions, and generally being the best of friends.

To David Rolnick for planting the seed that became Climate Change AI, and for your attention to detail, perserverance, mentorship, and friendship.
To David Rolnick, Priya Donti, Alex Lacoste, Lynn Kaack, Kelly Kochanski, Nikola Milojevik-Dupont, Jan Drgona, and all the rest of the team, you are an inspiring and wonderful group to work with!

To the founders and the current editorial team at JMLR - I'm proud and happy to work in a field founded on open science, and it's been an absolute pleasure to work with all of you and discover it's also founded on collegiality and a great sense of humour.

To all the above, to all my coauthors, and to Orestes Manzanilla, Gabriel Huang, Remi Le Priol, Cesar Laurent, Francesco Visin, Xavier Bouthillier, Jessica Thompson, Ishmael Belgazi, Adriana Romero, Hugo Larochelle, Sandeep Subramanian, Mohammad Pezeshki, Valentin Thomas, Li Yao, Sungjin Ahn, Mehdi Mirza, Sarath Chandar, Yaroslav Ganin, Michael Noukouvich, Jonathan Binas, Joseph Paul Cohen, Brady Neal, Devon Hjelm, Dendi Suhubdy, Akram Erraqabi, Gerry Che, Jose Gallego, Ryan Lowe, Jelena Luketina, Sander Dieleman, Zachary Liption, Zac Kenton, Konrad Zolna, Raymond Chua, Rim Assouel, Jean Harb, Genevieve Fried, Emily Denton, Timnit Gebru, Sumana Basu, Khimya Ketarpal, Audrey Durand, Tess Berthier, Anna Harutyunyan, Jordan Ash, Jan Leike, Christoph Carr, Miles Brundage, Gustavo Lacerda, Chaz Firestone, Simon Osindero, Sasha Luccioni, Sekou-Oumar Kaba, Aristide Baratin, Vikram Voleti, Niki Howe, Victor Schmidt, Tristan Deleu, Martin Weiss, Prateek Gupta, Shems Saleh and many others I am sure I will feel terrible about forgetting: I've missed you these last few years and am so looking forward to seeing you again ... thank you all for wise words, good conversations, and just for being you.

I have been incredibly fortunate in a number of opportunities for internships, conference attendance, jobs, and more. I'm sincerely grateful for these, and for general mentorship and advice from such greats as: Jennifer Chayes, Yoshua Bengio, Graham Taylor, Laurent Charlin, Jeff Clune, David Rolnick, Irina Rish, Hugo Larochelle, Ross Goroshin, Nicolas LeRoux, Jason Weston, Ioannis Mitliagkis, Aishwarya Agrawal, Berhnard Scholkopf, Francis Bach, David Blei, Victoria Krakovna, Max Tegmark, Gillian Hadfield, Shahar Avin, Gauthier Gidel, Audrey Durand, Gretchen Krueger, Yarin Gal, and Chris Pal.

I'm sincerely grateful to NSERC and IVADO for supporting my project(s) with the full-ride funding that made everything possible, and to CIFAR, Compute Canada and the other funding agencies that make not-for-profit science a reality.

To Montreal - your trees and parks and cafes and art and music and festivals and bikefriendliness and so much more. Montreal will always be home to me. Also known as Tiohtià:ke and Mooniyang, Montreal sits on unceded Indigenous lands. I gratefully thank the Kanien’kehá:ka Nation for their stewardship of these lands. Recognizing myself as a settler, I acknowledge my responsibility and connection to the land, and make this acknowledgement as one small step toward decolonization, truth, and reconciliation.

Thank you so much to my thesis committee: Michel Desmarais, Chris Pal, Aishwarya Agrawal, and Meg Mitchell - for making the time in such a busy and difficult time, and for your thoughtful comments, suggestions, and questions!

Thank you Chris. You are the best supervisor I can imagine; you're everything I hope to be as a professor. I feel I've always had just the right amounts of support, guidance, encouragement, freedom, and humour, always in an atmosphere of respect and collaboration. My spaceship continues on good course thanks to you. I will be sharing many a wise metaphor of yours for many a year to come.

Finally, thank you to David Krueger, for taking on all the ups and downs and sidewayses of this adventure with me.           
%
\chapter*{RÉSUMÉ}\thispagestyle{headings}
\addcontentsline{toc}{compteur}{RÉSUMÉ}
\vspace{-0.25cm}
L'apprentissage automatique formalise le problème de faire en sorte que les ordinateurs peuvent apprendre d'expériences comme optimiser la performance mesur\'ee avec une ou des m\'etriques sur une tache d\'efinie pour un ensemble de données. Cela contraste avec l'exigence d'un comportement déterminé en avance (c.-à-d. par règles). La formalisation de ce problème a permis de grands progrès dans de nombreuses applications ayant un impact important dans le monde réel, notamment la traduction, la reconnaissance vocale, les voitures autonomes et la découverte de médicaments.
Cependant, les instanciations pratiques de ce formalisme font de nombreuses hypothèses non-realiste pour les données réels - par exemple, que les données sont indépendantes et identiquement distribuées (i.i.d.) - dont la solidité est rarement étudiée. En réalisant de grands progrès en si peu de temps, le domaine a développé de nombreuses normes et standards ad hoc, axés sur une gamme de taches relativement restreinte.
Alors que les applications d'apprentissage automatique, en particulier dans les systèmes d'intelligence artificielle, deviennent de plus en plus répandues dans le monde réel, nous devons examiner de manière critique ces normes et hypothèses.
Il y a beaucoup de choses que nous ne comprenons toujours pas sur comment et pourquoi les réseaux profonds entraînés avec la descente de gradient sont capables de généraliser aussi bien qu'ils le font, pourquoi ils échouent quand ils le font et comment ils fonctionnent sur des données hors distribution.
Dans cette thèse, je couvre certains de mes travaux visant à mieux comprendre la généralisation de réseaux profonds, j'identifie plusieurs façons dont les hypothèses et les problèmes rencontrés ne parviennent pas à se généraliser au monde réel, et je propose des moyens de remédier à ces échecs dans la pratique. 

\newpage

\chapter*{ABSTRACT}\thispagestyle{headings}
\addcontentsline{toc}{compteur}{ABSTRACT}
\begin{otherlanguage}{english}
\vspace{-0.25cm}
Machine learning (ML) formalizes the problem of getting computers to learn from experience as optimization of performance according to some metric(s) on a set of data examples. This is in contrast to requiring behaviour specified in advance (e.g. by hard-coded rules). Formalization of this problem  has enabled great progress in many applications with large real-world impact, including translation, speech recognition, self-driving cars, and drug discovery.
But practical instantiations of this formalism make many assumptions - for example, that data are i.i.d.: independent and identically distributed - whose soundness is seldom investigated. And in making great progress in such a short time, the field has developed many norms and ad-hoc standards, focused on a relatively small range of problem settings.
As applications of ML, particularly in artificial intelligence (AI) systems, become more pervasive in the real world, we need to critically examine these assumptions, norms, and problem settings, as well as the methods that have become de-facto standards. 
There is much we still do not understand about how and why deep networks trained with stochastic gradient descent are able to generalize as well as they do, why they fail when they do, and how they will perform on out-of-distribution data. 
In this thesis I cover some of my work towards better understanding deep net generalization, identify several ways assumptions and problem settings fail to generalize to the real world, and propose ways to address those failures in practice. 




\end{otherlanguage}

\if{false}
\chapter*{SUMMARY}\label{sec:summary}
\addcontentsline{toc}{compteur}{SUMMARY}

    I introduce my work with a review of artificial intelligence (AI), machine learning (ML), and deep learning concepts and terminology, including a review of recent literature in understanding generalization with deep networks. 
    
The next chapter summarizes two works exploring memorization and generalization behaviour of deep networks [1,2], which discuss a series of experiments contrasting learning behaviour on noise vs. real data distributions. These works provide strong empirical evidence for important elements of folk wisdom about deep nets, including the dependence of deep network learning behaviour on the distribution of training data, and that simple patterns which generalize well are learned first, before memorization/overfitting.  The findings from these highly-cited works have been supported in a wave of empirical and theoretical follow-up studies.
    
After that, I cover work from [3] on a regularization method called Zoneout. Dropout, which randomly sets some activations of a deep network to 0, is a very widely used regularization technique for feed-forward neural networks, enabling them generalize well on a much wider range of atemporal real-world data (e.g.images). But in recurrent neural networks (RNNs), widely used for temporal data, where activations from one timestep become inputs for the next timestep, dropout causes instability and vanishing gradients. Our work proposes a simple alternative, to randomly set activations to their previous value, enabling RNNs to get all the regularization benefits of dropout. Zoneout was widely adopted in real-world applications of RNNs.
    
Next, I present an empirical exploration of learned representations for video data. In this work we propose and investigate a simple predictive task, fill-in-the-blank, and introduce a large dataset for this task. We demonstrate through extensive experiments, including human evaluations, that this simple task significantly improves video description quality, as well as 2D and 3D representations of video data. The usefulness of this simple task has been confirmed in many other works, notably in language modelling, where it is called masking. Large masked language models currently form the basis of Google search results.
    
The next chapter summarizes two works examining real-world consequences of distributional shift, and proposes two methods for responsible AI development that could help prevent or address these consequences: unit tests and sandboxes. First I summarize [5], where we define auto-induced distributional shift (ADS) as shift caused by the presence or actions of an algorithm/agent. We look at conditions under which an incentive for ADS could be revealed by a learning setup, and demonstrate the potential real-world consequences of incentives for ADS in content recommendation systems. 
    While ADS is not categorically undesirable, in some settings like content recommendation, it may be an unacceptable means of solving the problem, but our performance metrics fail to specify this. For example, an algorithm might be able to do well according to predictive accuracy by changing user interests to make them easier to predict, or driving away hard-to-predict users. We develop simple unit tests for reinforcement learning and online supervised learning systems, which detect whether a system may be vulnerable to incentives to achieve performance via ADS. Unit tests like this could form an important component of testing or auditing real-world AI systems. 
    
    Next, I summarize work in developing a detailed agent-based epidemiological simulator, tuned to real virological and behavioural data for Covid-19. This simulator serves as a sandbox for comparing and testing novel digital contact tracing (DCT) methods and other public health interventions. This approach allowed us to discover and deal with two important issues for DCT methods: (1) generating sufficient training data to be robust to a variety of real-world conditions, and (2) a successful DCT method will cause a shift in data, by preventing cases of the disease, and training needs to account for this. Sandboxing with an agent-based simulator is a general approach for evaluating representation learning systems prior to real-word deployment. In the context of digital health, among other benefits this approach could contribute to more efficient design of randomized control trials.
    
   Finally, I cover [5], which develops and tests a novel approach to digital contact tracing (DCT) for Covid-19. This approach casts the problem of DCT as distributed inference of individual-level risk of Covid-19, measured as individual infectiousness. In order to estimate this quantity privately on individual smartphones, we leverage the simulator from [6] to provide rich training data, including variables such as viral load which are not feasibly available at scale in real data, and train set-based predictors offline. We show that this approach provides an excellent trade-off in managing disease: reducing cases more than any tested method, while requiring very little quarantine of healthy individuals. This latter point is important for preventing and addressing pandemic fatigue, which has been shown to erode trust and compliance with public health measures.
   
   I conclude with some observations and takeaways from these works for effective real-world generalization, and identify next steps for research improving generalization of AI systems in realistic settings.

\newpage 
\textbf{In summary}, this thesis contains content verbatim from [2][3][4][7] in Chapters 2, 3, 4, 6 respectively, and selected content adapted from [1] and [5,6] for Chapters 2 and 5 respectively. Each chapter is self-contained, with background and appendices presented within the chapter when relevant.

\begin{itemize}\footnotesize
    \item[{[1]}] Devansh Arpit†, Stanislaw Jastrzebski†, Nicolas Ballas†, David Krueger†, Emmanuel Bengio, Maxinder S Kanwal, Tegan Maharaj, Asja Fischer, Aaron Courville, Yoshua Bengio, Simon Lacoste-Julien. 2017. A closer look at memorization in deep networks. International Conference on Machine Learning (ICML).

    \item[{[2]}] Tegan Maharaj, David Krueger, and Tim Coojimans.  2017.  Memorization in recurrent neural networks. International Conference on Machine Learning (ICML) workshop on Principled Approaches to Deep Learning (PADL).

    \item[{[3]}] David Krueger†, Tegan Maharaj†, János Krámar, Mohammad Pezeshki, Nicolas Ballas, Nan Rosemary Ke, Anirudh Goyal, Yoshua Bengio, Aaron Courville, Christopher Pal. 2017. Zoneout: Regularizing RNNs by randomly preserving hidden activations. International Conference on Learning Representations (ICLR).

    \item[{[4]}] Tegan Maharaj, Nicolas Ballas, Anna Rohrbach, Aaron Courville, Christopher Pal. 2017. A dataset and exploration of models for understanding video data through fill-in-the-blank question-answering. IEEE Conference on Computer Vision and Pattern Recognition (CVPR).

    \item[{[5]}] David Krueger, Tegan Maharaj, Jan Leike. 2019. Revealing the Incentive to Cause Distributional Shift. International Conference on Learning Representations (ICLR) SafeML workshop.

    \item[{[6]}] Prateek Gupta†, Tegan Maharaj†, Martin Weiss†, Nasim Rahaman†, Hannah Alsdurf††, Nanor Minoyan††, Soren Harnois-Leblanc††, Joanna Merckx††, Andrew Williams††, Victor Schmidt, Pierre-Luc St-Charles, Akshay Patel, Yang Zhang, David L. Buckeridge, Christopher Pal, Bernhard Scholkopf, Yoshua Bengio. 2021. Proactive Contact Tracing. Public Library of Science (PLoS) Medecine  (in submission).

    \item[{[7]}] Yoshua Bengio†, Prateek Gupta†, Tegan Maharaj†, Nasim Rahaman†, Martin Weiss†, Tristan Deleu, Eilif Muller, Meng Qu, Victor Schmidt, Pierre-Luc St-Charles, et al. 2020. Predicting infectiousness for proactive contact tracing. International Conference on Learning Representations (ICLR).

\end{itemize}
\vspace{-2mm}
\hspace{5MM} † indicates equal contribution
\vspace{3mm}

\textbf{A note on publication venues}: Conferences, including NeurIPS, ICML, ICLR, and CVPR, are among the top scholarly venues for research in the field of machine learning, comparable to journals in many other academic fields. They publish full-length papers with full peer-review. Workshops are a component of conferences which present short papers (around 4 pages), typically early results, to a more specialized audience than the overall conference. They are peer-reviewed, but non-archival.

\fi      

{\setlength{\parskip}{0pt}
\ifthenelse{\equal{\Langue}{english}}{
	\renewcommand\contentsname{TABLE OF CONTENTS}
}{
	\renewcommand\contentsname{TABLE DES MATIÈRES}
}
\tableofcontents

\mainmatter

\chapter*{PREAMBLE}
\addcontentsline{toc}{compteur}{PREAMBLE}
\pagestyle{pagenumber}

\section*{A. Original contributions \& relation to theme}
\addcontentsline{toc}{section}{A. Original contributions \& relation to theme}\label{sec:summary}

The theme of my thesis is real-world generalization with representation learning. My work in this thesis makes the following original contributions in this theme:

\vspace{2mm}
\textit{Note: I emphasize here individual contributions I made in collaborative works, but these of course depend on, and cannot be cleanly separated from, the contributions of other authors and of the works as a whole, for which credit is due to all the authors as a team. More detailed information on my contributions to each work is provided at the beginning of each section in the thesis.}

\begin{enumerate}[itemsep=2pt]
    \item \textbf{Review} of recent progress understanding the factors influencing real-world generalization of deep networks, contextualizing the works presented in this thesis and providing a foundation for future work in this fast-growing area. (CH. \ref{intro_gen}, \ref{intro_und})
   
    \item \textbf{Empirical exploration} of memorization and generalization with recurrent neural networks, generating insight on similarities and differences from feed-forward networks (and thereby the effects of depth on memorization). Theory works tend to focus on atemporal data, but many data of real-world interest (e.g. many data relating to climate, social science, and health) have important temporal variation. The `empirical theory' approach of this paper is generally useful for generating the kinds of insights and intuition that are helpful for designing systems that generalize in the real world. Understanding when neural networks will memorize (vs. learn underlying patterns of) training examples also helps us better design new methods and training regimes.  (CH. \ref{sec:memgen})
    
    \item Development and empirical evaluation of a novel regularization method for recurrent neural networks, \textbf{Zoneout}, with insight into the role of regularization for gradient propagation. Dropout became so quickly and widely adopted because of how well it encourages generalization, even in relatively low data regimes, common with real-world data. Zoneout allows the same benefits for temporal data. (CH. \ref{sec:zoneout})
    
    \item Proposal and comprehensive evaluation of a \textbf{fill-in-the-blank} dataset for benchmarking video models, with baseline results and challenge evaluation platform. Demonstration that the simple predictive task of filling-in-the-blank on captions (similar to and preceding masked language modelling) improves representations of video data and is better aligned to human evaluations than commonly-used word-overlap measures such as BLEU. Designing better-aligned tasks and methods of evaluation are important for ensuring real-world generalization. (CH. \ref{sec:moviefib})
    
    \item Co-definition of the term \textbf{auto-induced distributional shift (ADS)}, and exposition of the role it can play in content recommendation. The assumption that data are sampled iid (independently and identically distributed) is often made in machine learning theory, but it rarely holds for real-world data. However, the practical consequences of this assumption failing to hold are seldom investigated. We show the impact this can have in content recommendation, which is increasingly prevalent in peoples' daily life via search, social media, etc. Investigating the practical consequences of assumptions is an important aspect of ensuring good generalization in the real world.
    (CH. \ref{sec:dist})
    
    \item Proposal of the approach of \textbf{unit-testing} for recognizing behaviours that are difficult to specify in advance of training, but easy to recognize in a trained model. Contribution to developing unit-tests for models that might undesirably or unexpectedly achieve performance via ADS. Our unit test provides a practical way to assess when algorithms might follow an incentive to achieve performance via ADS - i.e., when they might generalize in an undesirable way in the real world. (CH. \ref{sec:dist})
    
    \item Proposal of the approach of \textbf{sandboxing} for training, evaluating, and comparing algorithms prior to real-world deployment by using Agent-Based Models (ABM) simulators; instantiation and testing of this approach in the development of a detailed ABM for simulation of Covid-19 epidemiology (COVI-AgentSim). The Covid-19 pandemic has had a huge impact on daily life for most people. New technologies like digital contact tracing could help lessen this impact and protect people in this and future pandemics. But like any intervention relating to health and welfare, it is essential that they be thoroughly tested. In general the approach of sandboxing with an ABM is helpful for evaluating the generalization capabilities (among other aspects) of a model prior to deployment in the real world, and specifically COVI-AgentSim helps us develop better DCT methods for Covid-19. (CH. \ref{sec:dist})
    
    \item Co-development of the method of \textbf{predictive contact tracing}, including formalizing the problem as distributed inference of infectiousness, and use of COVI-AgentSim as an evaluation sandbox, allowing us to compare models as well as to recognize and address issues of ADS in contact tracing.  This provides a good example of the kinds of important phenomena we can discover and address via sandboxing - we often do not know in advance how to speficy which kinds of generalization are desirable, but sandboxing can help us target desirable properties in practice. Distributed inference performed as we did is also generally useful for large-scale privacy-preserving use of representation learning for individual-level risk estimation, which could be important for many real-world tasks such as quantifying negative externalities from pollution, climate change, etc. (CH. \ref{sec:covi})
\end{enumerate}

\section*{B. Notes on scope (what this thesis is not)} \label{sec:Limitations}
\addcontentsline{toc}{section}{B. Notes on scope (what this thesis is not)}

There are many aspects of generalizing in the real world that are not addressed by this thesis. I focus on the techniques that have dominated real-world AI applications in recent years, and the dominant paradigms of theory for understanding A: i.e. deep neural networks used in mostly-supervised settings, described in the language of probability and statistical machine learning. This thesis does not cover, for example, tree-based methods, bayesian ML, causal ML, information theory, complex systems theory, evolutionary methods, or reinforcement learning. Nor does it cover much of the extensive data science work required to build datasets suitable for large-scale learning (although this is touched on in Chapter 4).

Perhaps most notably though, I focus here on the technical side of addressing what are fundamentally socio-technical systems.  A solely technical approach to real-world generalization (and by extension, responsible AI development) is doomed to failure. Artificial intelligence systems are human systems, built by humans and used by, on, and for humans. 
Experience and information about the world are not limited to or necessarily easily cast as sets of data examples; untidy data and other ways of knowing about the world are sorely under-represented in ML, AI, deep learning, and data science. 

However, in order to make concrete progress, I believe it's necessary and useful to engage with the dominant paradigms of the field, and this thesis reports my work in doing that. While the last chapters touch on some aspects of systems modeling, alignment, and human-computer interaction, these are hugely important areas I believe will see an explosion of interest in years to come. I hope that this thesis comes to fit in a larger context of multi-disciplinary, multi-stakeholder efforts towards more informed, responsible, and beneficial use of AI.

\section*{C. Notes on notation}
\addcontentsline{toc}{section}{C. Notes on notation}

 Because deep networks very typically operate on vectors of data, and not scalar-valued data, for consistency I use vector notation throughout, without loss of generality to the scalar case as scalars can be considered as 1-D vectors. Boldface italics are used to indicate a vector-valued function, e.g. \textbf{\textit{f}}. Capital letters in boldface are used for matrices, e.g. \textbf{W}, lower-case letters in boldface for vectors, e.g. \textbf{x}, boldface with an index for a particular element of the vector, e.g. $\textbf{x}_1$, and italic non-boldface to indicate any-and-every element of the vector, e.g. $\textit{y}$. This vector notation is prioritized over the conventions of probability, where random variables are typically denoted with capital (non-boldface) characters and values of random variables indicated with lower-case (non-boldface) type.  A boldface \textbf{1} or other number is used to indicate a vector where each element is that number. To avoid excessive boldface, function names with more than one letter are not bolded, even if they return a vector. For example, $\textit{exp}^\textbf{x}$ indicates a vector where each element is the number $e$ to the power of the corresponding element of \textbf{x}.

\section*{D. Outline of content in this thesis}
\addcontentsline{toc}{section}{D. Outline of content in this thesis}

\textit{For a full summary, see SUMMARY in the previous section. }

The introduction first presents a background to AI and ML, with a focus on generalization, including a review of recent works. Each chapter then first gives the full citation for article(s) covered in that chapter, and some contextual notes, before reproducing or summarizing the article(s). Each chapter is self-contained, with introduction and literature review and any appendices included as they were published. 

I begin with work empirically examining generalization behaviour of deep networks in \textbf{Chapter 2}, where we observe that generalization and learning behaviour depend on the data distribution, and regularization contributes significantly to generalization at the expense of memorization \cite{arpit2017,memRNN}.  

\textbf{Chapter 3} then integrates my co-first author work on an explicit regularizer for recurrent neural networks which confers the benefits of dropout while improving gradient flow to early timesteps \cite{zoneout}. 

\textbf{Chapter 4} integrates my first-author work examining evaluation of video models, and proposing the simple predictive task of `fill-in-the-blank' (in language modelling, now known as masking) as a means of improving learned representations \cite{moviefib}.

\textbf{Chapter 5} covers 2 works addressing auto-induced distributional shift (ADS) (i.e. shift caused by AI systems) in content recommendation \cite{hiads} and Covid-19 epidemiology \cite{gupta2020covisim} respectively. These works also serve to introduce two practical techniques for ensuring better real-world generalization: unit-testing and sandboxing. 

Finally, \textbf{Chapter 6} integrates the co-first author paper \cite{gupta2020covisim}  in which we cast the problem of digital contact tracing (DCT) as distributed inference of infectiousness, and leverage our ABM sandbox to train predictors offline. We show that this strategy, leveraging the weak signals available on smartphones, enables proactive recommendations that effectively prevent spread of disease while requiring very little quarantine of healthy individuals.

I conclude by identifying some key takeaways and approaches from my work for real-world generalization, and discuss next steps.


\Chapter{INTRODUCTION}\label{sec:Introduction}   

Machine learning (ML) and artificial intelligence (AI) are increasingly deployed in real-world settings, affecting many aspects of our daily lives:  text correction, prediction, and suggestion on smartphones; search, newsfeed, bots, game AI, and content recommendation algorithms; route, traffic, and delivery planning; and increasingly, science and medical applications such as drug discovery, disease diagnosis, personalized health recommendations, physics simulations, and weather prediction. AI has been hailed as the second industrial revolution \cite{revolution}, likely to fundamentally change the way society works and the types of things most people do every day. 

As is typical of useful and profitable technology, our fundamental understanding of these systems has lagged behind their deployment. We've made a lot of progress developing understanding and addressing issues with problematic systems in the last few years, but for high-risk/high-impact settings, a post-hoc (do first, ask questions later) approach is not sufficient or appropriate. Our lack of fundamental understanding has been a barrier to applying AI systems in many areas of science, and has created real-world harms in areas where deployment has been less cautious (for example, discrimination against minority groups in hiring \cite{amazon}, recidivism prediction \cite{compas}). These are complex, nuanced, socio-technical settings, and we need new tools, language, and scientific approaches to address them. In order to use AI responsibly, we need to understand the principles by which AI systems work, their interaction with the people and environment they're deployed in, and their potential risks and benefits. These are the defining problems motivating my work. 

Regrettably, these problems are too large to be solved by a single dissertation.\footnote{The algorithm for making scientific progress via dissertations is colloquially known as `grad student descent'; my thesis presents the vector of progress I have made toward the objective of responsible AI.} 
My aim is to present some concrete steps in useful directions, focused on improving our understanding and evaluation of AI behaviour, and critically examining the assumptions we make in designing AI systems.

This introduction first reviews some basics from AI, ML, and deep learning in Sections \ref{sec:AI}, \ref{sec:ML}, \ref{sec:DL} and then provides a review of more recent developments in the field relevant for understanding real-world generalization of deep networks in Sections \ref{intro_gen}, \ref{intro_und}.




\section{Artificial Intelligence} \label{sec:AI} 

Artificial Intelligence (AI) systems are automated or computer-based systems which perform a task or tasks an average person would typically describe as requiring intelligence. The most prevalent approach to AI in the early days of the field - the 60s and 70s - was via rule-based systems of symbols and logic operations on those symbols. This is often called Symbolic AI, and sometimes GOFAI: Good Old-Fashioned AI (see \cite{gofai} for an introduction). This approach was initially fruitful for many applications (e.g.~ performing search-like operations on databases), and researchers famously estimated it should only take a few months until an AI would be able to recognize the species of a bird from a photograph \cite{dartmouth}. 

But it proved much more difficult than anticipated to use logic-based systems for this type of task: One where the relationship between the \textbf{input} format (e.g.~ the pixels that make up the photo of the bird) and the desired \textbf{output} from the AI system (e.g.~ an English-language word that is the species of that bird) is easy to demonstrate to a person via \textbf{examples} of input-output pairs, but hard for a person to write down as a set of rules. This is the type of task that machine learning systems (particularly deep learning) excel at.

\begin{figure}[hbt]
\center
\includegraphics[width=0.45\textwidth]{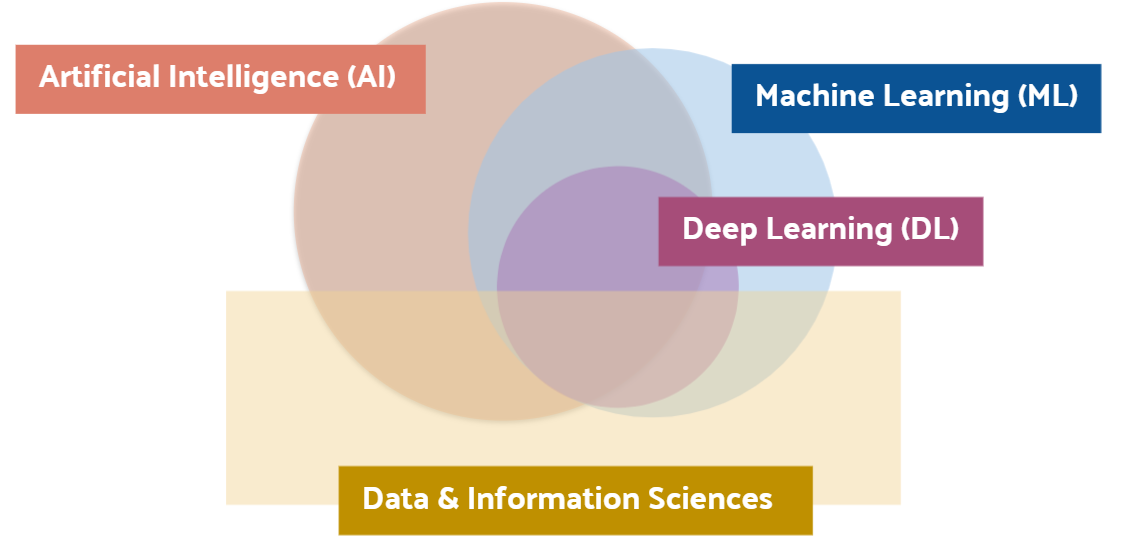} 
\caption{A conceptual diagram showing the relationship of some popularly-used terms / fields of research: Artificial Intelligence (AI) (Sec. \ref{sec:AI}), Machine Learning (ML) (Sec. \ref{sec:ML}), Deep Learning (DL) (Sec. \ref{sec:DL}), and Data \& Information Sciences. This thesis sits at the intersection of these four related fields.}
\label{fig:venn}
\end{figure}

\section{Machine Learning} \label{sec:ML}

Machine learning (ML) is the science and practice of getting a computer (machine) to arrive at an algorithm based on \textit{experience}, rather than predetermined rules. As mentioned in the preceding section, there are forms of AI which do not employ ML. There are also tasks we could do with ML which we would not consider as requiring intelligence. However, this thesis is largely concerned with the intersection (ML used for AI), which has seen so much growth and attention in recent years that these terms are often used interchangably- see Figure \ref{fig:venn}.

 A bit more formally, ML is often defined as a process/algorithm/model for performing a task $\mathcal{T}$ which adjusts its behaviour to improve at the task based on experience $\mathcal{E}$ and a measure of success at the task, or performance metric, $\mathcal{P}$ (see \cite{Bishop2006} for an introduction to ML).  Typically, the notion of experience is operationalized as a set of input-output pairs, called examples or samples - this is so common that sometimes $\mathcal{E}$ is taken to mean examples, rather than experience. 
The set of input-output pairs we use as examples are called a dataset, and typically denoted $\mathcal{D} =( $\textbf{x, y}), with a particular example indicated by a subscript e.g.~ ($\textbf{x}_1, \textbf{y}_1$), or indexed with $i$ out of $n$ total examples, and all examples with some label indicated $(\textbf{x}, y$).

We usually write $\hat{\textbf{y}}$ for the actual output (prediction) of a model $\textbf{\textit{f}}$, and compare $\hat{\textbf{y}}$ to ${\textbf{y}}$ in order to evaluate its performance.\footnote{Of course, not all tasks are easy to demonstrate via pairs like this: it might be difficult to choose a format for $\textbf{x}$, or there might be more than one `correct' answer $\textbf{y}$ for an input $x$. And it's not always easy to find a good way to compare  $\hat{\textbf{y}}$ to ${\textbf{y}}$. Some of these issues will be addressed by works in this thesis.} Sometimes we write $f(\textbf{x})$ instead of $\hat{\textbf{y}}$. 

The comparison between $\hat{\textbf{y}}$ and ${\textbf{y}}$ is performed by the loss function $\mathcal{L}$, which gives us the performance measure $\mathcal{P}$ (typically the loss function is synonymous with the performance measure, and the letter P stands for so many things that it can be confusing; $\mathcal{L}$ is much more commonly used). $\mathcal{L}$ measures some numerical distance between $\hat{\textbf{y}}$ and ${\textbf{y}}$, for example the squared difference. The expected value of a set of losses is called the \textbf{risk}, usually denoted $R$; for a discrete set of $n$ examples this is calculated as the expected loss. For losses which measure accuracy, the loss measures the average number of \textbf{errors} made, and the terms error, loss, and risk are used interchangeably. e.g.~ For the mean-squared differences loss we calculate \textbf{mean-squared error}:
\begin{align}
    \mathcal{L}_{MSE} := \frac{1}{n} \sum_{i=1}^n (\hat{\textbf{y}}_i-\textbf{y}_{i})^2 
\end{align}

Learning consists in adjusting the model's \textbf{parameters} (usually denoted $\theta$, or $\textbf{\textit{f}}_{\theta}$ if we want to make it clear that they belong to the model), according to the chosen loss. This is often described as learning by `trial and error'. \textbf{Hyperparameters} are parameters relating to the training or selection of the model, but not the parameters of the model itself.

\subsection{Typical assumptions}
In order to describe, evaluate, and predict properties of machine learning algorithms, some assumptions are typically made. The canonical assumptions made  are collectively referred to as the \textbf{i.i.d. assumptions} (sometimes written i.i.d. for short): that all data examples in the dataset were sampled independently and are identically distributed - i.e.~ come from the same underlying distribution. 

Another common, though less frequently formalized, assumption is that the dataset we have sampled is representative of the underlying data distribution we care about. I call this the \textbf{representativeness assumption}. This could also be described as an assumption that any bias introduced by the sampling procedure is random (not correlated with features of the data), and therefore in expectation over a large enough dataset can be ignored.

\subsection{Gradient-based Learning} \label{sec:gradientbasedlearning} 

The problem of making adjustments that we can be sure will eventually lead to a good model is classically viewed as an \textbf{optimization} problem, and many advances in ML have come from the field of optimization. In \textbf{gradient-based optimization}, a cornerstone of large-scale ML systems, parameters are adjusted based on the derivative of the loss function with respect to the parameters (the vector of partial derivatives of a function with respect to a vector, in this case the parameters, is called a gradient). 

When using gradient-based learning, it is typical to make only small adjustments in the direction of the gradient, a. The size of these is controlled by a hyperparameter called the \textbf{learning rate}. Because the adjustments at each iteration/epoch are small, it is typical to make multiple passes through the dataset (sometimes called multi-epoch training, although it is so much the norm it is usually not named).

\subsection{Typical Tasks in Machine Learning} \label{sec:mltasks} 

It is typical to express different tasks in Machine Learning in probabilistic terms; this can help clarify and compare different tasks.

In \textbf{supervised learning}, there is a an explicit `target' $y$ for each example $x$; if $y$ is categorical we call $y$ a label and the task \textbf{classification}. In probabilistic terms the task is to learn the conditional probability of a label $y$ given example $x$. Discriminative approaches are those which model this conditional probability directly; e.g.~ logistic regression, many simple neural networks. In general, \textbf{discriminative} approaches seek the class y that is most likely given the example $x$, i.e.~ $\text{argmax } P(\textbf{y}|\textbf{x})$.

In contrast, \textbf{generative} approaches to supervised learning model the \textbf{data generating process}: the joint probability $P(\textbf{x},\textbf{y})$. With this joint density, we can sample or, if tractable, use Bayes' rule to compute $P(\textbf{y}|\textbf{x})$: 
%
\begin{align}
P(\textbf{y}|\textbf{x})= \frac{P(\textbf{x},\textbf{y})}{P(\textbf{x})} = \frac{P(\textbf{x}|\textbf{y})P(\textbf{y})}{P\textbf{(x})}
\end{align} 

This can be read as ``the probability of a data example $x$ given that you choose class $y$, multiplied by the prior probability of class $y$, divided by the marginal probability of $x$ (often called the `evidence')''. $P(\textbf{x})$ is the same regardless of the class $y$, so in classification problems with a fixed dataset, we seek $\text{argmax }P(\textbf{x}|\textbf{y})P(\textbf{y})$.

In \textbf{unsupervised learning}, no explicit target is given; the task is to  model the density $P(\textbf{x})$:
\begin{align}
P(\textbf{x}) &= P(\textbf{x};\theta)
\intertext{or equivalently,}
%
\log P(\textbf{x}) &= \log P(\textbf{x};\theta)
\end{align} 

In order to capture or distill useful information out of what may be very high-dimensional $\textbf{x}$, we may introduce \textbf{latent} (unobserved) variables $\textbf{z}$, and a distributions over those latent variables, in order to find tractable proxies to $P(\textbf{x},\textbf{z})$, or factorize $P(\textbf{x})$ (e.g.~ as a product of conditionals) such that it becomes tractable. 

In \textbf{semi-supervised learning}, only some data examples are labelled, so typically a combination of generative and discriminative approaches are used. 

In \textbf{self-supervised learning}, automated methods are used to obtain labels that can then be predicted. A common example (a common task before the term self-supervised learning was applied to it) is next-step prediction for temporal data, where targets are simply obtained by shifting the input. Sometimes self-supervised is considered a form of unsupervised learning, and self-supervision could help in semi-supervised tasks. If there remains a primary objective that is supplemented with self-supervised tasks, these are called \textbf{auxilliary tasks}. 
Self-supervised learning has been important for many large-scale tasks such as language modelling.

\textbf{Transfer learning } takes a model learned on one distribution $p(x_1)$ and seeks to leverage this model for good performance on a different (possibly related) distribution $p(x_2)$. Pretraining followed by fine-tuning is a common example of transfer learning; see \ref{sec:fine}.

In \textbf{multi-task learning}, proposed by \cite{caruana} different tasks are learned in parallel using a shared representation for different tasks. Sometimes viewed as a type of multi-task learning, in \textbf{meta-learning} (sometimes written metalearning),  models explicitly optimize for novel test-set performance, by breaking the optimization procedure into two `levels' or `loops'. In this view, the inner loop learns to perform an `object level task' (e.g.~ accuracy on a supervised learning task), while the outer loop learns to optimize performance of the inner loop algorithm over multiple train-test sets. In other words, the meta-objective seeks the function for the inner loop that minimizes expected performance across tasks $\mathcal{T}$:
\begin{align}
    \min_\theta \sum_{\mathcal{T}} \mathcal{L}_\mathcal{T}(\textbf{\textit{f}}_\theta)
\end{align}

\textbf{Multimodality learning}\footnote{I prefer and encourage the term \textbf{multimodality} over multimodal, in order to avoid confusion with multimodal distributions (distributions which have more than one mode).} considers learning of data distributions from multiple different modalities.
 Intuitively, it's useful to think of a modality as a sensor type, or medium in which data is encoded; for example, audio, visual, and textual are all modalities. 
 Authors in \cite{multimodal_dl} identify 3 approaches to multimodality data: (1) fusion, where both (or all) modalities are encoded separately and then these representations are `fused' (usually by concatenation); (2) cross modal, where the representation in one modality is used as supplemental information in order to perform a task in the other modality; and (3) shared representation, where the modalities are viewed as components of a shared representation. 




 \textbf{Multiview learning} considers different perspectives on the same data example \textit{within} a modality (e.g.~ different camera angles on a scene).

In \textbf{online} or \textbf{continual} learning, models receive a continuous stream of experience (input) for a given task as opposed to a fixed-size dataset, and model parameters are updated correspondingly. In \textbf{lifelong} learning the task itself may change over time (sometimes the terms continual and lifelong learning are used interchangably). 

\subsection{Distributional Shift} \label{sec:distshift}

\textbf{Distributional shift} is any change in the data distribution over time, or between examples used to train the model as compared to the examples it will be tested with or see in practice. 

In supervised learning this is often called dataset shift: change in the joint distribution of $P(\mathbf{x},y)$ between the training and test sets \cite{datsetshift, candela}. Two ways we could end up with dataset shift are via 
\textbf{covariate shift}: changing $P(\mathbf{x})$, while the conditional probability of a label given an input does not change, or  %
    %
    %
%
\textbf{concept shift}: changing $P(y | \mathbf{x})$; i.e.~ the relationship between the label and input is what changes. 

In general, distributional shift violates one of the fundamental assumptions typically made in machine learning: that all data examples are sampled \textbf{independently and identically distributed}. These assumptions are collectively referred to as the i.i.d, i.i.d., or iid assumptions.

\subsection{Statistical Learning Theory} \label{sec:statisticallearningtheory}

Much of the theoretical understanding we have of ML systems has come from formalizing learning-as-optimization in the language of probability; this field is called statistical learning theory. The most common framework is called called \textbf{empirical risk minimization (ERM)} (for an intriduction to statistical learning theory, see \cite{understandML}). The \textbf{underlying data distribution} $p$, sometimes called the \textbf{true distribution} (distribution for the full possible `population' of examples) has an expected loss (risk) $\mathbb{E}(\mathcal{L})$ referred to as the \textbf{true risk} $R^{true}$. The true risk tells us on average, out of all possible models, how much does it `hurt' to use our particular model $\textbf{\textit{f}}$:
\begin{align}
\mathcal{E}_{true} := \mathbb{E}_p[  \mathcal{L}( \textbf{\textit{f}}(\textbf{x}), \textbf{y})] = \int \int p(\textbf{x},\textbf{y}) \mathcal{L}(\textbf{\textit{f}}(\textbf{x}),\textbf{y}) d\textbf{x}d\textbf{y}
\end{align}

But since we have only a sample from the underlying distribution, we can't calculate the true risk; we can only calculate the \textbf{empirical} risk $\hat{R}^{e}$, i.e.~ the expected loss on our dataset. If we assume examples are drawn \textbf{independently and identically distributed (iid)} from the underlying data distribution, i.e.~ that the dataset is representative, then the empirical risk $\hat{R}^{e}$$\hat{\mathbb{E}}[\mathcal{L}(\hat{\textbf{y}},\textbf{y})]$ should give us a good estimate of the true risk, i.e.~:
\begin{align}
R^{true} &\approx R^{empirical} \\
\mathbb{E}_p[\mathcal{L}(\textbf{\textit{f}}(\textbf{x}),\textbf{y})] &\approx\mathbb{E}_{data}[\mathcal{L}(\textbf{\textit{f}}(\textbf{x}),\textbf{y})]
\end{align}

Performing ERM is the process of choosing the function $\textbf{\textit{f}}^*$ (from a class of possible functions $\mathcal{F}$) that minimizes the empirical risk $\hat{\mathcal{E}}(\textbf{\textit{f}})$; 
\begin{align}
\textbf{\textit{f}}^* = \argmin_{\textbf{\textit{f}} \in F} \hat{\mathcal{E}}(\textbf{\textit{f}})
\end{align}

Early work in statistical learning theory (in general, and therefore also for neural networks) focused 
on the ability to fit a desired (``target'') function. The (in)ability to do this is called \textbf{approximation error} - how close the function a model class is capable of computing is to the target function. A related consideration is \textbf{estimation error} - how close the function we actually get is to the function that is theoretically possible for the model class. A measure of error which takes both into account is called \textbf{integrated error}. An important early result shows that neural networks with one hidden layer are universal approximators \cite{approximator}, meaning that they are at least theoretically capable of representing any function. But how likely were they to be able to capture this function in practice? A typical approach to answering this question is to find mathematical formulas, depending on various characteristics of the algorithm (most simply, the number of parameters), which give upper and lower bounds on the approximation and estimation errors. Progress in understanding learning theory could be made by 'tightening' the bounds (estimating narrower and more accurate ranges), and improvement in an algorithm could be measured by calculating how close its performance came in practice to the theoretical bound (e.g.~ by minimizing estimation error).  

An important characteristic in learning theory (and computer science generally) is \textbf{complexity}. There are many measures of complexity, and even having  a definition they are not always straightforward to compute. For example, the Kolmogorov complexity, also called \textbf{algorithmic complexity} (in a particular language) of something is the length of the shortest computer program that outputs that thing. ``Shortest'' can be measured in time or other resources required for computation (e.g.~ CPU cores), the number of characters or bits required to write down the program, etc., and it may be non-trivial to prove that a certain algorithm is definitely the ``shortest''. A measure of complexity which is relatively more straightforward to calculate and is important in statistical learning theory is the Vapnik–Chervonenkis (VC) complexity

\textbf{VC dimension} is the size of the largest set of examples that can be 'shattered', which means classified perfectly. As described in \cite{Vapnik} derive a generalization bound based on this measure which is widely used for simple machine learning classifiers, but vacuous (not at all tight, i.e.~ not useful for predicting generalization error) for typical neural network. The bound is only nonvacuous for models with many fewer parameters than the (input) dimensionality of data points they are trained on, whereas typically neural networks have orders of magnitude more parameters than input dimensionality.


\subsection{Generalization Basics} \label{sec:generalization_basics} 

The key thing that makes bird-picture labelling and similar tasks difficult for rule-based systems is that we care about generalization. \textbf{Generalization} is the ability of a system to perform well on examples that are similar (but not identical) to those already seen by the system.

Generalization performance is estimated via performance on a set sampled iid from the same underlying data distribution as the data used to learn $\textbf{\textit{f}}$. In order to measure generalization quantitatively, we typically separate our dataset $\mathcal{D}$ into three subsets (usually randomly, in order to preserve the assumption that all are sampled iid from the underlying data distribution): 
\begin{itemize}
    \item The \textbf{training set} $\mathcal{D}^{(train)}$ used to adjust parameters of the model,
    \item The \textbf{test set} $\mathcal{D}^{(test)}$, never used during model conception or training, only at the end, so as to provide an unbiased\footnote{Of course, only unbiased if the iid assumptions hold} evaluation of the model,
   \item The \textbf{validation set} $\mathcal{D}^{(val)}$ used as an estimation of test-set performance over the course of training, and also used to tune hyperparameters.  
\end{itemize}

\textbf{Generalization error} is the empirical risk on the test set, i.e.~:
\begin{align}
     = \mathbb{E} [\mathcal{L}(\textbf{\textit{f}}(\textbf{x}^{(test)}), \textbf{y}^{(test)} ) ]
\end{align}

and the \textbf{generalization gap} is the difference between training performance and test performance :
\begin{align}
    = \mathbb{E}[\mathcal{L}(\textbf{\textit{f}}(\textbf{x}^{(train)}), \textbf{y}^{(train)} ) ] - 
    \mathbb{E} [\mathcal{L}(\textbf{\textit{f}}(\textbf{x}^{(test)}), \textbf{y}^{(test)} ) ]
\end{align}

\subsubsection{Capacity}
Fitting a model to data is the process of adjusting parameters to match observed data. 
If a model performs well on examples it has been trained on, but fails to predict good values for examples it hasn't seen before, we say it generalizes poorly; specifically, the model has \textbf{overfit} the training data. If the model performs poorly on both training and test examples, we describe it as \textbf{underfit}.\footnote{if the model performs poorly on the training data but well on the test data, we could still describe it as underfit, but more salient is that there is probably some unaccounted-for distributional shift between training and test data.} 

Two primary reasons for underfitting are (1) lack of model \textbf{capacity}, the intrinsic or theoretical ability of the model to accurately capture the data distribution, and (2) failure of optimization, the practical reality of whether the training procedure was able to find the correct solution. 
Summarizing these two, we can examine the \textbf{effective capacity} of a model: the capacity in practice given the realities of limited training time, compute, etc.   In classic machine learning theory, capacity is measured by counting parameters.

A key differentiation made in many modern works is between \textbf{optimization failure}, i.e.~ failure of the learning algorithm in training the network, vs. \textbf{generalization failure}, i.e.~ failure of the patterns learned during training to generalize to the test set. These terms are often taken as being equivalent to underfitting and overfitting repectively, but where underfitting and overfitting are phenomenological descriptions based on how models perform on the bias-variace tradeoff, optimization and generalization failures are mechanistic descriptions of what went wrong, such that we observe phenomena like over- or under-fitting. For example, generalization failure can happen \textit{without} overfitting/memorization -  if the patterns useful for fitting the training set are simply less useful for the test set,  because of random bias introduced by small dataset size, or because the data are not truly i.i.d. - i.e.~ a failure to select the model (among equally-fit-on-training models) that generalizes well. The term optimization failure is usually used in this context to separate this kind of failure to generalize from failures of generalization that due to underfitting. 

\textbf{Regularization} is anything that benefits or encourages generalization; it can be \textbf{explicit} (e.g.~ a term in the loss function) or \textbf{implict} (e.g.~ provided by the structure of the network).  

Two common examples of explicit regularization are L1 and L2 (aka weight decay), which penalize the L1 and L2 norm of the weights. This prevents the parameter values from growing too large too quickly. This prevents instability, as large weights will result in large gradients, in turn leading to large 'swings' in performance.

A common example of implicit regularization is \textbf{early stopping}, wherein we stop training a model if validation performance ceases to improve. This is contrast to most earlier work, which focused on convergence, i.e.~ training to 0 error. Without being at convergence, we lack a clean theoretical description and guarantees about the function computed by the network. But empirically, early stopping greatly improves generalization performance. Part of the reason for this can be understood as equivalent to L2 normalization - stopping training early means that weights don't have a chance to grow large. 


Many regularization methods can be understood as adding different noise distributions in different ways to the learning process. An important work examining this theoretically and empirically is that of denoising autoencoders \cite{DAE}. This work shows that by adding noise to the input, a representation of $\textbf{x}$ can be learned in a unsupervised fashion by making the target the reconstruction of $\textbf{x}$ from the noised version. 

Notably, the simple technique of randomly setting some proportion of activations to 0, called \textbf{dropout} \cite{srivastava2014dropout}, was incredibly helpful for good generalization performance. Many explanations of the benefits of dropout have been proposed; the inital rationale was that dropping out activations prevented co-adaptation of features, and co-adaptations might generalize poorly. \cite{wager_adapt} show an approximate equivalence of the implicit regularizartion of dropout to a data-dependent form of explicit L2 regularization.  

Another explanation is provided by the observation that the dropping-out creates ad-hoc smaller networks; a network of $n$ neurons will have $2^n$ possible subnetworks, increasing the effective capacity of the model.  An elaboration of this explanation offered by \cite{bachman} is that this is equivalent to training a kind of ensemble of smaller networks. That `kind' is one they define as a \textbf{pseudoensemble}: whereas the members of a true ensemble are not related to one another, the models in a pseudoensemble are created by perturbing a model according to a noise process (in the case of dropout, the random zeroing of activations).

\subsection{Curse of Dimensionality}

It would be much easier to write or infer rules for labelling an image from pixels if the pixel values only differed between images in ways that are easy to know in advance or predict. But if the input format is pixels, rules have to refer to pixel values, directly or indirectly, and because there are very many pixels that can vary in vary many ways, it's hard to write rules that will generalize to all these variations. This problem is often described as the \textbf{underlying data distribution} of images being very \textbf{high-dimensional}. 
 In \cite{curse}, authors showed that as dimensionality increases, covering this kind of distribution should require exponentially more data examples: the so-called \textbf{curse of dimensionality}\footnote{This term is also sometimes used to refer more generally to any issues that arise in high dimensions that do not arise or are hard to predict from what happens in low dimensions. Low dimensions are typically considered to be 1\-, 2\-, or 3-dimensional data, high dimensions anything greater; typically much greater)}. 

Another issue is that these dimensions depend on one another in \textbf{non-linear} ways. Linear means you can map the relationship between two variables on a straight line in a plot; a change in one variable means a straightforwardly predictable change in the other; non-linear means anything else (e.g.~ maybe the relationship is roughly linear for certain values, but levels off or even changes direction after a certain range, like the relationship between amount of sunlight received and the size of a plant).

\section{Deep Learning} \label{sec:DL} 

\subsection{Neural Networks} \label{sec:NNs} 

A \textbf{neural network} is a type of non-linear model inspired by computational neuroscience models of neurons in animal brains. A neural net is organized in layers of neurons, with the first layer performing a computation on the input, and successive layers taking the output of the previous layer as input. The network is $\textbf{f}$, and the last layer's output is the prediction $\hat{y}$, also written $\textbf{\textit{f}}(x)$. Each neuron has a parameter, called the \textbf{weight} because it determines how much the output of that neuron will contribute to the overall output for the layer (how much `weight' it has). The set of weights of the connections between nodes are the matrix $\textbf{W}$. The computation performed at each neuron is to take its input, multiply it by the current parameter value of that neuron, sometimes add a \textbf{bias} term usually denoted $\textbf{b}$, and finally pass that value through a \textbf{non-linearity}, also called an \textbf{activation function}, to obtain the output value for that neuron. Confusingly, some descriptions of neural networks use $\textbf{\textit{f}}$ to denote the activation function, but it is more common to use $\textbf{\textit{g}}$, in order to avoid this confusion. The weights and biases are the parameters of the network; we sometimes use $\theta$ to refer to them collectively. Since each neuron performs the same type of calculation, it is easy to express the operations of the neural network with matrix operations, rather than writing out what happens for each individual neuron. Written this way, the function a single-layer neural network computes is:
\begin{align}
    \textbf{\textit{f}}(\textbf{x}) = \textbf{\textit{g}}(\textbf{W}^\top \textbf{x} + \textbf{b})
\end{align}

For multilayer neural networks, the function each layer computes is a \textit{composition} of the previous layers' compuations, i.e.~ for a 3 layer network with numbers denoting the layer, we would have:
\begin{align}
   \textbf{\textit{f}}(\textbf{x}) = \textbf{\textit{f}}^{(3)}(\textbf{\textit{f}}^{(2)}(\textbf{\textit{f}}^{(1)}(\textbf{x})))
\end{align}

The number of function compositions, i.e.~ the number of layers, is typically called the depth of the network; this is where the term \textbf{deep network} comes from. We often call the first layer the \textbf{input layer}, last layer the \textbf{output layer}, and any middle layers \textbf{hidden layers}. In symbols, we often use $\textbf{\textit{h}}_0$ for the input layer, $\textbf{\textit{h}}_i, i=[1,2,\dots,n]$ for the $n$ hidden layers, and we call the output of any given layer the \textbf{hidden state} of that layer.
\begin{align}
   \textbf{ \textit{f}}(\textbf{x}) = \textbf{\textit{g}}(\textbf{W}^\top \textbf{x} + \textbf{b})
\end{align}

\begin{figure}[hbt]
\center
\includegraphics[width=0.45\textwidth]{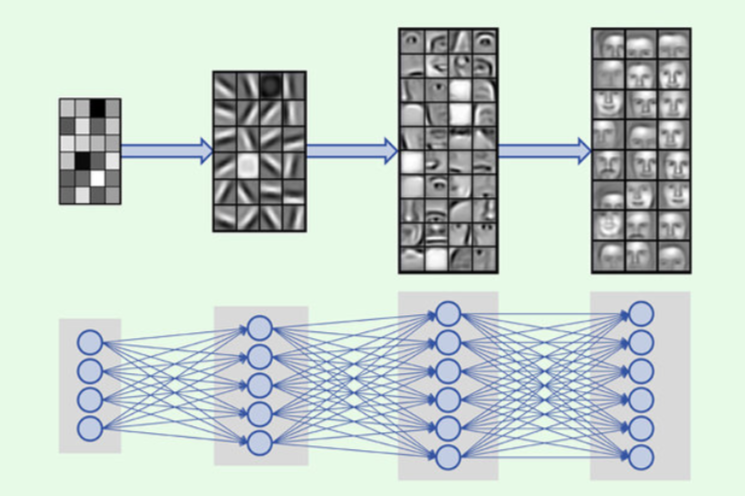}
\caption{A diagram showing (\textbf{bottom}) The layers of a neural network, and (\textbf{top}) a visualization of the features learned by a deep neural network, illustrating that the features at one layer are composed of those from the previous layer, i.e.~ the representation of a given input image hierarchical. Figure reproduced from \cite{deep09}.} 
\label{fig:qual_example}
\end{figure}

Early neural networks, e.g.~ \cite{lotteryticket} typically had around 5-10 of neurons per layer, and around 2-3 layers. In the prevailing wisdom at the time, stemming from Occam's Razor that a model should be as simple/minimal as possible, this was already a suspiciously large number of parameters; it was also already a challenge to find computationally practical training algorithms for could adjusting these parameters.
A first important step toward our current neural net ecosystem was the recognition that the \textbf{backpropagation algorithm} \cite{backprop}, which uses the chain-rule to calculate derivatives for compositional functions, provided a way to use gradient-based learning for neural networks  \cite{deepbelief}. Software frameworks like Theano \cite{Theano}, in which the architecture of the network is set up as a computational graph such that derivatives can be automatically computed, were a vital innovation in practical feasibility of training neural networks.

\subsection{Representation Learning}
When a person looks at a photograph of a bird, they use much more information than the pixels of the photograph to identify the bird (indeed, they may ignore the vast majority of the pixels) -- their previous knowledge about birds, other birds they've seen, information about the location the picture was taken in, their ability to recognize and mentally separate bird from background, etc. All of this information constitutes a \textbf{representation} or \textbf{internal representation} of the image that we form unconsciously, and it is this representation which makes it easy for humans to output the correct species for birds they are familiar with. 

The widespread application of AI in recent years has largely been driven by advances in \textbf{deep representation learning}, wherein \textbf{features} (important characteristics of data, often represented as the columns in a table of data) are learned by adjusting the parameters of a deep (many-layered) neural network using gradient-based learning. This is in contrast to previous approaches where features were hand-engineered. Because of the layered structure of deep nets, the representation learned is hierarchically structured, and therefore \textit{compositional} -- learning patterns that generalize well to one example help you learn something about other examples. e.g.~ learning a good representation of an image of a bird involves learning how colours and contrast compose edges, edges compose shapes, shapes compose structures like wings or beaks or tailfeathers, and these structures compose into different species of bird.

This is what allows deep representations to `defeat the curse of dimensionality', as humans and other creatures are able to: even though the input data distribution may be very high-dimensional, the representation transforms the input to much lower-dimensional manifold that is easily classified.

This is often called just deep learning or representation learning, depending on which aspect is being emphasized. To be clear, I note learned representations of a dataset need not be deep, and deep learning need not use learned representations, but in practice these are nearly synonymous.

The model used to learn deep representations are prototypically deep neural networks, and the learning algorithm employed in deep learning is prototypically a variant of \textbf{stochastic gradient descent (SGD)}. In SGD, noise is introduced to the optimization procedure by calculating the gradient not from the full `batch' of data (i.e.~ the full dataset), but smaller subsets at a time, called \textbf{minibatches}. One minibatch's training is called an \textbf{iteration}, going through the full dataset via minibatches is called an \textbf{epoch} of training. Originally minibatching was done simply for computational efficiency in working with increasingly large datasets, but it was quickly observed that the noise actually improved generalization performance, by providing implicit generalization. In contrast to minibatching, updating parameters after a full pass through data (i.e.~ using a minibatch size equal to the size of the network) is often called \textbf{full-batch gradient descent}.

\subsection{Some common layer types in deep nets}

The definition of a neural network above is quite general; it does not specify exactly what the functions $\textbf{\textit{f}}_i$  for the $i \in n$ layers should be, or how each layer is structured. They need not be structured the same way, and interactions of different functions can encode useful priors and/or make fitting the function more tractable. I cover here a few common types of neural network layers, but many others are possible.

\subsubsection{Feed-forward layers}
Also called a \textbf{fully connected} layer, this is a `basic' layer whose parameters are organized as a vector, in the most straighforward interpretation of the general formula above: $\textbf{\textit{f}}(\textbf{x}) = \textbf{\textit{g}}(\textbf{W}^\top \textbf{x} + \textbf{b})$. This section uses $\textbf{z}$ to denote the intermediate calculation of $\textbf{W}^\top \textbf{x} + \textbf{b}$ which the activation function acts on.\footnote{Not to be confused with the widespread use of $\textbf{z}$ for latent variables in probabilistic graphical models and deep learning models inspired by them, e.g.~ Variational Auto-Encoders (VAEs)} 
A network made only of feed-forward layers is sometimes called an \textbf{multilayer perceptron (MLP)} (regardless of the activation function used), though technically a \textbf{perceptron} is a layer which uses a \textbf{step function}, i.e.~ a threshold, as an activation function. This was based on inspiration from early models of neurons, which `activate' only above a certain threshold. In early days of neural networks, the most common activation function for feed-forward layers were softened versions of a step function; either a \textbf{sigmoid}, which returns a value between $0$ and $1$: 
\begin{align}
\textbf{\textit{g}}(\textbf{z}) = \bm\sigma(\textbf{z}) = \frac{\textbf{1}} {\textbf{1} + \textbf{\textit{e}}^{-\textbf{z}}}
\end{align}

or \textbf{hyperbolic tangent (\textbf{tanh})}, which returns a value between $-1$ and $1$:
\begin{align}
    \textbf{\textit{g}}(\textbf{z}) = \textit{tanh}(\textbf{z}) = \frac{\textbf{\textit{e}}^z - \textbf{\textit{e}}^{-\textbf{z}}}{\textbf{\textit{e}}^\textbf{z} + \textbf{\textit{e}}^{-\textbf{z}}} = \frac{\textbf{1} - \textbf{\textit{e}}^{-2\textbf{z}}}{\textbf{1} + \textbf{\textit{e}}^{-2\textbf{z}}}
\end{align}

The most common modern activation function is a \textbf{rectified linear} function, called \textbf{relu} for short, which outputs either 0 or \textbf{z}; whichever is greater:
\begin{align}
\textbf{\textit{g}}(\textbf{z}) = \textit{relu}(\textbf{z}) = max(0, \textbf{z})
\end{align}

This has a number of practical benefits, including computational efficiency, some of which are discussed  in Section \ref{intro_gen}.

If the activation function is linear, i.e.~ $\textbf{\textit{g}}(\textbf{z}) = \textbf{z}$, the layer performs an \textbf{affine transformation} of the input, sometimes called \textbf{embedding} the input (in a learned vector space). This this is a common first-layer structure.

A \textbf{softmax} is one of the most commonly used activation functions for the last layer (output layer) of a neural network, typically used for classification tasks. The softmax transforms the typically-continuous-valued hidden states to a probability distribution over categories, by using a number of units equal to the number of categories $K$ and normalizing across those inputs:
\begin{align}
    \bm\sigma(\textbf{z}_i) = \textit{softmax}(\textbf{z}_i) = \frac{\textbf{\textit{e}}^{\textbf{z}_{i}}}{\sum_{j=1}^K \textbf{\textit{e}}^{\textbf{z}_{j}}} \quad \text{for $i=1,2,\dots,K$}
\end{align}

This output vector can then be compared to the target vector, typically encoded as what's called a \textbf{one-hot vector}, i.e.~ a vector of zeros as long as the number of categories, with a 1 at the index of the correct label. 

\subsubsection{Convolutional neural networks}

Called \textbf{convnets} for short, these networks were the first to achieve widespread success on image-related tasks, first for recognizing handwritten digits for mail routing \cite{lecun}, and later when Alexnet, a 5-layer neural network trained with \cite{alexnet} surpassed all hand-engineered methods on the benchmark dataset Imagenet \cite{deng2009imagenet}. Convolution is an operation that performs an integral on two functions of a real-valued argument that can be thought of as a weighted average in a sliding window; this weighting function is typically called a \textbf{kernel} or \textbf{filter}. The weights  can be viewed as a filter because they `select' (make prominent) certain inputs (those that are weighted more strongly). Especially when the weights are learned, the output of the convolution is often called a \textbf{feature map}, because convolving the filter with the input is like passing a detector over the input for the feature encoded by the weights, so the output expresses the extent to which that feature is present at a given part of the input. The portion of the input that the filter `sees' is called the \textbf{receptive field}.

For example, for a real-valued argument of time $t$, a function $\textbf{\textit{x}}(t)$ providing location of an object at time $t$, and a weighting function $\textbf{\textit{w}}(a)$ that provides the sliding window, i.e.~ the weights at age $a$, the convolved output $\textbf{\textit{s}}(t)$ is given by:
\begin{align}
    \textbf{\textit{s}}(\textbf{t}) = \int \textbf{\textbf{\textit{x}}}(a) \textbf{\textit{w}}(t-a) da
\end{align}

To simplify, and to reflect the way convolution is implemented in matrix algebra, the operation of convolution is often denoted $*$, and written similarly to multiplication with $w$ with time as an argument:
\begin{align}
    \textbf{\textit{s}}(\textbf{t}) = (\textbf{x}*\textbf{w})(t)
\end{align}

For time-series data, the input function would give an input $\textbf{x}$ that might be an estimate of position from a sensor, and the weighting function would give $\textbf{w}$ which would be a weight vector over the last $a$ timesteps; $\textbf{\textit{s}}(t)$  would thereby give a smoothed estimate of the position averaged over the last $\textbf{a}$ timesteps according to the weights in $\textbf{w}$ (filtering e.g.~ for the most recent timesteps). For 2D data such as images, $\textbf{w}$ would usually be a 2D matrix \textbf{W}. The choice of \textbf{W} is not an accident here - an equivalence can be shown between convolution and feed-forward networks in terms of the function computed \cite{convmlp}.  The learned weights or output feature maps are often visualized in order to interpret (understand) the predictions of the network. 

Typically in convnets, a convolutional layer is followed by a \textbf{pooling} layer, which downsamples (lowers the resolution) of the incoming layer. Typically this is done by \textbf{maxpooling}, i.e.~ taking only the maximum activation from the previous layer as the output for the next layer, but other pooling schemes exist. 
\begin{align}
    \textbf{\textit{f}}(\textbf{x}) = pool(\textbf{x}*\textbf{W})
\end{align}

Taken together, the operations of convolution and pooling structurally encode a prior of translation invariance. A convolutional layer can be viewed as a fully-connected neural network with an infinitely strong prior on the weights forcing the weight for each neuron of a layer to be identical to their neighbours, but shifted in space, and with weights set to zero outside the receptive field: this enforces that each layer should only learn local interactions that are translation-equivariant. Similarly, pooling enforces an infinitely strong prior that each unit should be invariant to small translations. For a full examination of these properties with intuitive visualizations, see \cite{distill}. As noted by \cite{dlbook}, these priors will cause a model to underfit if they are a poor fit to the kind of patterns that exist in the data or the specified task; for instance if performing a task required synthesizing information from different corners of an image. Typically, convolutional nets interleave convolutional, pooling, and fully connected layers. Correctly representing the indices of these operations can be tricky in the mathematical notation we have used thus far, and it is common to use notation inspired from graphical models for clarity, e.g.~:

\hspace{4cm}\texttt{X -> EMB -> [CONV-> RELU -> POOL]*3 -> FC}

A number of python libraries for neural networks, such as pytorch and keras, use syntax similar to this notation, making it a straighforward way to communicate the structure of many networks.

\subsubsection{Recurrent neural networks}
A recurrent neural network enforces the prior of sequential input ordering, by processing the elements of input in temporal order and additionally taking the network's own hidden state from the previous time-step as input, where $\textbf{W}$ and $\textbf{U}$ are the weight matrices associated with the hidden state and input respectively, and \textbf{$b$} is a bias term as in feed-forward networks:
\begin{align}
    \textbf{\textit{h}}_{t} = \textbf{\textit{f}}(\textbf{W} \textbf{\textit{h}}_{t-1} +\textbf{ U} \textbf{x}_t + \textbf{b})
\end{align}

The advantage this gives is similar to the advantage of convolution, often called \textbf{parameter sharing}: reusing the same parameters for different elements of the input (in convnets, the filter that's passed over the input; in RNNs the network that is applied at each timestep). If we used a feed-forward network to process the same input, the patterns for each element of the input would need to be learned separately at each input location. Whereas for convolution the output is a sequence where each element is a function of a small neighbourhood in the input, in recurrence each element is a function of previous members of the output -- the ``internal calculations'' (hidden states) of the RNN influence future outputs. This property is sometimes called \textbf{statefulness}. It has the computational disadvantage that it can't be parallelized. Convolution and recurrence can also be combined; either by having stateful filters, or by having some layers of the recurrent network be convolutional. 

It is common to call something an RNN if any of the layers are recurrent; in fact most commonly when people use the term they are referring to a network with a feed-forward input layer (often an embedding layer), then the recurrent layer, then a feed-forward output layer. Note it is something of a misnomer to call the recurrent layer a layer, since it is in fact a potentially-infinitely-deep stack of layers (as deep as there are elements of input); for this reason it's often called a \textbf{cell}. This is often denoted with a self-directed arrow in graphical model notation as shown in Figure \ref{fig:my_label}

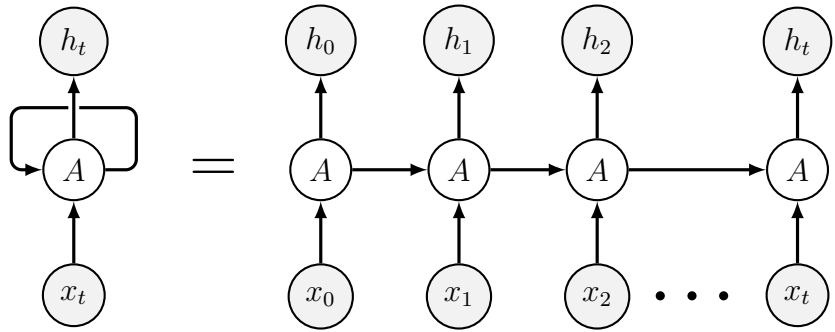
\begin{figure}
    \centering

\begin{tikzpicture}[item/.style={circle,draw,thick,align=center},
itemc/.style={item,on chain,join}]
 \begin{scope}[start chain=going right,nodes=itemc,every
 join/.style={-latex,very thick},local bounding box=chain]
 \path node (A0) {$A$} node (A1) {$A$} node (A2) {$A$} node[xshift=2em] (At)
 {$A$};
 \end{scope}
 \node[left=1em of chain,scale=2] (eq) {$=$};
 \node[left=2em of eq,item] (AL) {$A$};
 \path (AL.west) ++ (-1em,2em) coordinate (aux);
 \draw[very thick,-latex,rounded corners] (AL.east) -| ++ (1em,2em) -- (aux) 
 |- (AL.west);
 \foreach \X in {0,1,2,t} 
 {\draw[very thick,-latex] (A\X.north) -- ++ (0,2em)
 node[above,item,fill=gray!10] (h\X) {$h_\X$};
 \draw[very thick,latex-] (A\X.south) -- ++ (0,-2em)
 node[below,item,fill=gray!10] (x\X) {$x_\X$};}
 \draw[white,line width=0.8ex] (AL.north) -- ++ (0,1.9em);
 \draw[very thick,-latex] (AL.north) -- ++ (0,2em)
 node[above,item,fill=gray!10] {$h_t$};
 \draw[very thick,latex-] (AL.south) -- ++ (0,-2em)
 node[below,item,fill=gray!10] {$x_t$};
 \path (x2) -- (xt) node[midway,scale=2,font=\bfseries] {\dots};
\end{tikzpicture}

    \caption{Graphical model representation of a recurrent neural network, on the left showing the compact notation with a reflexive arrow on the recurrent cell $A$, the right showing the `unrolled' network. For a simple RNN (often called a \textbf{vanilla RNN} $A$ is computed as $\textbf{f}(\textbf{Wh}_{t−1} + \textbf{Ux}_t + \textbf{b})$ as described above.  }
    \label{fig:my_label}
    
\end{figure}
Recurrence is a general property; many structures for the recurrent cell $A$ are possible. An issue with using backpropagation in RNNs is that gradients can \textbf{vanish} (go to zero) or \textbf{explode} (go to infinity) as a result of repeated multiplications at each successive timestep \cite{bengio1994learning}. One of the first architectures successful in addressing this issue is a type of recurrent neural network was the Long Short-term Memory (LSTM) unit \cite{hochreiter1997long}, and a simplified version called a Gated Recurrent Unit (GRUs) \cite{GRU}; for an introduction and comparison of these architectures see \cite{olah}.

After processing an entire sequence with an RNN, the final output $\textbf{y}_t$ is often referred to as an \textbf{encoding} of the input. An \textbf{encoder} is a network which performs this encoding, denoted here as $h^{enc}$; a network which takes this encoding and produces an output sequence is called a \textbf{decoder}, denoted here $\textbf{h}^{dec}$:
\begin{align}
    \textbf{\textit{h}}^{enc}_t & = \textbf{\textit{g}}(\textbf{W}^{enc} \textbf{x}_t + \textbf{U}^{enc} \textbf{\textit{h}}^{enc}_{t-1} + \textbf{b}^{enc})\\
    \textbf{\textit{h}}^{dec}_t & = \textbf{\textit{g}}(\textbf{W}^{dec} \textbf{x}_t + \textbf{U}^{dec} \textbf{\textit{h}}^{dec}_{t-1} + \textbf{b}^{dec})
\end{align}

The encoder and decoder could use the same parameter matrices, described as \textbf{tied weights}. This encoder-decoder architecture was classically used for translation, where the parameter matrices would be different, each learning features of a different language.

\subsubsection{Attention}

Coming out of the literature on translation is an important innovation which softens the RNN's strict prior on sequential ordering: attention. As the name implies, attention determines which elements of the input are weighted (attended to) less or more; which could be (but is not limited to being) the most recent inputs most. With attention weights denoted $\bm\alpha$, $i$ indexing the decoder steps and $j$ indexing the encoder steps, attention is computed via a softmax over values returned by an attention-specific function $\textbf{\textit{a}}$, sometimes called the attention scoring function. The attention-specific function has two main variants; the originally proposed additive variant \cite{bahdanau} uses an additonal weight vector $\textbf{v}$, and a multiplicative dot-product version\cite{Luong} which does not:%
\begin{align}
    &\textbf{\textit{a}}^{add} = \textbf{v}^\top \textit{tanh}(\textbf{W}[\textbf{\textit{h}}^{dec}_{i-1}; \textbf{\textit{h}}^{enc}_j]) \\
&\textbf{\textit{a}}^{mul} = \textbf{\textit{h}}^{dec}_{i-1}{}^\top \textbf{W} \textbf{\textit{h}}^{enc}_j
\end{align}

The output $\textbf{h}_t$ for a given timestep labelled with $\textbf{y}_t$ is computed as: 
\begin{align}
    \textbf{z}_{ij}    & = \textbf{\textit{a}}(\textbf{h}^{dec}_{i-1}, \textbf{h}^{enc}_j)                      \\
    \bm\alpha_{ij}     & = \frac{\textit{exp}(\textbf{z}_{ij})}{\sum_{k=1}^{T_x} \textit{exp}(\textbf{z}_{ik})} \\
    \textbf{c}_i       & = \sum_{j=1}^{T_x} \bm\alpha_{ij}\textbf{h}^{enc}_j                                    \\
    \textbf{h}^{dec}_t & = \textit{tanh}(\textbf{W} [\textbf{h}^{dec}_{t-1};\textbf{y}_t;\textbf{c}_t])         \\
    \textbf{h}_t       & = \textit{softmax}(V \textbf{h}^{dec})
\end{align}

Both the additive and multiplicative versions, because they use a softmax, are termed \textbf{soft attention} (we get a distribution of weights $\alpha$ that tells the decoder how much to pay attention to various elements of the input). Using an argmax in place of the softmax, i.e.~ outputting a one-hot vector of attention weights, can give some gains in computational efficiency, but because the argmax is not differentiable, various methods for estimating or a gradient need to be use, such as the straight-through estimator \cite{straighthrough1}.

In \textbf{self-attention}, instead of separate encoder and decoder networks, attention is over previous states of the same RNN. Inspired by a content-addressable model of memory, attention can be viewed as a weighted sum of values for a query vector $\textbf{q}$ and key-value pairs $(\textbf{K,V})$ -- for example in the setting of video retrieval, $q$ might be the search terms, $K$ the tags/variables for videos such as location, category, etc., and $V$ the retrieved corresponding videos. In self-attention, the $\textbf{q}$ corresponds to the decoder's previous-step hidden state $\textbf{h}^{dec}_{i-1}$ and the key-value pairs are combined in the encoder hidden states $\textbf{h}^{enc}_{j=0 to T}$. Models which use self-attention are often called \textbf{autoregressive}, because the model is fit (regressed) based on its own state.

\subsubsection{Transformers}

Following the success of attention for translation tasks and in a series of architectures such as memory networks \cite{memnet} and neural Turing machines \cite{ntm}, a paper claiming ``Attention is all you need'' proposed a novel architecture for sequential data with no `hard-coded' sequential prior; relying on positional embeddings to learn any position-related information and composed of layers of two different types of attention: scaled dot-product attention:
\begin{align}
    \textit{Attention}(\textbf{\textbf{Q}},\textbf{ K}, \textbf{V}) = \textit{softmax}(\frac{\textbf{QK}^\textbf{T}}{\sqrt{d_k}})\textbf{V}
\end{align}
and multihead attention:
\begin{align}
    \textit{MultiHead}(\textbf{Q, K, V}) &= \textit{Concat}(\textit{head}_1, \textit{head}_2, \dots, \textit{head}_h)\textit{\textbf{W}}^{O}\\
 \textit{head}_i &= \textit{Attention}(\textbf{Q} \textbf{W}^{B_i}, \textbf{K} \textbf{W}^{\textbf{K}_i}, \textbf{V} \textbf{W}^{V_i})
\end{align}

Where $\textbf{W}^{O}$ indicates the output weights and $\textbf{W}^{Q}$, $\textbf{W}^{K}$, $\textbf{W}^{V}$ the weight matrices for the queries, keys, and values respectively, and $d_k$ is the dimensionality of the queries and keys.

These self-attention blocks (heads) can be easily parallelized, providing a significant computational benefit over typical recurrent architectures. As a result, transformers have become the backbone of many large-scale language models, for example BERT \cite{bert} and GPT-3 \cite{gpt3}.

\subsubsection{Pretraining}\label{sec:fine}

Software frameworks make it relatively simple to save and share the weights of a trained model. A popular way to leverage a trained model for a new but related task is by \textbf{fine-tuning} the network to that task: loading pretrained weights in to a model, and switching out the last layer(s) (sometimes also called a ``head'' in this context) to be appropriate for the new task. The resulting  network can either be trained \textbf{end-to-end}, meaning all weights, including the pre-trained ones, are adjusted during the new task, or the pretrained weights can be \textbf{frozen}, meaning they contribute to the forward pass but are not adjusted during the backward pass of training. This forces the new, task-specific information to be learned by the added layer(s). 

As increasingly large-scale models are trained on huge amounts of data with compute resources inaccessible to the average researcher, the approach of pretraining/fine-tuning has become increasingly popular - very large pretrained models which are intended to be multi-use are often referred to as \textbf{foundation models}. This has important implications for generalization theory, because traditional methods of study do not take into account the kind of structure and transformations introduced this way; generalization theory for transfer learning can help better understand performance of models which use pretrained layers.

\newpage
\section{Understanding Deep Net Generalization} \label{intro_gen}

\subsection{Background and terminology}

The\textbf{ No Free Lunch Theorem} states that assumptions of some kind need to be made in order to fit data, i.e.~ good performance must be `paid for' by making assumptions. This is often expanded to mean averaged over every possible data distribution, every model has the same performance -- i.e.~ the lunch of generalization on some distributions is not `free'; it's paid for by poorer performance on other distributions \cite{wolpert, understandML}. But we do not typically care about every possible data distribution; we care about those that are realistic in practice. In order to regularize the models we choose (or train) toward good generalization in the space of data distributions that are realistic, we often speak of encoding \textbf{good priors} in the model: structural biases for solutions that generalize well.


\textbf{Interpolation} means generalizing in between training examples. For example, if two training examples (with the same label) showed a light blue bird and a dark blue bird respectively, interpolative generalization would be correctly recognizing a medium-blue bird. This is often formalized as good performance within the convex hull (isoperimeter) of the support of the training data. \textbf{Extrapolation} means generalizing `outward' from training examples, i.e.~ outside the convex hull. For example, if training examples show that birds of many colours but consistent shape all have the same label, extrapolative generalization would be correctly ignoring colour. 
It has been argued that in high-dimensions, all generalization is effectively extrapolation \cite{lecun}, i.e.~ that there may not be a meaningful difference between these terms in practice. 


\textbf{Systematic generalization} is when generalization occurs according to a pattern learned by the system in a manner similar to following a rule, but in ways that can be composed (e.g.~ the classic example is numbers; if I tell you how to dax, you systematically generalize in order to know how to dax twice). It's worth noting that (1) the learning objective and dataset may not be set up to incentivise or test for systematic generalization; (2) there are many ways to generalize from training examples, and not all of them are necessarily `correct' - sometimes extrapolating a pattern which has worked for some examples will not work for others. See \cite{dima} for review of systematic generalization.


If what we care about is really systematic generalization, but we trained the model on a predictive task, this is an example of a \textbf{specification problem}: a mismatch between what we actually want and what we communicated to the model. This type of specification problem is called the \textbf{alignment problem}: the learning incentives or goals we set for the AI system are different from what we really want. But what we ``really want'' is typically hard to communicate or formalize. This leads to substantial problems when AI is used in the real world - including the reinforcement of systemic biases like racism, sexism, and ableism; risks of physical and psychological harms due to unexpected behaviour; and catastrophic and existential risks from powerful and/or weaponized AI systems. A commonality of these problems is that the approach of ``try first, ask questions later'' (i.e.~ finding problems by deploying a system and fixing problems after they occur) that has been common in much machine learning can be dangerous to the point of being unethical. 

The field of \textbf{alignment} seeks to get AI systems to understand and/or act in accordance with what we `really want' -- concretely, this might mean addressing specification problems, designing domain-specific losses, better leveraging expert feedback, encoding general or domain-specific priors, developing methods of testing AI systems prior to deployment, etc. Many failures of alignment can be viewed as failures to generalize `correctly'; this view can sometimes help identify avenues for improving alignment, by identifying the reasons for generalization failure. For a review of specification problems see \cite{specs, specificationprobs, vika}; for a list of concrete problems in AI safety see \cite{concrete_problems}, for informal introduction into the field of alignment see \cite{stuart}.


\textbf{Robustness} captures how performance degrades as a function of distance from the training data distribution (perfect robustness would mean no degradation over a specified distance). This is typically operationalized by adding different types of noise to input examples. If the noise distribution is specifically designed for a given model to have trouble performing well on it, this is called adversarial robustness\footnote{Due to historical precedence, robustness of deep networks considering other noise distributions is sometimes called non-adversarial robustness.}. In deep learning works, works studying robustness have gone hand in hand with methods for visualizing and understanding learning and generalization, in a field now often referred to as \textbf{interpretability}. 

Early visualization tools such as \cite{yosinski} helped demonstrate that features learned by neural networks were compositional, but they also demonstrated that features were often less interpretable than expected, particularly at higher (deeper) layers. Further work in this area identified adversarial perturbations, imperceptible to human eyes, which could reliably cause misclassifcation \cite{szegedy2013intriguing}. As the authors note, these results demonstrated that neural networks were not necessarily picking up on the patterns we would expect, posing a risk of failure in unexpected (and potentially unsafe) ways for real-world tasks. 

Similar techniques to interpretability, where the goal is to \textit{expose} information learned by a network, are also used for privacy, where the goal is to \textit{conceal} information learned by a network. \textbf{Differential privacy} studies how models can be trained to make useful predictions (e.g.~ about individual types or groups), without information of individuals able to be recovered from the model or its predictions.  \textbf{Federated learning} \cite{blaise}, wherein learning is decentralized across several networks (typically located on different servers), with limited communication between networks can also help with privacy. While this approach was primarily developed for computational efficiency in training many models, and may be used solely for this purpose, the setup lends itself to implementing privacy protections because of the control of flow of data. The federated networks also allow for training of more robust models, by regularizing one another (or a centralized model) similar to pseudo-ensembling \cite{pseudo_ensembles}.


The field of \textbf{out-of-distribution (OOD)} generalization studies how well (or not) a model generalizes to distributions other than those it was trained on. A related field of study is robust statistics, which seeks statistical measures that are not sensitive to the choice of underlying data distribution. 
A common approach in robust statistics is to assume that the test data distribution belongs to a (potentially infinite) family of distributions $\mathbb{P}^e$ where $e \in \mathcal{E}_{all}$ and $\mathcal{E}_{all}$ are often called environments. Under this assumption, one way to quantify OOD risk given by \cite{arjovsky}; the maximum risk across all environments:
\begin{align}
    R^{OOD}(\textbf{\textit{f}}) = max_{\mathcal{E}_{all}} R
\end{align}

 Relatedly, sometimes considered to be a type of OOD, \textbf{domain generalization} measures how well a given method performs on a set environments (different data distributions) from a domain (related set of environments). The domain is typically constructed/defined in an ad-hoc manner, e.g.~ by hand-chosen transformations of an input distribution, or by using 3 different image datasets. 
 Domain generalization often focuses on the impact of learning framework, i.e.~ alternatives to the empirical risk framework, such as invariant risk minimization\cite{invarisk}; see \cite{domgen} for review. In domain generalization, OOD risk might be quantified as the average, rather than the maximum, risk across environments. While in general (for potentially infinite environments) the average risk is intractable, for a smaller domain (often 2-5 environments) it is quite tractable to calculate.

\newpage

\section{Factors Influencing Real-World Generalization In Deep Networks} \label{intro_und}

Traditional measures of model capacity from statistical learning theory suggest that deep nets with their thousands of parameters ought to be massively over-parameterized and therefore generalize poorly. Influenced by Occam's Razor, and a desire for interpretable models, this perspective remained dominant for decades and led many to assume early work on neural networks and other connectionist approaches were ill-conceived. However, we now have abundant empirical evidence that these predictions of poor generalization are not borne out in practice, and new theoretical understanding is emerging to support the empirical findings. Here I review recent works (mostly from the last 5-10 years), organized under the following headings: 
\begin{enumerate}
    \item \textbf{Model structure and parameters}, including model architecture, especially depth; parameter initialization and regularization;
    \item \textbf{Optimization landscape}, including which minima (of the many possible) are incentivised by the loss, implications of explicit and implicit regularization on the landscape;
    \item \textbf{Data distribution}, including how underlying structure and patterns and (finite) sampling of data examples influence what can be learned and how models will thus generalize; 
    \item \textbf{Task definition and context}, including specification/task design, alignment, real-world deployment conditions, and interaction of the above items.
\end{enumerate}

 This ordering very approximately follows the recent history of learning and generalization theory, with many notable exceptions; also many works overlap  more than one of these categories (e.g.~ normalization could be considered as architecture, data, or interaction). Cross-cutting these categories is consideration of regularization: an important lens on understanding generalization is understanding regularization as these are two sides of the same coin. 
 
 This review does not attempt to be comprehensive (studies in this field number in the hundreds of thousands); rather to lay a groundwork of the main concepts, important results, and approaches to understanding generalization in deep learning, with a focus on those more relevant for real-world data. 
 



\subsection{Model Structure and Parameters}
Early results bounding approximation and estimation error for neural networks took  architecture into indirect account via the hypothesis class considered, as well as considering the data distribution as measured via the input dimensionality $d$. These works are summarized, and a new bound proposed, by \cite{barron1994}, who  found (at the time surprising) benefits of single-layer neural networks over traditional nonparametric models, when $d$ is large. Recent work by \cite{lin}, generalized by \cite{rolnick}, find congruent results for deep neural networks. They prove that the number of neurons required to fit natural distributions with $n$ variables is proportional to $exp(n)$ for single-layer networks, but proportional to $n$ for networks with $n$ layers - an exponential improvement in effective capacity. This exponential gain is also supported by earlier work by \cite{montufar}, who examine the symmetries and number of linear regions possible for the functions computed by deep networks. They show that for deep networks with piece-wise linear activation functions (e.g.~ rectifiers), the compositional structure allows patterns to be reused a number of times exponential in the depth of the network. They additionally show qualitatively that the `folding' structure created by deeply composing linear functions can relatively easily create complex patterns (reminiscent of a kaleidoscope).

These results about the importance of depth were supported empirically by the success of two major classes of deep network: recurrent neural networks (RNNs) and residual networks (ResNets). RNNs with gating for normalization, such as LSTMs, came to out-perform hand-engineered methods for predicting highly-structured time-series data, such as speech.\footnote{Recall that even single-layer RNNs can be viewed as an autoregressive feed-forward network as deep as the input dimension.} ResNets were the first ultra-deep (hundreds of layers) networks, which similarly addressed issues of vanishing gradients that had plagued earlier networks as depth radically increased, by a combination of residuals (concatenating the input with deeper layers) and normalization of hidden activations.

An important technique which gives us a lot of insight into the performance of neural networks is \textbf{distillation}, wherein a network is used or trained to produce a smaller (distilled) network with the same performance as the source network \cite{distill}. If the process of distillation doesn't transform or generate weights, only remove them, it is called \textbf{pruning}. \cite{lotteryticket} discuss pruned networks, observing that networks can frequently be pruned over 90\% without sacrificing accuracy, and noting the weights of the subnetwork are like a `   lottery ticket'. They pose and test the hypothesis that these subnetworks can be found in advance of training, rather than distilled after training, and propose an algorithm for identifying such 'winning tickets', showing that they can train faster and generalize better than the original network, at a fraction of the size of the original network. This is also supported by other empirical works showing that small language models can also generalize well \cite{schick}, as well as early work in neuro-inspired learning rules which showed that even random feedback weights can produce models that generalize \cite{lillicrap}.

\subsection{Optimization landscape}

Though theoretically motivated, deep networks were often difficult to train in practice. A popular method was \textbf{unsupervised pre-training}, e.g.~ \cite{deepbelief}. The benefits of this approach for generalization are empirically demonstrated and discussed in \cite{whypretrain}. In summary, the authors find that pretraining regularizes the network toward minima which generalize well, by encoding a helpful prior about the data distribution. 

Many of these benefits are being rediscovered in work on \textbf{foundation models}\cite{foundation}: enormous models trained (typically in an unsupervised or self-supervised manner) on massive amounts of data with vast compute resources (often only available at large for-profit companies). Use of these models for subsequent tasks amounts to a kind of unsupervised pretraining.  Motivated in part by Rich Sutton's Bitter Lesson \cite{bitterlesson} that the most successful methods are those that scale with compute, the nascent field of neural scaling laws seeks to predict and understand from large-scale experiments how learning and generalization performance scale with compute. Initial results show larger models are more sample-efficient (achieve good performance with less data), and for large transformer-based language models, and transfer learning performance scales with generalization performance.

For non-convex functions like those computed by neural networks, there is not necessarily a global minimum representing the best possible choice of parameters. Thus the process of training can be viewed as one of selecting among local minima (with early stopping, this applies to even convex optimization). Understanding and characterizing the optimization landscape can help understand how it is traversed by learning algorthmns, which minima are likely to be selected, and which give good generalization performance. 

Early stopping and stochastic (as opposed to full-batch) gradient descent were understood as important factors in generalization of deep networks, and many works explore the exact relationship. Early work showed that the noise introduced by mini-batching empirically improved performance, and proposed the noise introduced made the network robust to variance. In particular, there was a widespread view, summarized e.g.~ by \cite{hochtrieter97} that this kind of noise helped networks avoid \textbf{sharp minima}, i.e.~ solutions in the optimization landscape with near-discontinuities, causing instability, instead favouring \textbf{wide minima} which were more likely to generalize better. \cite{keskar} find strong empirical evidence for this view, showing that as batch size increases, networks are more likely to converge to sharp minima whereas small batch sizes resulted in convergence to flat minimizers. However, the relevance of sharpness and flatness of minima was disputed theoretically and empirically by \cite{laurent}, who showed that in high dimensions all flat minima can be made sharp with reparameterization that does not affect overall performance. \cite{trainlong} subsequently support this view by showing large batches can in fact be used to achieve good performance, as long as corresponding changes to the learning rate are made. These seemingly contradictory results were reconciled by \cite{stan}, showing theoretically and empirically that three primary factors influence the minima selected by SGD: learning rate, batch size, and gradient covariance; most of all the ratio of learning rate to batch size.

The lottery ticket hypothesis mentioned above made sense in light of preceding work exploring the dynamics of training with stochastic gradient descent (SGD). In particular, \cite{advanisaxe} studied dynamics of generalization error in the high-dimensional regime in which deep nets typically operate. They showed overfitting and memorization are most likely to occur when the number of data examples is approximately equal to the number of parameters in the network, and that generalization performance is benefitted both at small and large network sizes, via implicit regularization. In the case of overparameterzed networks, they identify two important reasons for the implicit regularization: one is that the statistical properties of training in this regime give better-conditioned input correlations, effectively making patterns easier to learn, and (2) networks with Relus have a large number of weights that move very little (what they call a frozen subspace), creating sparse subnetworks in the larger network. This is in line with findings from \cite{mohammad}, who find that deep networks are biased toward approximating simpler functions.

Other work on dynamics of generalization in high-dimensional consider generalization as a function of effective capacity. Traditional learning theory in the form of the bias-variance tradeoff considers a regime where test error first decreases and then increases as a function of model capacity (measured by number of parameters) or training time (measured by epochs), forming a U shape (i.e.~ a single `descent' into generalization). The authors of \cite{doubledescent} investigate an empirically-observed phenomenon where if model capacity or training time are increased further, a second, more gradual descent is observed, which eventually gets lower than the original. The authors conjecture that the double descent is a general phenomenon of the dependence of generalization on effective capacity. 

\subsection{Data distribution}

There is a consensus view that the compositional architecture expresses a prior over the space of functions that is useful for practical data distributions -- i.e.~ most real-world data are compositionally structured.

A lot of work, particularly in vision, has studied the effects of \textbf{data augmentation} on generalization: see \cite{dataaug} for review. Classically, data augmentation is used in situations of limited data/small datasets, but these works demonstrate empirically the generalization benefits of augmentations, even for large datasets. Data augmentation can be viewed as a form of non-adversarial noise which encourages robustness to the transformations applied in the augmentation. Study the predictions made by networks trained with different data augmentations can also help us interpret and understand the influence of data distribution characteristics on generalization, e.g.~ \cite{hendrycks_bench} This can have important implications for OOD generalzation, particularly in safety-critical domains. 

Similarly, many works have shown empirical benefits to robust generalization of pretraining and self-supervised learning, which can be viewed as forms of data augmentation \cite{hend_pretrain, hend_self, hend_augmix}. Very recent work proves a general robustess result for deep networks, showing that overparameterization is actually \textit{necessary} for smooth fitting of data distributions -- i.e.~ necessary for effective interpolative generalization \cite{robustnesslaw}. This gives theoretical weight to the folk wisdom that an over-parameterized model with regularization will have better generalization performance than a smaller network with no regularization. 

Many of the works mentioned in the above consider expressivity, not generalization, but their results are relevant for generalization theory because they demonstrate the importance of compositionality for pattern-learning. There was widespread folk wisdom that learning and composing patterns fundamentally favoured good generalization performance, because doing so was in some sense contrary to memorization. This was well-supported by empirical evidence (e.g.~ the huge gains in generalization performance that deep networks displayed over alternative methods on image benchmarks like ImageNet and CIFAR). However, learning-theoretic results still suggested that these networks were capable of memorizing training data, and by showing that over-parameterized networks were able to fit datasets of random noise, \cite{Zhang} et al. suggested that counter to folk wisdom, deep networks could indeed be functioning as giant lookup tables for training data. 

Our work \cite{arpit2017} refuted that conjecture and confirmed folk wisdom by showing empirically that learning behaviour is dependent on the data distribtution, and simple patterns that generalize are learned first before memorization, among other observations (see Sec\ref{sec:memgen}). In that work we called for data-dependent measures of generalization. 

These were soon provided by a landmark paper presenting non-vacuous bounds on generalization error by using a PAC-Bayes (Probabily Approximately Correct Bayes) framework \cite{nonvac}. Many  bounds are computed analytically based on simple measures of the data distribution such as input dimensionality or norms. In contrast, this work \textit{trains} a representation of the data to be used in a bound, jointly with the training of a (stochastic) neural network on real data, using stochastic gradient descent. This representation is a distribution over weights in a region of low error found by the optimization procedure of training, i.e.~ it is not only data-dependent but depends on the interaction of SGD, architecture, and optimization, in a realistic training setting, i.e.~ realistic to the hypothesis class of trained networks.   In this way the trained bound is actually a model of generalization performance, similar to a witness function or probes \cite{alain} trained to predict or make interpretable various aspects of network function. 
In similar work, \cite{hsu} use the process of distilling a network, together with a data augmentation as a source of stochasticity, to estimate generalization bounds.

Optimized non-vacuous data-dependent generalization bounds are an exciting direction for giving us a realistic picture of average generalization performance. However, particularly in the real world, sometimes it is not the average that matters, but the extreme cases. \cite{easyex} empirically studies which examples are easy or hard for a network. 

\subsection{Task definition and context}

Several recent works propose alternatives to the empirical risk framework for training deep models, with many such as \cite{mixup}, \textbf{invariant risk minimization (IRM)}\cite{invarisk}, and \textbf{risk extrapolation (ReX)} \cite{rex} emphasizing the shortcomings of ERM for encouraging generalization, especially (even slightly) out-of-distribution generalization: As shown by \cite{szegedy2013intriguing}, we can find images which differ only imperceptibly to the human eye but produce radically different classification. 

\textbf{Geometric deep learning} (GDL) is a notable new lens on understanding deep networks coming from an entirely different perspective than statistical learning. It explicitly studies the inductive biases and invariance properties of architectures (e.g.~ for convnets, translation-invariace) and how well these `match' those we expect from the data distribution \cite{geom}. A lot of work in GDL focuses on \textbf{graph neural networks} (GNNs), which can be viewed as a generalization of RNNs to dependencies other than temporal sequence, and are often used to learn representations of graph-structured data such as protein folds or social networks. See \cite{ameya} for an introduction. GNNs introduce many novel considerations for generalization, including generalization over the size of graph, studied by \cite{graphsize}.

Alignment is the process of getting AI systems to understand and act in accordance with what we `really want'. There is typically a `specification gap' between the way we define and operationalize a problem that an AI system solves (e.g.~ predicting which word someone is most likely to type) vs. what we `really want' (e.g.~ faster communication, with fewer errors) [20,21]. Sometimes the problems arising from this gap are minor inconveniences, but they can be much more serious, including amplifying existing bias and creating risks for already marginalized populations [22]. [23] present an overview of ways underspecification produces problems in real-world deployment settings. [18] perform an empirical exploration showing that even on simple OOD tasks, typical deep learning approaches fail, and [19] perform a theoretical analysis of the modes of failure. [24] summarizes concrete problems in AI Safety, and [25] propose addressing problems in practice by training AI systems to avoid `side effects'. Reliability engineering, unit tests, sandboxing or red-teaming, time-to-failure analysis are common practice in software engineering [26] and can in many cases be straightforwardly applied to AI systems. 


For real-world data, it is often unrealistic to assume train and test sets are sampled i.i.d, representatively of the population they will be deployed on. This is often called \textbf{dataset bias}, i.e.~ bias in the sampling procedure, particularly when it is desired to contrast it with \textbf{algorithmic bias}, i.e.~ bias due to the way an algorithm works. This separation of this view, often made by AI developers to declaim responsibility for adverse outcomes of systems trained on biased data (data's fault not the algorithm's fault), has been criticized as overly simplistic \cite{timnit}. As the procedure for computing non-vacuous bounds shows us, data and model are intimately linked in terms of predicting real-world performance. Additionally, the authors observe that training the witness actually regularizes the network -- modeling generalization performance in this way \textit{improves generalization}. This is very promising if it holds true other kinds and areas of generalization.

For very large language models trained on self-supervised tasks, an informal measure of generalization performance is transfer learning to new tasks - large-scale language models like GPT are used for many different tasks from reservation booking to poetry generation. \cite{gptprompt} explore a more formal way of doing this via \textbf{prompting}.

As the size of datasets increase beyond human capability to vet (e.g.~ substantial portions of all text on the internet), the methods described thus far for measuring and predicting generalization performance do not scale. 
The way models are trained can exacerbate bias in the data, but if the dataset is too large to vet we can't measure this or even identify it until it's a problem. \cite{parrots} discuss these issues and more, calling for investment into more careful data curation and pre-development practices.

\Chapter{\uppercase{Empirical investigations of memorization and learning behaviour}}\label{sec:memgen}

\begin{quote}\singlespacing

\cite{arpit2017} Devansh Arpit$\dagger$, Stanislaw Jastrzebski$\dagger$, Nicolas Ballas$\dagger$, David Krueger$\dagger$, Emmanuel Bengio, Maxinder S Kanwal, Tegan Maharaj, Asja Fischer, Aaron Courville, Yoshua Bengio, Simon Lacoste-Julien. 2017. A closer look at memorization in deep networks. \textit{International Conference on Machine Learning (ICML)}.

\cite{memRNN} Tegan Maharaj, David Krueger, and Tim Coojimans. 2017. Memorization in recurrent neural networks. \textit{International Conference on Machine Learning (ICML) workshop on Principled Approaches to Deep Learning (PADL)}

$\dagger$ denotes equal contribution.
\end{quote}

In early 2017, deep nets had just begun to be deployed widely in the real world, for example for image tagging/description of images from the internet. But the reasons for their success in generalizing well on these real-world images remained unclear - in particular, as noted by \cite{Zhang}, traditional statistical learning theory failed to explain why these massively overparameterized networks didn't simply memorize and thus overfit their training data.  
In this chapter I cover two works investigating this observation: I summarize \cite{arpit2017}, a published paper I contributed to examining memorization in feed-forward nets, and integrate \cite{memRNN}, a workshop paper examining memorization in recurrent neural networks (RNNs), of which I am first author.

Our work took a complementary direction to traditional statistical learning theory, one that is now sometimes called `empirical theory' or `science for deep learning'. In contrast to a more mathematical first-principles approach, this is more similar to the approach taken in studying biological intelligence: formulating testable hypotheses and designing experimental settings to elucidate phenomena. Specifically in this case, our approach was to operationalize the definition of memorization as whatever was done to fit noise data, and then compare learning and generalization behaviour on noise vs. real data. Further, I led a portion of the project which examined generalization in recurrent neural networks. Theory works tend to focus on atemporal data, but many data of real-world interest (e.g. many data relating to climate, social science, and health) have important temporal variation. 

Taken together, these works provide strong empirical evidence for important elements of folk wisdom about deep net generalization, including that learning behaviour depends on the data distribution, and that simple patterns which generalize well are learned first, before memorization/overfitting of more complex patterns occurs.  The findings from these highly-cited works have been supported in a wave of empirical and theoretical follow-up studies.



I contributed significantly to this large collaborative project, one of the most highly-cited works from Mila in recent years. With one of the lead co-authors of the project, David Krueger, I discussed the initial framing of the paper, and articulated the approach of defining memorization as behaviour on random data. With all other authors, I co-developed  the overall empirical approach of contrasting 
memorization and generalization behaviour to gain insight about deep network learning and reasons for good generalization performance. I contributed to discussion and refinement of experiments to run, including suggesting the second experiment, and determining and running experiments and making plots for the fourth experiment. I wrote significant portions of the paper, especially explanations of results, and edited drafts and final versions for clarity. I coded and ran all initial experiments with recurrent neural networks. We ultimately decided to move these results to a separate paper for space reasons, which is the workshop paper \cite{memRNN}. I created the poster presentation of our results, creating a list of our key insights and leading to a reorganization of content which significantly improved clarity.

The `empirical theory' approach of this paper is generally useful for generating the kinds of insights and intuition that are helpful for designing systems that generalize in the real world. Understanding when neural networks will memorize (vs. learn underlying patterns of) training examples also helps us better design new methods and training regimes.

\section{Memorization in feed-forward networks }
In this paper \cite{arpit2017} we identify behaviour on random noise as a useful operational definition of memorization, and perform a series of experiments demonstrating differences in learning behaviour for feed-forward networks trained on real data vs. random noise. 

In summary, we show the following for deep feedforward nets:
\begin{enumerate}
    \item Fitting noise requires more effective capacity 
    \item Training on noise gets harder, faster, when the dataset grows
    \item On real data, some examples are always/never fit immediately, and some examples have more/less impact on training (not so for noise)
    \item Simple patterns are learned first., before memorizing
    \item Regularization can effectively reduce memorization
\end{enumerate}

\subsection{Difference 1: Effective capacity needed to fit}

In the first experiment, shown in Figure \ref{fig:one} we examine performance as a function of capacity for different levels of noise in either the inputs or labels for 2-layer MLPs. Random inputs means a percentage of input examples are replaced with noise, random labels means a percentage of labels are replaced with a randomly chosen one.  For real data, performance is already very close to maximal with 4096 hidden units, but as noise is increased (for either random inputs or labels), higher capacity is needed to achieve maximal performance. This experiment shows greater effective capacity is required to fit noise than real data.

\begin{figure}[!h]
\centering
\includegraphics[width=5in]{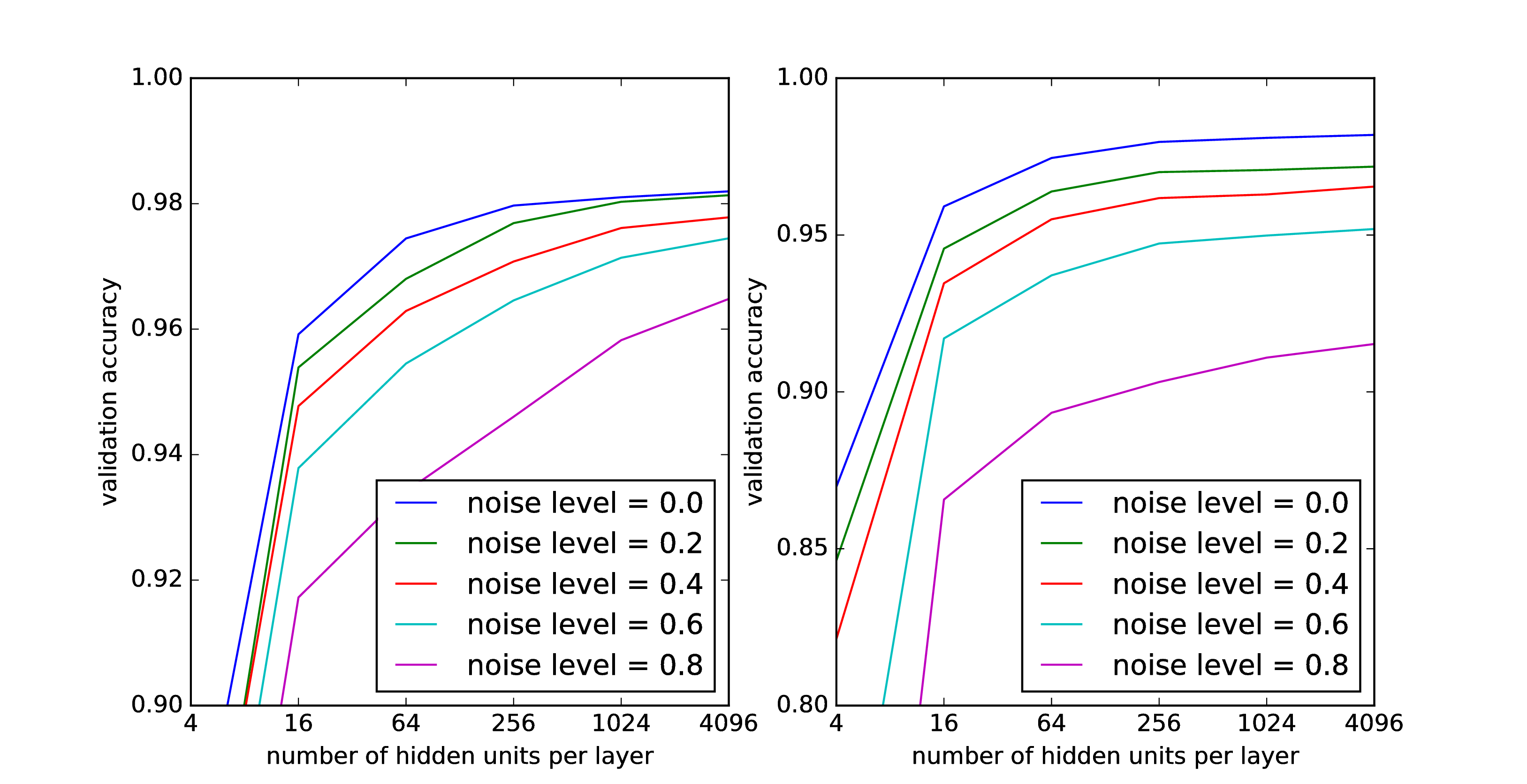}
\caption{Performance as a function of capacity in 2-layer MLPs trained on (noisy versions of) MNIST. For real data, performance is already very close to maximal with 4096 hidden units, but when there is noise in the dataset, higher capacity is needed. \textbf{Figure reproduced from \cite{arpit2017}}
}
\label{fig:one}
\end{figure}

\subsection{Difference 2: Scaling with dataset size}

In the second experiment, Figure \ref{fig:two} we plot change in normalized time to convergence as a function of dataset size, with capacity fixed. Because there are patterns underlying real data, increasing dataset size doesn't increase training time for real data as much as it does for noise; the network is able to leverage the patterns learned to achieve good performance. In other words,  training on noise gets harder, faster, when the dataset grows.

\begin{figure}[!h]
   \centering
    \includegraphics[width=2.5in]{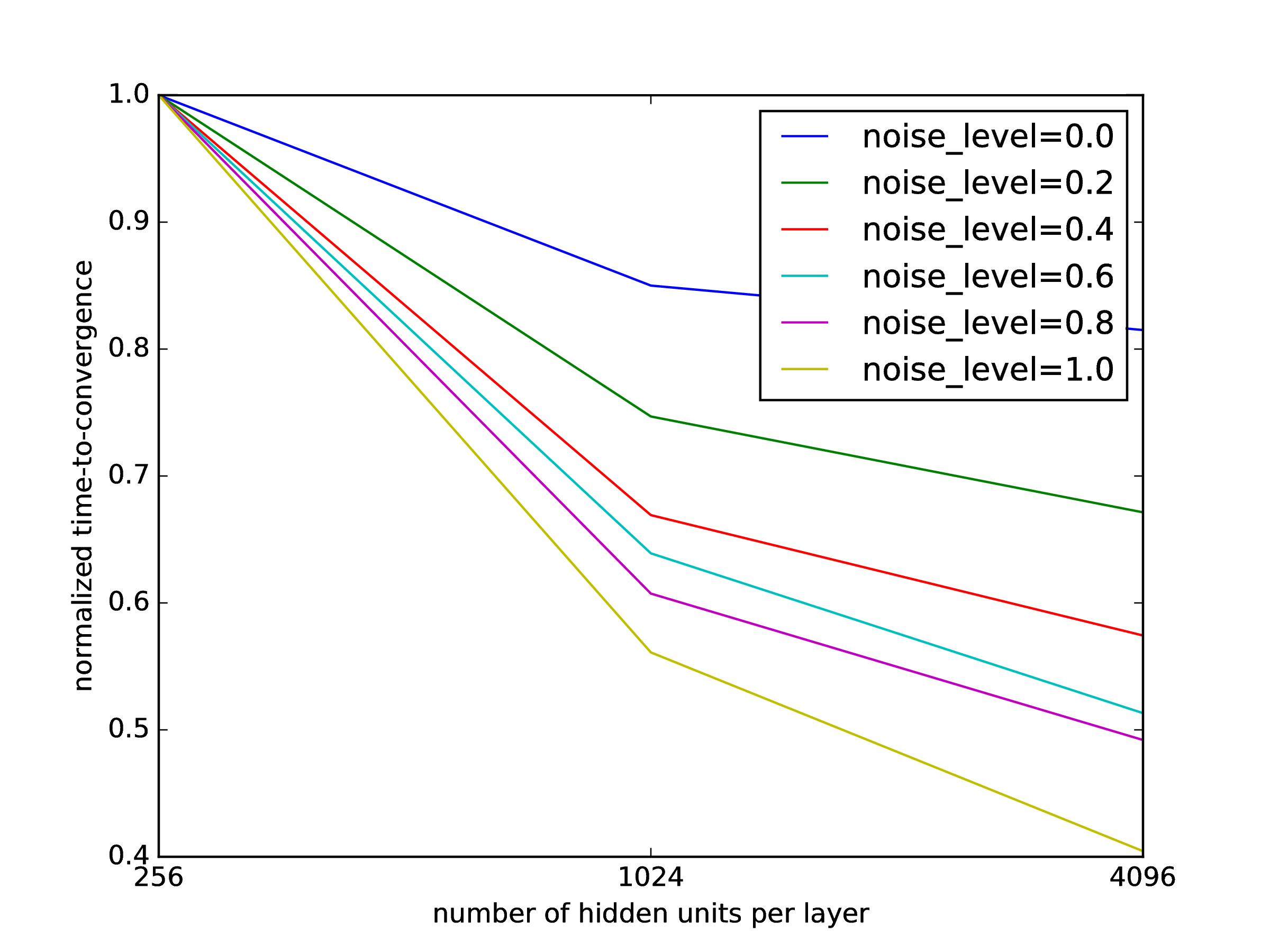}
\includegraphics[width=2.5in]{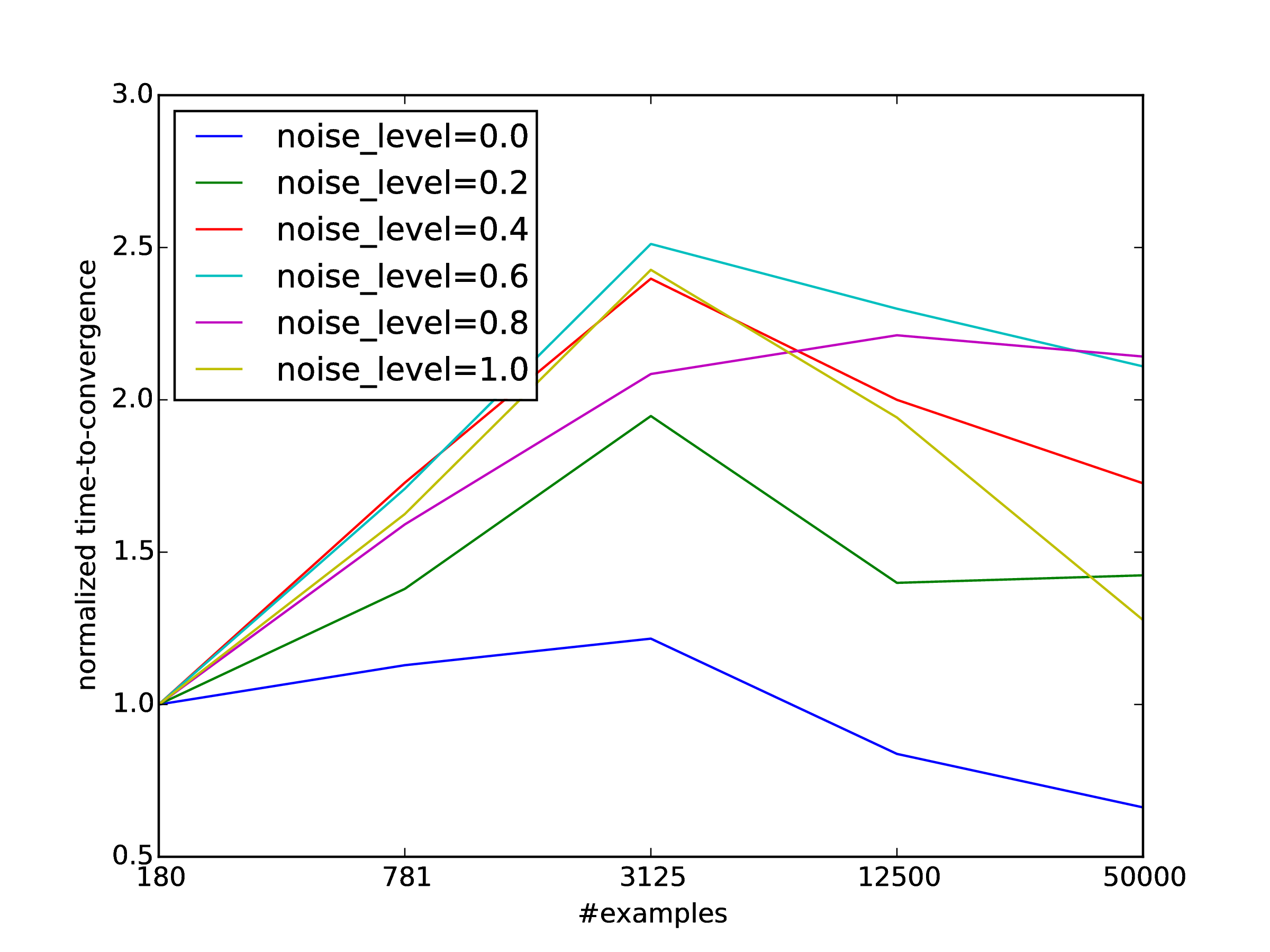}
   \caption{
Time to convergence as a function of capacity with dataset size fixed to 50000 (left), or dataset size with capacity fixed to 4096 units (right).
``Noise level'' denotes to the proportion of training points whose inputs are replaced by Gaussian noise.
Because of the patterns underlying real data, having more capacity/data does not decrease/increase training time as much as it does for noise data.
\textbf{Figure reproduced from \cite{arpit2017}}
}
   \label{fig:two}
\end{figure}
	
\subsection{Difference 3: Easiness of examples}

In the third set of experiments, we examine which examples are easy vs. hard for the network to classify. We first measure this in Figure \ref{fig:easy_hard.pdf}, by tracking which examples are correctly classified after one epoch of training. Easiness of examples (i.e. probability of being correctly classified after 1 epoch of training) varies much more for real data than for noise distributions, for which the probability of being correctly classified in the first epoch is random. 

\begin{figure}[!h]
  \center

  \includegraphics[width=3in]{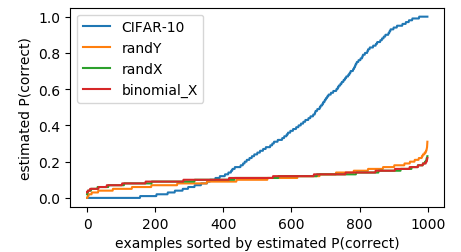}
  \caption{Average (over 100 experiments) misclassification rate for each of 1000 examples after one epoch of training.
           This measure of an example's difficulty is much more variable in real data.
       We conjecture this is because the easier examples are explained by some simple patterns, which are reliably learned within the first epoch of training.
	   We include 1000 points samples from a binomial distribution with $n=100$ and $p$ equal to the average estimated P(correct) for randX, and note that this curve closely resembles the randX curve, suggesting that random inputs are all equally difficult. \textbf{Figure reproduced from \cite{arpit2017}}
}
  \label{fig:easy_hard.pdf}
\end{figure}

We also in Figure \ref{fig:grad_sensitivity}  compute loss-sensitivity as the partial derivative of the loss $L$ with respect to example $x$ , averaged over training iterations $t$, and plot the Gini coefficient (a measure of roughness/disparity over categories) of the average loss-sensitivity over the course of training, on a 1000-example real dataset (14x14 MNIST). We show that this disparity (of loss-sensitivity, between different examples, over the course of training) is higher for real data.

\begin{figure*}[!ht]
  \centering
  \includegraphics[width=6.1cm]{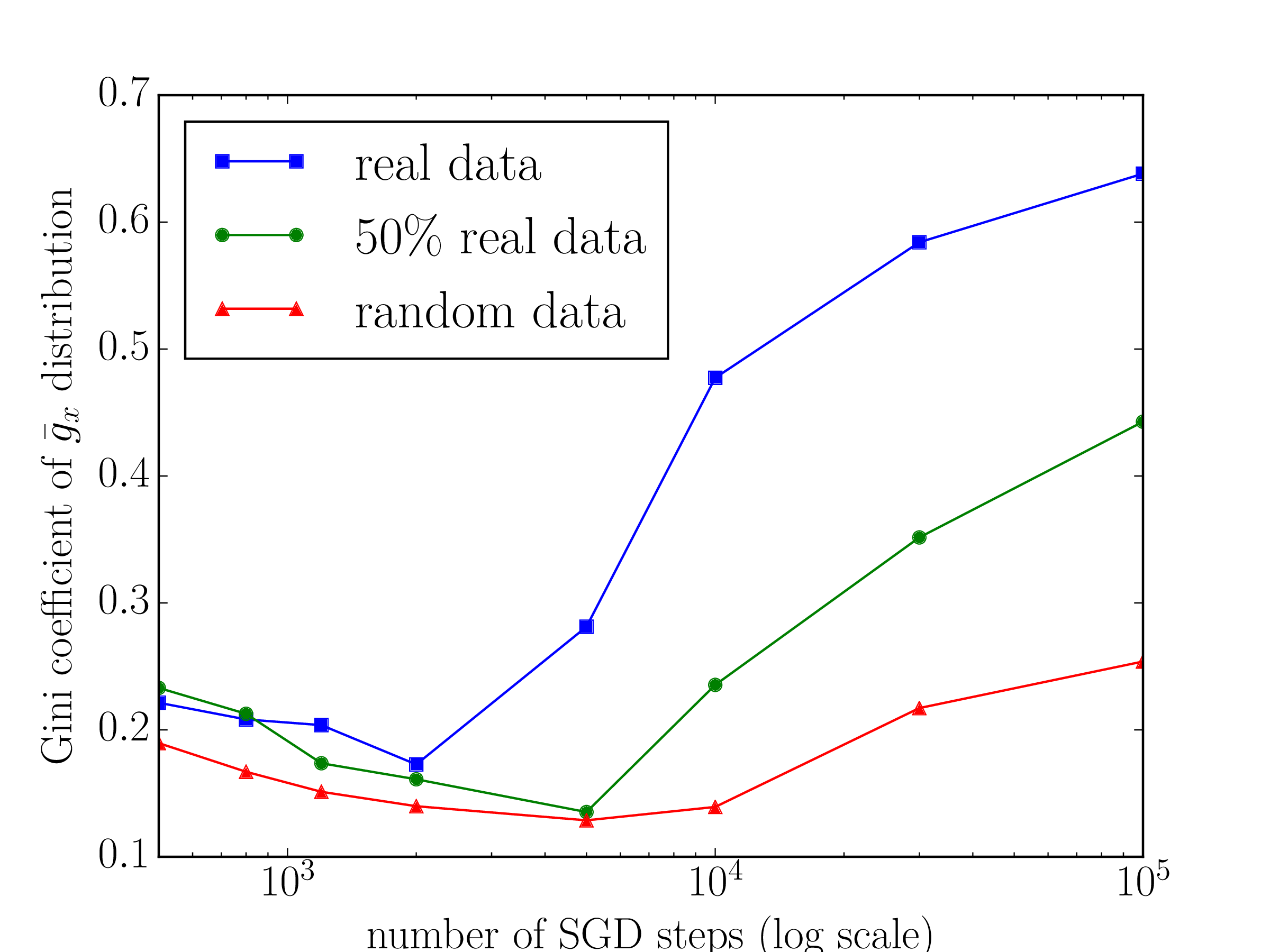}
  \hspace{-1em}
  \includegraphics[width=6.1cm]{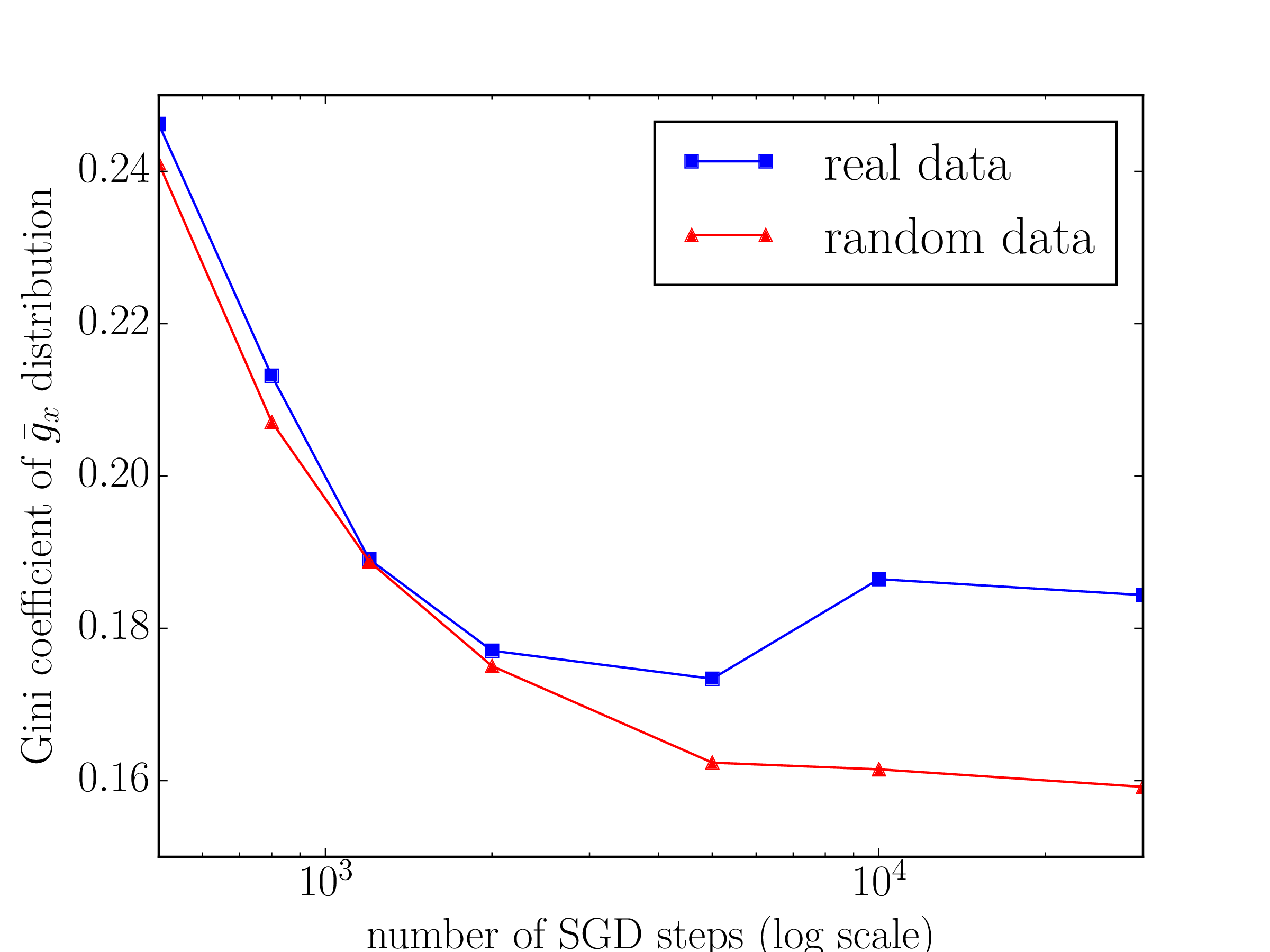}
  \caption{Plots of the Gini coefficient of $\bar{g}_\mathbf{x}$ over examples $\mathbf{x}$ as training progresses, for a 1000-example real dataset (14x14 MNIST) versus random data. On the left, $Y$ is the normal class label; on the right, there are as many classes as examples, the network has to learn to map each example to a unique class. \textbf{Figure reproduced from \cite{arpit2017}} 
    }
  \label{fig:grad_sensitivity}
\end{figure*}

In Figure \ref{fig:grad_class_sensitivity} We further compute and plot the per-class loss-sensitivity, showing that loss-sensitivity is more highly class-correlated for real data. These experiments and plots demonstrates that even before training, inductive biases of the architecture favour real data distributions, and that over the course of training some examples influence the loss more than others, in ways that contribute meaningfully to performance: On real data, some examples are always/never fit immediately, and some examples have more/less impact on training (not so for noise).

\begin{figure}[h]
  \centering
  \hspace{-2.0em}
  \includegraphics[width=3in]{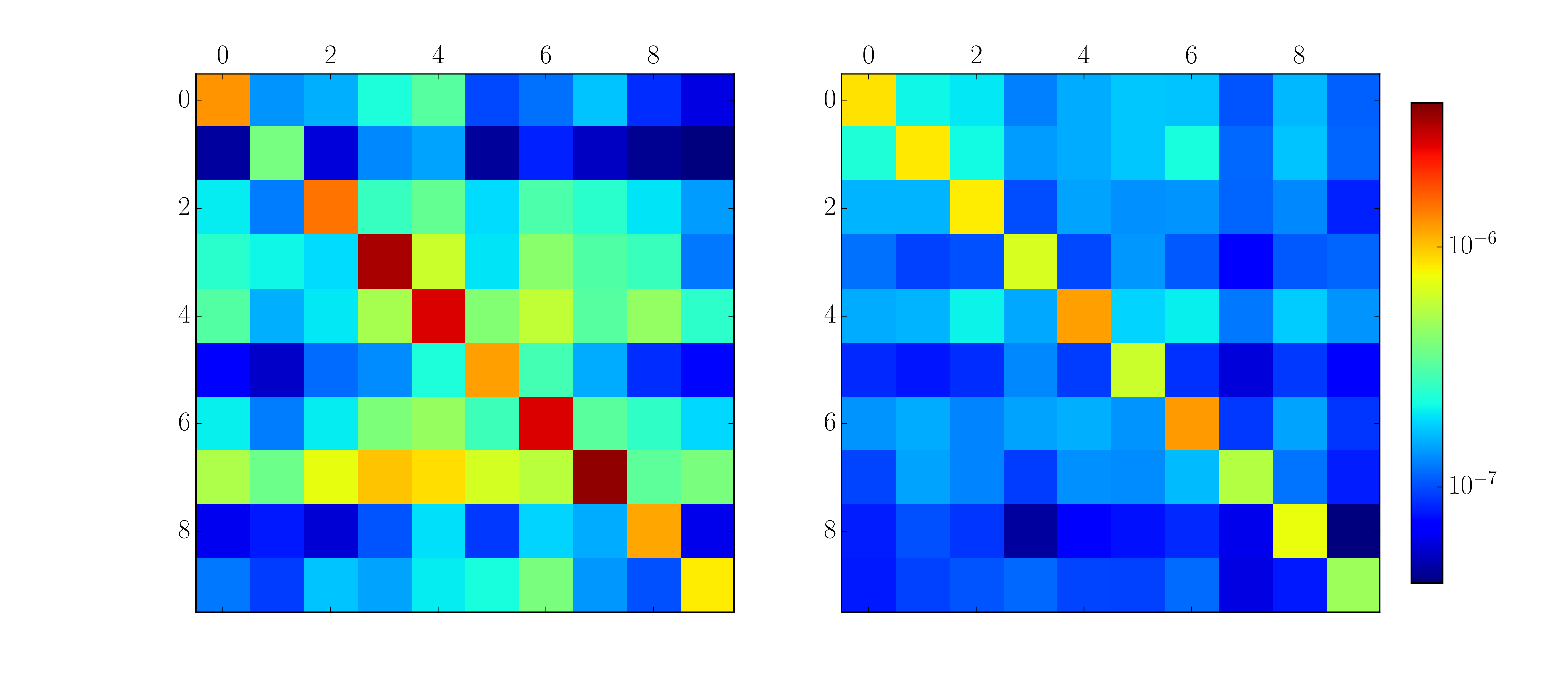}
  \caption{Plots of per-class $g_x$ (see previous figure; log scale), a cell $i,j$ represents the average $|\partial \mathcal{L}(y=i)/\partial x_{y=j}|$, i.e. the loss-sensitivity of examples of class $i$ w.r.t. training examples of class $j$. Left is real data, right is random data. \textbf{Figure reproduced from \cite{arpit2017}} 
  }
  \label{fig:grad_class_sensitivity}
\end{figure}

\subsection{Difference 4: Critical sample (differently classified but similar) ratio}
For the fourth experiment, we first define a  critical sample  as an example which has a nearby differently classified example. The ratio of critical samples is the proportion of examples for which a critical sample is found in radius $r$. This gives an idea of the number of decision boundaries in the function a network computes; i.e. how complicated that function is. We then compute the critical sample ratio for randomly chosen examples over the course of training on CIFAR-10, for noise input, noise labels, and real data. As measured by critical sample ratio, function complexity increases very rapidly for noise data, and increases eventually, to almost the same level, for noise labels. This slower rate for real data with noise labels as compared to noise data demonstrates that even without correct supervision from the loss, in the presence of real data the inductive bias of deep net architectures/training favours simple functions (which will tend to generalize better). For real data, function complexity increases at a rate similar to (but lower than) noise labels at first, but levels off at a much lower level (about half that) of than either noise data or labels - i.e. functions learned on real data are about half as complex as those on random data, as measured by the critical sample ratio.

This is further supported by a large suite of experiments where we gradually increase the amount of noise present in the data or labels, and show that in all cases, for real data, there is a bump in generalization performance early in training, before the network is fully trained and before overfitting occurs, as seen in Figure \ref{fig:comp} This demonstrates that the simple patterns learned early on enable the network to generalize, and it is only after substantially more training where networks enter the regime of memorization. Measuring the critical sample ratio across these experiments shows that critical samples provide a good basis for assessing generalization across tasks and data types. 

    \begin{figure}[h]
        \center
        
        \includegraphics[width=6.5in]{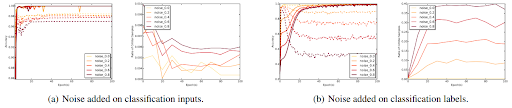}
        \caption{Accuracy (left in each pair, solid is train, dotted is validation) and Critical sample ratios (right in each pair) for MNIST (top row) and CIFAR-10 (bottom row) for different types of noise (inputs: left columns, labels: right columns), for different amounts of noise (real data is yellow, increasing noise is red). Critical samples provide a good basis for assessing generalization across tasks and data types. \textbf{Figure reproduced from \cite{arpit2017}} }
        \label{fig:comp}
    \end{figure}

\subsection{Difference 5: Effect of regularization}
Finally, in the fifth set of experiments in Figure \ref{fig:randY_capping} we examine the effects of different forms of explicit regularization on learning and generalization behaviour, and demonstrate that regularization can slow down and even prevent entering the memorization regime. Of the methods we experiment with, dropout was most effective at this.

    \begin{figure}[h]
        \center
       
        \includegraphics[width=2in]{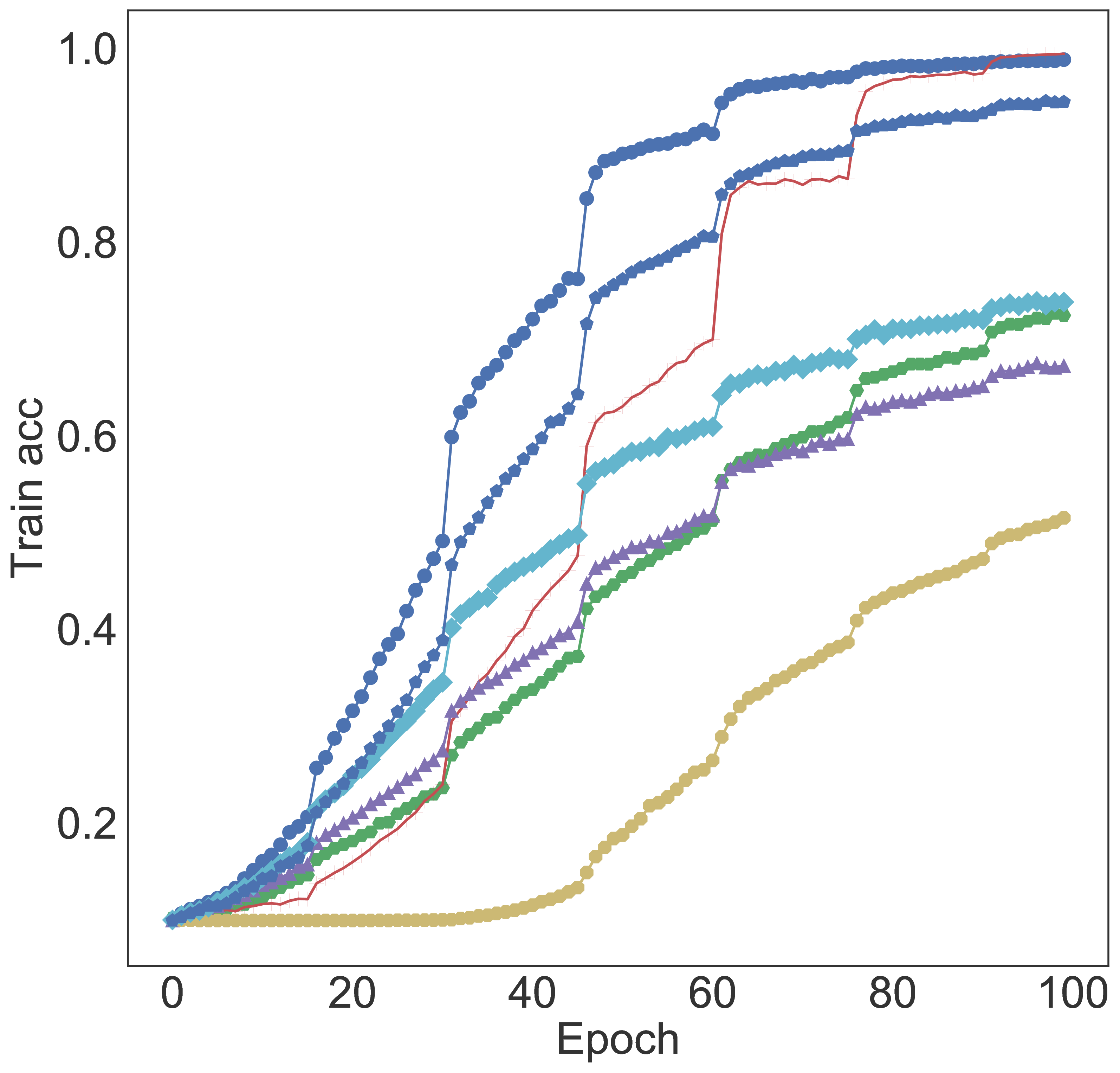}
        \includegraphics[width=2in]{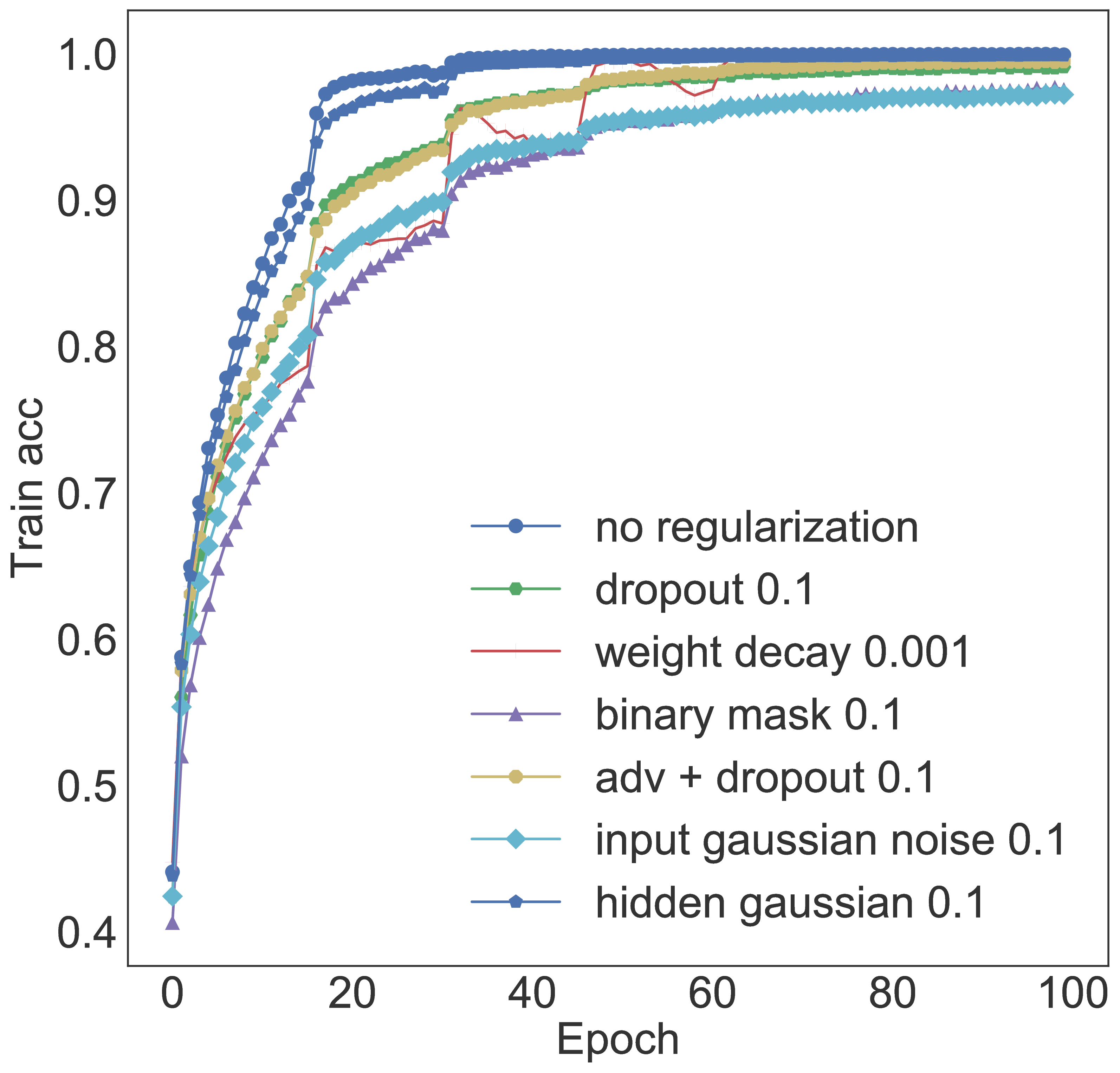}
        \caption{Training curves for different regularization techniques on random label (left) and real (right) data.  The vertical ordering of the curves is different for random labels than for real data, indicating differences in the propensity of different regularizers to slow-down memorization. \textbf{Figure reproduced from \cite{arpit2017}} }
         \label{fig:randY_capping}
    \end{figure}

\section{Memorization in RNNs}
\vspace{0.5cm}
\textbf{Abstract}: \\
We present work in progress on understanding generalization in deep networks by analyzing differences in learning behaviour of recurrent neural networks (RNNs) on noise data vs. real data - i.e., by investigating memorization.
There has been a recent surge of interest in explaining the generalization performance of deep neural networks (DNNs), partially spurred by the observation that feedforward DNNs can fit random noise datasets with 0 training error.
RNNs are typically extremely deep; analyzing learning behaviour in RNNs thus gives an interesting perspective for understanding memorization, generalization, and effective capacity in deep networks.
We demonstrate that fitting noise with RNNs is more difficult than in feedforward networks; standard gradient-based optimization fails to reach 0 training error. 
We make several other observations comparing and contrasting our results with previous work, and suggest a suite of experiments for future work.

%
%
\subsection{Introduction}
The question of whether deep networks fit data by finding patterns or simply memorizing examples has been the subject of recent work.
\cite{Zhang} point out that traditional measures of capacity do not explain the (very good) generalization performance of deep networks, given that these networks are typically massively over-parameterized, and are capable of fitting random noise. 
\cite{arpit2017} examine the question from another angle, demonstrating that deep networks have qualitatively different behavior on random vs noise data, and suggesting that DNNs generalization is due to their propensity to learn simple patterns first. 
\cite{arpit2017} call for a data-dependent notion of capacity to explain DNN generalization, and \cite{nonvac} provide one, establishing non-vacuous generalization bounds for stochastic DNNs. 

All these works, however, examine feed-forward networks; we extend this investigation to recurrent neural networks (RNNs), performing experiments on the tasks of character-level language modeling (Penn Treebank) and classification (sequential MNIST).
While some of the results in these previous work on feedforward networks carry over to RNNs, we also find significant differences.

{\bf Our findings so far are:}
\begin{enumerate}
\item Fitting random noise with RNNs is much harder than with feedforward nets.  While we were able to fit small sets of noise data to some extent, we fail to substantially reduce training error on either task when training on the full (noised) datasets.  This contrasts strongly with the results of \cite{Zhang} on feedforward nets.
\item Like \cite{Zhang} and \cite{arpit2017} we find that label noise is more difficult to fit than input noise.
\item Like \cite{arpit2017}, we find that models trained on mixed datasets (containing both real and noise examples) first learn patterns which generalize to the validation set (i.e. to unseen {\it real} data).
\item We find that easy examples exist in the random input version of the sequential MNIST task.  This contrasts with the results of \cite{arpit2017}; in their experiments with MLPs, random MNIST inputs appear to have equal difficulty.
\end{enumerate}

RNNs are known to be difficult to optimize in general, but the exceptional difficulty of fitting random data is surprising.
Having noted the difficulty of fitting random data with RNNs, our next step will be a more thorough attempt to fit RNNs on a broader range of synthetic tasks that allow us to control relevant factors such as sequence length and dataset size.
Going forward, we also plan to replicate more of the experiments of \cite{arpit2017} on RNNs.


Unlike for feedforward nets, for RNNs, the ability to memorize examples may actually be desirable.
At a high level, the distinguishing feature of RNNs is their ability to create a fixed-length representation of variable length data using a finite number of parameters.
From this viewpoint, an idealized RNN would perform lossless compression of any input sequence,
since the relevance of past and present inputs depends on future inputs and is hard to anticipate.
Meanwhile, the output mapping could take responsibility for discarding information stored in the current hidden state which is irrelevant to the current output. 
The difficulty of optimizing RNNs suggests that their learned compression is both 1) is lossy, and 2) strongly data-dependent.

\subsubsection{Structure of this paper}
We first define RNNs formally, then discuss concepts of depth, effective capacity, and memorization as they apply to RNNs, and motivate our empirical approach in line with previous work. 
We present and discuss results of our investigation of learning behaviour, generalization, and effective capacity as influenced by noise vs. real data, and conclude by proposing further experiments, conjectures to be investigated, and possible avenues for theoretical corroboration of our results.

\subsection{Background and Related Work}
\subsubsection{Recurrent neural network definitions and notation}
A recurrent neural network (RNN) is a neural network which processes sequential input $(x_1, x_2,\dots, x_t,\dots, x_n)$, using a nonlinear function $f$ to construct corresponding representations (called hidden states, or activations)  $(h_1, h_2, \dots, h_t, \dots, h_n)$, each of which depends on the input and on the previous timestep:
\begin{align} 
h_t = f(x_t, h_{t-1})
\end{align}
For a \textbf{simple RNN}, or `vanilla' RNN, this is most commonly implemented with two sets of recurrent parameters; $W$ for the input, and $U$ for the hidden-to-hidden transition, and where the activation function $f$ is usually a logistic sigmoid or hyperbolic tangent (tanh), sometimes rectified linear (ReLU).:
\begin{align} 
h_t = f(Wx_t, Uh_{t-1})
\end{align}
 RNNs are usually trained with stochastic gradient descent (SGD) via backpropagation through time (BPTT).
The repeated use of the same parameters for each timestep of input can be viewed as repeated applications of a transition operator with learned parameters. While this characteristically allows RNNs to maintain a 'memory' of information from past timesteps, because of repeated multiplications they are vulnerable to problems of both vanishing and exploding gradients, as demonstrated by \cite{bengio1994learning}. This observation motivates the use of gated architectures, as used in Long short-term memory networks (LSTM) \cite{hochreiter1997long} and Gated recurrent units (GRU). 

\subsubsection{Recurrence and Depth} 
RNNs are the deepest neural networks.
But unlike with feedforwards nets, the depth in recurrent networks mostly comes from the repeated application of the same transition operator.
This may cause an RNNs hidden activations (and their gradients) to increase or decrease exponentially \cite{bengio1994learning} (although inputs, noise, and nonlinearities may all counter-act this effect to some extent). 


\subsubsection{RNN capacity} 
Recurrent neural networks are theoretically capable of representing universal Turing machines \cite{siegelmann1995computational} (although in practice finite numerical precision limits their capacity), so effective capacity is particularly important when trying to look at RNNs.
We define effective capacity as in \cite{arpit2017}; effective capacity is an attribute of a {\it learning algorithm} (including a model and training procedure), and denotes the set of hypotheses which that algorithm could reach given {\it some} training set.

%
%

\subsection{Experiments and Discussion}
Experiments are designed to examine differences between noise vs. real data, in order to assess memorization behaviour. 
We perform experiments on sequential MNIST~\cite{le2015simple} (\textbf{sMNIST}) and character-level Penn Treebank~\cite{marcus1993building} \textbf{cPTB}). sMNIST is a classification task: given the sequence of pixels in a vector representing an image, classify the digit shown in that image. cPTB is a language-modeling task: given the sequence of characters thus far, predict the next. cPTB is usually measured in log-likelihood/bits-per-character (BPC)\footnote{BPC is a deterministic function of NLL; $BPC = \log_2(NLL)$}), but in order to assess memorization, we look at per-character and per-sequence \textit{accuracy} on cPTB.

Unless otherwise noted: all experiments for sMNIST involve a single-layer LSTM~\cite{hochreiter1997long} with orthogonal initialization and Layer Normalization~\cite{layernorm} optimized with RMSProp~\cite{rmsprop} with decay rate 0.5 and learning rate $1e^{-3}$.
All experiments for cPTB use single-layer LSTMs with 1000 units, initialized orthogonally, optimized with ADAM \cite{kingma2014adam}, with sequences of 100, gradient norm clipping set to 1, and learning rate of $2e^{-3}$
.

\subsubsection{Differences on noise vs. real data early in training}

Results on sMNIST, shown in \ref{fig:easy_ex_mnist}, are similar to findings of \cite{arpit2017} with feed-forward nets on MNIST in that some examples are easier than others, but differ in that this is also true for X noise (random data).

\begin{figure}[htb!]
  \centering
  \includegraphics[width=.45\textwidth]{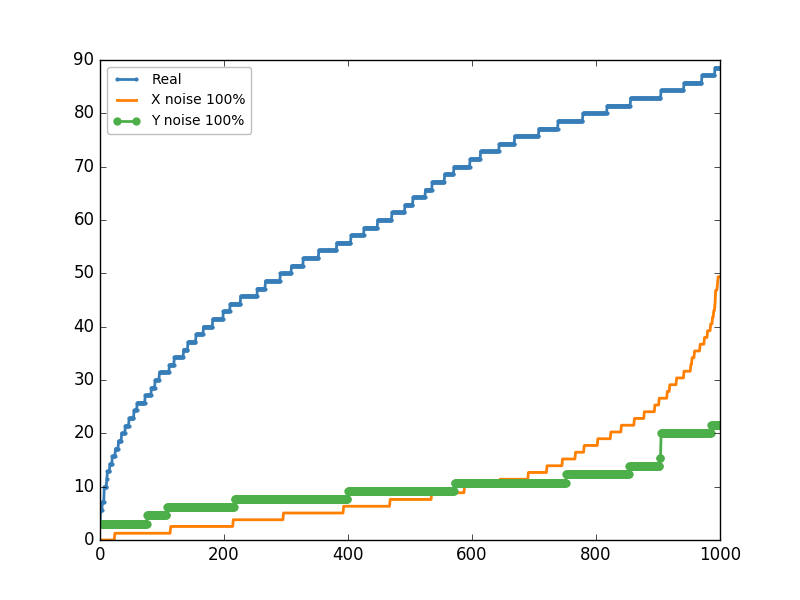}
  \caption{
Normalized percentage of times a given example is correctly classified after 100 epochs of training, averaged over 70 experiments with different random seeds on 1000 examples from sMNIST. 
More examples are consistently learned earlier for real data, and random data (X Noise, orange) is harder to fit than random labels (Y Noise, green). 
More significantly, we note that all three curves show evidence of variability in the difficulty of fitting examples, in contrast to previous results with feedforward nets.
}

  \label{fig:easy_ex_mnist}
\end{figure}




\subsubsection{Differences in learning behaviour and generalization}

First in Figure \ref{all_sizes_results} we plot. Observe that we can't fit noise on the full dataset or 1/8th, so we examine different noise levels only for 1/64th (3162 examples), 1/512th (790 examples), and 1/4096th (127 examples), shown in Figure \ref{noise_vals_results}.





\begin{figure}[htb!]
  \centering
  \begin{subfigure}[t]{0.47\textwidth}
    \includegraphics[width=\textwidth]{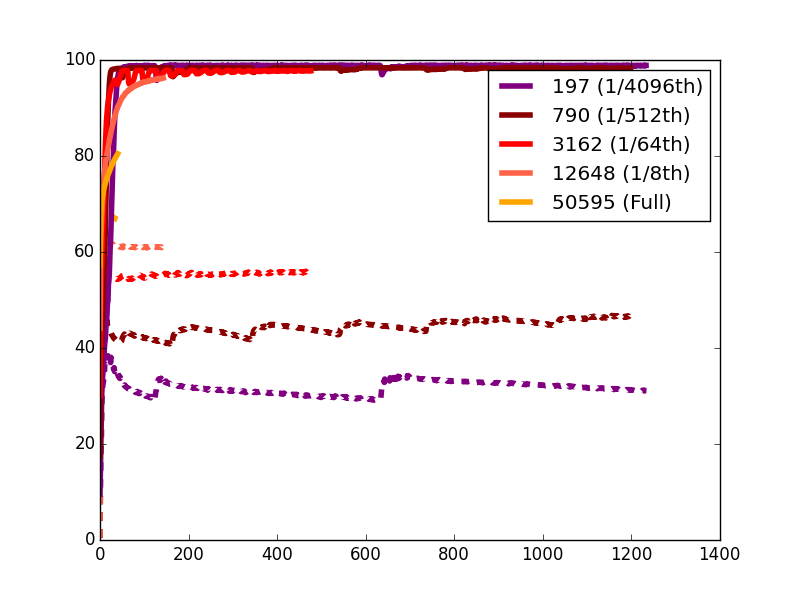}
  \end{subfigure}
  \begin{subfigure}[t]{0.47\textwidth}
    \includegraphics[width=\textwidth]{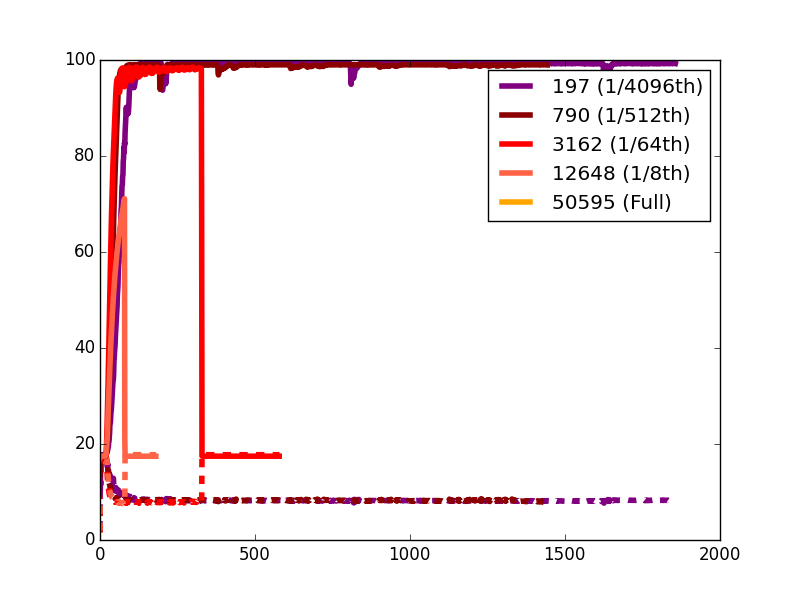}
  \end{subfigure}
    \begin{subfigure}[t]{0.47\textwidth}
    \includegraphics[width=\textwidth]{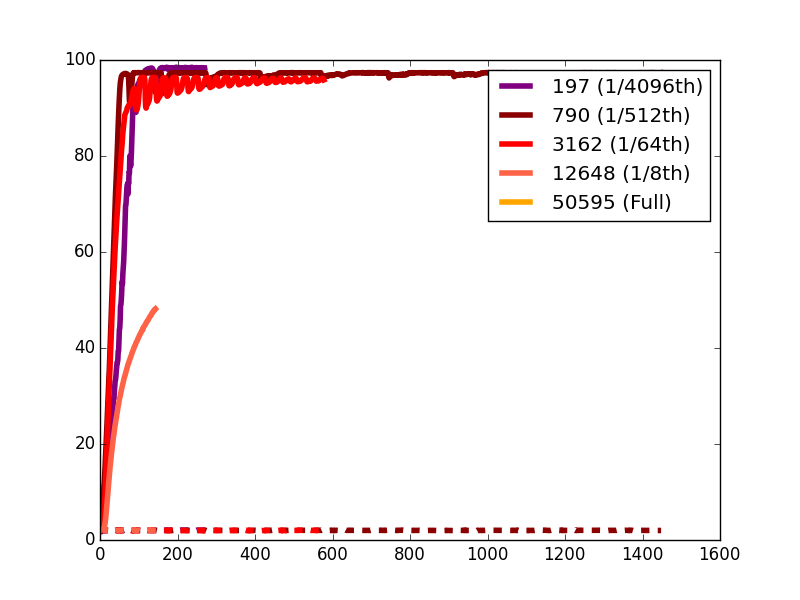}
  \end{subfigure}
  \caption{Training (solid) and validation (dotted) curves for differently sized subsets of the data in increasing powers of 8, for real data (top), 100\% X noise (middle), and 100\% Y noise (bottom) on cPTB.
Except on very small datasets (1/64th and less), optimization fails. on noise data.}
  \label{all_sizes_results}
\end{figure}

\begin{figure}[htb!]
  \centering
  \begin{subfigure}[h]{0.22\textwidth}
    \includegraphics[width=\textwidth]{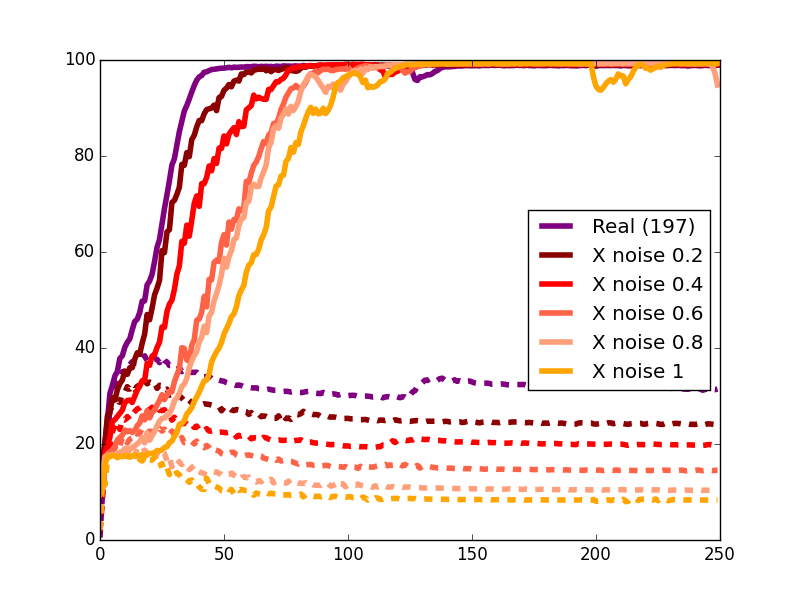}
  \end{subfigure}
  \begin{subfigure}[h]{0.22\textwidth}
    \includegraphics[width=\textwidth]{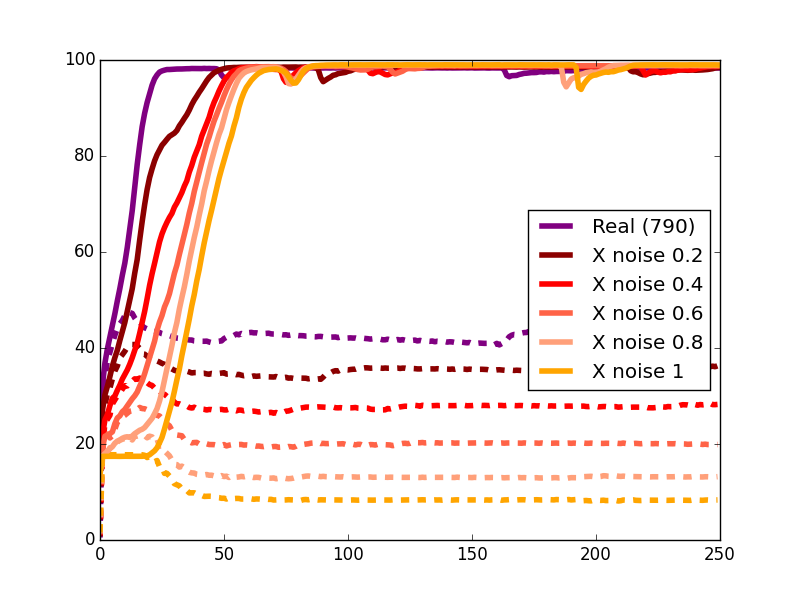}
  \end{subfigure}
    \begin{subfigure}[h]{0.22\textwidth}
    \includegraphics[width=\textwidth]{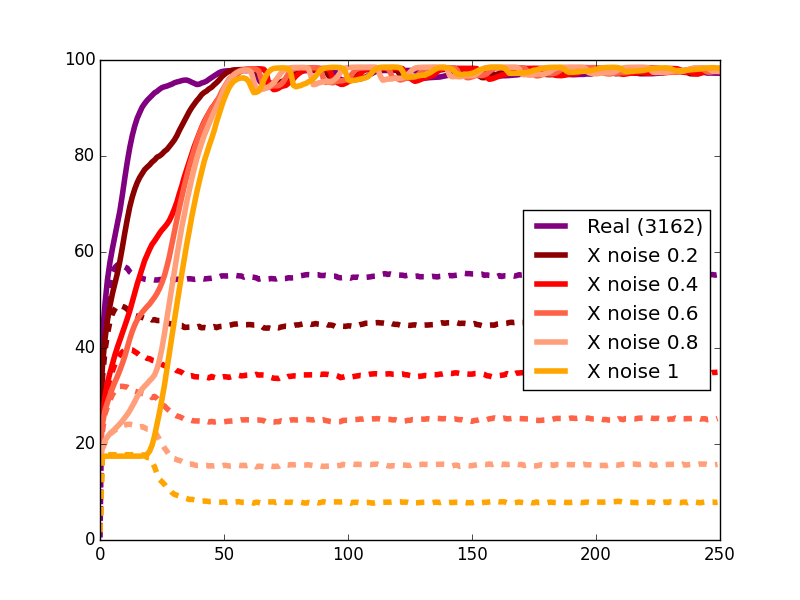}
  \end{subfigure}
  \begin{subfigure}[h]{0.22\textwidth}
    \includegraphics[width=\textwidth]{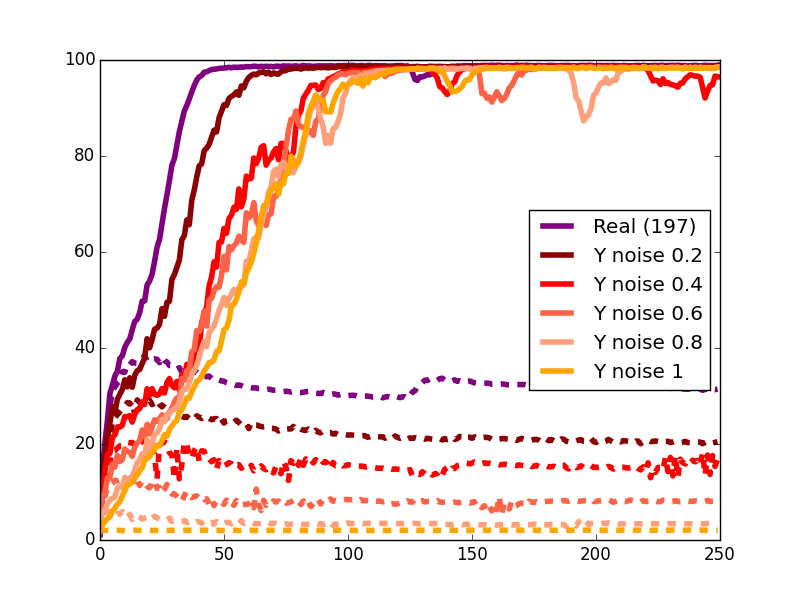}
  \end{subfigure}
    \begin{subfigure}[h]{0.22\textwidth}
    \includegraphics[width=\textwidth]{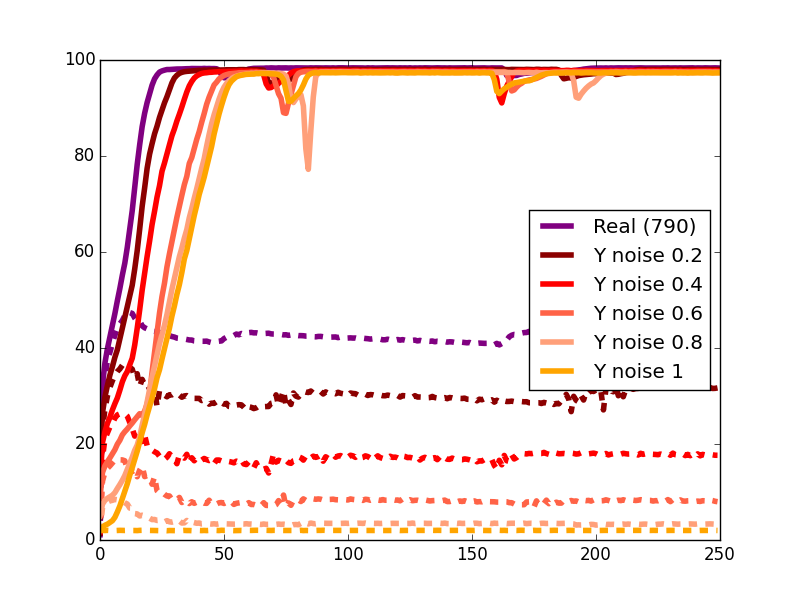}
  \end{subfigure}
  \begin{subfigure}[h]{0.22\textwidth}
    \includegraphics[width=\textwidth]{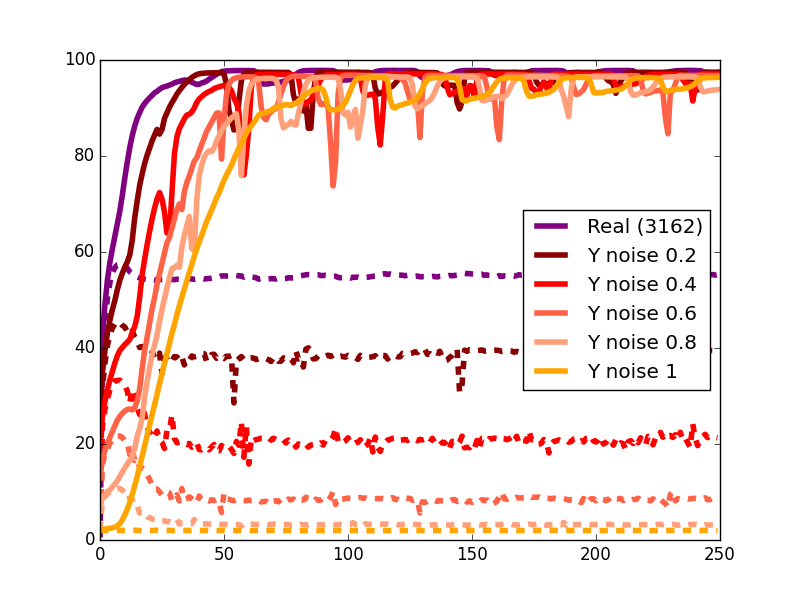}
  \end{subfigure}
  \caption{Training (solid) and validation (dotted) curves for different values of X noise (left column) and Y noise (right column) and differently sized subsets of the data (rows: [197, 790, and 3162] examples from top to bottom)}
  \label{noise_vals_results}
\end{figure}

\subsubsection{Differences in effective capacity}

Unlike typical feed-forward nets trained on common benchmark tasks like MNIST and CIFAR, typical RNNs trained on common benchmark tasks like sequential MNIST and Penn Treebank do not usually get close to 100\% training accuracy. We observe in Figure \ref{noise_vals_results} that it requires more effective capacity to fit noise.

Language models cannot hope to achieve 100\% accuracy on tasks, since the past sequence of characters does not determine the next character.
In a finite dataset, such ambiguities may not arise.
For instance, the Penn Treebank (PTB) corpus is composed of a single continuous example,\footnote{This is the way the corpus is formatted and used as an RNN benchmark. The full PTB dataset was developed primarily for parsing, and contains part-of-speech tags and other information that is discarded in this formatting.} and in principle, an RNN could be trained to memorize this string.
In practice, however, RNNs are trained on this task using truncated backprop through time (TBPTT), and the hidden state is often reset at regular intervals, for instance, after every 100 steps, or at the end of every sentence.

In future work, we plan to design tasks which do not suffer from such ambiguities.  
Designing synthetic tasks in such a way is straightforward, but removing ambiguities from cPTB requires more thought and is subject to design choices.
One simple idea which would ameliorate but not solve the problem is to measure performance on the end of training sequences only (where the model has more context).







\FloatBarrier

\subsection{Discussion}

Our results so far reveal some interesting differences between RNNs and feedforward nets.
The most significant is the difficulty of fitting random noise with RNNs.
The difficulty of fitting noise with RNNs suggests that these models might be inherently more reliant on learning patterns than feedforward models, and thus might generalize better.
Intuitively, the repeated application of the same transition operator might cause repeated subsequences within a given input to be processed in a similar way (although in principle, the difference in context could cause them to be treated entirely differently). 

On the other hand, 
the difficulty of fitting noise makes is attractive as challenge for optimization algorithms.
We would expect optimizers which can fit noise datasets to yield substantial improvements on training performance for real data as well.
This could be especially useful because RNN optimization is more difficult than feed-forward optimization, and remains a larger obstacle.
Still it is unclear whether this would improve generalization, since poor optimization may actually contribute to generalization in RNNs.
We hypothesize that better optimization would improve validation performance up to some point (dependent on other aspects of the problem and method), and degrade it thereafter.
If overfitting becomes a challenge, regularization might still allow one to reap the benefits of improved optimization.



\subsection{Future Work}

In future work we plan to 
evaluate RNN optimization on a broader range of synthetic tasks.
These experiments are designed to measure how sequence length, capacity, and dataset characteristics (categorical vs. real-vector values inputs/outputs; the number of examples, inputs, outputs) affect the ability of RNNs to fit a training set (e.g. with 100\% accuracy).

We propose using the following synthetic tasks:
\begin{enumerate}
\vspace{-2mm}\item {\bf next-step prediction}: given input and output sequences of length $N$, predict $y_n$ from $x_{1:n}, y_{1:n-1}$ (for all $1 \leq n \leq N$).  Language modeling is an instance of this setting where $y_n \doteq x_{n-1}$.
\vspace{-2mm}\item {\bf sequence-to-sequence (seq2seq)}: given an input $x \doteq x_{1:i}$ and output $y \doteq y_{1:j}$, predict $y_n$ from $x_{1:i}$ (for all $1 \leq n \leq N$).
\vspace{-2mm}\item {\bf vector-to-sequence / sequence-to-vector}: Special cases of seq2seq where the input / output (respectively) has only one time-step.  Sequential MNIST is an instance of sequence-to-vector. 
\end{enumerate}
\vspace{-2mm}
In every case, the inputs and targets can be independently chosen to be categorical or real-valued.
Combining all of the options yields a total of 16 synthetic task settings.
We note that previous work focused entirely on classification tasks on real-valued inputs, so changing the type of input and output for feedforward experiments would also be novel.

\Chapter{\uppercase{Regularization for recurrent neural networks}}\label{sec:zoneout}

\begin{quote}
    \singlespacing
    \cite{zoneout} David Krueger$^\dagger$, Tegan Maharaj$^\dagger$, J\'{a}nos Kr\'{a}mar, Mohammad Pezeshki, Nicolas Ballas, Nan Rosemary Ke, Anirudh Goyal, Yoshua Bengio, Aaron Courville, Christopher Pal. 2017. Zoneout: Regularizing RNNs by randomly preserving hidden activations. \textit{International Conference on Learning Representations (ICLR).}
    
    $\dagger$ denotes equal contribution.
\end{quote}

Dropout \cite{srivastava2014dropout} is a regularization method proposed in 2014 which saw widespread adoption for real-world tasks. It enabled good generalization even on relatively small datasets, and remains a cornerstone of practical deep learning. Many interpretations and rationales for Dropout's success have been explored, including prevention of co-adaptation of features, and the idea of training a pseudoensemble \cite{pseudo_ensembles} (an ensemble whose members are related in some way), i.e. that dropping certain activations amounts to subsampling the set of possible models that can be represented. 

As general and useful as Dropout is, it does not work well if used on the recurrent connections of an RNN, where it causes vanishing or exploding gradients. RNNs were and remain widely used for temporal tasks, and can often overfit on relatively small datasets, which are common in real-world applications such as health and environmental management. We sought a way to get similar properties as Dropout, but in a way that could be applied to RNNs.






This led to our work on Zoneout \cite{zoneout}, which is integrated into this chapter. In this work I contributed to discussions with Christopher Pal and David Krueger about regularizers with good inductive biases for temporal data, and the importance of noise (randomness). I discussed and agreed with the promise of the idea of random identity connections, as proposed by Anirudh Goyal in discussion with David.  

I checked initial versions of code written by David, and subsequently pair-programmed (with J\'{a}nos Kr\'{a}mar) a re-implementation of all initial experiments. I discussed initial results and oversaw a project merger and cooperation with a group (led by Mohammad Pezeshki) running similar experiments. I wrote additional scripts to enable large-scale experiments, created plots, and ran experiments which achieved state-of-the-art results on Penn-Treebank. I took over project management, and coordinated discussions, experimental progress, figure creation, and paper writing. I coded, ran, and plotted experiments examining gradient flow, with significant help from Nicolas Ballas. I contributed to related work research, re-implemented comparison baselines, wrote code and ran further experiments on other datasets, including Text8 where we also achieved state-of-the-art performance. I wrote a majority of the experimental sections of the paper, substantially revised and contributed to other sections, and created the main explanatory figure based on diagrams created by \cite{olah}. I contributed to responses in the reviewing phase, coded and ran a control experiment suggested by a reviewer (in appendix), and edited our camera-ready paper. Following a reproducibility checklist, I documented our code and created code snippets for open-source release, including commenting and re-running several experiments. I created our poster, and presented and co-presented our work at several workshops and other venues.

\vspace{0.5cm}
\textbf{Abstract}: \\
In this work, we propose zoneout, a novel method for regularizing RNNs.
At each timestep, zoneout stochastically forces some hidden units to maintain their previous values.
Like dropout, zoneout uses random noise to train a pseudo-ensemble, improving generalization.
But by preserving instead of dropping hidden units, gradient information and state information are more readily propagated through time, as in feedforward stochastic depth networks.
We perform an empirical investigation of various RNN regularizers, and find that zoneout gives significant performance improvements across tasks. We achieve competitive results with relatively simple models in character- and word-level language modelling on the Penn Treebank and Text8 datasets, and combining with recurrent batch normalization \cite{cooijmans2016recurrent} yields state-of-the-art results on permuted sequential MNIST.

\section{Introduction}

Regularizing neural nets can significantly improve performance, as indicated by the 
widespread 
use of early stopping, and success of regularization methods such as dropout and its recurrent variants \cite{hinton2012improving,srivastava2014dropout,zaremba2014recurrent,yarvin}.
In this paper, we address the issue of regularization in recurrent neural networks (RNNs) with a novel method called \textbf{zoneout}.

RNNs sequentially construct fixed-length representations of arbitrary-length sequences by folding new observations into their hidden state using an input-dependent transition operator.
The repeated application of the same transition operator at the different time steps of the sequence, however, can make the dynamics of an RNN sensitive to minor perturbations in the hidden state;
the transition dynamics can magnify components of these perturbations exponentially.
Zoneout aims to improve RNNs' robustness to perturbations in the hidden state in order to regularize transition dynamics.

Like dropout, zoneout injects noise during training.
But instead of setting some units' activations to 0 as in dropout, zoneout randomly replaces some units' activations with their activations from the previous timestep. 
As in dropout, we use the expectation of the random noise at test time. 
This results in a simple regularization approach which can be applied through time for any RNN architecture, and can be conceptually extended to any model whose state varies over time.
Compared with dropout, zoneout is appealing because it preserves information flow forwards and backwards through the network.
This helps combat the vanishing gradient problem \cite{hochreiter1991untersuchungen,bengio1994learning}, as we observe experimentally. 
We also empirically evaluate zoneout on classification using the permuted sequential MNIST dataset, and on language modelling using the Penn Treebank and Text8 datasets, demonstrating competitive or state of the art performance across tasks.
In particular, we show that zoneout performs competitively with other proposed regularization methods for RNNs, including recently-proposed dropout variants. 
Code for replicating all experiments can be found at: \texttt{http://github.com/teganmaharaj/zoneout}

\section{Related work}

\subsection{Relationship to dropout}
Zoneout can be seen as a selective application of dropout to some of the nodes in a modified computational graph, as shown in Figure~\ref{fig:zoneout_as_dropout}.  
In zoneout, instead of dropping out (being set to 0), units \emph{zone out} and are set to their previous value ($h_t = h_{t-1}$).
Zoneout, like dropout, can be viewed as a way to train a pseudo-ensemble \cite{pseudo_ensembles}, injecting noise using a stochastic ``identity-mask'' rather than a zero-mask. 
We conjecture that identity-masking is more appropriate for RNNs, since it makes it easier for the network to preserve information from previous timesteps going forward, and facilitates, rather than  hinders, the flow of gradient information going backward, as we demonstrate experimentally. 

\begin{figure}[htb]
\centering
\includegraphics[scale=1.1]{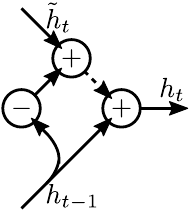}
\caption{
Zoneout as a special case of dropout; $\tilde h_t$ is the unit $h$'s hidden activation for the next time step (if not zoned out).
Zoneout can be seen as applying dropout on the hidden state delta, $\tilde h_t - h_{t-1}$.
When this update is dropped out (represented by the dashed line), $h_t$ becomes $h_{t-1}$. 
}
\label{fig:zoneout_as_dropout}
\end{figure}

\subsection{Dropout in RNNs}
Initially successful applications of dropout in RNNs \cite{pham13, zaremba2014recurrent} only applied 
dropout to feed-forward connections (``up the stack''), and not recurrent connections (``forward through time''), but several recent works \cite{elephant,rnnDrop,yarvin} propose methods that are not limited in this way.
\cite{rnn_fast_dropout} successfully apply fast dropout \cite{fast_dropout}, a deterministic approximation of dropout, to RNNs.

\cite{elephant} apply \textbf{recurrent dropout} to the {\it updates} to LSTM memory cells (or GRU states), i.e.\ they drop out the input/update gate in LSTM/GRU.
Like zoneout, their approach prevents the loss of long-term memories built up in the states/cells of GRUs/LSTMS, but zoneout does this by preserving units' activations \emph{exactly}.
This difference is most salient when zoning out the hidden states (not the memory cells) of an LSTM, for which there is no analogue in recurrent dropout.
Whereas saturated output gates or output nonlinearities would cause recurrent dropout to suffer from vanishing gradients \cite{bengio1994learning}, zoned-out units still propagate gradients effectively in this situation.
Furthermore, while the recurrent dropout method is specific to LSTMs and GRUs, zoneout generalizes to any model that sequentially builds distributed representations of its input, including vanilla RNNs.

Also motivated by preventing memory loss, \cite{rnnDrop} propose \textbf{rnnDrop}. This technique amounts to using the same dropout mask at every timestep, which the authors show results in improved performance on speech recognition in their experiments.
\cite{elephant} show, however, that past states' influence vanishes exponentially as a function of dropout probability when taking the expectation at test time in rnnDrop; 
this is problematic for tasks involving longer-term dependencies.

\cite{yarvin} propose another technique which uses the same mask at each timestep.
Motivated by variational inference, they drop out the rows of weight matrices in the input and output embeddings and LSTM gates, instead of dropping units' activations.
The proposed \textbf{variational RNN} technique achieves single-model state-of-the-art test perplexity of $73.4$ on word-level language modelling of Penn Treebank.

\subsection{Relationship to Stochastic Depth}
Zoneout can also be viewed as a per-unit version of \textbf{stochastic depth} \cite{stochastic_depth}, which  randomly drops entire layers of feed-forward residual networks (ResNets \cite{resnet}).
This is equivalent to zoning out all of the units of a layer at the same time.
In a typical RNN, there is a new input at each timestep, causing issues for a naive implementation of stochastic depth.
Zoning out an entire layer in an RNN means the input at the corresponding timestep is completely ignored, whereas zoning out individual units allows the RNN to take each element of its input sequence into account.
We also found that using residual connections in recurrent nets led to instability, presumably due to the parameter sharing in RNNs.
Concurrent with our work, \cite{swapout} propose zoneout for ResNets, calling it {\bf SkipForward}.
In their experiments, zoneout is outperformed by stochastic depth, dropout, and their proposed {\bf Swapout} technique, which randomly drops either or both of the identity or residual connections.
Unlike \cite{swapout}, we apply zoneout to RNNs, and find it outperforms stochastic depth and recurrent dropout.

\subsection{Selectively updating hidden units}
Like zoneout, {\bf clockwork RNNs}~\cite{koutnik2014clockwork} and {\bf hierarchical RNNs}~\cite{hierarchical_rnn} update only some units' activations at every timestep, but their updates are periodic, whereas zoneout's are stochastic.  
Inspired by clockwork RNNs, we experimented with zoneout variants that target different update rates or schedules for different units, but did not find any performance benefit. 
\textbf{Hierarchical multiscale LSTMs} \cite{hmrnn} learn update probabilities for different units using the straight-through estimator \cite{straighthrough1,straightthrough2}, and combined with recently-proposed Layer Normalization \cite{layernorm}, achieve competitive results on a variety of tasks.
As the authors note, their method can be interpreted as an input-dependent form of adaptive zoneout. 

In recent work, \cite{hypernets} use a hypernetwork to dynamically rescale the row-weights of a primary LSTM network, achieving state-of-the-art 1.21 BPC on character-level Penn Treebank when combined with layer normalization \cite{layernorm} in a two-layer network. 
This scaling can be viewed as an adaptive, differentiable version of the variational LSTM \cite{yarvin}, and could similarly be used to create an adaptive, differentiable version of zoneout.
Very recent work conditions zoneout probabilities on suprisal (a measure of the discrepancy between the predicted and actual state), and sets a new state of the art on enwik8 \cite{suprisalzoneout}.

\section{Zoneout and preliminaries}

We now explain zoneout in full detail, and compare with other forms of dropout in RNNs. We start by reviewing recurrent neural networks (RNNs).

\subsection{Recurrent Neural Networks}

Recurrent neural networks process data $x_1, x_2, \dots, x_T$ sequentially, constructing a corresponding sequence of representations, $h_1, h_2, \dots, h_T$.  
Each hidden state is trained (implicitly) to remember and emphasize all task-relevant aspects of the preceding inputs, and to incorporate new inputs via a  transition operator, $\mathcal{T}$, which converts the present hidden state and input into a new hidden state: $h_{t} = \mathcal{T} (h_{t-1}, x_t)$.
Zoneout modifies these dynamics by mixing the original transition operator $\mathcal{\tilde T}$ with the identity operator (as opposed to the null operator used in dropout), according to a vector of Bernoulli masks, $d_t$:
\begin{align*}
\mbox{Zoneout:}&&\mathcal{T}&=d_t\odot\mathcal{\tilde T}+(1-d_t)\odot 1
&\mbox{Dropout:}&&
\mathcal{T}&=d_t\odot\mathcal{\tilde T}+(1-d_t)\odot 0 
\end{align*}


\subsection{Long short-term memory}

In long short-term memory RNNs (LSTMs) \cite{hochreiter1997long}, the hidden state is divided into memory cell $c_t$, intended for internal long-term storage, and hidden state $h_t$, used as a transient representation of state at timestep $t$. 
In the most widely used formulation of an LSTM \cite{LSTM_forgetgate}, $c_t$ and $h_t$ are computed via a set of four ``gates'', including the forget gate, $f_t$, which directly connects  $c_t$ to the memories of the previous timestep  $c_{t-1}$, via an element-wise multiplication.  
Large values of the forget gate cause the cell to remember most (not all) of its previous value.  
The other gates control the flow of information in ($i_t, g_t$) and out ($o_t$) of the cell.
Each gate has a weight matrix and bias vector; for example the forget gate has $W_{xf}$, $W_{hf}$, and $b_f$. 
For brevity, we will write these as $W_x,W_h,b$.

An LSTM is defined as follows:
\begin{align*}
i_t, f_t, o_t &= \sigma(W_{x}x_t+W_{h}h_{t-1}+b)\\
g_t &= \tanh(W_{xg}x_t+W_{hg}h_{t-1}+b_g)\\
c_t &= f_t \odot c_{t-1} + i_t \odot g_t \\
h_t &= o_t \odot \tanh(c_t)
\end{align*}

A naive application of dropout in LSTMs would zero-mask either or both of the memory cells and hidden states, without changing the computation of the gates ($i,f,o,g$). Dropping memory cells, for example, changes the computation of $c_t$ as follows:
\begin{align*}
c_t &= d_t\odot (f_t \odot c_{t-1} + i_t \odot g_t)
\end{align*}

Alternatives abound, however; masks can be applied to any subset of the gates, cells, and states.  \cite{elephant}, for instance, zero-mask the input gate:
\begin{align*}
c_t &= (f_t \odot c_{t-1} + d_t \odot i_t \odot g_t)
\end{align*}

When the input gate is masked like this, there is no additive contribution from the input or hidden state, and the value of the memory cell simply decays according to the forget gate.


\begin{figure}[htb]
\begin{subfigure}{0.5\textwidth}
\includegraphics[scale=1.1]{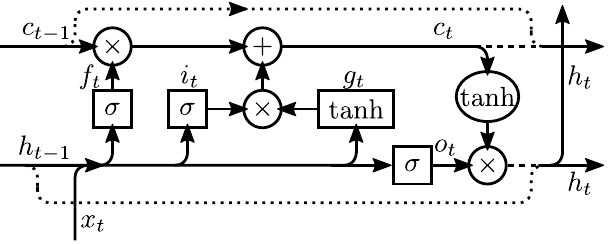}
\caption{}
\end{subfigure}
\begin{subfigure}{0.5\textwidth}
\raisebox{12.1pt}{\includegraphics[scale=1.1]{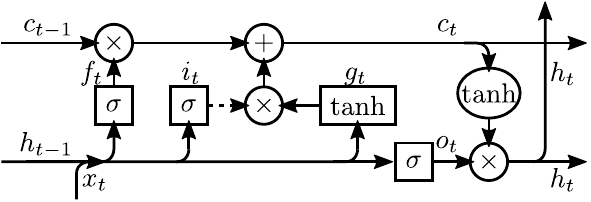}}
\caption{}
\end{subfigure}
\caption{(a) Zoneout, vs (b) the recurrent dropout strategy of \cite{elephant}  in an LSTM.
Dashed lines are zero-masked; in zoneout, the corresponding dotted lines are masked with the corresponding opposite zero-mask. Rectangular nodes are embedding layers.}
\label{zoneout_vs_elephant}
\end{figure}

In \textbf{zoneout}, the values of the hidden state and memory cell randomly either maintain their previous value or are updated as usual.  This introduces stochastic identity connections between subsequent time steps:
\begin{align*}
c_t &= d^c_t\odot c_{t-1} + (1-d^c_t)\odot \big(f_t \odot c_{t-1} + i_t \odot g_t\big)\\
h_t &= d^h_t\odot h_{t-1} + (1-d^h_t)\odot \big(o_t \odot \tanh\big(f_t \odot c_{t-1} + i_t \odot g_t\big)\big)
\end{align*}

We usually use different zoneout masks for cells and hiddens. 
We also experiment with a variant of recurrent dropout that reuses the input dropout mask to zoneout the corresponding output gates:
\begin{align*}
c_t &= (f_t \odot c_{t-1} + d_t \odot i_t \odot g_t) \\
h_t &= ((1 - d_t) \odot o_t + d_t \odot o_{t-1})\odot \tanh(c_t)
\end{align*}
The motivation for this variant is to prevent the network from being forced (by the output gate) to expose a memory cell which has not been updated, and hence may contain misleading information.




\section{Experiments and Discussion}

We evaluate zoneout's performance on the following tasks: (1) Character-level language modelling on the Penn Treebank corpus  \cite{marcus1993building}; (2) Word-level language modelling on the Penn Treebank corpus \cite{marcus1993building};  (3) Character-level language modelling on the Text8 corpus~\cite{Text8}; (4) Classification of hand-written digits on permuted sequential MNIST ($p$MNIST) \cite{le2015simple}.
We also investigate the gradient flow to past hidden states, using $p$MNIST.


\subsection{Penn Treebank Language Modelling Dataset}
The Penn Treebank language model corpus contains 1 million words. 
The model is trained to predict the next word (evaluated on perplexity) or character (evaluated on BPC: bits per character) in a sequence.
\footnote{
These metrics are deterministic functions of negative log-likelihood (NLL).  Specifically, perplexity is exponentiated NLL, and BPC (entropy) is NLL divided by the natural logarithm of 2.
}

\subsubsection{Character-level}
For the character-level task, we train networks with one layer of 1000 hidden units.
We train LSTMs with a learning rate of 0.002 on overlapping sequences of 100 in batches of 32, optimize using Adam, and clip gradients with threshold 1.  These settings match those used in \cite{cooijmans2016recurrent}. We also train GRUs and tanh-RNNs with the same parameters as above, except sequences are non-overlapping and we use learning rates of 0.001, and 0.0003 for GRUs and tanh-RNNs respectively.
Small values (0.1, 0.05) of zoneout significantly improve generalization performance for all three models. 
Intriguingly, we find zoneout increases training time for GRU and tanh-RNN, but \emph{decreases} training time for LSTMs.

We focus our investigation on LSTM units, where the dynamics of zoning out states, cells, or both provide interesting insight into zoneout's behaviour. 
Figure~\ref{charchar} shows our exploration of zoneout in LSTMs, for various zoneout probabilities of cells and/or hiddens. Zoneout on cells with probability 0.5 or zoneout on states with probability 0.05 both outperform the best-performing recurrent dropout ($p=0.25$). Combining $z_c=0.5$ and $z_h=0.05$ leads to our best-performing model, which achieves 1.27 BPC, competitive with recent state-of-the-art set by \cite{hypernets}.
We compare zoneout to recurrent dropout (for $p \in \{0.05, 0.2, 0.25, 0.5, 0.7\}$), weight noise ($\sigma=0.075$), norm stabilizer ($\beta=50$) \cite{norm_stabilizer}, and explore stochastic depth \cite{stochastic_depth} in a recurrent setting (analagous to zoning out an entire timestep). We also tried a shared-mask variant of zoneout as used in $p$MNIST experiments, where the same mask is used for both cells and hiddens. Neither stochastic depth or shared-mask zoneout performed as well as separate masks, sampled per unit. Figure~\ref{charchar} shows the best performance achieved with each regularizer, as well as an unregularized LSTM baseline.
Results are reported in Table \ref{tab:all_results}, and learning curves shown in Figure \ref{char_and_text8_results}.

Low zoneout probabilities (0.05-0.25) also improve over baseline in GRUs and tanh-RNNs, reducing BPC from 1.53 to 1.41 for GRU and 1.67 to 1.52 for tanh-RNN.
Similarly, low zoneout probabilities work best on the hidden states of LSTMs. For memory cells in LSTMs, however, higher probabilities (around 0.5) work well, perhaps because large forget-gate values approximate the effect of cells zoning out.
We conjecture that best performance is achieved with zoneout LSTMs because of the stability of having both state and cell. The probability that both will be zoned out is very low, but having one or the other zoned out carries information from the previous timestep forward, while having the other react 'normally' to new information. 

{
\begin{figure}[htb!]
  \centering
  \begin{subfigure}[t]{0.49\textwidth}
    \includegraphics[width=\textwidth]{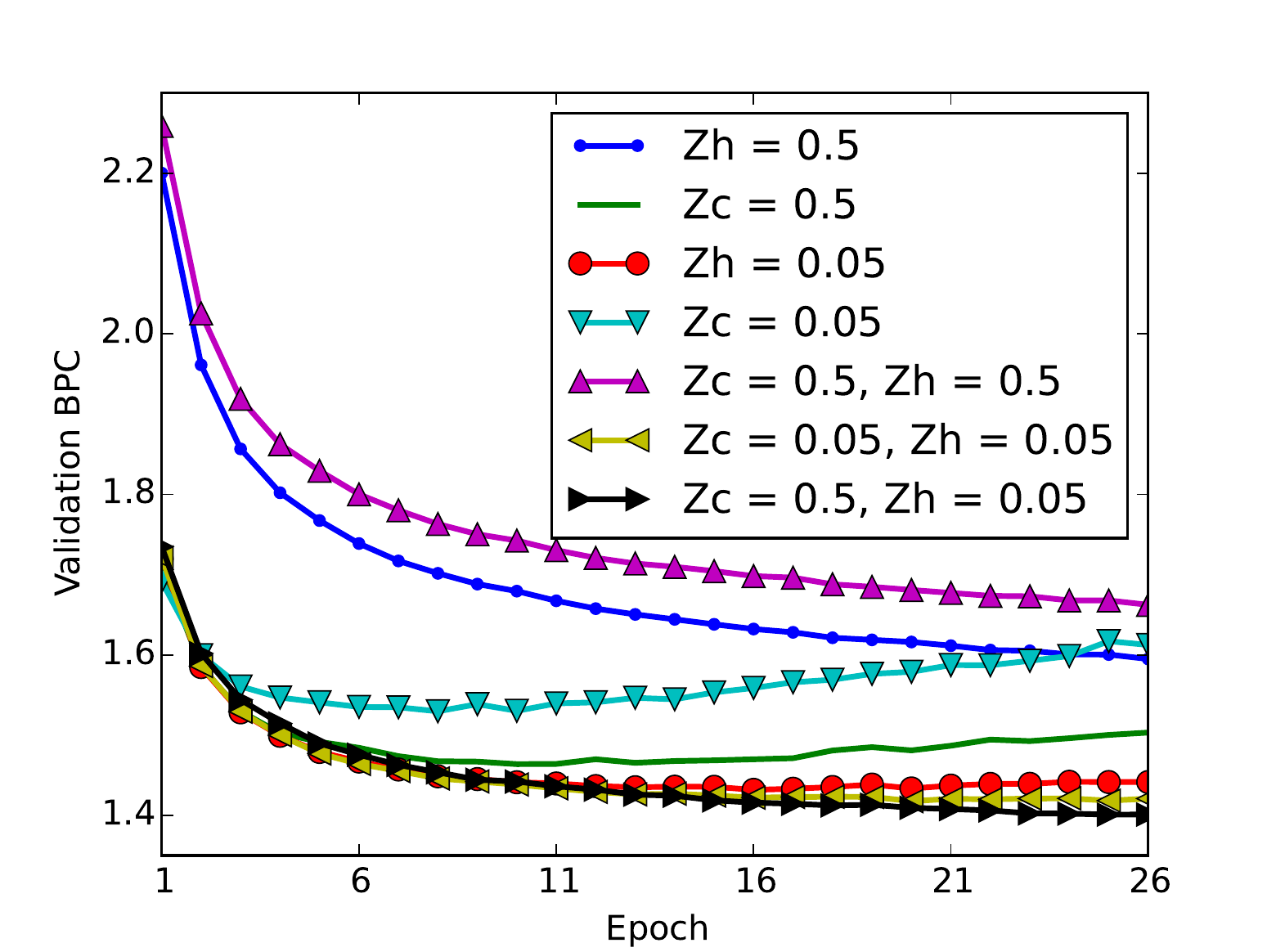}
  \end{subfigure}
  \begin{subfigure}[t]{0.49\textwidth}
 	    \includegraphics[width=\textwidth]{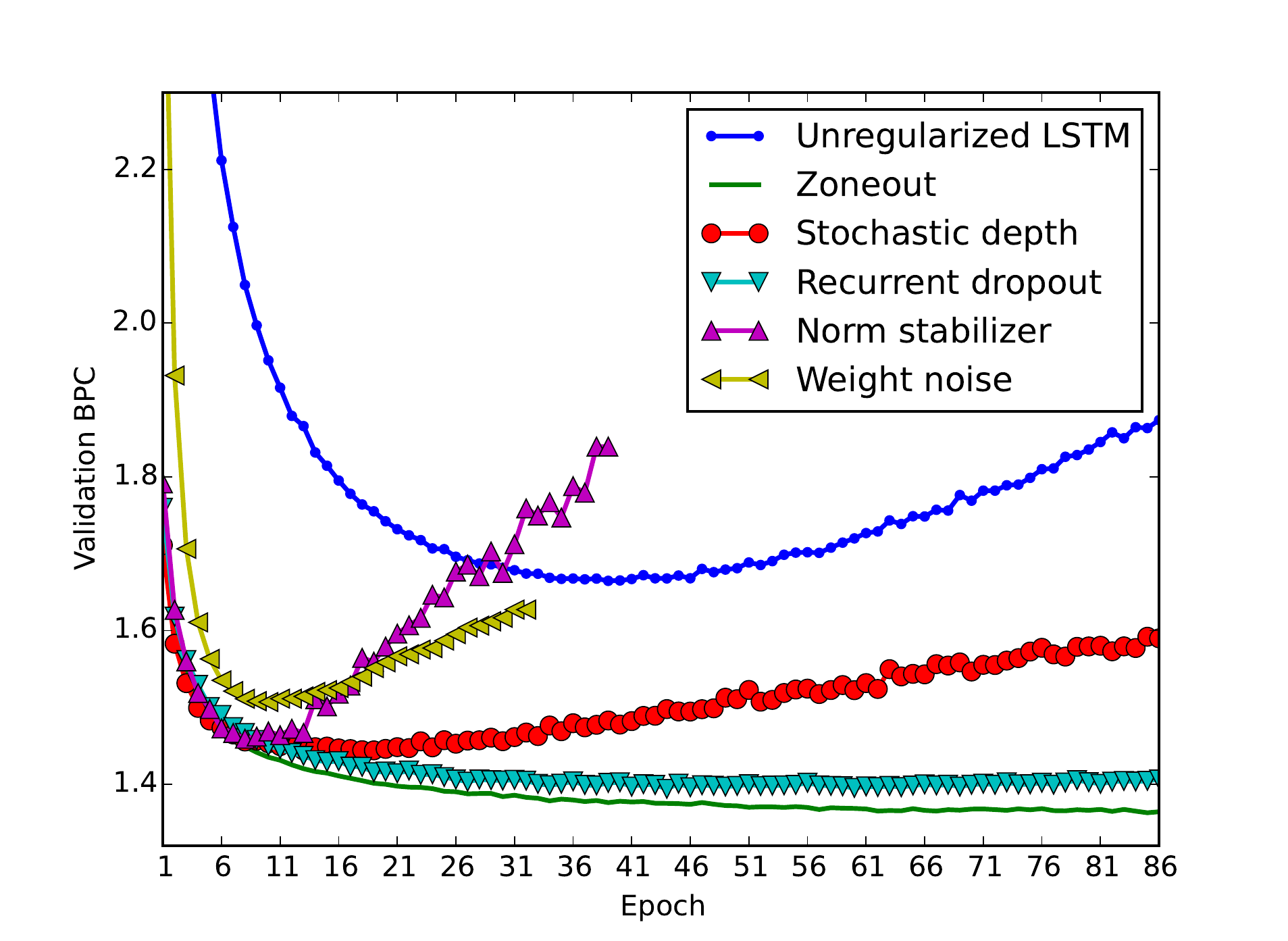}
  \end{subfigure}
  \caption{Validation BPC (bits per character) on Character-level Penn Treebank, for different probabilities of zoneout on cells $z_c$ and hidden states $z_h$ (left), and comparison of an unregularized LSTM, zoneout $z_c=0.5, z_h=0.05$, stochastic depth zoneout $z=0.05$, recurrent dropout $p=0.25$, norm stabilizer $\beta=50$, and weight noise $\sigma=0.075$ (right).}
  \label{charchar}
  \vspace{-3mm}
\end{figure}
}

{
\begin{figure}[htb!]
  \centering
  \begin{subfigure}[t]{0.51\textwidth}
    \includegraphics[width=\textwidth]{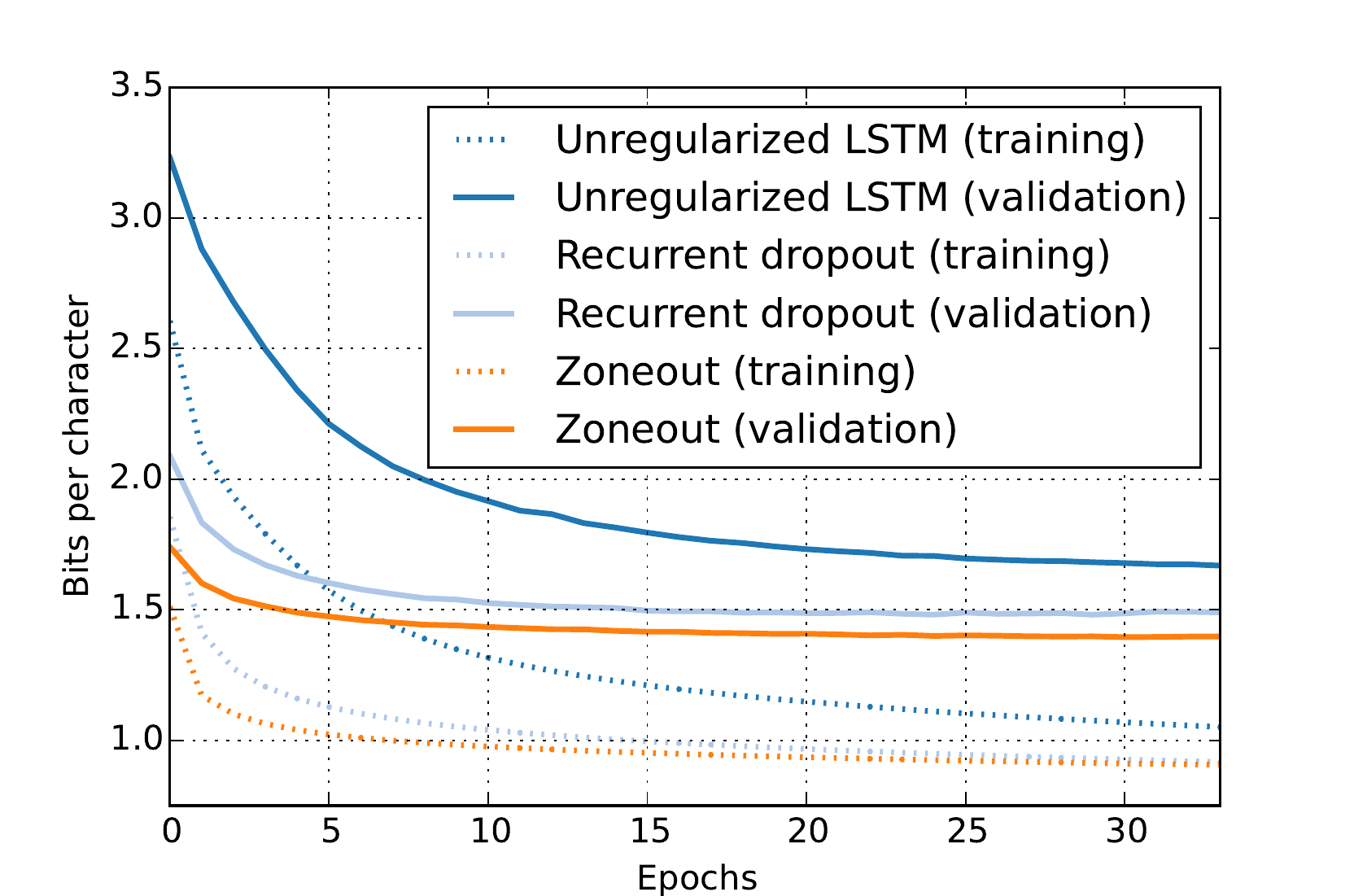}
  \end{subfigure}
  \begin{subfigure}[t]{0.48\textwidth}
    \includegraphics[width=\textwidth]{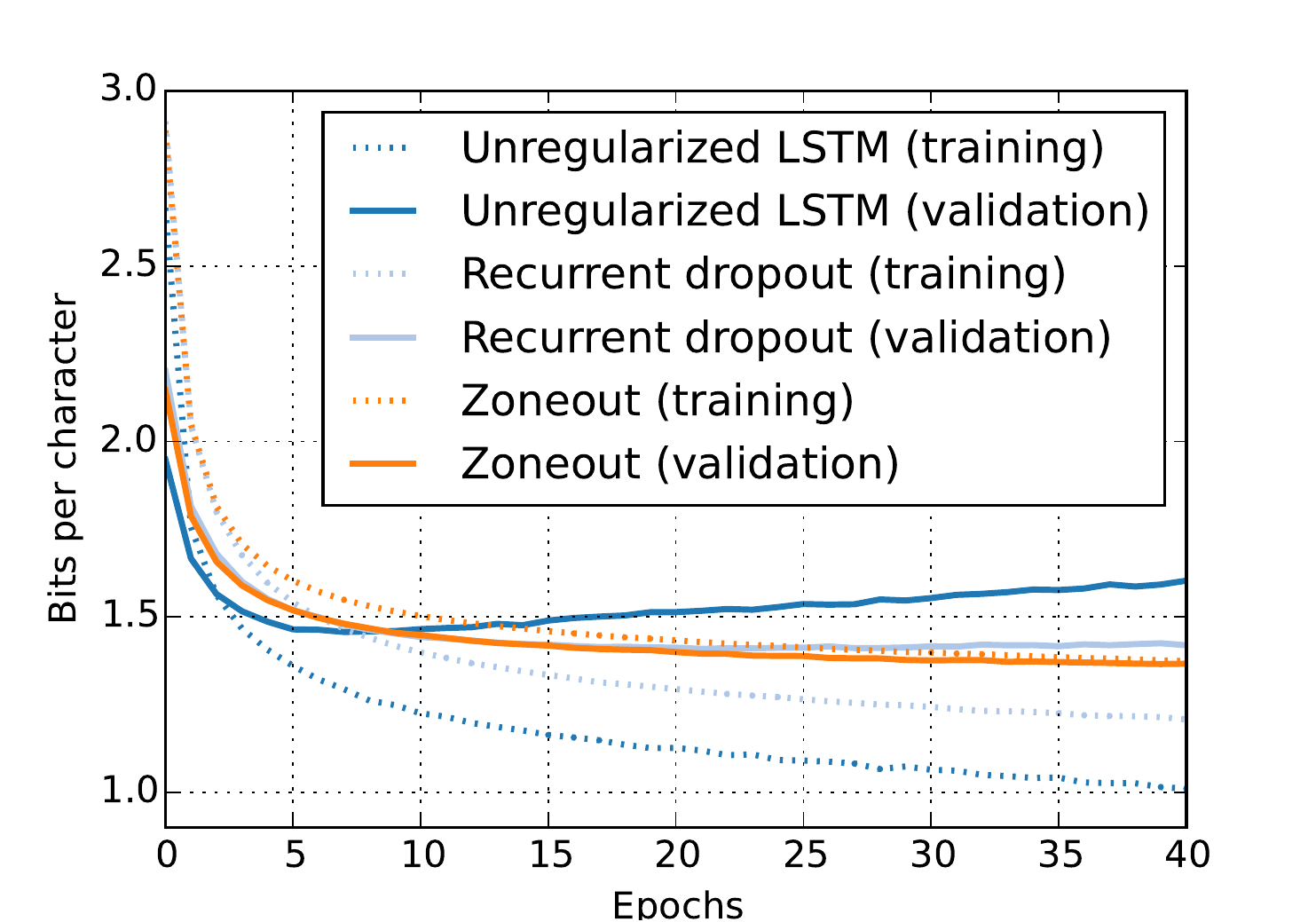}
  \end{subfigure}
  \caption{Training and validation bits-per-character (BPC) comparing LSTM regularization methods on character-level Penn Treebank (left) and Text8. (right)}
  \label{char_and_text8_results}
\end{figure}
}


\vspace{-4mm}
\subsubsection{Word-level}
For the word-level task, we replicate settings from \cite{zaremba2014recurrent}'s best single-model performance. This network has 2 layers of 1500 units, with weights initialized uniformly [-0.04, +0.04]. The model is trained for 14 epochs with learning rate 1, after which the learning rate is reduced by a factor of 1.15 after each epoch. Gradient norms are clipped at 10.  

With no dropout on the non-recurrent connections (i.e. zoneout as the only regularization), we do not achieve competitive results. We did not perform any search over models, and conjecture that the large model size requires regularization of the feed-forward connections. Adding zoneout ($z_c=0.25$ and $z_h=0.025$) on the recurrent connections to the model optimized for dropout on the non-recurrent connections however, we are able to improve test perplexity from 78.4 to 77.4.
We report the best performance achieved with a given technique in Table~\ref{tab:all_results}.

\subsection{Text8}
Enwik8 is a corpus made from the first $10^9$ bytes of Wikipedia dumped on Mar. 3, 2006. Text8 is a ``clean text'' version of this corpus; with html tags removed, numbers spelled out, symbols converted to spaces, all lower-cased. Both datasets were created and are hosted by \cite{Text8}. 

We use a single-layer network of 2000 units, initialized orthogonally, with batch size 128, learning rate 0.001, and sequence length 180. We optimize with Adam \cite{kingma2014adam}, clip gradients to a maximum norm of 1 \cite{pascanu2013construct}, and use early stopping, again matching the settings of \cite{cooijmans2016recurrent}. 
Results are reported in Table~\ref{tab:all_results}, and Figure~\ref{char_and_text8_results} shows training and validation learning curves for zoneout ($z_c=0.5, z_h=0.05$) compared to an unregularized LSTM and to recurrent dropout.

\subsection{Permuted sequential MNIST}

In sequential MNIST, pixels of an image representing a number [0-9] are presented one at a time, left to right, top to bottom. The task is to classify the number shown in the image. In $p$MNIST , the pixels are presented in a (fixed) random order.

We compare recurrent dropout and zoneout to an unregularized LSTM baseline.
All models have a single layer of 100 units, and are trained for 150 epochs using RMSProp \cite{rmsprop} with a decay rate of 0.5 for the moving average of gradient norms. 
The learning rate is set to 0.001 and the gradients are clipped to a maximum norm of 1 \cite{pascanu2013construct}.

As shown in Figure~\ref{fig:mnist_results} and Table~\ref{tab:mnist_results}, zoneout gives a significant performance boost compared to the LSTM baseline and outperforms recurrent dropout~\cite{elephant}, although recurrent batch normalization~\cite{cooijmans2016recurrent} outperforms all three. 
However, by adding zoneout to the recurrent batch normalized LSTM, we achieve state of the art performance. For this setting, the zoneout mask is shared between cells and states, and the recurrent dropout probability and zoneout probabilities are both set to 0.15.

\FloatBarrier

\begin{table*}[!htb]
\centering
\caption{Validation and test results of different models on the three language modelling tasks. Results are reported for the best-performing settings. Performance on Char-PTB and Text8 is measured in bits-per-character (BPC); Word-PTB is measured in perplexity.  For Char-PTB and Text8 all models are 1-layer unless otherwise noted; for Word-PTB all models are 2-layer. Results above the line are from our own implementation and experiments. Models below the line are: NR-dropout (non-recurrent dropout), V-Dropout (variational dropout), RBN (recurrent batchnorm), H-LSTM+LN (HyperLSTM + LayerNorm), 3-HM-LSTM+LN (3-layer Hierarchical Multiscale LSTM + LayerNorm).}
  \begin{tabular}{ccccccc}
  \toprule
  & \multicolumn{2}{c}{\bf Char-PTB}
  & \multicolumn{2}{c}{\bf Word-PTB}
  & \multicolumn{2}{c}{\bf Text8}\\
  \midrule
  {\bf Model} & {\bf Valid } & {\bf Test}& {\bf Valid } & {\bf Test}& {\bf Valid } & {\bf Test} \\ \midrule
Unregularized LSTM                     & 1.466 & 1.356          & 120.7 & 114.5          & 1.396 & 1.408          \\
Weight noise                           & 1.507 & 1.344          & --    & --             & 1.356 & 1.367          \\
Norm stabilizer                        & 1.459 & 1.352          & --    & --             & 1.382 & 1.398          \\
Stochastic depth                       & 1.432 & 1.343          & --    & --             & 1.337 & 1.343          \\
Recurrent dropout                      & 1.396 & 1.286          & ~91.6 & ~87.0          & 1.386 & 1.401          \\
Zoneout                                & 1.362 & 1.252          & ~81.4 & ~77.4          & 1.331 & 1.336          \\
\midrule
NR-dropout \cite{zaremba2014recurrent} & --    & --             & ~82.2 & ~ 78.4         & --    & --             \\
V-dropout \cite{yarvin}                & --    & --             & --    & ~\textbf{73.4} & --    & --             \\
RBN \cite{cooijmans2016recurrent}      & --    & 1.32~          & --    & --             & --    & 1.36~          \\
H-LSTM + LN \cite{hypernets}           & 1.281 & 1.250          & --    & --             & --    & --             \\
3-HM-LSTM + LN \cite{hmrnn}            & --    & \textbf{1.24}~ & --    & --             & --    & \textbf{1.29}~ \\
  \bottomrule
 \end{tabular}
\label{tab:char_explore}
\label{tab:all_results}
\end{table*}

\begin{table*}[!htb]
\centering
\caption{Error rates on the pMNIST digit classification task.  Zoneout outperforms recurrent dropout, and sets state of the art when combined with recurrent batch normalization.}
  \begin{tabular}{ccccccc}
  \toprule
 
  {\bf Model} & {\bf Valid } & {\bf Test} \\
  \midrule
Unregularized LSTM & 0.092 & 0.102 &\\
Recurrent dropout $p=0.5$& 0.083 & 0.075 \\
Zoneout $z_c=z_h=0.15$ & 0.063 & 0.069 \\
Recurrent batchnorm & - & 0.046 \\
Recurrent batchnorm \& Zoneout $z_c=z_h=0.15$ & 0.045 & \textbf{0.041}\\
  \bottomrule
  \end{tabular}
\label{tab:mnist_results}
\end{table*}

\begin{figure}[t]
  \centering
  \includegraphics[width=.45\textwidth]{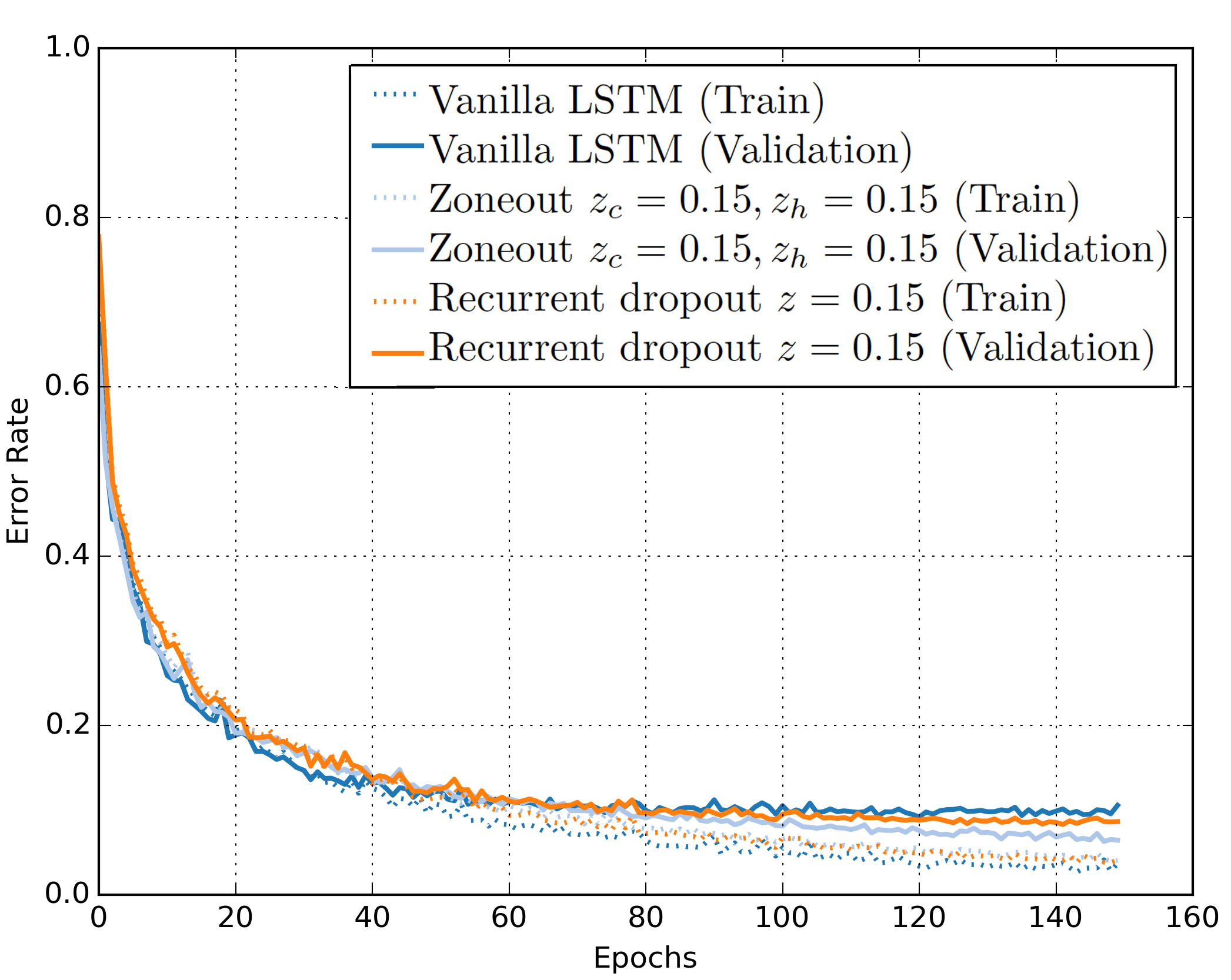}
  \caption{Training and validation error rates for an unregularized LSTM, recurrent dropout, and zoneout on the task of permuted sequential MNIST digit classification.}
  \label{fig:mnist_results}
\end{figure}

\subsection{Gradient flow}
We investigate the hypothesis that identity connections introduced by zoneout facilitate gradient flow to earlier timesteps.
Vanishing gradients are a perennial issue in RNNs. 
As effective as many techniques are for mitigating vanishing gradients (notably the LSTM architecture \cite{hochreiter1997long}), we can always imagine a longer sequence to train on, or a longer-term dependence we want to capture. 

We compare gradient flow in an unregularized LSTM to zoning out (stochastic identity-mapping) and dropping out (stochastic zero-mapping) the recurrent connections after one epoch of training on $p$MNIST. 
We compute the average gradient norms $\|{\frac{\partial L}{\partial c_t}}\|$ of loss $L$ with respect to cell activations $c_t$ at each timestep $t$, and for each method, normalize the average gradient norms by the sum of average gradient norms for all timesteps.

Figure~\ref{fig:grads_norm} shows that zoneout propagates gradient information to early timesteps much more effectively than dropout on the recurrent connections, and even more effectively than an unregularized LSTM. The same effect was observed for hidden states $h_t$.


 \begin{figure}[b]
  \centering
  \includegraphics[width=0.45\textwidth]{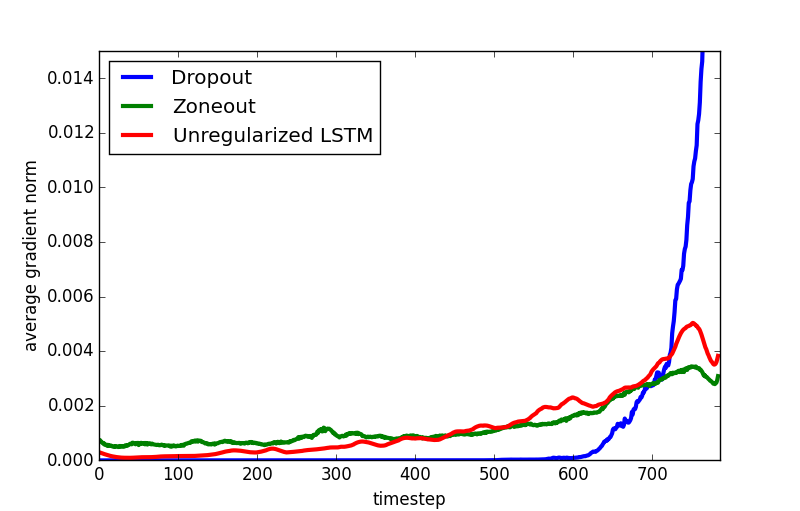}
  \caption{Normalized ${\sum\|\frac{\partial L}{\partial c_t}}\|$ of loss $L$ with respect to cell activations $c_t$ at each timestep $t$ for zoneout ($z_c=0.5$), dropout ($z_c=0.5$), and an unregularized LSTM on one epoch of $p$MNIST}.
  \label{fig:grads_norm}
 \end{figure}

\FloatBarrier


%

\subsection{Static identity connections experiment}
This experiment was suggested by AnonReviewer2 during the ICLR review process with the goal of disentangling the effects zoneout has (1) through noise injection in the training process and (2) through identity connections. Based on these results, we observe that noise injection is essential for obtaining the regularization benefits of zoneout. 

In this experiment, one zoneout mask is sampled at the beginning of training, and used for all examples. This means the identity connections introduced are static across training examples (but still different for each timestep). 
Using static identity connections resulted in slightly lower {\it training} (but not validation) error than zoneout, but worse performance than an unregularized LSTM on both train and validation sets, as shown in Figure ~\ref{fig:static_mask}.

 \begin{figure}[h]
  \centering
  \includegraphics[width=0.65\textwidth]{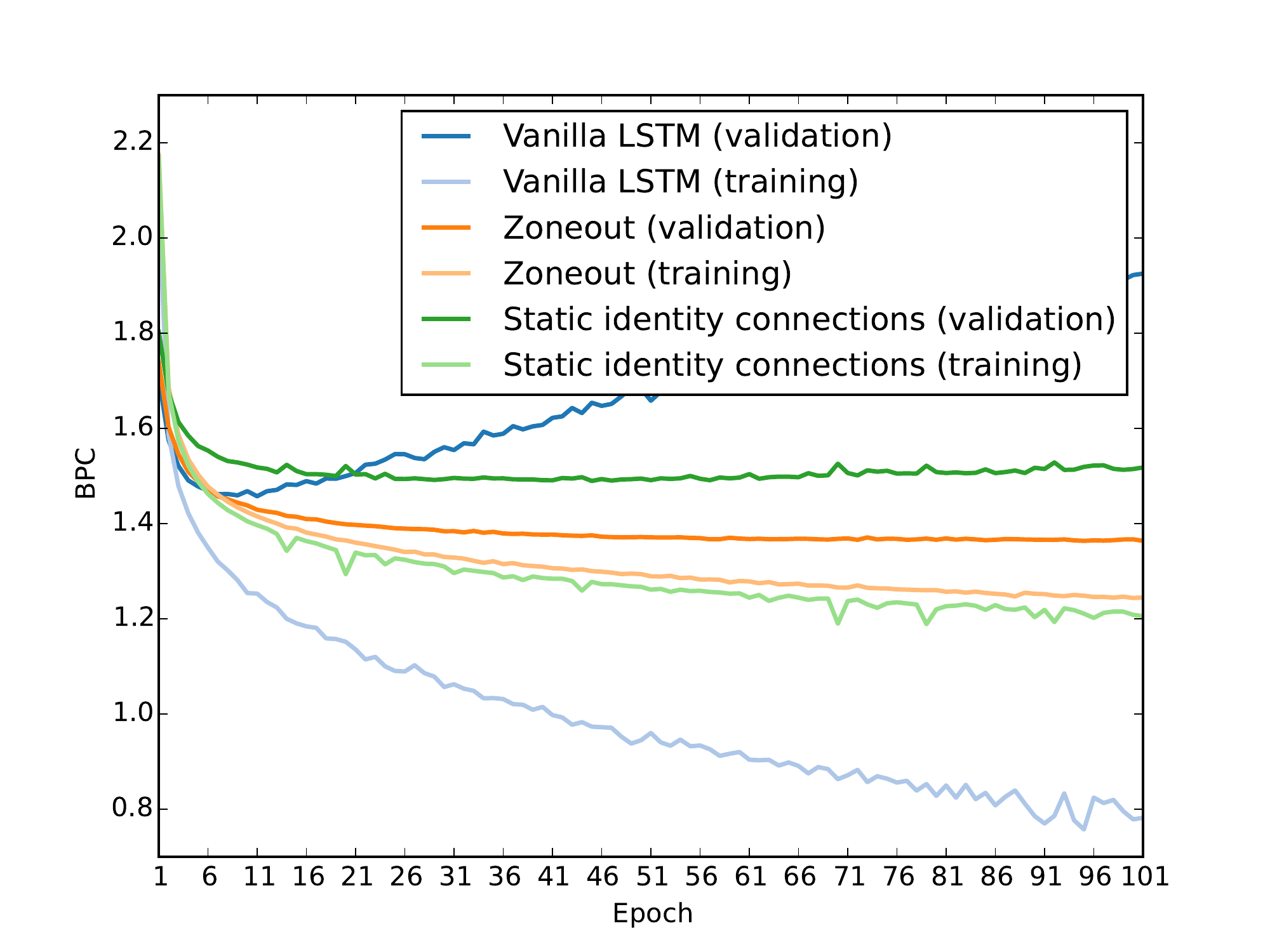}
  \caption{Training and validation curves for an LSTM with static identity connections compared to zoneout (both $Z_c=0.5$ and $Z_h=0.05$) and compared to a vanilla LSTM, showing that static identity connections fail to capture the benefits of zoneout.}
  \label{fig:static_mask}
 \end{figure}

\FloatBarrier
\section{Conclusion}

We have introduced zoneout, a novel and simple regularizer for RNNs, which stochastically preserves hidden units' activations.
Zoneout improves performance across tasks, outperforming many alternative regularizers to achieve results competitive with state of the art on the Penn Treebank and Text8 datasets, and state of the art results on $p$MNIST. 
While searching over zoneout probabilites allows us to tune zoneout to each task, low zoneout probabilities (0.05 - 0.2) on states reliably improve performance of existing models. 

We perform no hyperparameter search to achieve these results, simply using settings from the previous state of the art. 
Results on $p$MNIST and word-level Penn Treebank suggest that Zoneout works well in combination with other regularizers, such as recurrent batch normalization, and dropout on feedforward/embedding layers. 
We conjecture that the benefits of zoneout arise from two main factors: (1) Introducing stochasticity makes the network more robust to changes in the hidden state; (2) The identity connections improve the flow of information forward and backward through the network.



\subsubsection*{Acknowledgments}
We are grateful to Hugo Larochelle, Jan Chorowski, and students at MILA, especially \c{C}a\u{g}lar G\"ul\c{c}ehre, Marcin Moczulski, Chiheb Trabelsi, and Christopher Beckham, for helpful feedback and discussions.
We thank the developers of Theano \cite{Theano}, Fuel, and Blocks \cite{Blocks}.
We acknowledge the computing resources provided by ComputeCanada and CalculQuebec.
We also thank IBM and Samsung for their support. We would also like to acknowledge the work of Pranav Shyam on learning RNN hierarchies.
This research was developed with funding from the Defense Advanced Research Projects Agency (DARPA) and the Air Force Research Laboratory (AFRL). The views, opinions and/or findings expressed are those of the authors and should not be interpreted as representing the official views or policies of the Department of Defense or the U.S. Government.



\Chapter{\uppercase{A dataset and exploration of models for understanding video data through fill-in-the-blank question-answering}}\label{sec:moviefib}

\begin{quote}
\singlespacing
\cite{moviefib} Tegan Maharaj, Nicolas Ballas, Anna Rohrbach, Aaron Courville,  Christopher Pal. 2017. A dataset and exploration of models for understanding video data through fill-in-the-blank question-answering. \textit{IEEE Conference on Computer Vision and Pattern Recognition (CVPR)}.
\end{quote}


\if{false}
\section{Notes}
This section is included verbatim from my first-author paper \cite{moviefib}. 

\subsection{Full citation}

\begin{quote}
\singlespacing
\cite{moviefib} Tegan Maharaj, Nicolas Ballas, Anna Rohrbach, Aaron Courville,  Christopher Pal. 2017. A dataset and exploration of models for understanding video data through fill-in-the-blank question-answering. \textit{IEEE Conference on Computer Vision and Pattern Recognition (CVPR)}.
\end{quote}
\fi


In this work on video modelling, I hypothesized that a major reason for the relatively poor performance we were seeing was that evaluation metrics common at the time (e.g. BLEU) relied on word overlap with a ground-truth caption. Description is a notoriously under-determined problem: there are many potentially correct appropriate captions for an image or video, some of which may not overlap at all. Thus word-overlap measures do a poor job of specifying the intended task, and are even potentially counter-productive in learning a good representation. I proposed an alternative task, which I called \textbf{fill-in-the-blank question- answering (FIB-QA)}: labels for this task were automatically generated by removing certain parts of speech from a caption, and the language model’s task was to fill in the resulting blank, conditioned on frames of the video. In language modeling, FIB has come to be known as `masking’ or `masked language modelling'.

We tested a variety of deep video models, and performed an extensive human evaluation of the results. We found that representations improved markedly for the downstream captioning task, and importantly, in doing the human evaluation we found that accuracy on the FIB task corresponded well with human judgement.  This finding, and the utility of the overall approach of FIB for training good representations, have been confirmed in many works, perhaps most notably in the remarkable success of masked language models such as BERT \cite{bert}, which as of 2019 form the basis of Google search results.  
Apart from confirming the usefulness of masking as an auxiliary task for learning better representations, this result shows the utility of human evaluations for creating better-aligned evaluation metrics. I presented this work at CVPR 2017, and led the FIB track of the ‘Large Scale Movie Description Challenge’ for 2 years.

\vspace{0.5cm}
\textbf{Abstract}: \\
While deep convolutional neural networks frequently approach or exceed human-level performance in benchmark tasks involving static images, extending this success to moving images is not straightforward. Video understanding is of interest for many applications, including content recommendation, prediction, summarization, event/object detection, and understanding human visual perception. However, many domains lack sufficient data to explore and perfect video models. In order to address the need for a simple, quantitative benchmark for developing and understanding video, we present MovieFIB, a fill-in-the-blank question-answering dataset with over 300,000 examples, based on descriptive video annotations for the visually impaired. 
In addition to presenting statistics and a description of the dataset, we perform a detailed analysis of 5 different models' predictions, and compare these with human performance. We investigate the relative importance of language, static (2D) visual features, and moving (3D) visual features; the effects of increasing dataset size, the number of frames sampled; and of vocabulary size. We illustrate that: this task is not solvable by a language model alone; our model combining 2D and 3D visual information indeed provides the best result; all models perform significantly worse than human-level. We provide human evaluation for responses given by different models and find that accuracy on the MovieFIB evaluation corresponds well with human judgment. We suggest avenues for improving video models, and hope that the MovieFIB challenge can be useful for measuring and encouraging progress in this very interesting field.

\section{Introduction}

\begin{figure}
\center
\includegraphics[width=0.45\textwidth]{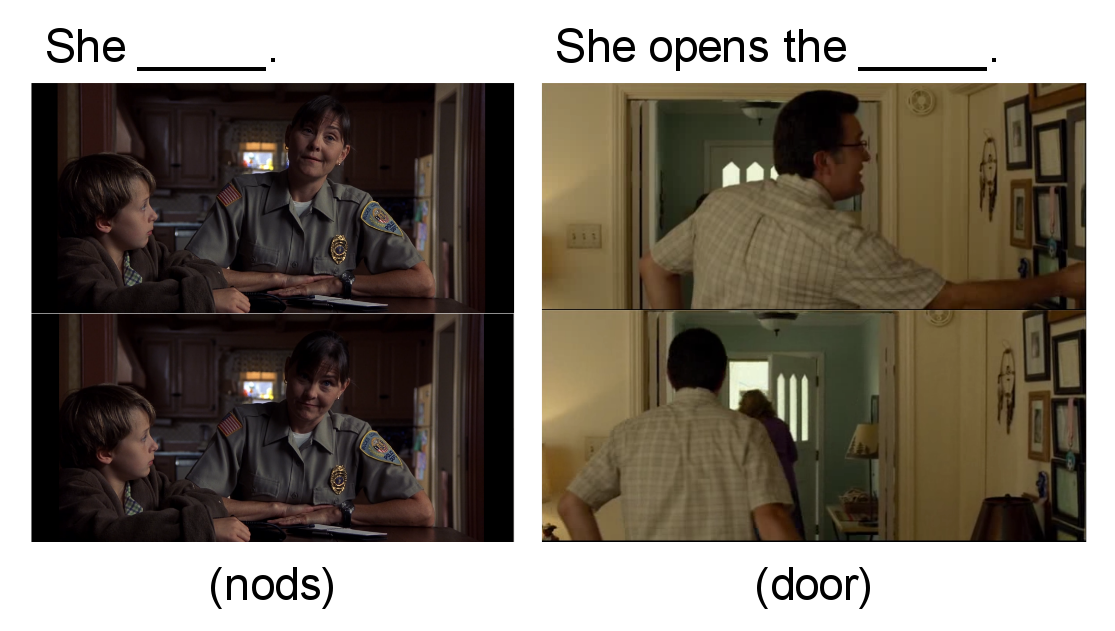}
\caption{Two examples from the training set of our fill-in-the-blank dataset.}
\label{fig:qual_example7}
\end{figure}

Most current work investigating multimodal question answering (QA) focuses either on the natural language aspects of the problem (e.g. \cite{WestonBCM15, MovieQA}), or QA in static images (e.g. \cite{VQA,malinowski2014multi,madlibs}). Our goal is to use QA to eliminate ambiguity in natural language evaluation, in order to target the benchmarking and development of video models. More specifically, we are interested in models of moving visual information which will be leveraged for a task in another modality - in this case, text-based QA. Our proposed dataset, \textbf{MovieFIB}, is used in the fill-in-the-blank track of LSMDC (Large Scale Movie Description and Understanding Challenge) \cite{lsmdc-challenge}.
 
\subsection{Video understanding}
A long-standing goal in computer vision research is complete understanding of visual scenes: recognizing entities, describing their attributes and relationships. In video data, this difficult task is complicated by the need to understand and remember temporal dynamics. 
The task of automatically translating videos containing rich and open-domain activities into natural language (or some other modality) requires tackling the above-mentioned challenges, which stand as open problems for computer vision. 

A key ingredient sparking the impressive recent progress in object category recognition ~\cite{krizhevsky2012imagenet} has been the development of large scale image recognition datasets ~\cite{deng2009imagenet}. Accordingly, several large video datasets have been proposed~\cite{rohrbach15cvpr, AtorabiM-VAD2015} to address the problem of translating video to natural language problem.
These datasets rely on transcriptions of Descriptive Video Services (DVS), also known as Audio Description (AD), included in movies as an aide for the visually impaired, to obtain text-based descriptions of movie scenes.
DVS provides an audio narration of the most important aspects of the visual information relevant to a movie which typically consists of descriptions of human actions, gestures, scenes, and character appearance~\cite{rohrbach15cvpr}.

While the extraction of scene descriptions from DVS has proven to be a reliable way to automatically associate video with text based descriptions, DVS  provides only one textual description per segment of video despite the fact that multiple descriptions for a given scene are often equally applicable and relevant.
This is problematic from an evaluation perspective. Standard evaluation metrics used for the video to natural language translation task, such as BLEU \cite{papineni2002bleu}, ROUGE \cite{lin2004rouge}, METEOR \cite{denkowski2014meteor} or CIDEr \cite{vedantam2015cider}, have been shown to not correlate well with human assessments when few target descriptions are available \cite{vedantam2015cider}. 
Therefore, it is questionable to rely on such automated metrics to evaluate and compare different approaches on those datasets.
\subsection{Our contributions} 
To address the issues with evaluating video models, we propose recasting the video description problem as a more straightforward classification task by reformulating description as a fill-in-the-blank question-answering (QA) problem. Specifically, given a video and its description with one word blanked-out, our goal is to predict the missing word as illustrated in Figure~\ref{fig:qual_example7}.

Our approach to creating fill-in-the-blank questions allows them to be easily generated automatically from a collection of video descriptions; it does not require extra manual work and can therefore be scaled to a large number of queries. 
Through this approach we have created over 300,000 fill-in-the-blank question-answer and video pairs.
The questions concern entities, actions and attributes. Answering such questions therefore implies that a model must obtain some level of understanding of the visual content of a scene, in order to be able to detect objects and people, aspects of their appearance, activities and interactions, as well as features of the general scene context of a video.

We compare performance of 7 models on MovieFIB; 5 run by us, 2 by independent works using our dataset, and additionally compare with an estimate of human performance. We have humans compare all models' responses. We discuss results, empirically demonstrate that classification accuracy on MovieFIB correlates well with human judgment, and suggest avenues for future work.

Dataset \& Challenge: \texttt{sites.google.com/site/ describingmovies/lsmdc-2016/download} 

Code: \texttt{github.com/teganmaharaj/movieFIB}

\section{Related work}

\subsection{Video Captioning}

The problem of bridging the gap between video and natural language has attracted significant recent attention. Early models tackling video captioning such as \cite{kojima2002, rohrbach2013},
focused on constrained domains with limited appearance of activities and objects in videos and depended heavily on hand-crafted video features, followed by a template-based or shallow statistical machine translation. However, recent models such as~\cite{Nicolas15,donahue2014long,venugopalan2014translating, yao2015describing}
have shifted towards a more general encoder-decoder neural approach to tackle the captioning problem for open-domain videos.
In such architectures, videos are usually encoded into a vector representation using a convolutional neural network, and then fed to a caption decoder usually implemented with a recurrent neural network.

The development of these encoder-decoder models has been made possible by the release of large scale datasets such as ~\cite{AtorabiM-VAD2015,rohrbach15cvpr}. 
In particular,~\cite{AtorabiM-VAD2015,rohrbach15cvpr} exploit Descriptive Video Service (DVS) data
to construct captioning datasets that have a large number of video clips. DVS is a type of narration designed for the visually impaired; it supplements the original dialogue and audio tracks of the movie by describing the visual content of a scene in detail, and is produced for many movies and TV shows. This type of description is very appealing for machine learning methods, because the things described tend to be those which are relevant to the plot, but they also stand alone as 'local' descriptions of events and objects with associated visual content. In \cite{lsmdc-challenge}, the authors create a dataset of 200 HD Hollywood movies split into 128,085 short (4-5 second) clips, aligned to transcribed DVS track and movie scripts. This dataset was used as the basis of the Large Scale Movie Description Challenge (LSMDC) presented in 2015 and 2016\footnote{https://sites.google.com/site/describingmovies/}.

While the development of these datasets has lead to new models which can produce impressive descriptions in terms of their syntactic and semantic quality, the evaluation of such techniques is challenging \cite{lsmdc-challenge}. Many different descriptions may be valid for a given image and as we have motivated above, commonly used metrics such as BLEU, METEOR, ROUGE-L and CIDEr have been found to correlate poorly with human judgments of description quality and utility \cite{lsmdc-challenge}.




\subsection{Image and Video QA}

One of the first large scale visual question-answering datasets is the visual question answering (VQA) challenge introduced in ~\cite{VQA}. It consists of 254,721 images from the MSCOCO \cite{lin2014microsoft} dataset, plus imagery of cartoon-like drawings from an abstract scene dataset \cite{zitnick2016adopting}. There are 3 questions per image for a total of 764,163 questions with 10 ground truth answers per question. The dataset includes questions with possible responses of yes, no, or maybe as well as open-ended and free-form questions and answers provided by humans. Other work has looked at algorithmically transforming MSCOCO descriptions into question format creating the COCO-QA dataset \cite{ren2015exploring}. The DAtaset for QUestion Answering on Real-world images (DAQUAR) was introduced in \cite{malinowski2014multi}. It was built on top of the NYU-Depth V2 dataset which consists of 1,449 RGBD images \cite{silberman2012indoor}. They collected 12,468 human question-answer pairs focusing on questions involving identifying 894 categories of objects, colors of objects and the number of objects in a scene. In \cite{madlibs}, the authors take a similar approach to ours by transforming the description task into fill-in-the-blank questions about images. 

Following this effort, \cite{zhu2015uncovering} compile various video description datasets including TACoS \cite{regneri2013grounding}, MPII-MD \cite{rohrbach15cvpr} and the TRECVID MEDTest 14 \cite{TRECVIDMED14}. As in our work, they reformulate the descriptions into QA tasks, and use an encoder-decoder RNN architecture for examining the performance of different approaches to solve this problem. Their work differs in approach from ours; they evaluate on questions describing the past, present, and future around a clip. It also differs in that they use a multiple choice format, and thus the selection of possible answers has an important impact on model performance. To avoid these issues here we work with an open vocabulary fill-in-the-blank format for our video QA formulation.

Other recent work develops MovieQA, a dataset and evaluation based on a QA formulation for story comprehension, using both video and text resources associated with movies \cite{MovieQA}. MovieQA is composed of 408 subtitled movies with summaries of the movie from Wikipedia, scripts obtained from the Internet Movie Script Database (IMSDb), which are available for almost half of the movies, and descriptive video service (DVS) annotations, available for 60 movies using the MPII-MD \cite{rohrbach15cvpr} annotations. The composition of MovieQA orients it heavily towards story understanding; there are 14,944 questions but only 6,462 are paired with video clips (see Table \ref{tab:movieQA_comparison}).

\begin{table}
  \center
  \caption{Comparison of statistics of the proposed MovieFIB dataset with the MovieQA\cite{MovieQA} dataset. Number of words includes the blank for MovieFIB.} 
  \small
  \begin{tabular}{lrrrr}
  \toprule
     \textbf{MovieQA dataset}& Train & Val & Test & Total \\
\midrule
\#Movies &93& 21& 26& 140 \\
\#Clips& 4,385& 1,098 &1,288& 6,771\\
Mean clip dur. (s)& 201.0 &198.5 &211.4& 202.7$\pm$216.2\\
\#QA& 4,318& 886& 1,258& 6,462\\
Mean \#words in Q& 9.3 &9.3& 9.5& 9.3$\pm$3.5\\
\midrule
     \textbf{MovieFIB dataset}& Train & Val & Test & Total \\
\midrule
\#Movies &153& 12& 17 & 180 \\
\#Clips& 101,046 &7,408& 10,053 &118,507 \\
Mean clip dur. (s)& 4.1 & 4.1 &4.2 & 4.1\\
\#QA& 296,960& 21,689& 30,349 &348,998\\
Mean \#words in Q& 9.94& 9.75& 8.67& 9.72\\
\bottomrule
  \end{tabular}
\label{tab:movieQA_comparison}
\end{table}


\section{MovieFIB: a fill-in-the-blank question-answering dataset }

In the following we describe the dataset creation process and provide some statistics and analysis.

\subsection{Creating the dataset} \label{makedataset}
The LSMDC 2016 description dataset \cite{lsmdc-challenge} forms the basis of our proposed fill-in-the-blank dataset (MovieFIB) and evaluation. Our procedure to generate a fill-in-the-blank question from an annotation is simple. For each annotation, we use a pretrained maximum-entropy parser \cite{maxentRatnaparkhi1996,maxentNLTK} from the Natural Language Toolkit (NLTK) \cite{nltk} to tag all words in the annotation with their part-of-speech (POS). We keep nouns, verbs, adjectives, and adverbs as candidate blanks, and filter candidates through a manually curated stop-list (see supplementary materials). Finally, we keep only words which occur $\geq 50$ times in the training set.
\subsection{Dataset statistics and analysis}
The procedure described in Section \ref{makedataset} gives us 348,998 examples: an average of 3 per original LSMDC annotation. We refer to the annotation with a blank (e.g. 'She \_\_\_\_\_ her head') as the \textbf{question} sentence, and the word which fills in the blank as the \textbf{answer}. We follow the training-validation-test split of the LSMDC2016 dataset and create 296,960 training, 21,689 validation, and 30,349 test QA pairs. Validation and test sets come from movies which are disjoint from the training set. We use only the public test set, so as not to provide ground truth for the blind test set used in the captioning challenge. Some examples from the training set are shown in Figure \ref{fig:qual_example7}, and Table~\ref{tab:movieQA_comparison} compares statistics of our dataset with the MovieQA dataset. For a more thorough comparison of video-text datasets, see \cite{lsmdc-challenge} 

Figure \ref{fig:histogram} is the histogram of answer counts for the training set, showing that most words occur 100-200 times, with a heavy tail of more frequent words going up to 12,541 for the most frequent word (her). For ease of viewing, we have binned the 20 most frequently-occurring words together in the last red bin. Figure \ref{fig:wordcloud} shows a word-cloud of the top 100 most frequently occurring words, with a list of the most frequent 20 words with their counts. In Figure \ref{fig:pospie} we examine the distribution by part-of-speech (POS) tag, showing the most frequent words for each of the categories.

\begin{figure}
\center
\includegraphics[width=0.48\textwidth]{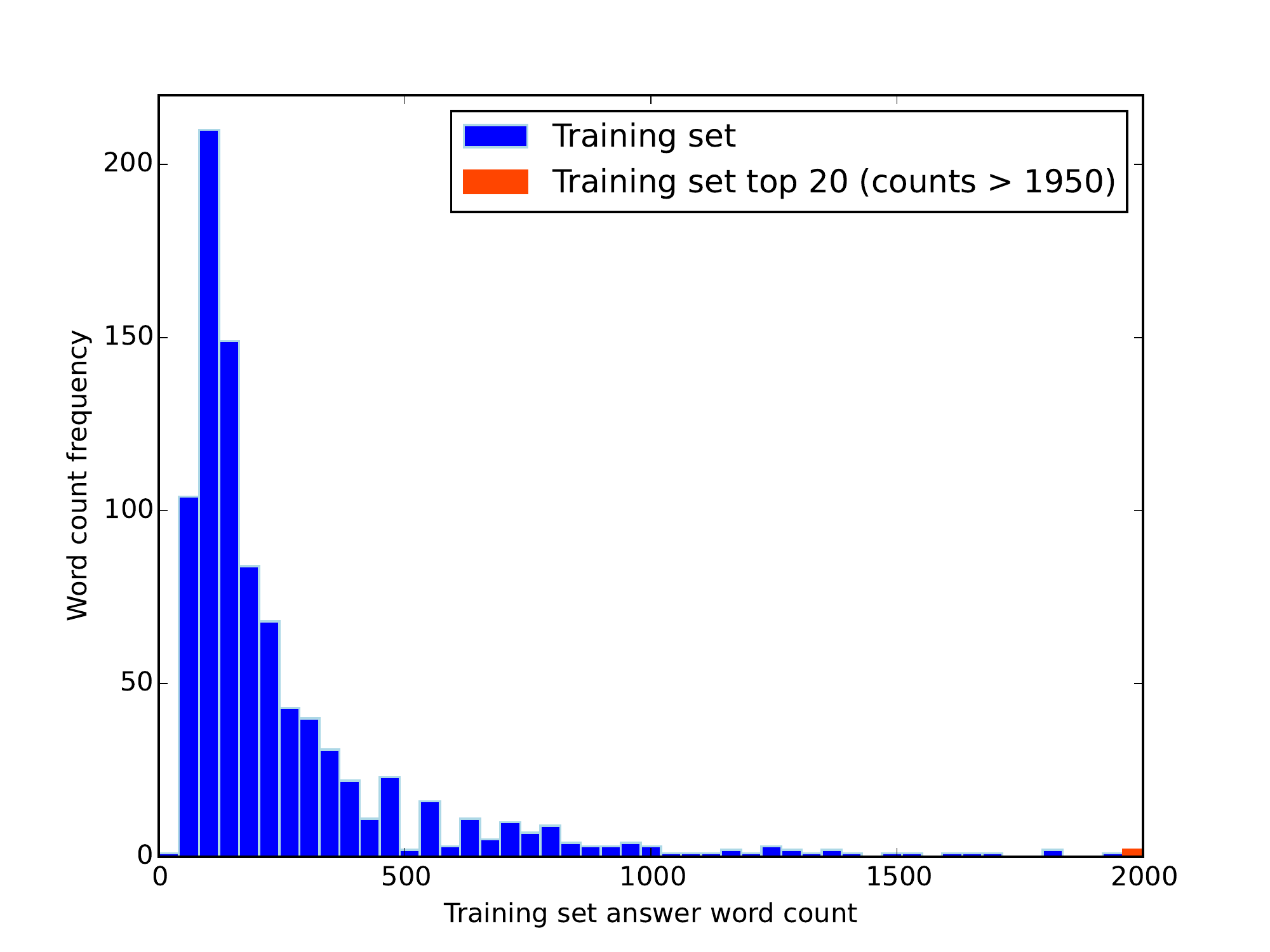}
\caption{Histogram showing frequencies of counts for answers (blanks) in the training set. Note that the last red bin of the histogram covers the interval [1,950 : 12,541], containing the 20 most frequent words which are listed in Figure \ref{fig:wordcloud}.}
\label{fig:histogram}
\end{figure}

\begin{figure}
\center
\includegraphics[width=0.48\textwidth]{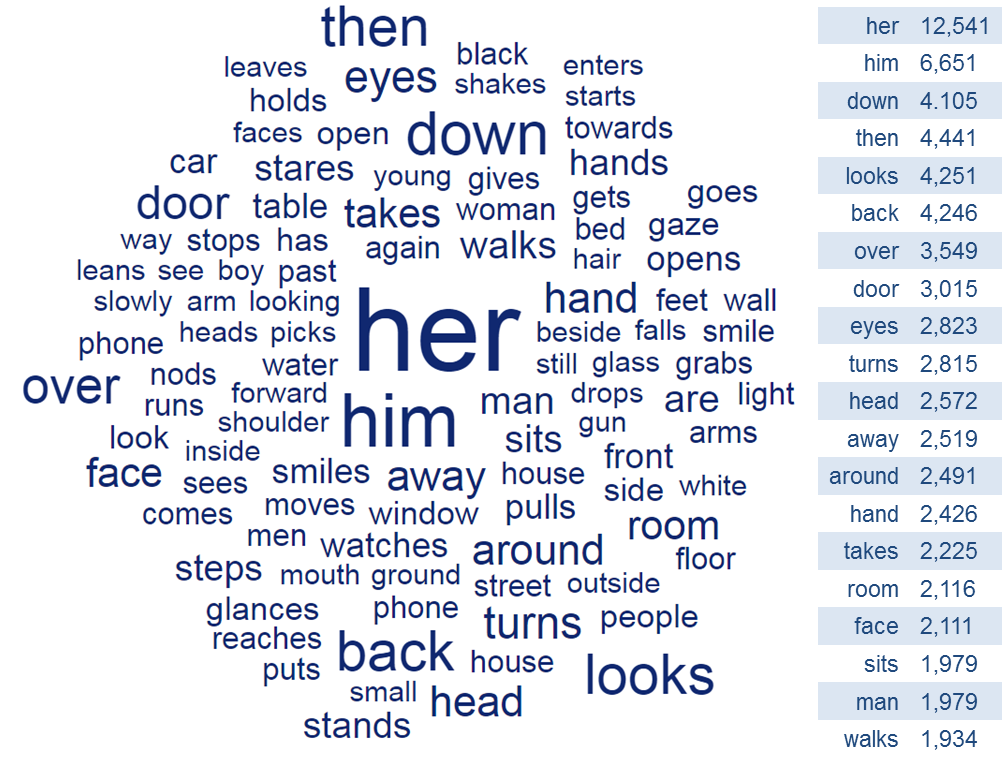}
\caption{Word cloud showing the top 100 most frequently-occurring words in the training set answers (font size scaled by frequency) and list with counts of the 20 most frequent answers.}
\label{fig:wordcloud}
\end{figure}

\begin{figure}
\center
\includegraphics[width=0.40\textwidth]{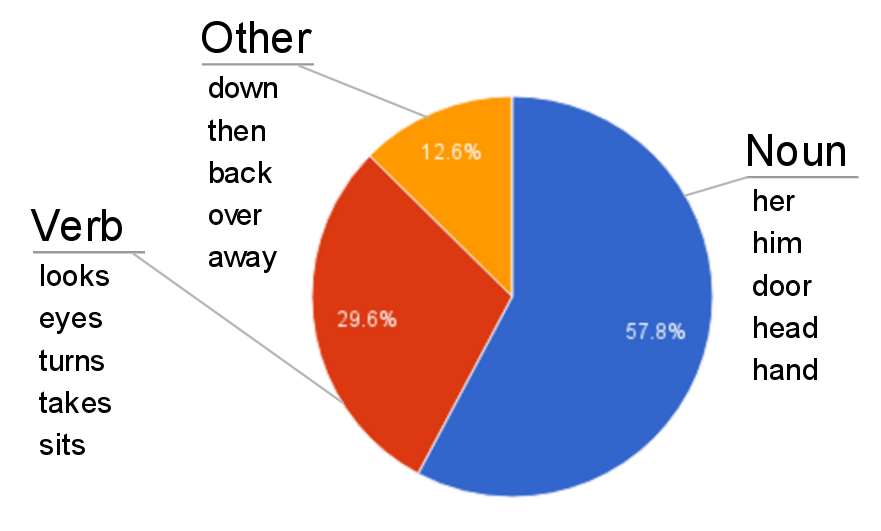}
\caption{Pie chart showing the answer words of the training set by POS-tag supercategory (noun, verb, or other), with the five most frequent words per category.} 
\label{fig:pospie}
\end{figure}


    

\section{Neural framework for video fill-in-the-blank question-answering}
\label{sec:framework}

In this section, we describe a general neural network-based approach to address the fill-in-the-blank video question-answering problem. This neural network provides a basis for all of our baseline models.

We consider a training set $(\vect v^i, \vect q^i, y^i)_{i \in (1\dots N)}$  with videos $\vect v^i$, questions $\vect q^i$ and their associated answers $y^i$.
Our goal is to learn a model that predicts $y^i$ given $\vect v^i$ and $\vect q^i$.

We first use encoder networks $\Phi_v$ and $\Phi_q$ to extract fixed-length representations from a video and a question respectively, as illustrated in Figure~\ref{fig:model}. The fixed length representations are then fed to a classifier network $f$ that outputs a probability distribution over the different answers, $p(y \mid \vect v^i, \vect q^i)) = f(\Phi_v(\vect v^i), \Phi_q(\vect q^i))_y$. $f$ is a single-layer MLP with a softmax. We estimate the model parameters $\TT$ composed of the encoder and classifier networks parameters $\TT = \{\TT_v, \TT_q, \TT_f\}$ by maximizing the model log-likelihood on the training set: 
\begin{equation}
\mathcal{L}(\TT) = \frac{1}{N} \sum_{i=1}^N \log p(y^i \mid \vect v^i, \vect q^i), \TT).
\end{equation}

\begin{figure}
\center
\includegraphics[width=0.40\textwidth]{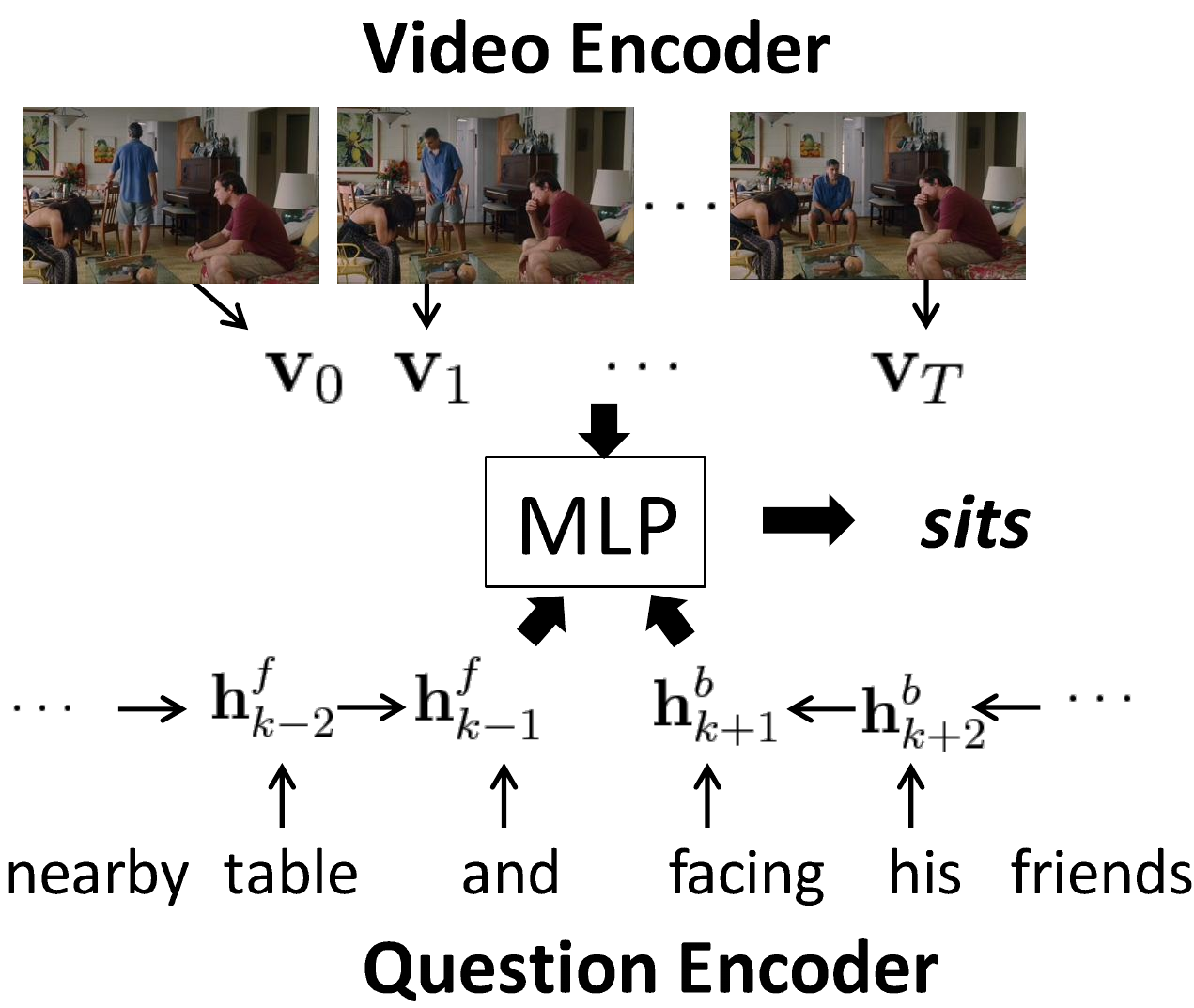}
\caption{Fill-in-the-blank model architecture, showing the video encoder $\Phi_v$, question encoder $\Phi_q$, and MLP classifier network $f$.}
\label{fig:model}
\end{figure}

\subsection{Question Encoder}
Recurrent neural networks have become the standard neural approach to encode text, as text data is composed of a variable-length sequence of symbols ~\cite{cho2014learning,sutskever2014sequence}. Given a sequence of words $\vect w_t$ composing a question $\vect q$, we define our encoder function as $\vect h_t = \Phi_q(\vect h_{t-1}, \vect w_t)$ with $\vect h_0$ being a learned parameter. 
For the fill-in-the-blank task, a question $\vect q$ composed by $l$ words can be written as $ \vect q = \{ \vect w_0, \ldots \vect w_{k-1}, \vect b, \vect w_{k+1}, \vect w_l \}$, where $\vect b$ is the symbol representing the blanked word.
To exploit this structure, we decompose our encoder $\Phi_q$ in two recurrent networks, one forward RNN $\Phi^f_q$ applied on the sequence $\{ \vect w_0, \ldots, \vect w_{k-1}\}$, and one  backward RNN applied on the reverse sequence $\{ \vect w_{l}, \ldots, \vect w_{k+1}\}$.
The forward hidden state $\vect h^f_{k-1}$ and backward hidden state $\vect h^b_{k+1}$ are concatenated
and provided as input to the classifier network $f$. A similar network structure for fill-in-the-blank QA is explored in~\cite{mazaheri2016video}.

Forward and backward functions $\Phi^f_q$ and  $\Phi^b_q$ could be implemented using vanilla RNNs,
however training such models using stochastic gradient descent is notoriously difficult
due to the exploding/vanishing gradients problems~\cite{bengio1994learning,hochreiter1991untersuchungen}.
Although solving gradient stability is fundamentally difficult~\cite{bengio1994learning}, effects can be mitigated through architectural variations such as LSTM~\cite{hochreiter1997long}, GRU~\cite{cho2014learning}. In this work, we rely on the Batch-Normalized variant of LSTM~\cite{cooijmans2016recurrent}:
\begin{eqnarray}
\left(\begin{array}{ccc}
\tilde{\vect{i}}_t \\
\tilde{\vect{f}}_t \\
\tilde{\vect{o}}_t \\
\tilde{\vect{g}}_t
\end{array}\right)
= 
 \mathrm{BN} (\mat{W}_w \vect{w}_t, \gamma_w) +
 \mathrm{BN} (\mat{W}_h \vect{h}_{t-1}, \gamma_h) +
 \vect{b}
\end{eqnarray}
where 
\begin{eqnarray}
\vect{c}_t &=& \sigma(\tilde{\vect{i}}_t) \ewprod \tanh(\tilde{\vect{g}_t}) + 
               \sigma(\tilde{\vect{f}}_t) \ewprod \vect{c}_{t-1} \\
\vect{h}_t &=& \sigma(\tilde{\vect{o}}_t) \ewprod \tanh(
 \mathrm{BN} (\vect{c}_t; \gamma_c) +
 \vect{b}_c
)
\end{eqnarray}
and where  
\begin{equation}
\mathrm{BN}(\vect{x} ; \gamma) =  \gamma \ewprod
\frac{\vect{x} -   \widehat{\mathbb{E  }}[\vect{x}]}
     {       \sqrt{\widehat{\mathrm{Var}}[\vect{x}] + \epsilon}}
\end{equation}
is the  batch-normalizing transform with $\widehat{\mathbb{E}}[\vect{x}], \widehat{\mathrm{Var}}[\vect{x}]$  being the activation mean and variance estimated from the mini-batch samples. $\vect{W}_h \in \reals^{d_h \times 4 d_h}, \vect{W}_w  \in \reals^{d_w \times 4 d_h}, \vect{b} \in \reals^{4 d_h}$
and the initial states $\vect{h}_0 \in \reals^{d_h}, \vect{c}_0 \in \reals^{d_h}$
are model parameters.
$\sigma$ is the logistic sigmoid function, $\tilde{\vect{i}}_t,
\tilde{\vect{f}}_t, \tilde{\vect{o}}_t$, and $\tilde{\vect{g}}_t$ are the LSTM gates, and the $\ewprod$ operator denotes the Hadamard product.


\subsection{Video Encoder}

Following recent work on video modeling~\cite{donahue2014long,srivastava2015unsupervised},
we use 2D (or 3D\footnote{In this work, 2D = (height,width) and 3D = (height,width,time)}, as indicated) convolutional neural networks which map each frame (or sequence of frames) into a
sequence vector. The video encoder $\Phi_v$ then extracts a fixed-length representation from the sequence of 2D frames composing a video. As described for the question encoder, we rely on the Batch-Normalized LSTM~\cite{cooijmans2016recurrent} to model the sequence of vectors.

\section{Experiments and Discussion}

First in Section \ref{model_comp} we describe 5 baseline models which investigate the relative importance of 2D vs. 3D features, as well as early vs. late fusion of text information (by initializing the video encoder with the question encoding and then finetuning). Using these models, we investigate the importance of various aspects of text and video preprocessing and the effects of dataset size in Section \ref{prepro_eff}. We then describe the setup and results of getting an estimate of human performance on MovieFIB in Section \ref{human_test}. Next, Section \ref{rel} describes the models and results of two independent works \cite{challenge1,mazaheri2016video} which use our dataset. All of these results are summarized in Table \ref{tab:results}. 
Finally, in Section \ref{human_eval} we perform a human evaluation of all of these different models' responses, and show that using the standard metric of accuracy for comparing models yields results which correlate well with human assessment.


\subsection{Comparison of baseline models}\label{model_comp}

\begin{table}
\caption{Accuracy results for single models, and estimated human performance (both human experiments are conducted with a subset of 569 examples from the test set). Finetuned indicates the question encoder was initialized with the parameters of the Text-only model. Vocabulary* indicates the output softmax was reduced to only consider words with frequency $\geq 50$ in the training set.} 
  \center
  \begin{tabular}{ccc}
  \toprule
    Model & Validation & Test\\
    \midrule
    Text-only & 33.8 & 34.4\\
    GoogleNet-2D  & 34.1 & 34.9\\
    C3D       & 34.0 & 34.5\\
    GoogleNet-2D Finetuned & 34.7 & 35.3\\
    GoogleNet-2D + C3D Finetuned &  35.0 & 35.7\\
    \midrule    
    Vocabulary* Text-only & 34.3 & 35.0\\
    Vocabulary* 2D + C3D Finetuned & 35.4 & 36.3\\
    \midrule
    Human text-only & - & 30.2\\
    Human text+video & - & 68.7\\
    \midrule
    VGG-2D-MergingLSTMs~\cite{mazaheri2016video} & - & 34.2\\
    ResNet-2D-biLSTM-attn~\cite{challenge1} & - & 38.0\\

    \bottomrule
  \end{tabular}
\label{tab:results}
\end{table}

\paragraph{Text Preprocessing.}
We preprocess the questions and answers with wordpunct tokenizer from the NLTK toolbox~\cite{nltk}.
We then lowercase all the word tokens, and end up with a vocabulary of 26,818 unique words. In Section \ref{prepro_eff} we analyze the impact of changing the vocabulary size.
\paragraph{Video Preprocessing.}
To leverage the video input, we investigate both 2D (static) and 3D (moving) visual features.
We rely on a GoogLeNet convolutional neural network \cite{szegedy2015going} that
has been pretrained on ImageNet~\cite{deng2009imagenet} to extract static features.
Features are extracted from the pool5/7x7 layer.
3D moving features are extracted using the C3D model~\cite{tran2015learning}, pretrained on  Sports-1M~\cite{karpathy2014large}. We apply the C3D on chunks of 16 consecutive frames in a video and retrieve the ``fc7''-layer activations. We do not finetune the 2D and 3D CNN parameters during training.

To reduce memory and computational requirements, we only consider a fixed number
of frames/temporal segments from the different videos. Unless otherwise specified, we consider 25 frames/temporal segments per video. Those frames/temporal segments are sampled randomly during training while being equally spaced during inference on the validation and test sets. We investigate the effects of sampling different numbers of frames in Section \ref{prepro_eff}.
\paragraph{Language, static visual (2D), and moving visual (3D) information.}
We test model variations for video fill-in-the-blank task based on the framework described in section~\ref{sec:framework}.
Specifically, we investigate the performance a baseline model using only the question encoder (i.e. a language model), which we call \textbf{Text-only} and  the impact of 2D and 3D features individually as well as their combination.
We train our baseline models using stochastic gradient descent with  Adam update rules~\cite{kingma2014adam}. Model hyperparameters can be found in the supplementary materials. Results are reported in Table~\ref{tab:results}.

While the Text-only baseline obtains reasonable results
by itself, adding a visual input in any form (2D, 3D, or combination) improves accuracy. We observe that the contributions of the different visual features seems complimentary, as  they can be combined to further improve performance. To illustrate this qualitatively, in Figure \ref{fig:qualex} we show two examples which the Text-only model gets wrong, but which GoogleNet-2D+C3D model gets right. Whereas MovieQA authors find that adding video information actually hurts performance \cite{MovieQA}, our experiments demonstrate the utility of our dataset for targeting video understanding, compared to MovieQA which targets story understanding.
\begin{figure}
\center
\includegraphics[width=0.45
\textwidth]{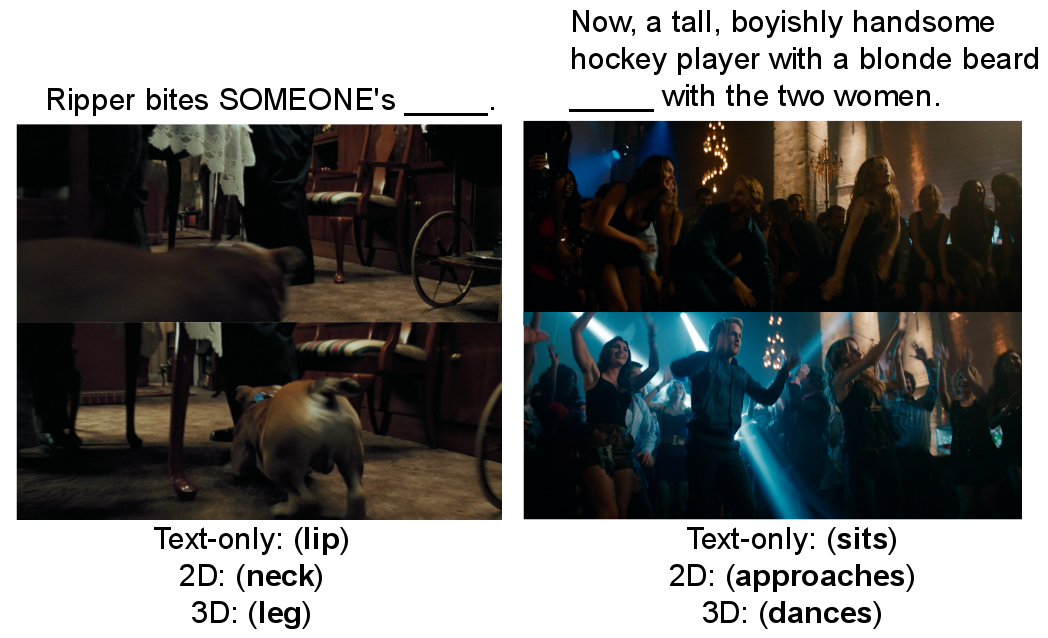}
\caption{Qualitative examples for the Text-only, 2D (GoogleNet-2D), and 3D (Googlenet-2D+C3D) showing the importance of visual information; in particular the importance of 3D features in recognizing actions.} 

\label{fig:qualex}
\end{figure}

We also compare models with parameters initialized randomly versus model having
the question encoder parameters initialized directly from the Text-only baseline, which we refer to as \textbf{Finetuned} in Table~\ref{tab:results}). Finetuned initialization leads to better results; we empirically observe that it tends to reduce the model overfitting.

\begin{figure}
\center
\includegraphics[width=0.38\textwidth]{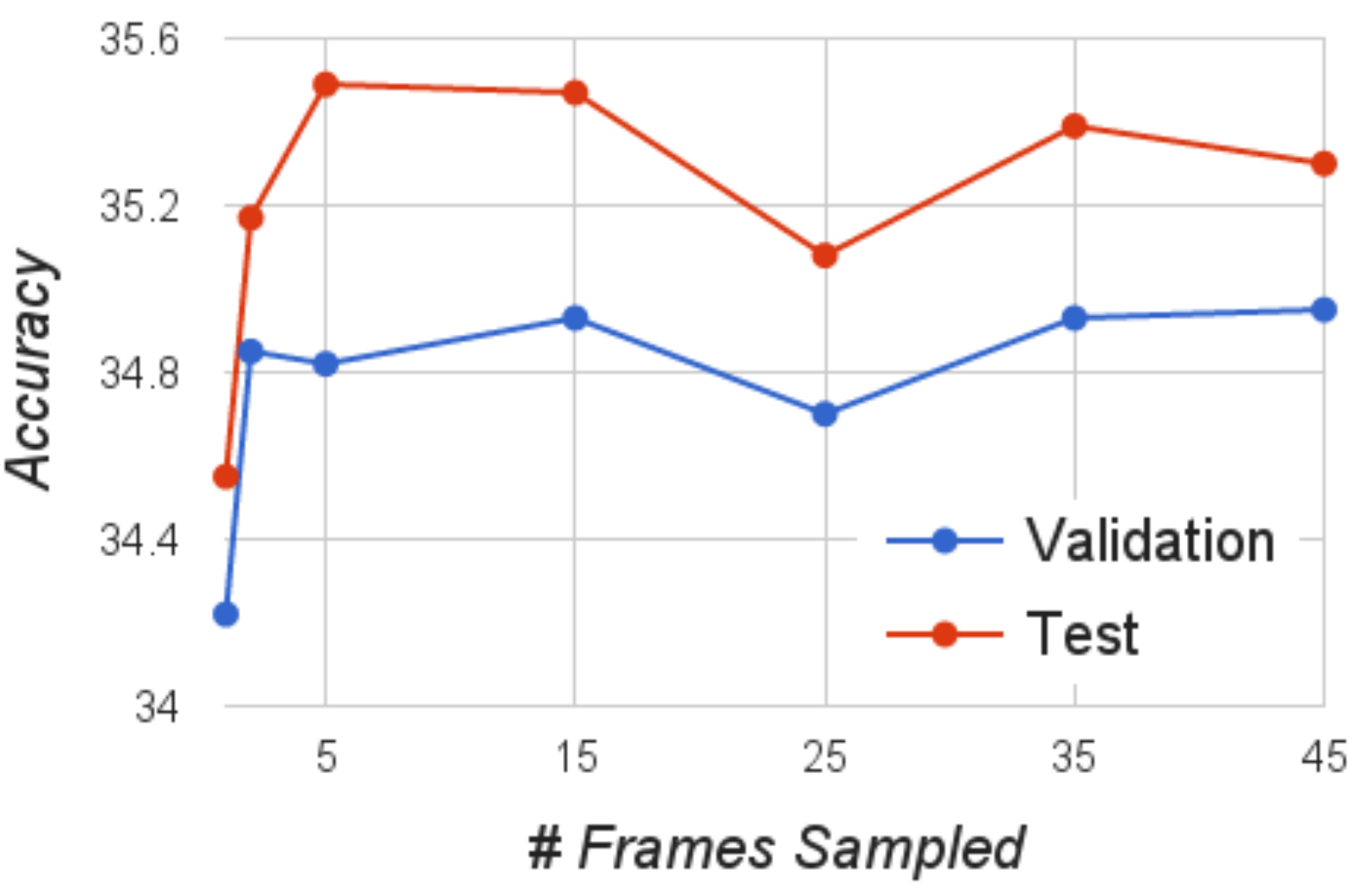}
\caption{Performance on the test set for GoogleNet-2D (finetuned) showing that comparable performance is achieved with only two sampled frames.} 
\label{fig:frames}
\end{figure}

\subsection{Effects of amount and preprocessing of data}\label{prepro_eff}

\paragraph{Vocabulary size} We first look at the impact of the input vocabulary size. In addition to the text preprocessing described in Section~\ref{model_comp}, we eliminate rare tokens from the vocabulary applied on the input question, to only consider words that occur more than 3 times in the training set. Rare words are replaced with an ``unknown'' token. This leads to vocabulary of size $18,663$. We also reduce the vocabulary size of the output softmax for answer words, considering only words present more than 50 times in the training set,
resulting in a vocabulary of size $3,994$. 
We denote this variant as ``Vocabulary*'' in Table~\ref{tab:results} and observe that reducing the vocabulary sizes results in improved performance, highlighting the importance of the text preprocessing.
\paragraph{Number of input frames}
We also investigate the importance of the number of input frames for the GoogleNet-2D baseline model. Results are reported in Figure~\ref{fig:frames}. We observe that the validation performance saturates quickly, as we almost reach the best performance with only 2 sampled frames from the videos on the validation set.

\paragraph{Effects of increasing dataset size}
As evidenced by performance on large datasets like ImageNet, the amount of training data available can be a huge factor in the success of deep learning models. We are interested to know if the dataset size is an important factor in the performance of video models, and specifically, if we should expect to see an increase in the performance of existing models simply by increasing the amount of training data available.

\begin{figure}
  \center
\includegraphics[width=0.48\textwidth]{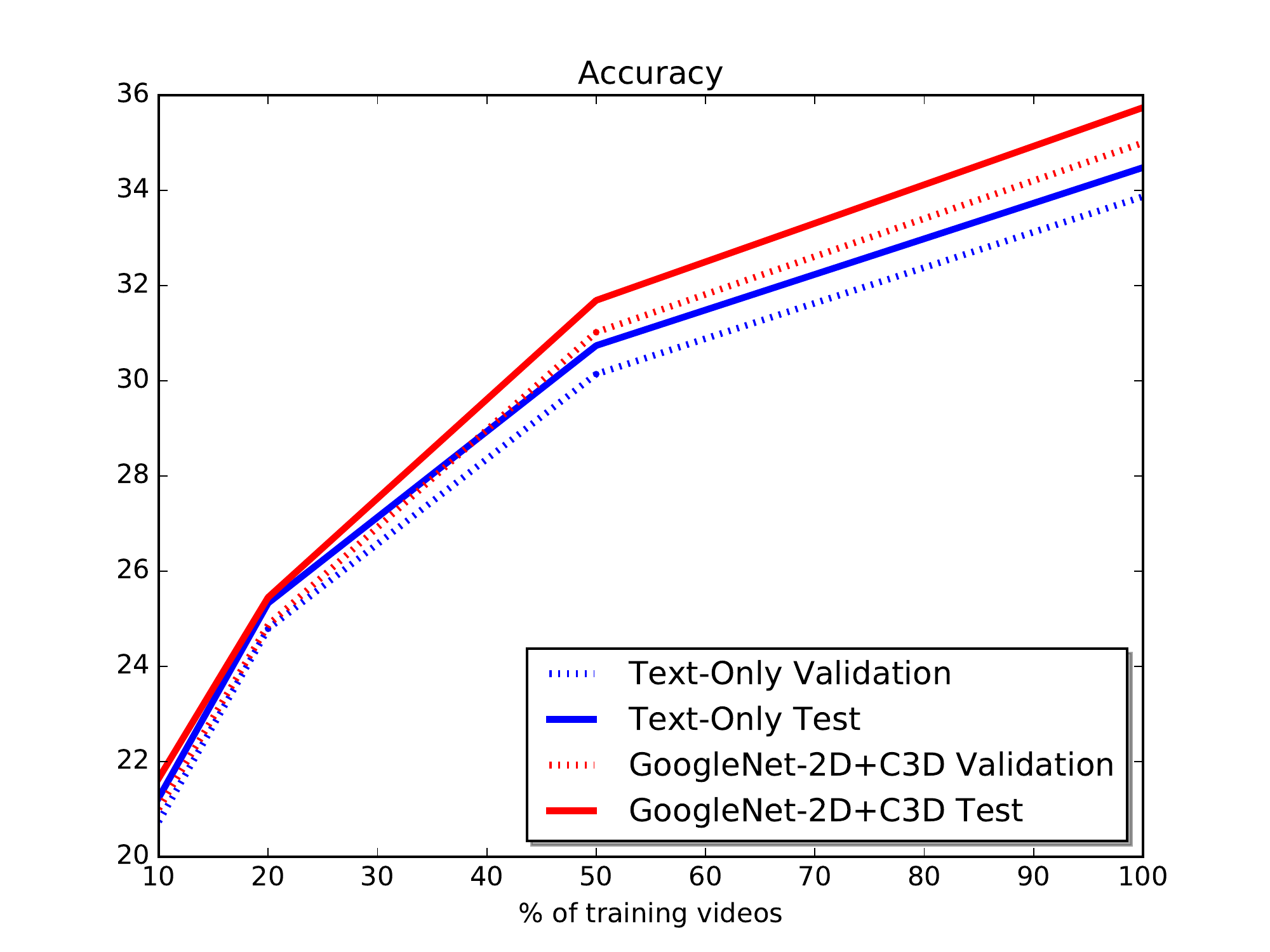}
\caption{Fill-in-the-blank accuracy results for the Text-only and GoogleNet-2D+C3D-finetuned models on validation and test sets, trained on varying percentages (10,20,50, and 100\%) of the training data, showing a larger gain in test performance relative to validation for the video model (Note that results for models trained with 100\% of training data are the same as reported in Table \ref{tab:results}).} 
\label{fig:data_impact}
\end{figure}

\begin{figure}
\center
\includegraphics[width=0.48\textwidth]{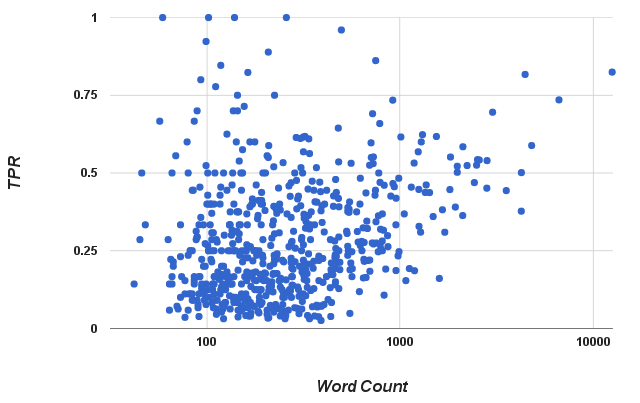}
\caption{The true positive rate (TPR) per answer word for the GoogleNet-2D+C3D-5frame model, plotted by answer word frequency in the training set (note log scale), showing that the TRP (aka recall, sensitivity) is correlated with answer word frequency.} 
\label{fig:acc_by_freq}
\end{figure}

\begin{figure*}[!ht]
\center
\includegraphics[width=0.96\textwidth]{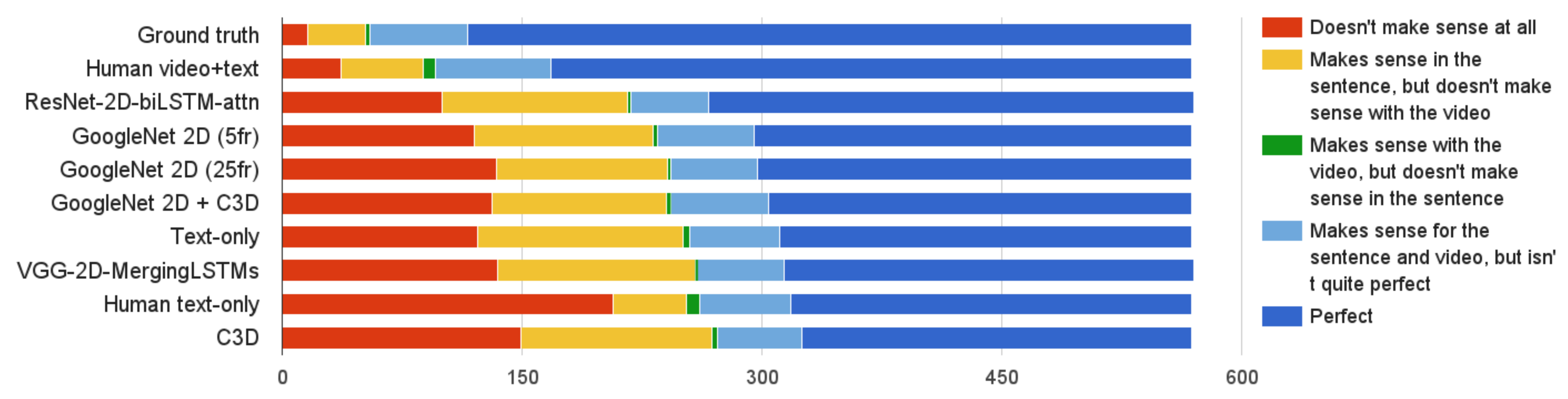}
\caption{Human evaluation of different models' answers.}
\label{fig:humaneval}
\end{figure*}

\begin{figure}
\center
\includegraphics[width=0.48\textwidth]{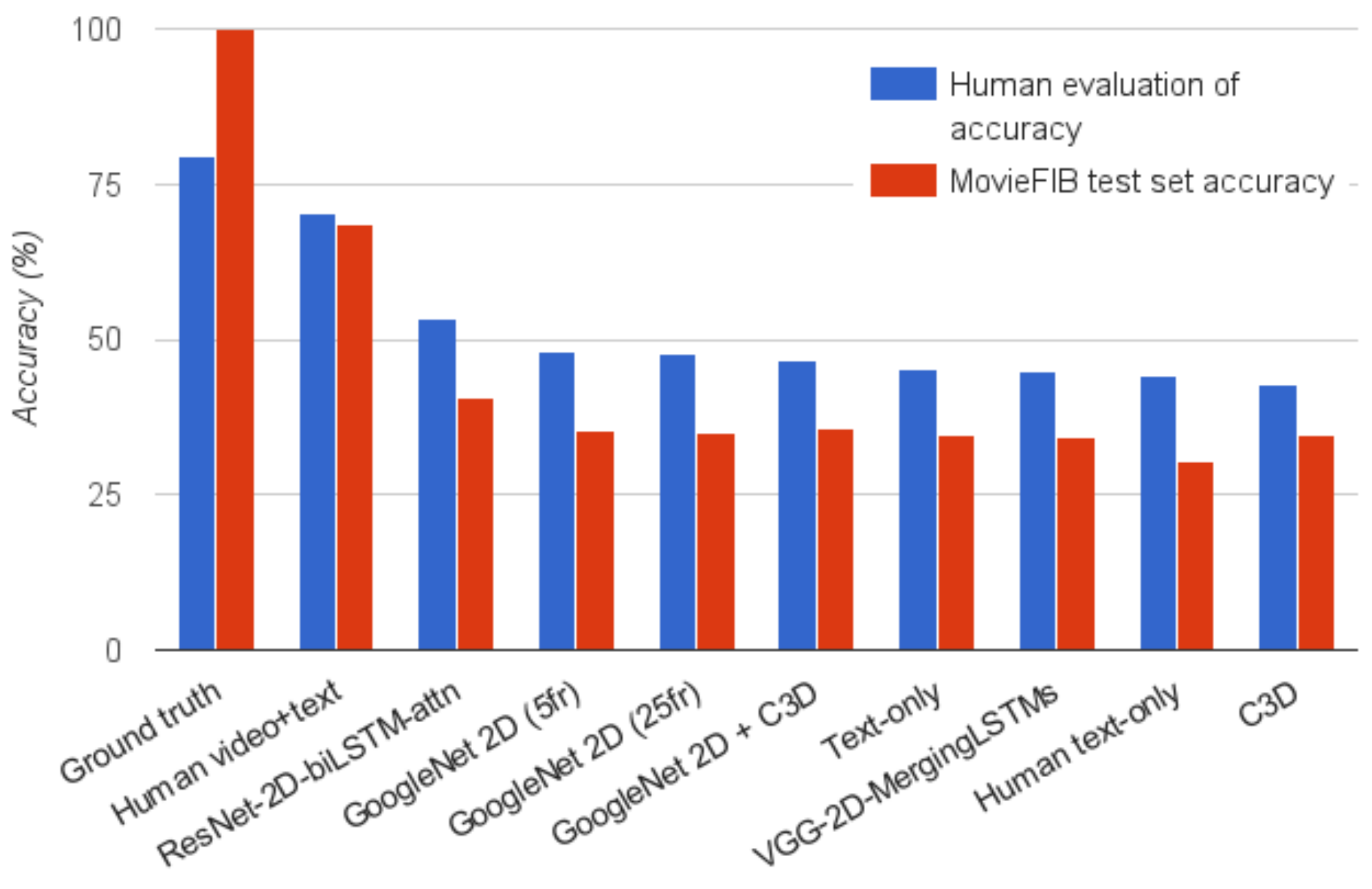}
\caption{Performance on the test set and performance according to human evaluation, demonstrating that these metrics correspond well.} 
\label{fig:scatter}
\end{figure}

Figure~\ref{fig:data_impact} reports the validation and test accuracies of Text-only and GoogLeNet-2D+C3D baselines as we increase the number of training videos.
It shows that at 10\% of training data (9,511 videos), Text-only and video models perform very similarly ($20.7\%$ accuracy for Text-only versus $21.0\%$ for GoogleNet-2D+C3D on the valid set).
It suggests that at 10\% of training data, there are not enough video examples for the model to leverage useful information that generalizes to unseen examples from the visual input.

However, we observe that increasing the amount of training data benefit more to the video-based model relatively to the Text-only model. As data increases the performance of the video model increases more rapidly than the Text-only model.
This suggests that existing video models are in fact able to gain some generalization
from the visual input given enough training examples.
Hence, Figure~\ref{fig:data_impact} highlights that further increasing the dataset size should be more beneficial for the video-based models.

Figure~\ref{fig:acc_by_freq} shows that per-word true positive rate (TPR) is highly correlated with answer prevalence in the training set, indicating that increasing the number of examples for each target would likely also increase performance. We plot here the results only for GoogleNet-2D + C3D for brevity, but similar correlations are seen for all models, which can be viewed in the supplementary material.


\subsection{Human performance on the test set}\label{human_test}
Table~\ref{tab:results} also reports human performance on a subset of the test set.
In order to obtain an estimate for human performance on the test set, we use Amazon Mechanical Turk to employ humans to answer a sample of 569 test questions, which is representative of the test set at a confidence of $95\%/{^+_{-}4}$. To mimic the information given to a neural network model, we require humans to fill in the blank using words from a predefined vocabulary in a searchable drop-down menu. In order to ensure quality of responses, we follow \cite{deng2009imagenet} in having 3 humans answer each question. If two or more humans answer the same for a given question, we take that as the answer; if all disagree, we randomly choose one response as the answer out of the 3 candidates.

We perform two experiments with this setup: \textbf{Human text-only} and \textbf{Human text+video}. In the text-only experiment, workers are only shown the question, and not the video clip, while in the text+video setting workers are given both the video clip and the question.
As in the automated models, we observe that adding video input drastically improves human performance. This confirms that visual information is of prime importance for solving this task. 

We also observe in Table~\ref{tab:results} that there is a significant gap between our
best automated model and the best human performance (on text+video), leaving room for improvement of future video models.
Interestingly, we notice that our text-only model outperforms the human text-only accuracy. Descriptive Video Service (DVS) annotations are written by movie industry professionals, and have a certain linguistic style which appears to induce some statistical regularities in the text data. 
Our text-only baseline, directly trained on DVS data, is able to exploit these statistical regularities, while a Mechanical Turk worker who is not familiar with the DVS style of writing may miss them.

\subsection{Related works using MovieFIB}\label{rel}
We have made the MovieFIB dataset publicly available, and two recent works have made use of it. We include the results of these models in our comparisons, and report the best single-model performance of these model in \ref{tab:results}.

In \cite{challenge1}, the authors use an LSTM on pretrained ImageNet features from layer conv5b of a ResNet \cite{resnet} to encode the video, with temporal attention on frames, and a bidirectional LSTM with semantic attention for encoding the question. We refer to this model as \textbf{ResNet-2D-biLSTM-attn}, and it achieves the highest reported accuracy on our dataset so far - 38.0\% accuracy for a single model, and 40.7 for an ensemble.

In \cite{mazaheri2016video}, the authors use a model similar to our baseline, encoding video with an LSTM on pretrained VGG \cite{VGG} features, combined with the output of two LSTMs running in opposite directions on the question by an MLP. We refer to this model as \textbf{VGG-2D-MergingLSTMs}. Their method differs from ours in that they first train a Word2Vec \cite{mikolov13nips} embedding space for the questions. Like us, they find that using a pretrained question encoding improves performance.


\subsection{Human evaluation of results}\label{human_eval}
We employ Mechanical Turk workers to rank the responses from the models described in Table \ref{tab:results}. Workers are given the clip and question, and a list of the different models' responses (including ground truth). Figure \ref{fig:humaneval} shows how humans evaluated different models' responses. Interestingly, humans evaluate that the ground truth is ``Perfect'' in about 80\% of examples, an additional 11\% ``Make sense for the sentence and video, but isn't quite perfect'', and for 3\% of ground truth answers (16 examples) workers say the ground truth ``Doesn't make sense at all''.  We observe that for most of these examples, the issue appears to be language style; for example ``He \_\_\_\_\_ her'' where the ground truth is ``eyes''. This may be an unfamiliar use of language for some workers, which is supported by the Human text-only results (see Section \ref{human_test}).
Figure \ref{fig:scatter} shows that accuracy tracks the human evaluation well on the test set, in other words, that accuracy on MovieFIB is a representative metric. 

\section{Conclusion}

We have presented MovieFIB, a fill-in-the-blank question-answering dataset, based on descriptive video annotations for the visually impaired, with over 300,000 question-answer and video pairs.

To explore our dataset, and to better understand the capabilities of video models in general, we have evaluated five different models and compared them with human performance, as well as with two independent works using our dataset \cite{mazaheri2016video,challenge1}.
We observe that using both visual and temporal information is of prime importance to model performance on this task. However, all models still perform significantly worse than human-level in their use of video.

We have studied the importance of quantity of training data, showing that models leveraging visual input benefit more than text-only models from an increase of the training samples. This suggests that performance could be further improved just by increasing the amount of training data.

Finally, we have performed a human evaluation of all models' responses on the dataset so far. These results show that accuracy on MovieFIB is a robust metric, corresponding well with human assessment.

We hope that the MovieFIB dataset will be useful to develop and evaluate models which better understand the moving-visual content of videos, and that it will encourage further research and progress in this field. 

For future work, we suggest: (1) transforming a difficult-to-evaluate task (e.g. 'translation' between modalities, generation, etc.) into a classification task is a broadly applicable idea, useful for benchmarking models; (2) exploring spatio-temporal attention; (3) determining which factors contribute most to improvement of video model performance - increasing data, refinement of existing architectures, development of novel spatio-temporal architectures, etc; (4) further investigation of multimodal fusion in video (e.g. better combining text and visual, leveraging audio).

\newpage 
\textbf{Appendix}

\section{Question generation stoplist}

When selecting words to blank to create fill-in-the-blank question-answer pairs, we performed some manual filtering to prevent undesirable words from becoming targets. This is the list of words which were disallowed for blanking (i.e., the stoplist):

\begin{tabular} {lll}
A &
Dr&
Elsewhere\\
IN&
Meanwhile&
Mr\\
Mrs&
Now&
OF\\
SOMEONE&
SOMEONE'&
SOMEONE's\\
THE&
a&
as\\
aside&
be&
before\\
can't&
does&
doesn't\\
in&
is&
it\\
it's&
it's&
later\\
meanwhile&
new&
not\\
now&
on&
other\\
other's&
something&
them\\
there&
up&
while\\
who&
who's&

\end{tabular}

\section{Model Specification}

We describe the different hyperparameters used in our experimental setting.

\subsection{Question Encoding}

Words composing a question are given as in one-hot encoding formats, leading to vectors having the same size than the vocabulary.
They are then projected into vector of size 512 using an embedding matrix. They are fed to the forward and backward BN-LSTM leading to hidden vectors of size 320.
The hidden vectors corresponding to the last timestep of the forward and backward BN-LSTM  are concatenanted to obtain the question respresentation.

\subsection{Video Encoding}

As specified in the main paper, we rely on a GoogLeNet convolutional neural network that
has been pretrained on ImageNet~\cite{szegedy2015going}  to extract static features.
Features are extracted from the pool5/7x7 layer.
3D moving features are extracted using the C3D model~\cite{tran2015learning}, pretrained on Sport 1 million~\cite{karpathy2014large}. We apply the C3D frames on chunk of 16 consecutive frames in a video and  retrieve the activations corresponding to the ``fc7'' layer. We don't finetune the 2D and 3D CNN parameters during training on the fill-in-the-blank task.
It leads to an input visual vector of size 1024 for GoogleNet, 4096 for C3D hence 5120 when we are concatenating  both GoogleNet and C3D.
If not specified otherwise, we extract visual vectors for 25 temporal chunks in the videos.

Vectors are then fed to a video encoding BN-LSTM of size 320.
The hidden vectors corresponding to the last timestep of the video LSTM is used as video representation.

\subsection{Classifier}

The classifier is a single layer MLP with a softmax activation function in order to output a probability distributions over the diverent candidate words.
We apply dropout on the MLP inputs with drop probability of $0.5$.

\subsection{Training}

We used Adam~\cite{kingma2014adam} with the gradient computed by
the backpropagation algorithm. We use a learning rate of $1e-3$, and a gradient clipping of $10$.
Early stopping of the model is performed based on the validation set performance.

\section{True Positive Rate (TPR) by Word Frequency for all models}

In the main paper we plot TPR (number of times a model correctly filled in the blank with a given word) by word frequency (how often that word appears in the training set) for GoogleNet-2D-C3D; here in Figure \ref{fig:tpr} we provide these plots for all models. These plots show that all models perform better on more frequent words, suggesting that increasing dataset size would be one way to improve model performance. 

\begin{figure*}[h]
\center
\includegraphics[width=.9
\textwidth]{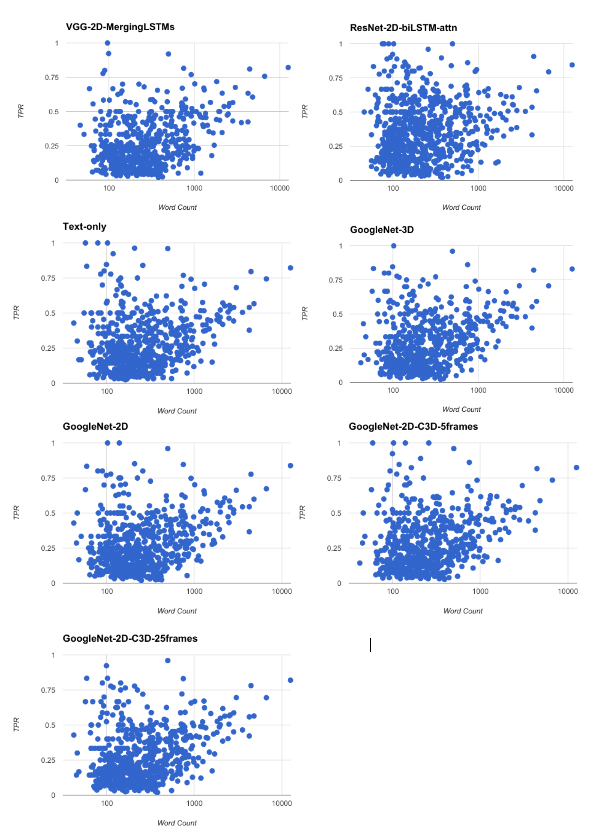}
\caption{True Positive Rate (TPR) per Word Frequency for all models.} 

\label{fig:tpr}
\end{figure*} 
\Chapter{\uppercase{Dealing with Distributional Shift}}\label{sec:dist}

\begin{quote}
    \singlespacing
\cite{hiads} David Krueger, Tegan Maharaj, Jan Leike.  2019.  Hidden incentives for auto-induced distributional shift. Safe Machine Learning Workshop at the International Conference on Machine Learning (ICML).

\cite{gupta2020covisim} Prateek Gupta$\dagger$, Tegan Maharaj$\dagger$, Martin Weiss$\dagger$, Nasim Rahaman$\dagger$, Hannah Alsdurf, Abhinav Sharma, Nanor Minoyan, Soren Harnois-Leblanc, Victor Schmidt, Pierre-Luc St Charles, Tristan Deleu, Andrew Williams, Akshay Patel, Meng Qu, Olexa Bilaniuk, Gaétan Marceau Caron, Pierre Luc Carrier, Satya Ortiz-Gagné, Marc-Andre Rousseau, David Buckeridge, Joumana Ghosn, Yang Zhang, Bernhard Schölkopf, Jian Tang, Irina Rish, Christopher Pal, Joanna Merckx, Eilif B Muller, Yoshua Bengio. COVI-AgentSim: An agent-based model for evaluating methods of digital contact tracing. 

$\dagger$ denotes equal contribution.
\end{quote}

This chapter summarizes \cite{hiads} and the approach of unit-testing, as well as \cite{gupta2020covisim} and the approach of sandboxing, both of which can help address issues with distributional shift in practice.

\if{false}
\subsection {Full citations:}
\begin{quote}
    \singlespacing
\cite{hiads} David Krueger, Tegan Maharaj, Jan Leike.  2019.  Hidden incentives for auto-induced distributional shift. Safe Machine Learning Workshop at the International Conference on Machine Learning (ICML).

\cite{gupta2020covisim} Prateek Gupta$\dagger$, Tegan Maharaj$\dagger$, Martin Weiss$\dagger$, Nasim Rahaman$\dagger$, Hannah Alsdurf, Abhinav Sharma, Nanor Minoyan, Soren Harnois-Leblanc, Victor Schmidt, Pierre-Luc St Charles, Tristan Deleu, Andrew Williams, Akshay Patel, Meng Qu, Olexa Bilaniuk, Gaétan Marceau Caron, Pierre Luc Carrier, Satya Ortiz-Gagné, Marc-Andre Rousseau, David Buckeridge, Joumana Ghosn, Yang Zhang, Bernhard Schölkopf, Jian Tang, Irina Rish, Christopher Pal, Joanna Merckx, Eilif B Muller, Yoshua Bengio. COVI-AgentSim: An agent-based model for evaluating methods of digital contact tracing. 
\end{quote}

$\dagger$ denotes equal contribution.

\subsection{Summary of contributions:}
\fi

In \cite{hiads}, I participated in early discussions of ways to ensure myopic behaviour, and conceptualized our work as unit-testing. I contributed to discussions where we recognized the relevance of myopic behaviour to distributional shift, and coined the term auto-induced distributional shift. I suggested and helped formalize, code, and run experiments in content recommendation, and wrote most sections of the paper relevant to these experiments. I helped run and plot all experiments, research related work, and revise drafts. I created our presentation and poster and many related plots.

My contributions to \cite{gupta2020covisim} began by reading and synthesizing background literature on Covid-19 epidemiology, virology, and contact tracing. Continuing in this role, I co-led the development of our Covid-19 simulator, which we called Covi-AgentSim. From familiarity with ecology literature, I suggested the use of an agent-based model, a key aspect of our work. With a small team, I helped write the first version of the simulator and documentation, including short explanatory documents, and co-led a large group of volunteers: assigning them coding tasks and reviewing and integrating their code. I wrote the majority of code relating to pre-existing medical conditions and disease-specific symptoms, in consultation with literature and relevant experts.

\section{Auto-induced distributional shift and diagnosis via unit-testing}


This work \cite{hiads} defines the term \textbf{auto-induced distributional shift (ADS)} as shift caused by the presence or actions of an algorithm. To illustrate, in Figure \ref{fig:ADS} we show a distribution of users with and without ADS.

\begin{figure}[!htb]\centering
\begin{subfigure}{0.48\linewidth}\centering
\includegraphics[trim=0 0 0 30, clip,  width=\linewidth]{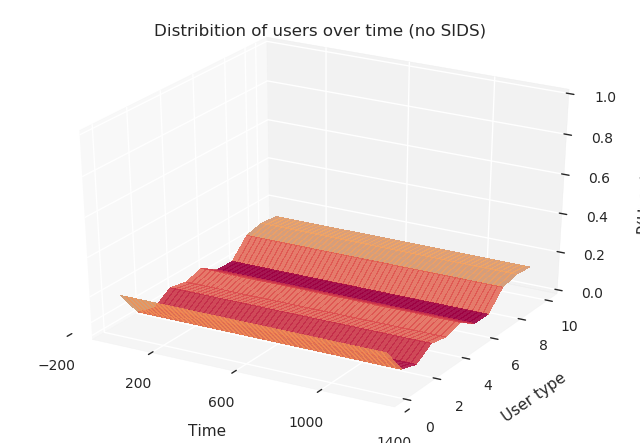}
\end{subfigure}\hfill\begin{subfigure}{0.48\linewidth}\centering
\includegraphics[trim=0 0 0 30, clip, width=\linewidth]{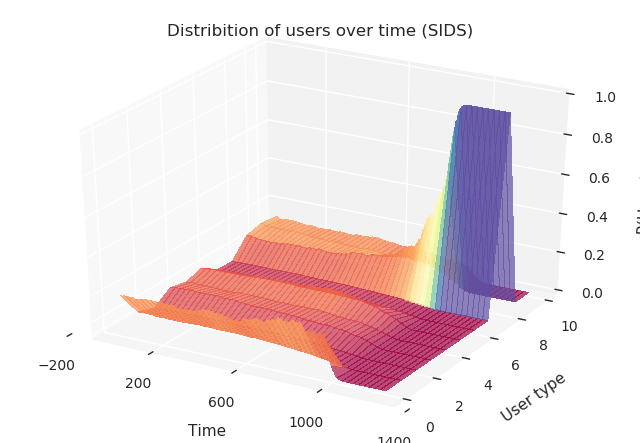}
\end{subfigure}
\caption{
Distributions of users over time.  \textbf{Left}: A distribution which remains constant over time, following the i.i.d assumption. \textbf{Right}: Auto-induced Distributional Shift (ADS) results in a change in the distribution of users in our content recommendation environment. \textbf{Figure reproduced from \cite{hiads}}
}
\label{fig:ADS}
\end{figure}

We note that while ADS are not categorically undesirable, assuming that they will not happen can lead to undesirable and potentially dangerous unexpected behaviour. Typical machine learning algorithms used in content recommendation, by making the iid assumptions, admit the possibility of undesirable effects of ADS, because performance metrics fail to specify whether or not it is acceptable to achieve good performance via ADS, and the performance metrics typically used can incentivise this behaviour. For example, a content recommendation system trained on accuracy is incentivised to learn to better predict user interests, but is also incentivised to drive away hard-to-predict users, and/or to change user interests in order to make preferences easier to predict (because both of these would increase performance according to the metric of accuracy). What we want is for the algorithm to achieve good performance only by the first method, but the learning setup fails to specify this. And the complexity of human preference, together with various human biases (e.g. the illusory truth effect) could mean that the latter two methods are actually easier than the former. 

Using a model content recommendation setting, with a set of user types each of which has a distribution of interests over a set of topics, using an MLP as the recommender system, we show that systems pick up on incentives to drive performance via ADS in Figure 5.2.

\begin{figure}[!htb]
\centering
\includegraphics[trim={0 1mm 0 3mm},clip,width=0.8\linewidth]{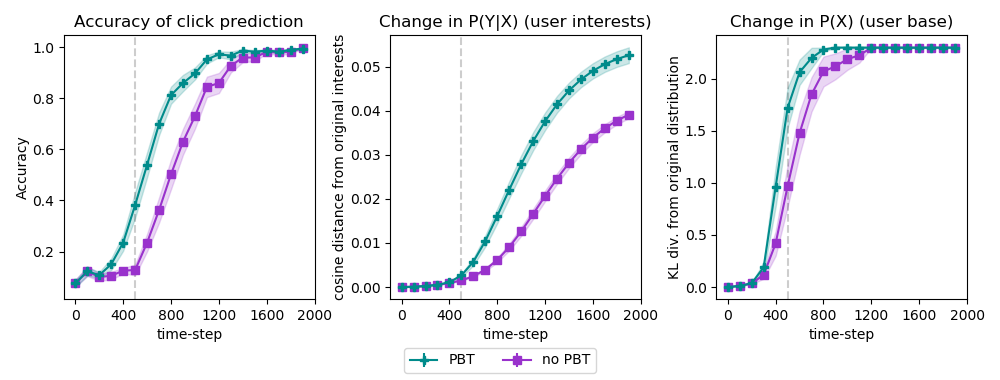}
\caption{
Content recommendation experiments.
\textbf{Left}: using Population Based Training (PBT) increases accuracy of predications faster, leads to a faster and larger drift in users' interests, $P(y|\mathbf{x})$, (\textbf{Center}); as well as the distribution of users, $P(\mathbf{x})$, (\textbf{Right}).
Shading shows std error over 20 runs. \textbf{Figure reproduced from \cite{hiads}}
}
\label{fig:content_rec}
\end{figure}

To address issues like this, when ADS may be undesirable, we propose the approach of unit testing for identifying high-level behaviours of trained models. Specifically, we develop unit tests in the form of a lightweight environment based on a version of an iterated prisoners dilemma (see Table 5.1). Non-myopic behaviour on the unit test indicates that the algorithm may follow incentives to achieve performance via shifting the distribution (`cooperating' with its future self).

\begin{table}[h!]
\caption{
Rewards for the RL unit test. %
Note that the myopic \texttt{defect} action always increases reward at the current time-step, but decreases reward at the next time-step -- the incentive is hidden from the point of view of a myopic learner.
A learner `fails' the unit test if the hidden incentive to cooperate is revealed, i.e.\ if we see more \texttt{cooperate (C)} actions than \texttt{defect (D)}. \textbf{Table reproduced from \cite{hiads}} }

\centering
\begin{tabular}{l|c|c }
 ~  & $a_t = \texttt{D}$ & $a_t = \texttt{C}$ \\
 \hline
\makecell{$s_t = a_{t-1} = \texttt{D}$ } & $-1/2$ & $-1$  \\ 
 \hline %
  \makecell{$s_t = a_{t-1}  = \texttt{C}$ } & $\>\>\>\,1/2$ & $\>\>\>0$\\
\end{tabular}
\label{tab:payoff}
\end{table}

We demonstrate that the unit test correctly identifies the algorithms which cause ADS, as modelled in the simple content recommendation environment described above. We identify elements which make algorithms more likely to fail the unit test, including meta-learning-like algorithms (including population-based training) with optimization over a longer time horizon, and larger population sizes. We also propose a mitigation strategy, of swapping learners between learning environments, and show in practice that this can prevent following incentives to achieve performance via ADS. However this strategy does not eliminate ADS in all situations; highlighting the need for alternative mitigation strategies, or alternative specifications and  learning strategies which take explicit account of (and thereby avoid) undesirable ADS.

\section{Recognizing and addressing distributional shift via sandboxing}

At the beginning of the Covid-19 pandemic, some places had great success controlling spread of disease via exhaustive test-and-trace programmes, which relied on contact tracing. Noting that the scale of the pandemic overwhelmed traditional manual contact tracing approaches, these programmes were supported by the development of \textbf{digital contact tracing}, wherein smartphone apps and other digital devices can be used to supplement manual contact tracing, recognizing close contacts via wifi or bluetooth proximity. Our work recognized that a wealth of other data useful for predicting risk of contracting a disease could be available on smartphones, if distributed inference algorithms could be developed for making use of it in a secure and private way. 

To enable the development of such algorithms, we first developed a detailed agent-based Covid-19 simulator \cite{gupta2020covisim}, in cooperation with epidemiologists, public health experts, virologists, doctors, and other relevant stakeholders. This simulator enabled us to simulate individual-level features (such as pre-existing medical conditions, or symptoms arising from other conditions) calibrated to real demographic, public health, and movement pattern statistics, and also individual-level latent variables (like effective viral load) that could be used as labels for training predictors, modelled according to the most recent available information about Covid-19 viral load progressions and resulting symptoms. In this way we could generate very realistic training data, without any risk of privacy violation. 

Using an agent-based model for the simulator also allowed us to \textbf{sandbox} different contact tracing algorithms: deploy them in a realistic but controlled setting, where we could examine their properties and address potential issues. An example of an issue we encountered is detailed in the following chapter: we trained risk predictors offline, but once deployed, if they are effective for their intended purpose (averting disease), they will induce a distribution of cases different from what was seen during training - i.e. they will cause ADS. We develop a simple method of dealing with this that we call iterative retraining (described in the following paper). 

\Chapter{\uppercase{Predicting infectiousness for proactive contact tracing}} \label{sec:covi}

\begin{quote}
    \singlespacing
    \cite{covidpct} Yoshua Bengio$^\dagger$, Prateek Gupta$^\dagger$, \textbf{Tegan Maharaj}$^\dagger$, Nasim Rahaman$^\dagger$, Martin Weiss$^\dagger$, Tristan Deleu, Eilif Muller, Meng Qu, Victor Schmidt, Pierre-Luc St-Charles, Hannah Alsdurf, Olexa Bilanuik, David Buckeridge, Gáetan Marceau Caron, Pierre-Luc Carrier, Joumana Ghosn, Satya Ortiz-Gagne, Chris Pal, Irina Rish, Bernhard Schölkopf, Abhinav Sharma, Jian Tang, Andrew Williams. 2020. Predicting infectiousness for proactive contact tracing. \textit{International Conference on Learning Representations (ICLR)}.

$\dagger$ denotes equal contribution.
\end{quote} 

\if{false}
\section{Notes}

This chapter is included verbatim from the  paper \cite{covidpct} of which I am co-first author. 

\subsection {Full citation:}

\begin{quote}
    \singlespacing
    Yoshua Bengio$^\dagger$, Prateek Gupta$^\dagger$, \textbf{Tegan Maharaj}$^\dagger$, Nasim Rahaman$^\dagger$, Martin Weiss$^\dagger$, Tristan Deleu, Eilif Muller, Meng Qu, Victor Schmidt, Pierre-Luc St-Charles, Hannah Alsdurf, Olexa Bilanuik, David Buckeridge, Gáetan Marceau Caron, Pierre-Luc Carrier, Joumana Ghosn, Satya Ortiz-Gagne, Chris Pal, Irina Rish, Bernhard Schölkopf, Abhinav Sharma, Jian Tang, Andrew Williams. 2020. Predicting infectiousness for proactive contact tracing. \textit{International Conference on Learning Representations (ICLR)}.

\end{quote} 
$\dagger$ denotes equal contribution.

\subsection{Summary of contributions:}
\fi
In early 2021, I was working on a project using agent-based models (ABMs) to produce training data to learn representations of real-world ecosystems. In the early stages of the pandemic, Yoshua asked the lab if anyone was available to contribute to a large project he was leading called COVI, and I jumped at the opportunity to apply my recent work to this very real-world problem.

My contributions to the COVI project began with epidemiological research for Covi-AgentSim, the agent-based Covid-19 simulator described in the previous section. I became project manager for the simulator and machine learning aspects of the COVI app, onboarding new members to the codebase, scheduling meetings, helping identify coding and experimental priorities, reviewing code, onboarding epidemiologists and other experts to the project,  writing summaries of our work for communication with the rest of the multidisciplinary team, and presenting our work-in-progress. I contributed early on to the framing of risk as predicted infectiousness. I also continued reading and summarizing relevant literature and incorporating it into portions of the code, checking outputs of the simulator with published statistics and relevant experts, developing plots, brainstorming about ML model details, helping formalize our approach as distributed inference and refining notation, and discussing experimental results with the team. I wrote the first draft of our paper, including a survey of related work and a glossary, managed and contributed to revisions, helped identify experiments to be run and helped run them (including trouble-shooting and fixing code), helped design and create figures and plots, managed and helped write responses during review phases, and ultimately re-focused/re-organized and substantially re-wrote much of the draft to arrive at the final text. I wrote the script for our oral presentation, and created and revised our poster.

To train the models in this paper, we leverage the sandbox as developed in the preceeding work, providing several examples of the important phenomena we can discover and address via sandboxing.

This approach also allows us to perform distributed inference in an entirely private way, offline,  as well as providing variables such as viral load which are not feasibly available at scale in real data. This general approach of training and sandboxing predictors with an ABM could be applied to many settings in the real world where individual-level  behaviour is important in determining dynamics, and/or where privacy is of great concern: including climate change and environmental and health applications.

We show that this approach provides an excellent trade-off in managing disease: reducing cases more than any tested method, while requiring very little quarantine of healthy individuals. This latter point is important for preventing and addressing pandemic fatigue, which has been shown to erode trust and compliance with public health measures.


\vspace{0.5cm}
\textbf{Abstract}: \\

The COVID-19 pandemic has spread rapidly worldwide, overwhelming manual contact tracing in many countries and resulting in widespread lockdowns for emergency containment.  Large-scale \textbf{digital contact tracing (DCT)}\footnote{All \textbf{bolded} terms are defined in the Glossary; Appendix 1.} has emerged as a potential solution to resume economic and social activity while minimizing  spread of the virus. Various DCT methods have been proposed, each making trade-offs between privacy, mobility restrictions, and public health.
The most common approach, \textbf{binary contact tracing (BCT)}, models infection as a binary event, informed only by an individual's test results, with corresponding binary recommendations that either all or none of the individual's contacts quarantine. 
BCT ignores the inherent uncertainty in contacts and the infection process, which could be used to tailor messaging to high-risk individuals, and prompt proactive testing or earlier warnings. It also does not make use of observations such as symptoms or pre-existing medical conditions, which could be used to make more accurate infectiousness predictions.  
In this paper, we use a recently-proposed COVID-19 epidemiological simulator to develop and test methods that can be deployed to a smartphone to locally and proactively predict an individual's infectiousness (risk of infecting others) based on their contact history and other information, while respecting strong privacy constraints. Predictions are used to provide personalized recommendations to the individual via an app, as well as to send anonymized messages to the individual's contacts, who use this information to better predict their own infectiousness, an approach we call \textbf{proactive contact tracing (PCT)}. 

\section{Introduction} 
\label{sec:introduction}
Until pharmaceutical interventions such as a vaccine become available, control of the COVID-19 pandemic relies on nonpharmaceutical interventions such as lockdown and social distancing. While these have often been successful in limiting spread of the disease in the short term, these restrictive measures have important negative social, mental health, and economic impacts.  \textbf{Digital contact tracing (DCT)}, a technique to track the spread of the virus among individuals in a population using smartphones, is an attractive potential solution to help reduce growth in the number of cases and thereby allow  more economic and social activities to resume while keeping the number of cases low.
Most currently deployed DCT solutions use \textbf{binary contact tracing (BCT)}, which sends a quarantine recommendation to all recent contacts of a person after a positive test result. 
While BCT is simple and fast to deploy, and most importantly can help curb spread of the disease \cite{lowadoption}, epidemiological simulations by \cite{hinch2020effective} suggest that using
only one bit of information about the infection status can lead to quarantining many healthy individuals while failing to quarantine infectious individuals. Relying only on positive test results as a trigger is also inefficient for a number of reasons:
 (i) Tests have high false negative rates~\cite{li2020false}; 
 (ii) Tests are administered late, only after symptoms appear, leaving the asymptomatic population, estimated 20\%-30\% of cases
 \cite{gandhi2020asymptomatic}, likely untested; 
 (iii) It is estimated that infectiousness is highest \textit{before} symptoms appear, well before someone would get a test~\cite{heneghan2020sars}, thus allowing them to infect others before being traced, 
 (iv) Results typically take at least 1-2 days, and
 (v) In many places, tests are in limited supply. 
 
Recognizing the issues with test-based tracing, \cite{gupta2020covisim} propose a rule-based system leveraging other input clues potentially available on a smartphone (e.g. symptoms, pre-existing medical conditions), an approach they call \textbf{feature-based contact tracing (FCT)}. 
Probabilistic (non-binary) approaches, using variants of belief propagation in graphical models, e.g. \cite{baker2020probabilistic,satorras2020neural, briers2020risk}, could also make use of features other than test results to improve over BCT, although these approaches rely on knowing the social graph, either centrally or via distributed  exchanges between nodes. The latter solution may require many bits exchanged between nodes (for precise probability distributions), which is
 challenging both in terms of privacy and bandwidth. 
Building on these works, we propose a novel FCT methodology we call \textbf{proactive contact tracing (PCT)}, in which we use the type of features proposed by \cite{gupta2020covisim} as inputs to a predictor trained to output \textit{proactive} (before current-day) estimates of expected infectiousness (i.e. risk of having infected others
in the past and of infecting them in the future). 
The challenges of privacy and bandwidth motivated our particular form of \textbf{distributed inference} where we pretrain the predictor offline and do not assume that the messages exchanged are probability distributions, but instead just informative inputs to the node-level predictor of infectiousness.

We use a recently proposed COVID-19 agent-based simulation testbed \cite{gupta2020covisim} called COVI-AgentSim to compare PCT to other contact tracing methods under a wide variety of conditions. We develop deep learning predictors for PCT in concert with a professional app-development company, ensuring inference models are appropriate for legacy smartphones. By leveraging the rich individual-level data produced by COVI-AgentSim to train predictors offline, we are able to perform individual-level infectiousness predictions locally to the smartphone, with sensitive personal data never required to leave the device. We find deep learning based methods to be consistently able to reduce the spread of the disease more effectively, at lower cost to mobility, and at lower adoption rates than other predictors.  These results suggest that deep learning enabled PCT could be deployed in a smartphone app to help produce a better trade-off between the spread of the virus and the economic cost of mobility constraints than other DCT methods, while enforcing strong privacy constraints.

\subsection{Summary of technical contributions}
\begin{enumerate}[leftmargin=0.8cm]
\item We examine the consequential problem of COVID-19 infectiousness prediction and propose a new method for contact tracing, called proactive contact tracing (see Sec. \ref{sec:probabilistic-contact-tracing}).  

\item In order to perform distributed inference with deep learning models, we develop an architectural scaffold whose core is any set-based neural network. 
We embed two recently proposed networks, namely Deep Sets \cite{zaheer2017deep} and Set Transformers \cite{lee2018set} and evaluate the resulting models via the COVI-AgentSim testbed \cite{gupta2020covisim} (see Sec. \ref{sec:distributed_inference}). 

\item To our knowledge the combination of techniques in this pipeline is entirely novel, and of potential interest in other settings where privacy, safety, and domain shift are of concern. Our training pipeline consists of training an ML infectiousness predictor on the domain-randomized output of an agent-based epidemiological model, in several loops of retraining to mitigate issues with (i) non-stationarity and (ii) distributional shift due to predictions made by one phone influencing the input for the predictions of other phones (see Sec. \ref{subsec:training-procedure}).

\item To our knowledge this is the first work to apply and benchmark a deep learning approach for probabilistic contact tracing and infectiousness risk assessment. 
We find such models are able to leverage weak signals and patterns in noisy, heterogeneous data to 
 better estimate infectiousness compared to binary contact tracing and rule-based methods (see Sec. \ref{sec:experiments}).

\end{enumerate}


\section{Proactive Contact Tracing}
\label{sec:probabilistic-contact-tracing}
\textbf{Proactive contact tracing (PCT)} is an approach to digital contact tracing which leverages the rich suite of features potentially available on a smartphone (including information about symptoms, preexisting conditions, age and lifestyle habits if willingly reported) to compute proactive estimates of an individual's expected infectiousness. These estimates are used to (a) provide an individualized recommendation and (b) propagate a graded risk message to other people who have been in contact with that individual (see Fig.~\ref{fig:PCToverview}). This stands in contrast with existing approaches for contact tracing, which are either binary (recommending all-or-nothing quarantine to contacts), or require centralized storage of the contact graph or other transfers of information which are incompatible with privacy constraints in many societies. 
Further, the estimator runs locally on the individual's device, such that any sensitive information volunteered does not need to leave the device. 

In Section~\ref{sec:problem_setup}, we formally define the general problem PCT solves. In Section~\ref{sec:privacy_approach}, we describe how privacy considerations inform and shape the design of the proposed framework and implementation. Finally, in Section~\ref{sec:distributed_inference}, we introduce deep-learning based estimators of expected infectiousness, which we show in Section~\ref{sec:experiments} to outperform DCT baselines by a large margin. 

\begin{figure}[h]
\centering
\includegraphics[width=\textwidth]{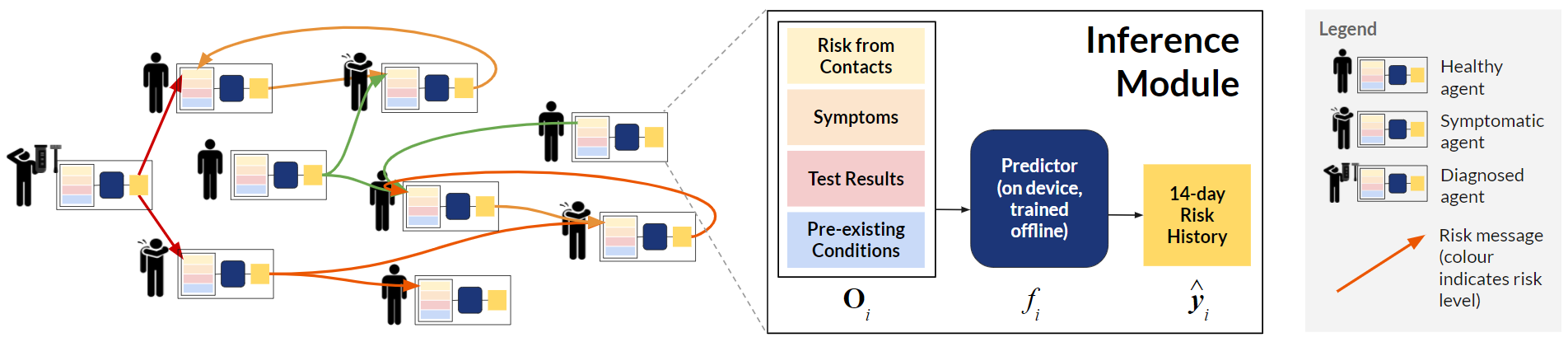}
\caption{\textbf{Proactive contact tracing overview.} Diagram showing \textbf{Left:} the propagation of anonymized, graded (non-binary) risk messages between users and \textbf{Inset:} overview of the inference module deployed to each user's phone. The inference module for agent $i$ takes in observables $\mathbf{O}_i$, and uses a pretrained predictor $f$ to estimate that agent's risk (expected infectiousness)  for each of the last 14 days. Anonymized selected elements of this risk vector are sent as messages to appropriate contacts, allowing them to proactively update their own estimate of expected infectiousness. } 
\label{fig:PCToverview}
\end{figure} 



\subsection{Problem Setup} \label{sec:problem_setup}
%
We wish to estimate \textbf{infectiousness} $y_i^{d'}$ of an agent $i$ on day $d'$, given access to \emph{locally observable information} $\vec{O}_i^d$ now on day $d\geq d'$ and over the past $d_{max}$ days ($d'\geq d-d_{max}$)
Some of the information available on day $d$ is static, including reported age, sex, pre-existing conditions, and lifestyle habits (e.g. smoking), denoted $g_i$, the
health profile. Other information is measured
each day: the health status $h_i^d$ of
reported symptoms and known test results. Finally, $\vec{O}_i^d$
also includes information about encounters in the last
$d_{max}$ days, grouped in $e_i^d$ for day $d$. Thus:
\begin{align}
    \vec{O}_i^d = (g_i, h_i^d, e_i^d, h_i^{d-1}, e_i^{d-1},
    \ldots h_i^{d-d_{max}}, e_i^{d-d_{max}})
\end{align}
 The information $e_i^{d'}$ about the encounters  from day $d'$ is subject to privacy constraints detailed in Section~\ref{sec:privacy_approach} but provides
 indications about the estimated infectiousness of the persons
 encountered at $d'$, given the last available
 information by these contacts as of day $d$, hence
 these contacts try to estimate their past infectiousness a posteriori.
 Our goal is thus to model the \textit{history} of the agent's infectiousness (in the last $d_{max}$ days), which is what enables the recommendations of PCT to be \emph{proactive} and makes it possible for an infected asymptomatic person to receive a warning from their contact even before they develop symptoms, because
 their contact obtained sufficient evidence that they
 were contagious on the day of their encounter.
 Formally, we wish to model 
$P_{\theta}(\vec{y}_i^{d} | \vec{O}_i^{d})$,
%
where $\vec{y}_i^d=(y_i^d,y_i^{d-1},\ldots,y_i^{d-d_{max}})$ is the
vector of present and past infectiousness of agent $i$ and 
$\theta$ specifies the parameters of the predictive
model. In our experiments we only estimate conditional
expectations with a predictor $f_\theta$, with $\vec{\hat y}_i^d = (\hat y_i^d, \dots, \hat y_i^{d - d_{max}}) = f_\theta(\vec{O}_i^d)$ an estimate of the conditional expected per-day present and past infectiousness $\mathbb{E}_{P_\theta}[\vec{y}_i^d | \vec{O}_i^d]$. 
%

The predicted expected values are used in two ways. First, they are used to generate messages transmitted on day $d$ to contacts involved in encounters on day $d' \in (d-d_{max},d)$. 
These messages contain the estimates $\hat{y}_i^{d'}$ of the expected infectiousness of $i$ at day $d'$,  quantized to 4 bits for privacy reasons discussed in section~\ref{sec:privacy_approach}.
Second, the prediction for today $\hat y_i^d$ is also used to form 
a discrete recommendation level $\zeta_i^d \in \{0, 1, \dots, n\}$ regarding the behavior of agent $i$ at day $d$ via a recommendation mapping $\psi$, i.e. $\zeta_i^d = \psi(\hat y_i^d)$. \footnote{The recommendations for each level are those proposed by \cite{gupta2020covisim}, hand-tuned by behavioural experts to lead to a reduction in the number of contacts; details in Appendix 5. Full compliance with recommendations is not assumed.}  
%
At $\zeta_i^d = 0$ agent $i$ is not subjected to any restrictions, $\zeta_i^d = 1$ is baseline restrictions of a post-lockdown scenario (as in summer 2020 in many countries), $\zeta_i^d = n$ is full quarantine (also the behaviour recommended for contacts of positively diagnosed agents under BCT), and intermediate levels interpolate between levels 1 and $n$.

Here we make two important observations about contact tracing: (1) There is a trade-off between decelerating spread of disease, measured by the \textbf{reproduction number $R$}, or as number of cases,\footnote{Note that the the number of cases is highly non-stationary; it grows exponentially over time, even where $R$, which is in the exponent, is constant.} and minimizing the degree of restriction on agents, e.g., measured by the average number of contacts between agents.
Managing this tradeoff is a social choice which involves not just epidemiology but also economics, politics, and the particular weight different people and nations may put on individual freedoms, economic productivity and public health. The purpose of PCT is to improve the corresponding \textbf{Pareto frontier}. A solution which performs well on this problem will encode a policy that contains the infection while applying minimal restrictions on healthy individuals, but it may be that different methods are more appropriate depending on where society stands on that tradeoff. 
(2) A significant challenge comes from the feedback loop between agents: observables $\vec{O}_i^d$ of agent $i$ depend on the predicted infectiousness histories and the pattern of contacts generated by the behavior $\zeta_j$ of \textit{other agents} $j$. This is compounded by privacy restrictions that prevent us from knowing which agent sent which message; we discuss our proposed solution in the following section. 

\vspace{-0.25 cm}

\subsection{Privacy-preserving PCT}\label{sec:privacy_approach}
\vspace{-0.25 cm}

One of the primary concerns with the transmission and centralization of contact information is that someone with malicious intent could identify and track people. Even small amounts of personal data could allow someone to infer the identities of individuals by cross-referencing with other sources, a process called \textbf{re-identification} \cite{el2011systematic}.  Minimizing the number of bits being transmitted
and avoiding information that makes it easy to triangulate people is a protection
against \textbf{big brother attacks}, where a central authority with access to the data
can abuse its power, as well as against \textbf{little brother attacks}, where malicious
individuals (e.g. someone you encounter) could use your information against you.
Little brother attacks include \textbf{vigilante attacks}: harassment, violence, hate crimes, or stigmatization against individuals which could occur if infection status was revealed to others. Sadly, there have been a number of such attacks related to COVID-19 \cite{russell2020rise}. To address this, PCT operates with 
(1) no central storage of the contact graph; 
(2) de-identification \cite{10.1142/S0218488502001648} and encryption of all data leaving the phones;
(3) informed and optional consent to share information (only for the purpose of improving the predictor); and 
(4) distributed inference which can achieve accurate predictions without the need for a central authority. 
Current solutions~\cite{DCT-list} share many of these goals, but are restricted to only binary CT. Our solution achieves these goals while providing individual-level graded recommendations.

Each app records \textbf{contacts} which are defined by
\cite{canada-public-health-prolonged-exposure} as being an encounter which lasts at least 15 minutes with a proximity under 2 meters. Once every 6 hours, each application processes contacts, predicts infectiousness for days $d$ to $d_{max}$, and may update its current recommended behavior. If the newly predicted infectiousness history differs from the previous prediction on some day (e.g. typically because the agent enters new symptoms, receives a negative or positive test result, or receives a significantly updated infectiousness estimate), then the app creates and sends small update messages to all relevant contacts. These heavily quantized messages reduce the network bandwidth required for message passing as well as provide additional privacy to the individual. We follow the communication and security protocol for these messages introduced by \cite{whitepaper}. We note that a naive method would tend to \textit{over-estimate} infectiousness in these conditions, because repeated encounters with the same person (a very common situation, for example people living in the same household) should carry a \textit{lower} predicted infectiousness than the same number of encounters with different people. To mitigate this over-estimation while not identifying anyone, messages are clustered based on time of receipt and risk level as in  \cite{gupta2020covisim}.
\section{Methodology for Infectiousness Estimation}
\label{sec:methodology-for-infectiousness-estimation}
\subsection{Distributed Inference} \label{sec:distributed_inference}
Distributed inference networks seek to estimate marginals by passing messages between nodes of the graph to make
predictions which are consistent with each other (and with the local evidence available at each node). Because messages in our framework are highly constrained by privacy concerns in that we are prevented from passing distributions or continuous values between identifiable nodes, typical distributed inference approaches (e.g. loopy belief propagation \cite{murphy1999loopy}, variational message passing \cite{winn2005variational}, or expectation propagation \cite{Minka2001_Expectation_Propagation}) are not readily applicable. We instead propose an approach to distributed inference which uses a trained ML predictor for estimating these marginals
(or the corresponding expectations). The predictor $f$ is trained offline on simulated data to predict expected infectiousness from the locally available 
information $\vec{O}_i^d$ for agent $i$ on day $d$. The choice of ML predictor is informed by several factors. First and foremost, we expect the input $e_i^d$ (a component of $\mathbf{O}_i^d$) to be variable-sized, because we do not know \emph{a priori} how many contacts an individual will have on any day. Further, privacy constraints dictate that the contacts within the day cannot be temporally ordered, implying that the quantity $e_i^d$ is set-valued. Second, given that we are training the predictor on a large amount of domain-randomized data from many epidemiological scenarios (see Section~\ref{subsec:training-procedure}), we require the architecture to be sufficiently expressive. Finally, we require the predictor to be easily deployable on edge devices like legacy smartphones.

To these ends, we construct a general architectural scaffold in which any neural network that maps between sets may be used. In this work, we experiment with Set Transformers \cite{lee2018set} and a variant of Deep Sets \cite{zaheer2017deep}. The former is a variant of Transformers \cite{vaswani2017attention}, which model pairwise interactions between all set elements via multi-head dot-product attention and can therefore be quite expressive, but scales quadratically with elements. The latter relies on iterative max-pooling of features along the elements of the set followed by a broadcasting of the aggregated features, and scales linearly with the number of set elements (see Figure~\ref{fig:model-architecture}).

In both cases, we first compute two categories of embeddings: per-day and per-encounter. Per-day, we have embedding MLP modules $\phi_{h}$ of health status, $\phi_{g}$ for the health profile, and a linear map $\phi_{\delta d}$ for the day-offset $\delta d = d - d'$. 
Per-encounter, the cluster matrix $e_i^d$ can be expressed as the set $\{(r^{d}_{j \to i}, n^{d}_{j \to i})\}_{j \in K_i^d}$ where $j$ is in the set $K_i^d$ of putative anonymous persons encountered by $i$ at $d$, $r_{j \to i}$ is the risk level sent from $j$ to $i$ and $n_{j \to i}$ is the number of repeated encounters between $i$ and $j$ at $d$. Accordingly, we define the risk level embedding function $\phi_e^{(r)}$ and $\phi_e^{(n)}$ for the number of repeated encounters. $\phi_e^{(r)}$ is parameterized by an embedding matrix, while $\phi_e^{(n)}$ by the vector 
%
%
%
\begin{align}
(\phi_{e}^{(n)}(n))_{2i} = \sin{\left(\nicefrac{n}{10000^i}\right)} \; \text{and} \; (\phi_{e}^{(n)}(n))_{2i + 1} = \cos{\left(\nicefrac{n}{10000^i}\right)}
\end{align}
which resembles the positional encoding in \cite{vaswani2017attention} and counteracts the spectral bias of downstream MLPs \cite{rahaman2019spectral,mildenhall2020nerf}. We tried alternative encoding schemes like thermometer encodings \cite{buckman2018thermometer}, but they did not perform better in our experiments. Next, we join the per-day and per-encounter embeddings to obtain $\mathcal{D}_i^d$ and $\mathcal{E}_i^d$ respectively. Where $\otimes$ is the concatenation operation, we have: 
\begin{align}
\label{eq:joined-embeddings}
\mathcal{D}_{i}^{d} &= \{ \phi_{h}(h_{i}^{d'}) \otimes \phi_{g}(g_i) \otimes \phi_{\delta d}(d' - d) \,|\, d' \in \{d,\dots,d-d_{max}\} \} \\
\mathcal{E}_{i}^{d} &= \{ \phi_{e}^{(r)}(r^{d'}_{j \rightarrow i}) \otimes \phi_{e}^{(n)}(n^{d'}_{j \rightarrow i}) \otimes \phi_{h}(h_{i}^{d'}) \otimes \phi_{\delta d}(d' - d) \,|\, j \in K_i^{d'}, \, \forall \, d' \in \{d,\dots,d-d_{max}\} \} \nonumber
\end{align}
The union of $\mathcal{D}_i^d$ and $\mathcal{E}_i^d$ forms the input to a set neural network $f_{S}$, that predicts infectiousness:
\begin{align}
\vec{\hat{y}}_i^d = f_{S}(\mathcal{O}_i^d) \text{ where } \mathcal{O}_i^d = \mathcal{D}_i^d \cup \mathcal{E}_i^d \text{ and } \vec{\hat{y}}_i^d \in \mathbb{R}_{+}^{d_{max}}
\end{align}

\begin{figure}[t]
\centering
\begin{subfigure}[c]{0.52\textwidth}
  \centering
  \includegraphics[width=\linewidth]{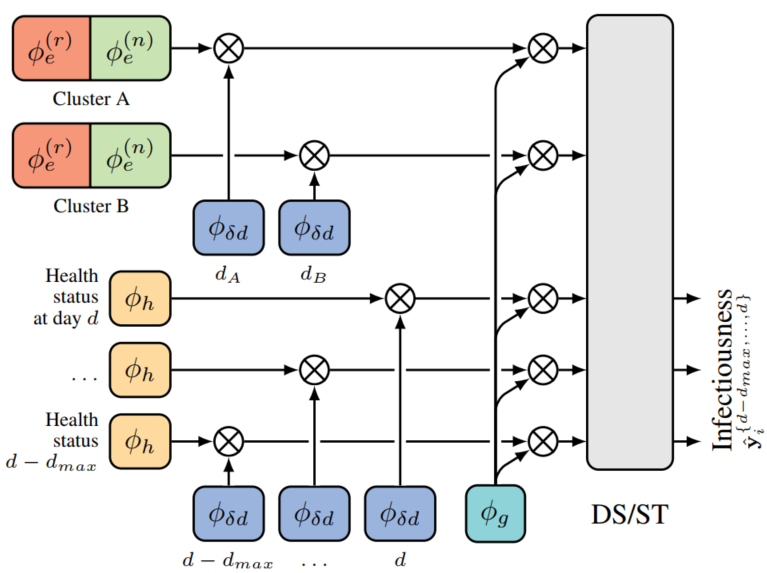}
  \label{fig:model-architecture-1}
\end{subfigure}\hfill
\begin{subfigure}[c]{0.47\textwidth}
  \centering
  \includegraphics[width=\linewidth]{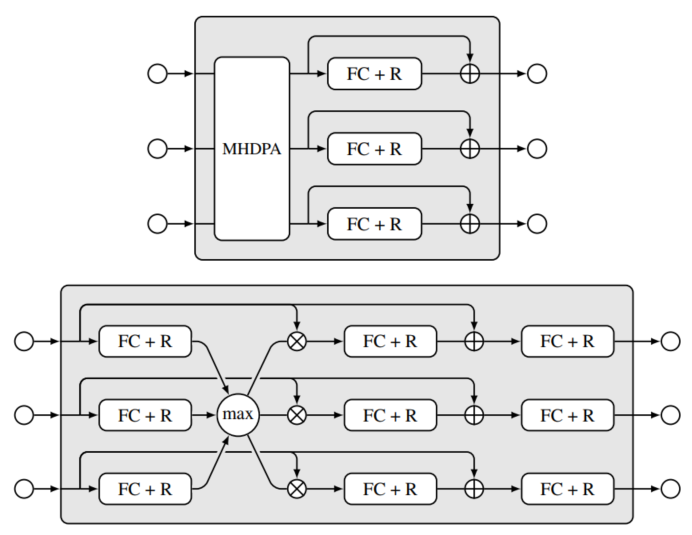}\\[5pt]
  \label{fig:model-architecture-2}
\end{subfigure}
\caption{
\textbf{PCT model architecture.} Diagram showing \textbf{Left}: The embedding network combining pre-existing conditions \protect\tikz\protect\draw[fill=BlueGreen!50] circle(0.8ex);, day offsets \protect\tikz\protect\draw[fill=NavyBlue!40] circle(0.8ex);, risk message clusters \protect\tikz\protect\draw[fill=Red!50] circle(0.8ex); \protect\tikz\protect\draw[fill=YellowGreen!50] circle(0.8ex);, and symptoms information \protect\tikz\protect\draw[fill=Dandelion!50] circle(0.8ex); to be fed into a stack of 5 either Deep-Set (DS) or Set Transformer (ST) blocks  \protect\tikz\protect\draw[fill=gray!20] circle(0.8ex);. \textbf{Right-top}: (ST) self-attention block featuring Multi-Head Dot Product Attention (MHDPA), Fully-Connected layers with ReLU (FC + R) and residual connections. \textbf{Right-bottom:} (DS) set processing block. Here, the $\otimes$ denotes concatenation and the $\oplus$ addition operation. 
}
\label{fig:model-architecture}
\end{figure}


The trunk of both models -- deep-set (DS-PCT) and set-transformer (ST-PCT) -- is a sequence of 5 set processing blocks  (see Figure \ref{fig:model-architecture}). 
A subset of outputs from these blocks (corresponding to $\mathcal{D}_i^d$) is processed by a final MLP to yield $\mathbf{\hat y}_i^d$.
As a training objective for agent $i$, we minimize the Mean Squared Error (MSE) between $\vec{y}_i^{d}$ (which is generated by the simulator) and the prediction $\hat{\vec{y}}_i^d$. We treat each agent $i$ as a sample in batch to obtain the sample loss
%
$L_i = \text{MSE}(\vec{y}_i^d, \vec{\hat{y}}_i^d) = \frac{1}{d_{max}}\sum_{d'= d - d_{max}}^{d}(y_i^{d'}-\hat{y}_i^{d'})^2.$
The net loss is the sum of $L_i$ over all agents $i$. 
%
\subsection{Training Procedure}
\label{subsec:training-procedure}
Unlike existing contact tracing methods like BCT and the NHSx contact tracing app \cite{briers2020risk} that rely on simple and hand-designed heuristics to estimate the risk of infection, our core hypothesis is that methods using a machine-learning based predictor can learn from patterns in rich but noisy signals that might be available locally on a smartphone. 
In order to test this hypothesis prior to deployment in the real world, we require a simulator built with the objective to serve as a testbed for app-based contact tracing methods. We opt to use COVI-AgentSim \cite{gupta2020covisim}, which features an agent specific virology model together with realistic agent behaviour, mobility, contact and app-usage patterns. 
With the simulator, we generate large datasets of $\mathcal{O}(10^7)$ samples comprising the input and target variables defined in sections \ref{sec:problem_setup} and \ref{sec:distributed_inference}. Most importantly for our prediction task, this includes a continuous-valued infectiousness parameter as target for each agent. 

We use $240$ runs of the simulator to generate this dataset, where each simulation is configured with parameters randomly sampled from selected intervals (See Appendix \ref{app:training}). These parameters include the adoption rate of the contact tracing app, initial fraction of exposed agents in the population, the likelihood of agents ignoring the recommendations, and strength of social distancing and mobility restriction measures. 
There are two reasons for sampling over parameter ranges: First, the intervals these parameters are sampled from reflect both the intrinsic uncertainty in the parameters, as well as the practical setting and limitations of an app-based contact tracing method (e.g. we only care about realistic levels of app usage). Second, randomly sampling these key parameters significantly improves the diversity of the dataset, which in turn yields predictors that can be more robust when deployed in the real world. The overall technique resembles that of \textbf{domain-randomization} \cite{tobin2017domain, sadeghi2016cad2rl} in the Sim2Real literature, where its efficacy is well studied for transfer from RGB images to robotic control, e.g. \cite{chebotar2018closing, openai2018learning}. 


We use $200$ runs for training and the remaining $40$ for validation (full training and other reproducibility details are in Appendix \ref{app:experimental_details}).
The model with the best validation score is selected for \emph{online} evaluation, wherein we embed it in the simulator to measure the reduction in the $R$ as a function of the average number of contacts per agent per day. While the evaluation protocol is described in section~\ref{sec:experiments}, we now discuss how we mitigate a fundamental challenge that we share with offline reinforcement learning methods, that of \textbf{auto-induced distribution shift} when the model is used in the simulation loop \cite{levine2020offline, hiads}.

To understand the problem, consider the case where the model is trained on simulation runs where PCT is driven by  ground-truth infectiousness values, i.e. an \emph{oracle} predictor. 
The predictions made by an oracle will in general differ from ones made by a trained model, as will resulting dynamics of spread of disease. This leads to a distribution over contacts and epidemiological scenarios that the model has not encountered during training. 
%
For example, oracle-driven PCT might even be successful in eliminating the disease in its early phases -- a scenario unlikely to occur in model-driven simulations. For similar reasons, oracle-driven simulations will also be less diverse than model-driven ones.  
To mitigate these issues, we adopt the following strategy: 
First, we generate an initial dataset with simulations driven by a \emph{noisy-oracle}, i.e. we add multiplicative and additive noise to the ground-truth infectiousness to partially emulate the output distribution resulting from trained models. The corresponding noise levels are subject to domain-randomization (as described above), resulting in a dataset with some diversity in epidemiological scenarios and contact patterns. Having trained the predictor on this dataset (until early-stopped), we generate another dataset from simulations driven by the thus-far trained predictor, in place of the noisy-oracle.
We then fine-tune the predictor on the new dataset (until early stopped) to obtain the final predictor. This process can be repeated multiple times, in what we call \textbf{iterative retraining}. We find  three steps yields a good trade-off between performance and compute requirement. 

\label{sec:agent-based-epi-simulator}
\section{Experiments}
\label{sec:experiments}
We evaluate the proposed PCT methods, \textbf{ST-PCT }and \textbf{DS-PCT}, and benchmark them against: test-based BCT; a rule-based FCT \textbf{Heuristic} method proposed by \cite{gupta2020covisim}\footnote{While we did experiment with additional methods like linear regression and MLPs, they did not improve the performance over the rule-based FCT heuristic proposed by \cite{gupta2020covisim}.};
and a baseline \textbf{No Tracing (NT)} scenario which corresponds to recommendation level 1 (some social distancing).


\textit{\underline{\textbf{EXP1}}:} In Figure~\ref{fig:paretos}, 
we plot the Pareto frontier between spread of disease ($R$) and the amount of restriction imposed. To traverse the frontier at 60\% adoption rate, we sweep through multiple values of the simulator's \textbf{global mobility scaling factor}, a parameter which controls the strength of social distancing and mobility restriction measures. 
Each method uses $3000$ agents for $50$ days with $12$ random seeds, and we plot the resulting number of contacts per day per human.
We fit a Gaussian Process Regressor to the simulation outcomes and highlight the average reduction in $R$ obtained in the vicinity of $R \approx 1$, finding 
DS-PCT and ST-PCT reduce $R$ to below 1 at a much lower cost to mobility on average. 
For subsequent plots, we select the number of contacts per day per human such that the no-tracing baseline yields a realistic post-lockdown $R$ of around $1.2$ \cite{brisson2020}\footnote{To estimate this number, we use the GP regression fit in figure~\ref{fig:paretos} and consider the $x$ value for which the mean of the NT process is at $1.2$. We find an estimate at $5.61$, and select only the runs yielding effective number of contacts that lie within the interval $(5.61 - 0.5, 5.61 + 0.5)$.}.

\begin{figure}[htp]
\centering
\begin{subfigure}{.49\textwidth}
  \centering
  \includegraphics[width=1\linewidth]{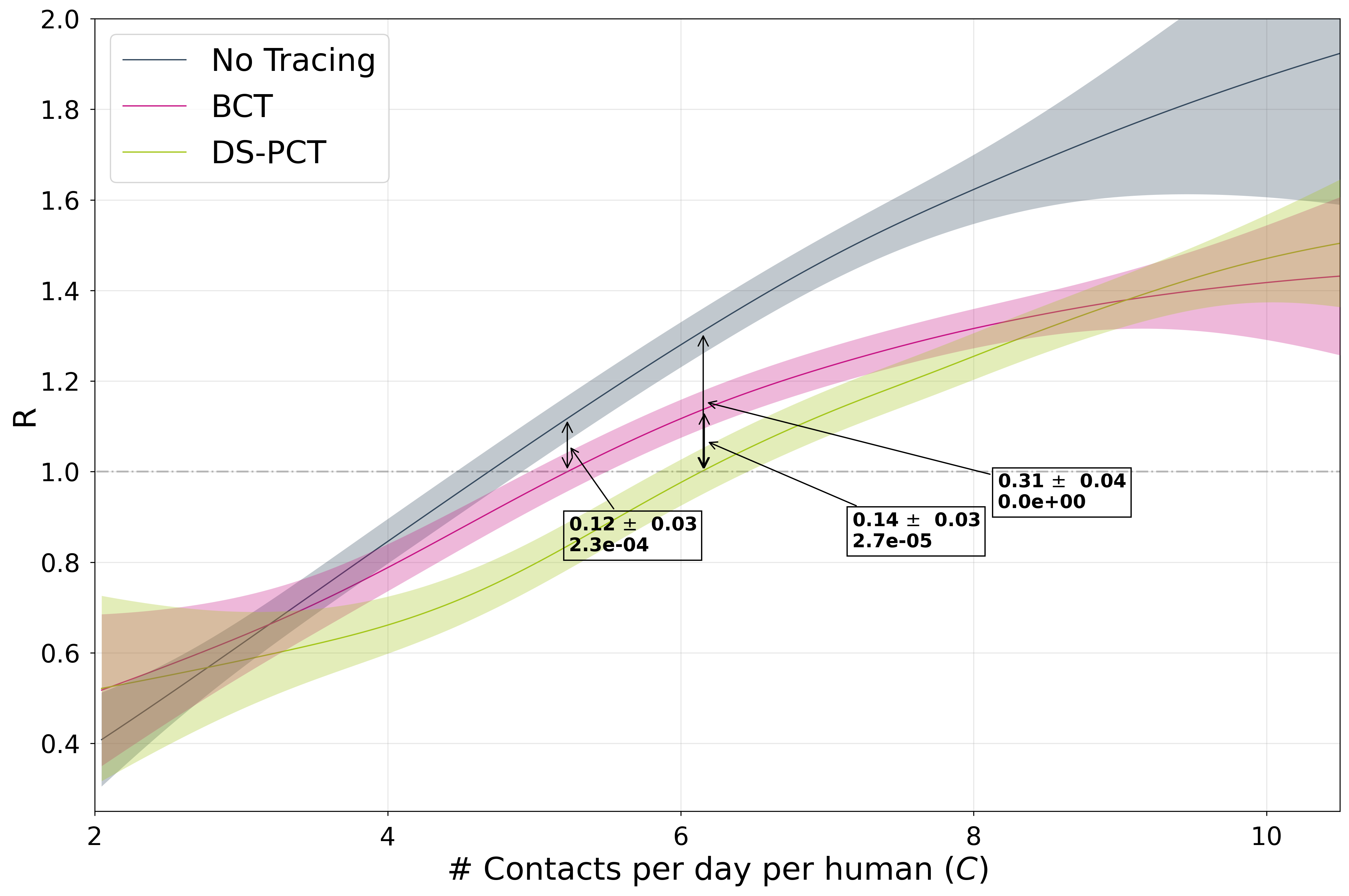}
  \label{fig:ds_pareto}
\end{subfigure}
\begin{subfigure}{.49\textwidth}
  \centering
  \includegraphics[width=1\linewidth]{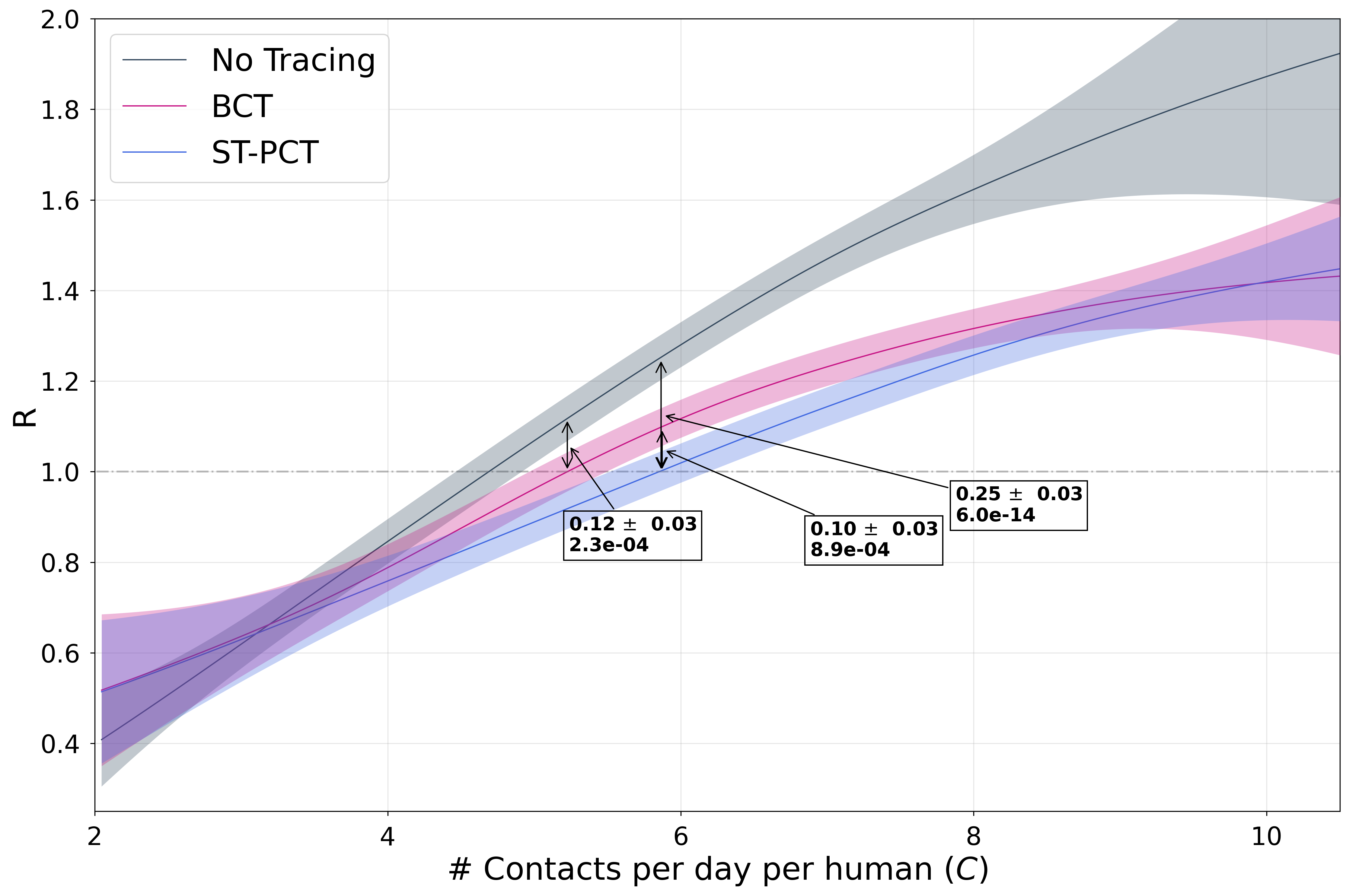}
  \label{fig:st_pareto}
\end{subfigure}
\caption{\textbf{Pareto frontier of mobility and disease spread:} reproduction number $R$ as a function of mobility. In boxes, average difference +/- standard error, p-value under null hypothesis of no difference. 
Note that small differences in $R$ over time produce large changes to the number of cases. 
\textbf{Gist:} All methods span a wide range of $R$ values in their Pareto frontier, including values for the No Tracing scenario which have $R$ below 1, achieved by imposing strong restrictions on mobility (e.g. a lockdown). 
However DCT methods are able to reduce $R$ at much lower cost to average mobility. Compared to NT, BCT has a 12\% advantage in $R$, DS-PCT (\textbf{left}) 31\% and ST-PCT 25\% (\textbf{right)}.
}
\label{fig:paretos}
\end{figure}

\textit{\underline{\textbf{EXP2}}:} Figure \ref{fig:case-counts} compares various DCT methods in terms of their cumulative cases and their fraction of \textbf{false quarantine}:  the number of healthy agents the method wrongly recommends to quarantine.
Again, all DCT methods have a clear advantage over the no-tracing baseline, while the number of agents wrongly recommended quarantine is much lower for ML-enabled PCT. \footnote{Note that the baseline no tracing method also has false quarantines, because household members of an infected individual are also recommended quarantine, irrespective of whether they are infected.} 
\begin{figure}[htp]
\centering
\begin{subfigure}{.49\textwidth}
\includegraphics[width=1\linewidth]{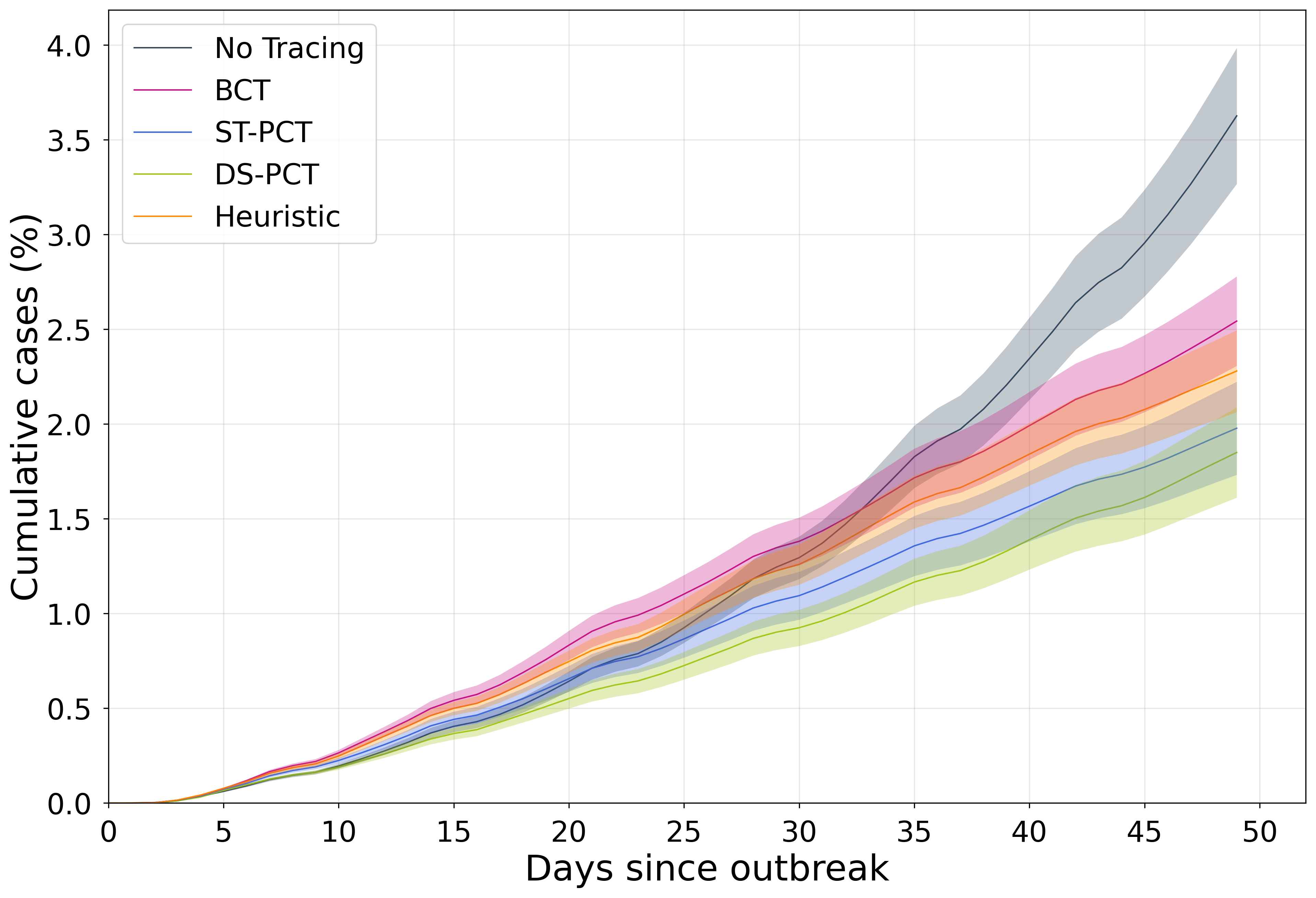}
\end{subfigure}
\begin{subfigure}{.49\textwidth}
\includegraphics[width=1\linewidth]{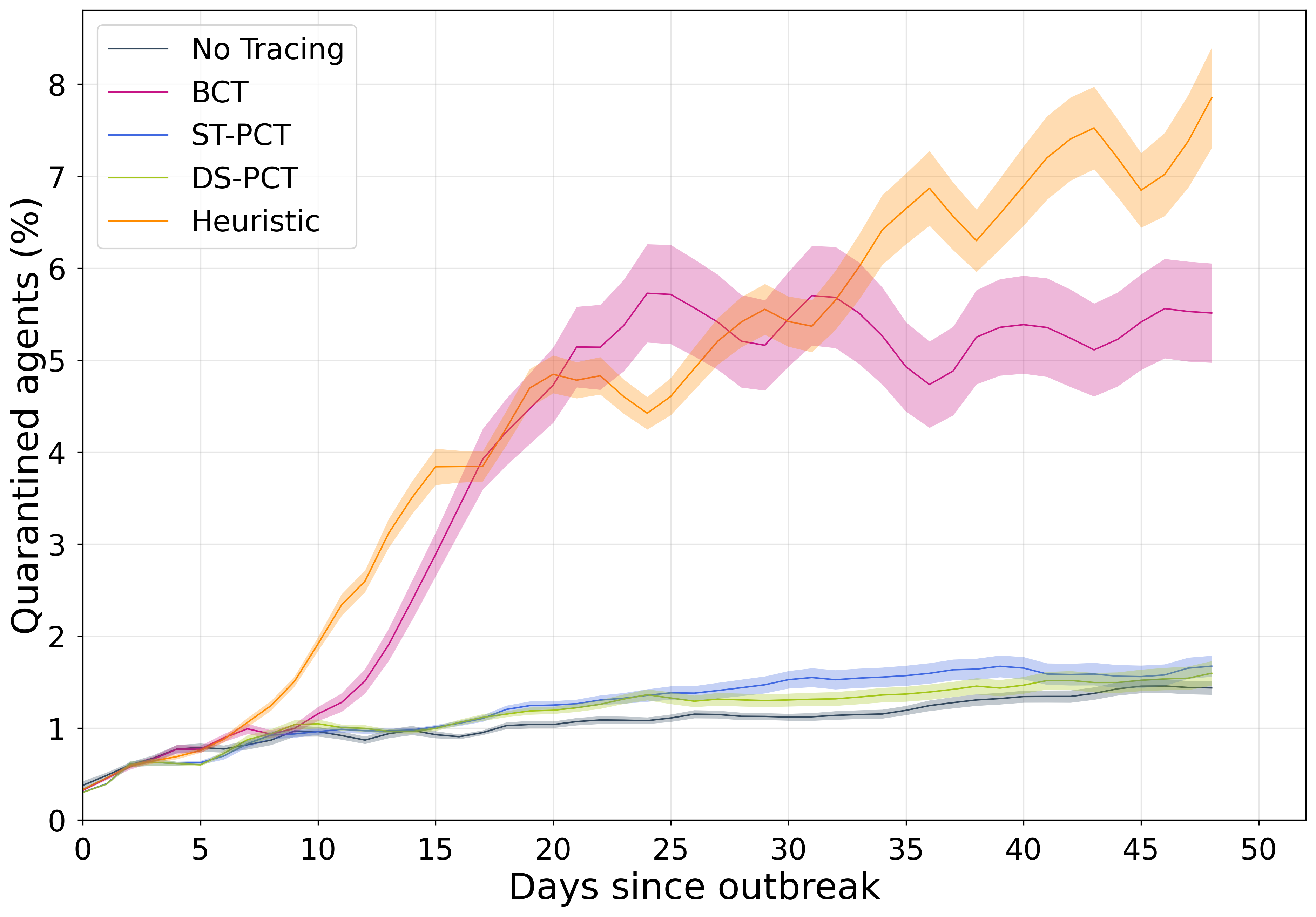}
\end{subfigure}
\caption{\textbf{Left:} Cumulative case counts for each method, 60\% adoption, 50 days, with all runs normalized to $5.61 \pm 0.5$ effective contacts per day per agent. \textbf{Right:} Mobility restriction for the same experiments (fraction of quarantines). \textbf{Gist:} For a similar number of cumulative cases to other DCT methods (left), PCT methods impose very little mobility restriction (right), close to NT.}
\label{fig:case-counts}    
\end{figure}
%

\textit{\underline{\textbf{EXP3}}:} In Figure~\ref{fig:comparisons} \textbf{Left} we plot the bootstrapped distribution of mean $R$ for different DCT methods. Recall $R$ is an estimate of how many other agents an infectious agent will infects,
i.e. even a numerically small improvement in $R$ could yield an exponential improvement in the number of cases. 
We find that both PCT methods yield a clear improvement over BCT and the rule-based heuristic, all of which  significantly improve over the no-tracing baseline. We hypothesize this is because deep neural networks are better able to capture the non-linear relationship between features available on the phone, interaction patterns between agents, and individual infectiousness. 

\textit{\underline{\textbf{EXP4}}:} In Figure~\ref{fig:comparisons} \textbf{Right} we investigate the effect of \textbf{iterative retraining} on the machine-learning based methods, DS-PCT and ST-PCT. We evaluate all 3 iterations of each method in the same experimental setup as in Figure~\ref{fig:comparisons} and find that DS-PCT benefits from the 3 iterations, while ST-PCT saturates at the second iteration. We hypothesize that this is a form of overfitting, given that the set transformer (ST) models all-to-all interactions and is therefore more expressive than deep-sets (DS).
\begin{figure}[h!]
\centering
\begin{subfigure}{.49\textwidth}
\includegraphics[width=1\linewidth]{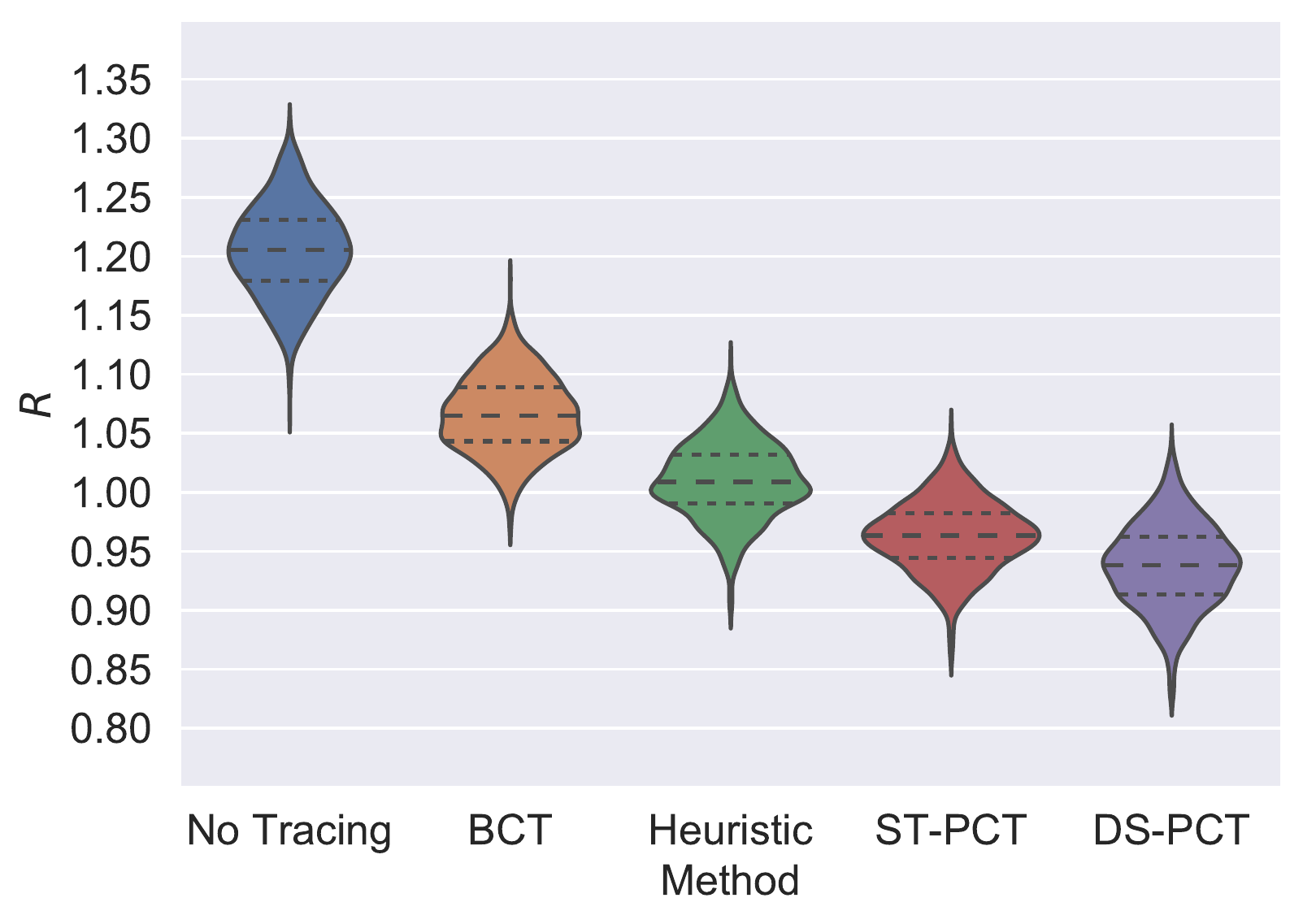}
\label{fig:_comparisons}    
\end{subfigure}
\begin{subfigure}{.49\textwidth}
\includegraphics[width=1\linewidth]{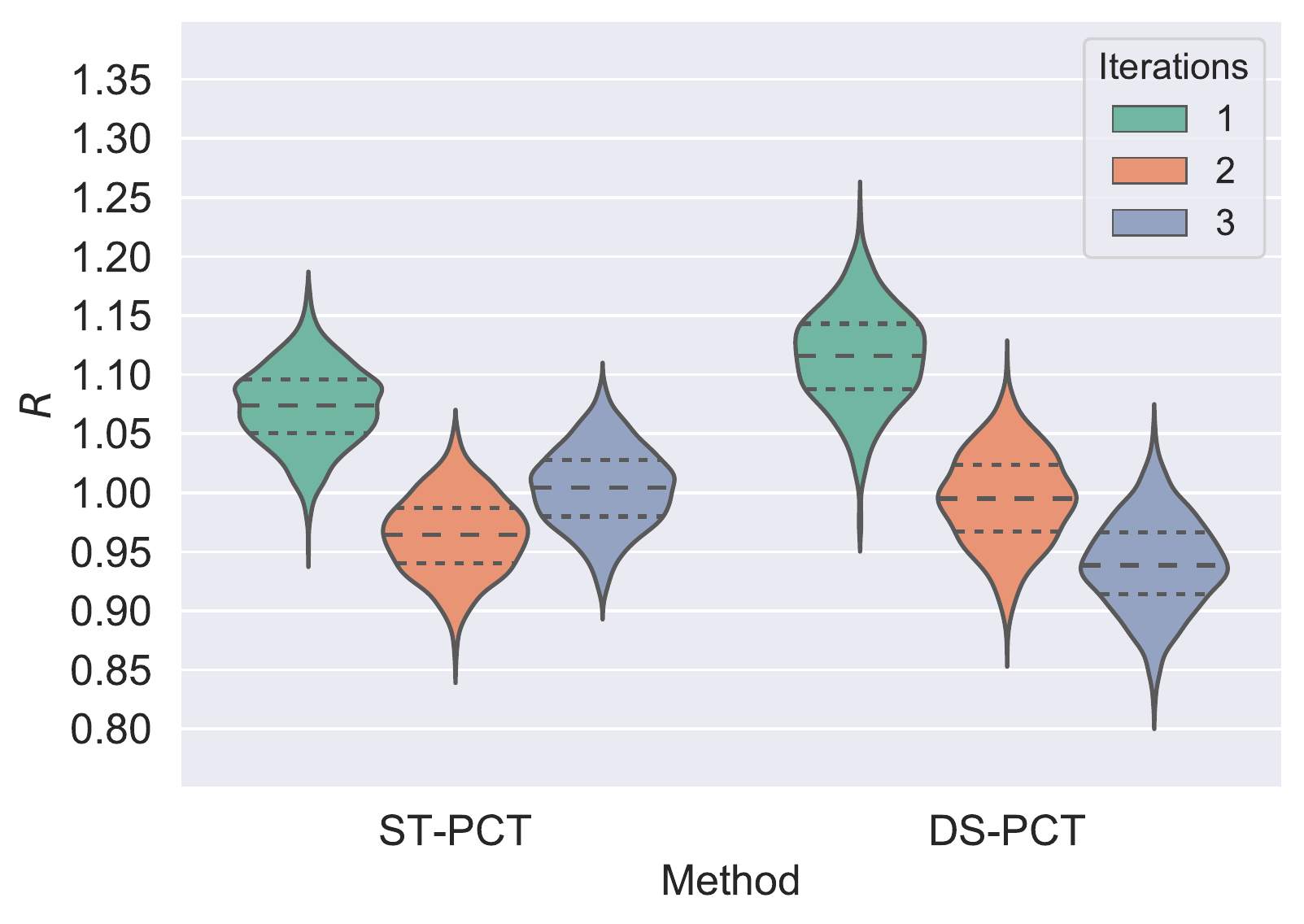}
\label{fig:_iterations}
\end{subfigure}
\caption{\textbf{Method Comparison and Retraining}. \textbf{Left:} Distribution (median and quartiles) of $R$ at 60\% adoption where the ML methods outperform the Heuristic and BCT methods proposed by \cite{gupta2020covisim}. \textbf{Right:} Iterative re-training significantly improves model performance. Upon the second re-training, however, it seems that ST-PCT begins to overfit. }
\label{fig:comparisons}
\end{figure}
%
%

%
%

\textbf{\textit{\underline{EXP5}}:}  In Figure~\ref{fig:adoption}, we analyze the sensitivity of the various methods to adoption rate, which measures what percent of the population actively uses the CT app. Adoption rate is an important parameter for DCT methods, as it directly determines the effectiveness of an app. We visualize the effect of varying the adoption rate on the reproduction number $R$. Similarly to prior work \cite{lowadoption}, we find that all DCT methods improve over the no-tracing baseline even at low adoptions
and PCT methods dominate at all levels of adoption.
    
\begin{figure}[h ]
    {\caption{\textbf{Adoption rate comparison}. We compare  all methods for adoption rates between 0\% (NT) and 60\%. 
    \textbf{Gist:} All methods are able to improve over NT, even at low adoption rates. At 30\% and 45\%, ST-PCT performs the best by a relatively wide margin while DS-PCT outperforms it at 60\%. 
    }\label{fig:adoption}}
    {\includegraphics[width=0.95\textwidth]{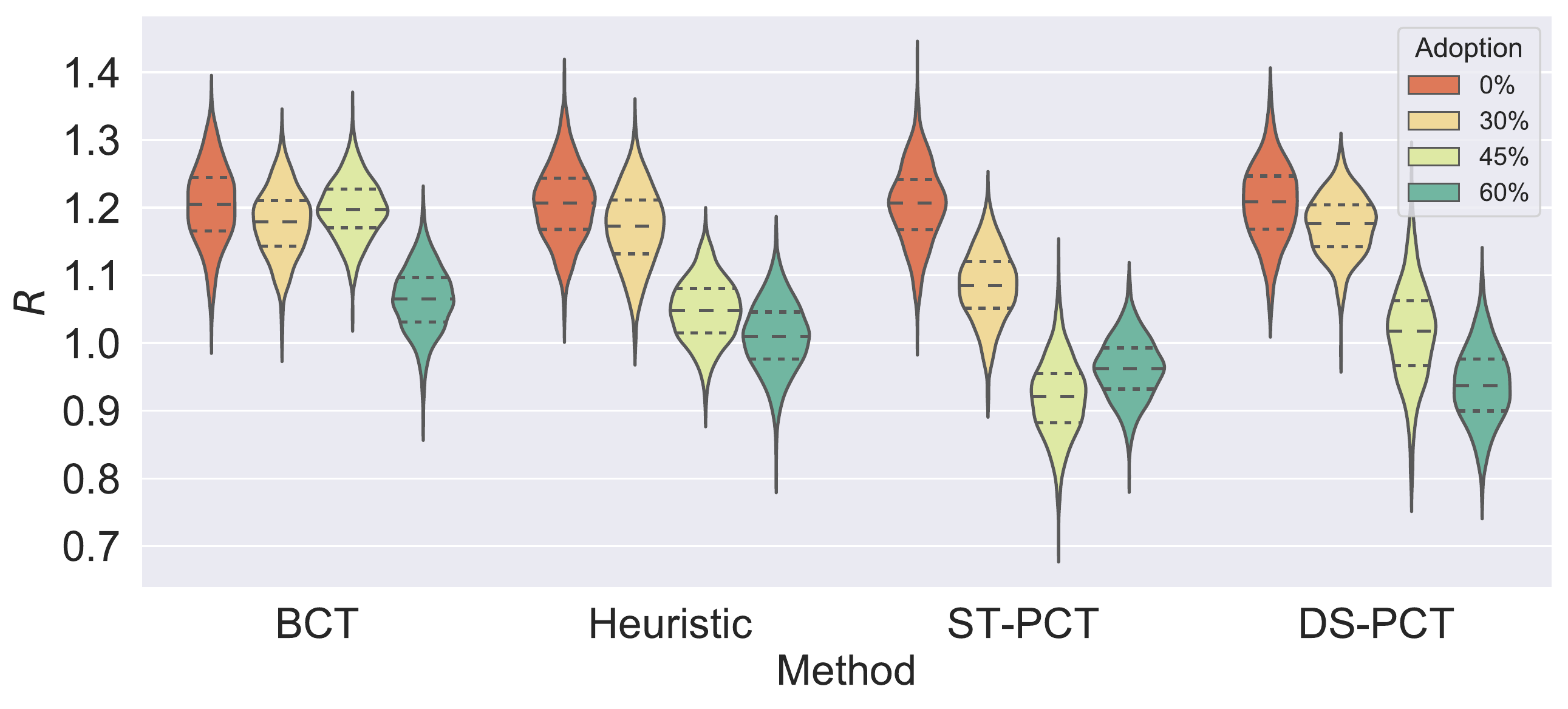}}
    
\end{figure}
 \FloatBarrier
\section{Conclusion}
Our results demonstrate the potential benefit of digital contact tracing approaches for saving lives while reducing mobility restrictions and preserving privacy, independently confirming previous reports \cite{ferretti2020quantifying, lowadoption}.
Of all methods in our study,  
we find that deep learning based PCT provides the best trade-off between restrictions on mobility and reducing the spread of disease under a range of settings, making it a potentially powerful tool for saving lives in a safe deconfinement.
%
%
This area of research holds many interesting avenues of future work in machine learning, including: (1) The comparison of methods for fitting parameters of the epidemiological simulator data using spatial information (not done here to avoid the need for GPS), (2) Using reinforcement learning to learn (rather than expert hand-tune) the mapping from estimated infectiousness and individual-level features (number of contacts, age, etc.), and thereby target desirable outcomes for heterogeneous populations.  
%

Accurate and practical models for contact tracing are only a small part of the non\hyp{}pharmaceutical efforts against the pandemic, a response to which is a complex venture necessitating cooperation between public health, government, individual citizens, and scientists of many kinds - epidemiologists, sociologists, behavioural psychologists, virologists, and machine learning researchers, among many others. We hope this work can play a role in fostering this necessary collaboration.

\section*{Acknowledgements}
\vspace{-4pt}
We thank all members of the broader COVI project \url{https://covicanada.org} for their dedication and teamwork; it was a pleasure and honour to work with such a great team. We also thank the Mila and Empirical Inference communities, in particular Xavier Bouthillier, Vincent Mai, Gabriele Prato, and Georgios Arvanitidis, for detailed and helpful feedback on early drafts.
The authors gratefully acknowledge the following funding sources: NSERC, IVADO, CIFAR.
This project could not have been completed without the resources of MPI-IS cluster, Compute Canada \& Calcul Quebec, in particular the Beluga cluster. In gratitude we have donated to a project to help understand and protect the St. Lawrence beluga whales for whom the cluster is named \url{https://baleinesendirect.org/}

\newpage

\newpage
Appendix
\vspace{1cm}
\section{Glossary of bolded terms} \label{app:glossary}

\paragraph{\textbf{Contact tracing}} Finding the people who have been in contact with an infected person, typically using information such as test results and phone surveys, and recommending that they quarantine themselves to prevent further spread of the disease.

\paragraph{\textbf{Manual contact tracing} } Method for contact tracing using trained professionals  who interview diagnosed individuals to identify people that have come into contact with an infected person and recommend a change in their behavior (or further testing).

\paragraph{\textbf{Digital contact tracing (DCT)}} Predominantly smartphone-based methods of identifying individuals at high risk of contracting an infectious disease based on their interactions, mobility patterns, and available medical information such as test results.

\paragraph{\textbf{Binary contact tracing (BCT)}} Methods which use binary information (e.g. tested positive or negative) to perform contact tracing, thus putting users in binary risk categories (at-risk or not-at-risk). 

\paragraph{\textbf{Proactive contact tracing (PCT)}} DCT methods which produce graded risk information to mitigate spread of disease, thus putting users in graded risk categories (more-at-risk or less-at-risk), potentially using non-binary clues like reported symptoms which are available earlier than test results.

\paragraph{\textbf{Agent-based epidemiological models (ABMs)} } Also called individual-based, this type of simulation defines rules of behaviour for individual agents, and by stepping forward in time, contact patterns between agents are generated in a ``bottom up'' fashion. Contrast with \textbf{Population-level epidemiological models}.

\paragraph{\textbf{Risk of infection} } Expected infectiousness, or probability of infection for a susceptible agent by an infectious agent given a qualifying contact (15+ minutes at under 2 meters).

\paragraph{\textbf{Contact}} An encounter between 2 agents which lasts at least 15 minutes with a distance under 2 meters \cite{canada-public-health-prolonged-exposure}.

\paragraph{\textbf{Risk Level}} A version of the risk of infection quantized into 16 bins (for
reducing the number of bits being exchanged,
for better privacy protection). The thresholds are selected by running the domain randomization with separate seeds and grouping risk messages such that there are an approximately uniform number in each bin.

\paragraph{\textbf{Risk mapping}} A table which maps floating point risk scalars into one of 16 discrete risk levels. 

\paragraph{\textbf{Recommendation mapping}} A table of which recommendation level should be associated with a given risk level.
Each recommendation level comes with a series
of specific recommendations
(e.g. ``Limit contact with others'', ``wash hands frequently'', ``wear a mask when near others'', 
``avoid public transportation'', etc) which should be shown for a given risk level. Reminders of these recommendations may be sent via push notification more or less often depending
on the recommendation level.

\paragraph{\textbf{Population-level epidemiological models}} This type of model fits global statistics of a population with a mathematical model, typically sets of ordinary differential equations. The equations track some statistics over time, typically counts of people in each of a few ``compartments'', e.g. the number of Infected and Recovered, and the parameters of the equations are often tuned to match statistics collected from real data. It does not give any information about the individual-level contact patterns which
give rise to these statistics.

\paragraph{\textbf{Compartment model (SEIR)} } A compartment model tracks the counts of agents in each of several mutually-exclusive categories called ``compartments''. In an SEIR model, the 4 compartments are \textbf{Susceptible, Exposed, Infectious, and Recovered} (see entries for each of these words).

\paragraph{\textbf{Susceptible}} At risk of catching disease, but not infected.

\paragraph{\textbf{Exposed or infected}} Infected with the disease, (i.e. potentially carries some \textbf{viral load}).

\paragraph{\textbf{Infectious}} Carries sufficient viral load to transmit the disease to others. In real life, this typically means that the virus has multiplied sufficiently to overwhelm the immune system.

\paragraph{\textbf{Recovered}} No longer carries measurable viral load, after having been
infected. In real life this is typically measured by two successive negative \textbf{lab tests}.

\paragraph{\textbf{Viral load}} Viral load is the number of actual viral RNA in a person as measured by a \textbf{lab test}.

\paragraph{\textbf{Effective viral load}} A term we introduce representing a number between 0 and 1 which we use as a proxy for viral load. It could be converted to an actual viral load via multiplying by the maximum amount of viral RNA detectable by a lab test.

\paragraph{\textbf{Infectiousness}} Degree to which an agent is able to transmit viruses to another agent. A factor in determining whether an encounter between a susceptible and an infectious agent results in the susceptible agent being exposed (it is a property of the infector agent; this does not consider environmental factors or the other agent's susceptibility; see \textbf{Transmission probability}).

\paragraph{\textbf{Attack rate}}  The proportion of individuals in a population who are infected over some time period.

\paragraph{\textbf{Contamination duration} } How long a location remains infectious after an infected person has visited it and shed viruses there.

\paragraph{\textbf{Transmission probability} } The chance that one agent infects another during an encounter. \textbf{Infectiousness} of the infected agent, \textbf{susceptibility} of the other agent, both their behaviour (e.g. usage of masks) and other environmental factors all play a role in determining this probability. 

\paragraph{\textbf{PCR Test}} Also, colloquially referred to as a lab test. In this context, this means a Polymerase Chain Reaction (PCR) test used to amplify viral DNA samples, typically obtained via a nasopharangeal swab.

\paragraph{\textbf{Asymptomatic}} When an infected person shows no symptoms. For COVID-19, many people remain asymptomatic for the entire course of their disease.

\paragraph{\textbf{Incubation days} }  The number of days from exposure (infection) until an agent shows symptoms. In real life this is the average amount of time it takes the viral RNA to multiply sufficiently in / burst out of host cells and cause an immune response, which is what produces the symptoms.

\paragraph{\textbf{Reproduction number ($R$)} } The average number of other agents an agent infects, measured over a certain window of time. We follow \cite{gupta2020covisim} and approximate $R$ by computing the infection tree and taking the ratio of the number of infected children divided by the number of parents who are recovered infectors.

\paragraph{\textbf{Serial interval} } The average number of days between successive symptom onsets (i.e. between the \textbf{incubation} time of the infector and the infectee).

\paragraph{\textbf{Generation time}} The number of days between successive infections (i.e. between the exposure of the infector and the infectee). Generation time is difficult
to measure directly so one can estimate it
from the {\bf serial interval}.

\paragraph{\textbf{Domain randomization}} A method of generating a training dataset by sampling from several distributions, each
corresponding to a different setting of
parameters of a simulator.
 This allows to generate a dataset which covers a potentially wide range of settings, improving generalization to environments which
 may not be well covered by the simulator
 in any one particular setting chosen
 a priori. 

\paragraph{\textbf{Oracle Predictor} } Baseline for the \textbf{Set Transformer} which uses ground-truth infectiousness levels as ``predictions''. Provides an expected upper-bound performance for the predictors and can be used
to provide risk level inputs when pre-training a
predictors.

\paragraph{\textbf{Multi-Layer Perceptron (MLP)} } A risk prediction model which concatenates all inputs (after heuristically aggregating to a fixed size vector the set-valued inputs), feeds these through several fully-connected neural network layers and is trained to predict infectiousness. See Section \ref{app:architectures}.

\paragraph{\textbf{Set Transformer (ST) Model} } A risk prediction model which uses a modified set transformer to process and attend to inputs, and is trained to predict infectiousness. See Section \ref{app:architectures}.

\paragraph{\textbf{Deep Set (DS) Model}} Similar structure as the Set Transformer, but using pooling instead of self-attention. This allows it to use less memory and compute as compared to the Set Transformer.

\paragraph{\textbf{Heuristic (HCT) Model} } A rule-based method for predicting risk histories and corresponding current behavior levels developed in \cite{gupta2020covisim}.

\paragraph{\textbf{Adoption rate} } The proportion of the total population which uses a digital contact tracing application.

\paragraph{\textbf{Domain randomization} } The technique of varying the parameters of a simulation environment to create a broad distribution of training data.

\paragraph{\textbf{Re-identification} } The process of matching anonymous data with publicly available data so as to determine which person is the owner.

\paragraph{\textbf{Big brother attacks} } ``Big brother'' is a reference to the 1984 book by George Orwell where a highly regulated government surveilled and controlled its citizenry. The term ``big brother attacks'' references the idea that an organized group (e.g., state, federal, academic or financial institution) may try to gain control of your data.

\paragraph{\textbf{Little brother attacks} } Little brother attacks references the idea that potential security threats are also posed by smaller actors like individuals, criminals, and data brokers. 

\paragraph{\textbf{Vigilante attacks} } Extra-judicial violence by an individual. In the context of this paper, specifically as a result of a recommendation shown to the user or gained through other means.

\paragraph{\textbf{Degree of Restriction} } Ratio between the number of people given a recommendation in the highest level of restriction (i.e. quarantine) relative to other categories.

\paragraph{\textbf{Global mobility scaling factor} } A simulator parameter enabling us to vary the amount of contacts which may lead to infection in the simulator. It allows us to scale the mobility to simulate pre-lockdown or post-lockdown environments.

\paragraph{\textbf{Auto-induced distribution shift} } A change in distribution of data observed by an agent or algorithm as a result of the agent or algorithm's actions.

\paragraph{\textbf{Iterative re-training} } The process of generating data using some method in a simulator, training a model on this data in a supervised manner, then evaluating the model in the simulator to produce more data.

\paragraph{\textbf{Pareto frontier} } The Pareto frontier is a way of evaluating a tradeoffs in a set of policies and environments with multi-dimensional outputs (e.g., viral spread and mobility). 

\newpage

\section{Motivating Example}
\begin{figure}[ht]\centering
   \includegraphics[width=\linewidth]{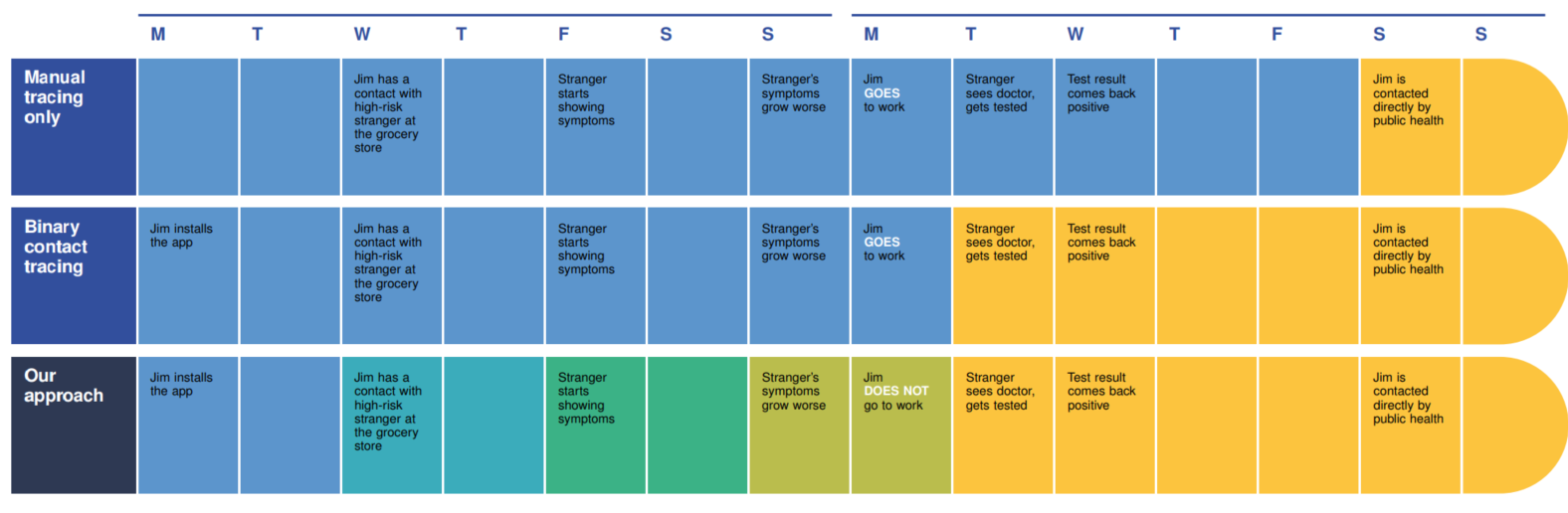}
    \caption{\textbf{Motivating example comparing manual, binary, and proactive contact tracing}: This example shows the potential effectiveness of early warnings in controlling the spread of the infection. Manual tracing is delayed because of the time
    between diagnosis and calling all contacts. Both manual and digital contact
    tracing are sending late signals because they only make use of the strongest
    possible signal (positive diagnosis). The proposed ML approach takes advantage
    of reported symptoms and the propagation of risk signals between phones to
    obtain much earlier signals.}
    \label{fig:early-awareness-example}
\end{figure}

\section{Comparison of approach to related work}\label{sec:rel-work}
\noindent{\bf Epidemiological Modeling and Simulations.} Given that modeling contact-tracing requires capturing past interactions, it is mathematically complicated to consider the dynamic nature of network interactions.
An agent-based model (simulator), on the other hand, gives us maximum flexibility to incorporate real data and/or assumptions easily
and emulate the effect of personalized
policies (such as resulting from the proposed
app).
Models with binary contagion and random-walk mobility are ubiquitous  \cite{stevens2020}.
The appeal of such models is their simplicity; they are easy to code, fast to run, and can give a general picture of some aspects of disease spread. There are other models with increasing complexity of either the mobility model, contagion model, agent demographics, or some combination of these, e.g. \cite{verity,gleam,wood2020planning, lorch2020spatiotemporal,hinch2020effective}.

GLEAM (Global Epidemic And Mobility model) \cite{gleam} is an off-the-shelf simulation platform for epidemics which offers mobility patterns and demographic information, and uses a generic format for defining how a disease depends on these two things.
Similarly, FRED (Framework for Reconstructing Epidemiological Dynamics) provides an open-source, agent-based model with realistic social networks and US demographics. \cite{FRED}, and \cite{wood2020planning} build an intervention-planning tool on top of this simulator. These works are comparable to our simulator only; they do not do any contact tracing or individual risk prediction. 

The work presented in \cite{lorch2020spatiotemporal} and \cite{ferretti2020quantifying} is closely related to ours.
A detailed mobility model for interactions is presented  in \cite{lorch2020spatiotemporal}, but the epidemiological model of the disease is much simpler there.
First, the contact graph is built based on their mobility model. The next step is an  implementation of various policy interventions by health authorities, which includes contact tracing. This is done instead of  building a contact graph that takes into account policy interventions and contact tracing apps at an individual level. Next, most similarly to our work, \cite{ferretti2020quantifying} proposes an app-based tracing and recommendations. 
However, their epidemiological model is a very simple differential-equation   model, and interventions for controlling the disease are computed analytically.


\noindent{\bf Risk estimation approaches.} While the applications deployed so far are based primarily on binary contact tracing as discussed above,   some probabilistic risk estimation approaches, similar to ours, have been developed for other diseases or applications, and have begun to be applied to COVID-19.  For example,   \cite{baker2020probabilistic} uses the susceptible-infected-recovered (SIR) model and describes the dynamical process of infection propagation using the dynamical message passing equations from \cite{lokhov2014inferring};   the probability of each node (person) to be in a specific state (S,I or R) is estimated via the dynamic message-passing (DMP) algorithm, which
belongs to the family of  local message-passing methods similar to belief propagation (BP) algorithm  \cite{yedidia2001generalized}  for  estimating marginal probability distributions over the network nodes; despite BP being only an approximate inference method,   not  guaranteed to converge to the correct marginals when the underlying graphical model has (undirected) cycles, it  demonstrated remarkable performance in various applications. A similar approach based on BP was also discussed in \cite{KevinModel}. 
Furthermore, there are recent extensions of belief propagation approach graph neural nets \cite{satorras2020neural}. Also, another recent work uses Gibbs sampling and SEIR model for their test-trace-isolate approach \cite{CRISP}.
However, such approaches typically rely on the   knowledge of the social  interaction graph, which is not available in our case due to privacy and security constraints.

To the best of our knowledge, our work is the first to use an approach based on  detailed agent-based epidemiological model together with a model of phone app messaging to generate simulated data for training an ML-based predictor of individual-level risk.

\textbf{Digital contact tracing for COVID-19}
\cite{hinch2020effective, hellewell2020feasibility, aleta2020modeling, grantz2020maximizing} study, either via simulations or mathematical models, the conditions under which BCT can be effective. Toward addressing the issues with BCT, ~\cite{hinch2020effective} show in simulation that using self-reported symptoms in addition to test results can greatly help control an outbreak.
Probabilistic (non-binary) approaches to the problem of contact tracing (e.g. \cite{baker2020probabilistic,satorras2020neural, briers2020risk}) typically assume full access to location histories and contact graph, an unacceptable violation of privacy in most places in the world. As a result, these methods are most often used for predicting overall patterns of disease spread.

\textbf{Agent-Based Models as a generative process}
Our use of an agent-based simulator \cite{gupta2020covisim} as a generative model allows us to generate fine-grained (continuous) values for expected infectiousness with realistic contact patterns. 
Most works performing probabilistic inference for disease modeling use a simple differential equation generative model, which does not characterize individual behaviour but rather the dynamics of transition between each of several mutually exclusive disease states.
Such models make many simplifying assumptions, such as contact patterns based on random walks, which make them unsuitable for individual-level prediction of infectiousness; they are typically used instead to infer latent variables such as infectiousness that would be consistent with the population-level statistics generated by the differential equation model, and/or to predict statistics of spread of the disease in a population, e.g. (ref, ref).
While agent-based models are widely used in epidemiological literature to model the spread of disease (see e.g. (ref) for review), to our knowledge we are the first to use an ABM as a generative model for training a deep learning-based infectiousness predictor. 

\textbf{Distributed inference and belief propagation}
Belief propagation in graphical models is often used for disease spread modeling, e.g. \cite{fan}. Some recent works have applied this to COVID-19; for example, \cite{baker2020probabilistic} use the susceptible-infected-recovered (SIR) model and describe the process of infection propagation using the dynamical message passing equations from \cite{lokhov2014inferring}. 
A work concurrent with ours follows a similar justification for modeling a latent parameter of expected infectiousness, using an SEIR model with inference via \cite{CRISP}. However, these approaches rely on a centralized social graph or a large number of bits exchanged between nodes, which is challenging both in terms of privacy and bandwidth. This challenge motivated our particular form of distributed inference where we pretrain the predictor and do not assume that the messages exchanged are probability distributions, but instead just informative input to the node-level predictor.

\section{Experimental details} \label{app:experimental_details}


\subsection{ML architectures and baseline details:} \label{app:architectures}

\paragraph{Binary contact tracing} 
quarantines app-users who had high-risk encounters with an app-user who receives a positive PCR test. Under BCT1, if Alice gets a positive test result, then every user who encountered Alice within 14 days of her receiving the positive test result is sent a message which places them in app-recommendation level 3 (quarantine) for 14 days. Formally, $\zeta_i^d = \psi(\hat y_i^d) = 3$.


\paragraph{Set Transformer} 
Recall that in Section~\ref{sec:methodology-for-infectiousness-estimation} we proposed two parameterizations of the model (DS and ST in Figure \ref{fig:model-architecture}). In the first proposal, we use a set transformer (ST) to model interactions between the elements in the set $\mathbb{D}_i^d \cup \mathbb{E}_i^d$. We now describe the precise architecture of the model used. 

The model comprises 5 embedding modules, namely: the health status embedding $\phi_{hs}$, health profile embedding $\phi_{hp}$, day offset embedding $\phi_{do}$, risk message embedding $\phi_{e}^{(r)}$ and an embedding $\phi_{e}^{(n)}$ of the number of repeated encounters. 

The model was trained for 160 epochs on a domain randomized dataset (see below) comprising $\sim 10^7$ samples. We used a batch-size of 1024, resulting in $\sim 80k$ training steps. The learning rate schedule is such that the first $2.5k$ steps are used for linear learning-rate warmup, wherein the learning rate is linearly increased from $0$ to $2 \times 10^{-4}$, followed by a cosine annealing schedule that decays the learning rate from $2 \times 10^{-4}$ to $8 \times 10^{-6}$ in $50k$ steps. 


\subsection{Domain randomization} \label{app:training}

Inspired by research in hyper-parameter search \cite{JMLR:v13:bergstra12a} and recent advances in deep reinforcement learning \cite{tobin2017domain} we created the transformer's training data by sampling uniformly in the following ranges:

\begin{enumerate}
    \item Adoption rate $\in [30 - 60]$
    \item Carefulness $\in [0.5 - 0.8]$ 
    \item Initial proportion of exposed people $\in [0.002, 0.006]$
    \item Oracle additive noise $\in [0.05 - 0.15]$
    \item Oracle multiplicative noise $\in [0.2 - 0.8]$
    \item Global mobility scaling factor $\in [0.3 - 0.9]$
    \item Symptoms dropout: Likelihood of not reporting some symptoms $\in [0.1, 0.6]$
    \item Symptoms drop-in: Likelihood of falsely reporting symptoms $\in [0.0001, 0.001]$
    \item Quarantine dropout (test) $\in [0.01, 0.03]$: likelihood of not quarantining when recommended to quarantine due to a positive test 
    \item Quarantine dropout (household) $\in [0.02, 0.05]$: likelihood of not quarantining when recommended to quarantine because a household member got a positive test 
    \item All-levels dropout $\in [0.01, 0.05]$: likelihood of not following app-recommended behavior and instead exhibit pre-pandemic behavior 
\end{enumerate}

\subsection{Training time} 
Our ML experiments use approximately 250 days training time on GPUs while simulations required approximately 41 days of CPU time. All CPU time was run on compute using renewable resources.

\newpage

\section*{Relevant Background on COVI-AgentSim }

Details on the simulator can be found in \cite{gupta2020covisim}, and the code is open-source at \url{https://github.com/mila-iqia/COVI-AgentSim}. Here we summarize some of the most relevant details about the simulator for ease of reference.

\subsection{App adoption}
COVI-AgentSim models app adoption proportional to smartphone usage statistics, shown in Table \ref{tab:adoption}.

\begin{table}[h]
\begin{center}
\begin{tabular}{c|c}
\% of population with app & Uptake required to get that \%  \\
\hline
~1 & ~1.50 \\
30 & 42.15 \\
40 & 56.18 \\
60 & 84.15 \\
70 & 98.31 
\end{tabular}
\end{center}
\caption{
\textbf{Adoption Rate vs Uptake}: The left column show the total percentage of the population with the app, while the right column shows the proportion of \textit{smartphone users} with the app. }
\label{tab:adoption}
\end{table}

\vspace{-4pt}
\subsection{Recommendation levels and containment protocol}

In \cite{gupta2020covisim}, the implementation of binary contact tracing uses a simple, binary containment protocol: quarantine at home for 14 days if in contact with someone who has had a positive test result, otherwise do nothing. In contrast, the graded recommendation levels used by feature-based contact tracing methods including Heuristic and Proactive Contact Tracing are not binary, and do not model an explicit containment protocol. 
Instead, \cite{gupta2020covisim} models the effects of behavioural recommendations through reducing the number of daily effective contacts with respect to pre-confinement contact rates defined by \cite{Prem2017ProjectingSC}.

We denote by $C_l$ the number of contacts that occurred at a location $l$ during pre-confinement. 
While operating under recommendation level 0, an agent visiting location $l$ will on average contact $C_l$ other agents.
Post-confinement contact patterns surveyed in Quebec during summer 2020~\cite{brisson2020} are used to scale down the number of pre-confinement interactions across different locations, the percentage reduction per location are denoted $\alpha_l$.
Table \ref{tab:reduction} show the number of contacts and percentage reduction as a result of confinement.

\begin{table}[h]
\begin{center}
\begin{tabular}{l|l|c}
Location  &  $C_l$  & $\alpha_l$ \%  \\
\hline
Household & 2.7 & 0.30 \\
Workplace & 10 & 0.80 \\
School & 6 & 0.80 \\
Other & 3.1 & 0.50 \\
\end{tabular}
\end{center}
\caption{\textbf{Daily contacts per location type:} This table shows for each location $l$ the pre-confinement mean number of daily contacts $C_l$ and the reduction in number of contacts $\alpha_l$ based on data collected in the Region of Montréal.}
\label{tab:reduction}
\end{table}

There are 5 recommendation levels described by Table \ref{tab:recommendation} with their corresponding expected number of contacts.
When these recommendations are followed, they modulate the number of effective contacts that agents have while at locations.
Recommendation level 3 scales down the contacts at a location by $\alpha_l$. 
Levels 1 and 2 are successively obtained by reducing the location dependent contacts by half from level 3, while level 4 is used to impose full quarantine i.e. no contacts.
Level 0 is not used except to represent the behavior when an agent does not adhere to the recommendations modeled by a daily dropout parameter.

\begin{table}[ht]
\begin{center}
\begin{tabular}{l|l|c}
Recommendation\\ Level  &  Description  & Effective contacts  \\
\hline
Level 0 & Pre-pandemic behaviour & $C_l$ \\
Level 1 & Intermediate recommendations reducing contacts & $1/4 * ( 1 - \alpha_l) * C_l$ \\
Level 2 & Stronger recommendations reducing contacts & $1/2 * ( 1 - \alpha_l) * C_l$ \\
Level 3 & Post-confinement behaviour & $ (1 - \alpha_l) * C_l $ \\
Level 4 & Quarantine/self-isolation & $0$
\end{tabular}
\end{center}
\caption{\textbf{Recommendation level and corresponding effective contacts:} Recommendation levels and corresponding expected number of daily effective contacts for a location type $l$. }
\label{tab:recommendation}
\end{table}


\subsection{Infectivity (transmission) model}

Effective viral load (EVL) varies [0., 1.], and represents the 'severity' of viral load the agent is currently experiencing (which takes into account both the viral load and the interaction of the virus with the host's immune system), as shown in Figure \ref{fig:viral-load}. Each agent's disease progresses according to this curve, and the probability of transmission is also proportional to EVL.

\begin{figure}[h]
    \centering
    \includegraphics[width=\textwidth]{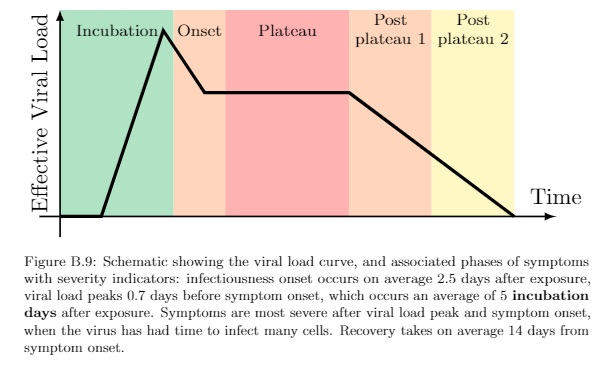}
    \caption{\textbf{Effective Viral Load }[Reproduced with permission from \cite{gupta2020covisim}]: Schematic showing the viral load curve, and associated phases of symptoms with severity indicators: infectiousness onset occurs on average 2.5 days after exposure, viral load peaks 0.7 days before symptom onset, which occurs an average of 5 \textbf{incubation days} after exposure. Symptoms are most severe after viral load peak and symptom onset, when the virus has had time to infect many cells. Recovery takes on average 14 days from symptom onset.}
    \label{fig:viral-load}
\end{figure}


\subsection{Tailoring the simulator to regional demographics and other diseases}

Many aspects of the simulator are customizable to a particular region's statistics, including individual agent features (pre-existing medical conditions, biological sex, etc.), contact patterns,  and prevalence of other diseases (which affect the predictions of COVID-19 by introducing other reasons for an agent to experience symptoms. The simulator provides details of fit to real data for the region of Montreal, Canada, with the confounding symptom-producing conditions of flu, allergies, and colds. The process for tailoring the simulator to other regions is detailed in their work, but briefly, configuration files with the appropriate statistics need to be provided in \url{https://github.com/mila-iqia/COVI-AgentSim/blob/master/src/covid19sim/configs/}.

\Chapter{GENERAL DISCUSSION} 

AI and ML algorithms are often described (and developed) in the language of probability, particularly the framework of empirical risk minimization. In this framework, several assumptions are typically made: that examples in the dataset are representative of an underlying distribution of interest, and that they are sampled independently and identically distributed from that underlying distribution. 

While making these assumptions has been of great practical use in developing deep learning systems for a wide rage of tasks, it is clear that they do not hold for many real-world data. 





Taken together, the reviewed works in this thesis paint a picture of implicit regularization from various sources performing implicit model selection, shaping the hypothesis space such that certain patterns and invariances are quicker and easier to learn. Specifically, in the presence of real data distributions, there is an early \textbf{pattern-learning regime}, mapping the space of possiblities, characterized by good early generalization performance and inconsistency in easiness of examples, followed by an \textbf{interpolation regime}, characterized by memorization and consistency in easiness of examples. Some reasults, such as \cite{doubledescent}, suggest these two phases, or others, may repeat in the learning dyanamics of large, deep networks.

Compositional representations allow for greater effective capacity, and this capacity is necessary for smooth interpolation between data points \cite{robustnesslaw}, which is in turn necessary for robust out-of-distribution generalization.  However, compositionality is not sufficient to ensure good generalization \cite{arpit2017, Zhang, szegedy2013intriguing}. Scaling laws \cite{scalinglaws} and the Bitter Lesson \cite{bitterlesson} suggest good performance according to loss and accuracy metrics will scale when scaling up data and compute. But the specification gap, issues with bias, distributional shift, and other failures of assumptions also scale.
 
The implicit regularization given by mini-batching, frozen subspaces of weights, active learning in the near-online setting of massive data regimes, etc. all provide something of a counterpoint to the bitter lesson of performance scaling with compute: resource constraints can often lead to better generalization. This gives us hope that if we could find effective ways to ensure or specify alignment, i.e. to regularize or constrain the model toward desirable solutions, it could actually lead to more performant models via improved generalization.

Perhaps the most important conclusion from \cite{arpit2017} is that learning behaviour depends on the data distribution; when patterns are present in the data, deep networks will first learn and exploit them to generalize well, before memorizing. While this was a commonly held belief among AI practitioners, recent works had called this perspective into question, and we were the first to provide solid evidence for this folk wisdom. This work and its collaborators sparked a wave of empirical investigations into learning theory for deep networks, an area previously dominated by theoretical results that translated poorly to the real-world usages of deep nets in AI systems.

Designing auxilliary predictive tasks, sandboxing, and unit-testing are all examples of methodology that can help us  ensure better generalization in practice for real-world AI systems. These practical methods complement and inform more idealized theoretical analyses such as generalization bounds and scaling laws. They can also provide a basis for important regulation and norm-setting in AI research and practice. For example, unit tests for high-level behaviours of AI systems could be used as an auditing tool, and sandboxes can help perform sensitivity analyses to better understand which methods are appropriate, dangerous, or performant for different applications.


There is not one clean and simple explanation for or way to guarantee good generalization performance for real-world data.  
As the language of probability lacks the tools to distinguish interventional distributions from observational conditionals, it even more generally lacks a way to easily describe context, structure, and relationships in data. The fundamental particle of probability is a random variable - one whose value is determined by a random process. Most variables in real-world data are far from randomly determined. While statistical learning theory has great value for describing and understanding AI systems, we also need other ways of knowing and communicating about (and with) AI systems in order to ensure robust generalization and alignment.



%


\Chapter{CONCLUSION}\label{sec:Conclusion}

In summary, this thesis (I) presents an introduction to the fields of artificial intelligence, machine learning, and deep learning (II) reviews recent work in understanding generalization behaviour of deep networks, with a focus on real, large-scale data distributions (III) presents my original contributions to the field in understanding memorization and generalization behaviour of deep nets; developing improved metrics and methods of evaluation; and applying novel methods to the real-world problem of digital contact tracing; and (IV) discusses insights from this body of work. Here I conclude by discussing limitations, takeaways, challenges, and opportunities for future work.

\section{Limitations}\label{sec:Limitations2}
An overall limitation of this work, as mentioned in the notes on context in the Preamble, is that it focuses on technical aspects of fundamentally socio-techincal issues. It also focuses on works understanding deep neural networks, with little consideration of other machine learning techniques. While deep learning is by far the dominant paradigm in real-world AI systems currently, more detailed comparison with alternatives could shed more light on which aspects of DL have made it so successful, and perhaps help us better understand generalization and learning behaviour. Further, while some pure-theory works are reviewed, this thesis focuses more on empirical insights (and resulting theory). While this can give a more realistic picture of real-world behaviour, and potentially identify phenomena that would not be encountered in more idealized settings, there is also great value in purely theoretical or mathematical inquiry.

The individual works also have particular limitations.

In Chapter 2, we compare behavior on random labels and random inputs for image data, where random inputs are created by some proportion of the images in the dataset being replaced with randomly-sampled Gaussian noise. Comparing to different (and more realistic) noise distributions could have given us a fuller picture and would be an interesting direction for future work. Similarly, in the feed-forward case, we ran experiments only on images; other modalities could potentially present other interesting results, as seen in our initial study of recurrent neural networks. Our RNN study also has its own limitations; the experiments we run are not as extensive as those run for feed-forward nets. For instance, this initial study did not examine the critical sample ratio or effects of different regularizers in deep recurrent networks.

In Chapter 3, the regularization method we proposed is limited in practical use to recurrent neural networks; the analogous method (per-unit stochastic depth) we test for feed-forward networks is out-performed by other regularization methods such as dropout and layer-wise stochastic depth. It would be an interesting study for future work to test analogous ideas to Zoneout on newer architectures such as Transformers.

In Chapter 4, while the main objective of our work is to address limitations of word-overlap measures common for evaluating captions and video representation quality, this work is also limited by requiring a fixed, pre-determined vocabulary. Later works in masked language modeling with transformers, e.g. \cite{bert}, address this shortcoming with bytepair encodings.

In Chapter 5, our work in content recommendation uses a very minimal environment to showcase the effects of ADS, but a more large-scale or real-world experiment could help to better illustrate the phenomena we are interested in to really show the utility of the unit tests. Also, as mentioned in that work, the environment swapping mitigation strategy does not appear to solve issues in all contexts. Alternative mitigation strategies are necessary, as is a more detailed understanding of incentives and how to manage them. In the work on Covid-19, our strategy for dealing with distributional shift, which we call iterative refinement, is not very data efficient (it requires retraining the model each time). Alternative strategies which could adapt to the shift in real time would greatly improve data efficiency.

In Chapter 6, the main limitation of this work is that it should not be considered a full study into proactive contact tracing for epidemic management; only an initial study on the ML-related aspects. The next stage for this research should be a randomized control trial in a real-world experiment. Variables of especial interest are the adoption rate and other aspects of human behaviour; while we consulted with behavioural experts, we found that all methods were fairly sensitive to adoption rate, and in general human behaviours are complex, with many influences and potential external factors that simulation studies are insufficient to address.


\section{Takeaways, Challenges, and Opportunities}

The AI systems that have achieved great success at bird-species labelling and other similar tasks do so via machine learning, in which the process/algorithm/model for performing a task adjusts its behaviour to improve at the task based on examples and a measure of success at the task. In particular, the most successful type of machine learning for high-dimensional tasks has been deep learning, in which a hierarchical (compositional) representation of data is learned from examples, prototypically by using a deep neural network trained with stochastic gradient descent of a predictive objective. 

Deep learning-based AI systems have been extremely successful at tasks in primarily four domains: images (i.e. convolutional neural networks), text (i.e. RNNs/LSTMs, Transformers), audio/speech (i.e. RNNs/LSTMs, dialated convolutional autoregressive networks),  and games (i.e. deep RL). There are several common features of these domains and others where deep learning is successful: (i) they exhibit a lot of structure, with hierarchical and/or compositional dependencies in both space and time, (ii) they are relatively easy to collect very large, well-formatted, labelled or self-supervised datasets for (typically via consumer internet use)  (iii) the `deployment' conditions of the model are relatively similar to the data they're trained on. 

Many of the problems that interest me, such as environmental impact assessment; large-scale planning and decision-support systems; AI alignment; and risk analysis for climate change, sustainability, epidemiology, etc., exhibit similar structure, motivating the use of deep learning techniques.  However, for these problems large well-formatted datasets are frequently more difficult to collect, and failure of the iid assumptions (differences in real-world deployment conditions from those of training) have important impacts.

To address these kinds of issues and responsibly use AI, we need to be able to:
(1) understand and predict generalization performance, including out-of-distribution; 
(2) better specify tasks of interest, and/or align the performance metrics we use;
(3) empirically evaluate systems before deployment in case 1 and 2 don't work.

My work makes progress toward these objectives by (1) examining memorization and generalization behaviour in both feed-forward and recurrent nets (CH2), improving regularization (CH3) and performance under conditions of distributional shift (CH5); (2) examining and evaluating alternative task definitions (CH4, CH6); . See Section \ref{sec:summary} for a detailed summary of results by Chapter.

A key to effectively using AI-based methods for many real-world tasks is to ground the task in a prediction that can be made, for which supervised training data is available or can be generated, without over-optimizing to the point of overfitting or gaming the predictive metric. 
Identifying the predictive task(s) and regularization that are appropriate for a given real-world task is time-consuming and difficult, requiring experts in both domain sciences and in AI. Studying these tasks holistically could inspire more general-purpose and informed ways to produce an aligned specification for a given real-world task, and/or useful evaluation metrics and strategies.

Improved specification is necessary but not sufficient to ensure alignment of AI systems, and in practice, poorly aligned systems cause real-world harms. Approaches from reliability and software engineering, such as unit-testing and sand-boxing, can help identify problems early on and enable third-party auditing of AI systems, and technical approaches to AI safety can help improve alignment in practice. Developing measures of alignment in particular settings could encourage progress on these problems, and provide more general insight on how to improve specifications. Unit tests require expertise to design, and don't exist for many undesirable behaviours. Theoretical results about the performance of AI systems are not typically accessible to non-experts, which can be a significant barrier to translating theory to practice for real-world systems. Better scientific communication (including novel or improved visualizations), illustration through examples, and involvement of diverse stakeholders earlier in the design process of AI systems, can all help with this.

A more systemic view of the deployment of AI/ML models will not only help resolve practical performance, bias, and safety issues with real-world systems, it will help us more fundamentally understand the principles of learning, generalization, and intelligence. 
Understanding real-world generalization is one (important) component of making AI development more responsible, equitable, useful, and safe.



\backmatter
\ifthenelse{\equal{\Langue}{english}}{
	\renewcommand\bibname{REFERENCES}
	\bibliography{Document}
	\bibliographystyle{IEEEtran}			
}{
	\renewcommand\bibname{RÉFÉRENCES}
	\bibliography{Document}
	\bibliographystyle{IEEEtran-francais}    
}

\end{document}